\documentclass{article}

\usepackage[english]{babel}

\usepackage[letterpaper,top=2cm,bottom=2cm,left=3cm,right=3cm,marginparwidth=1.75cm]{geometry}
\usepackage{diagbox}
\usepackage{amsmath, amsfonts, amsthm}
\usepackage{graphicx,soul}
\usepackage[colorlinks=true, allcolors=blue]{hyperref}
\newtheorem{theorem}{Theorem}[section]

\newtheorem{corollary}{Corollary}[section]
\newtheorem{proposition}{Proposition}[section]
\usepackage{amsfonts}
\usepackage{color,xcolor}
\theoremstyle{definition}
\newtheorem{definition}[theorem]{Definition}

\usepackage{multirow}
\usepackage{soul}
\theoremstyle{remark}
\newtheorem{remark}[theorem]{Remark}

\usepackage{algorithm}
\usepackage{algorithmic}
\usepackage{ulem}
\numberwithin{equation}{section}
\usepackage{amssymb}

\newcommand{\cC}{\mathcal{C}}

\newcommand{\cO}{\mathcal{O}}
\newcommand{\bx}{\mathbf{x}}

\newcommand{\by}{\mathbf{y}}

\newcommand{\bP}{\mathbf{P}}

\newcommand{\calH}{\mathcal{H}}

\usepackage{graphicx}
\graphicspath{{../figures/}}

\begin{document}

\title{A Formalization of Image Vectorization by Region Merging}
\author{Roy Y. He, Sung Ha Kang, Jean-Michel Morel}

\maketitle

 \begin{abstract} 
Image vectorization converts raster images into vector graphics composed of regions separated by curves. Typical vectorization methods first define the regions by grouping similar colored regions via color quantization, then approximate their boundaries by B\'{e}zier curves. In that way, the raster input is converted into an SVG format parameterizing the regions' colors and the B\'{e}zier control points.  This compact representation has many graphical applications thanks to its universality and resolution-independence. 
In this paper, we remark that  image vectorization is nothing but an image segmentation, and that it can be built by fine to coarse region merging. 
Our  analysis of the problem leads us to propose a vectorization method alternating region merging and curve smoothing. We formalize the method by alternate operations on the dual and primal graph induced from any domain partition. 
In that way, we address a   limitation of current vectorization methods, which separate the update of regional information from curve approximation.  
We formalize region merging methods by associating them with various gain functionals, including the classic Beaulieu-Goldberg and Mumford-Shah  functionals.  More generally, we introduce and compare  region merging criteria involving region number,  scale, area, and internal standard deviation.  We also show that the curve smoothing, implicit in all vectorization methods, can be performed by the shape-preserving affine scale space. We extend this flow to a network of curves and give a sufficient condition for the topological preservation of the segmentation.
The general vectorization method that follows from this analysis shows explainable behaviors,  explicitly controlled by a few intuitive  parameters. It is experimentally compared to state-of-the-art software and proved to have comparable or superior fidelity and cost efficiency.
\end{abstract}

\section{Introduction}
 
Image vectorization is a powerful technique that yields tangible benefits to many applications, including shape analysis~\cite{he2021finding,he2022silhouette,he2023topology}, content editing~\cite{orzan2008diffusion}, line-drawing reconstruction~\cite{hilaire2006robust,noris2013topology}, and topological data analysis~\cite{ali2023survey}. A raster image is represented by its values on a rectangular grid whose nodes are called  \textit{pixels}.  Such a grid-dependent pixel-level representation is  inefficient~\cite{attneave1957physical,elder1999edges,torralba2009many} and cumbersome for geometric transformations and analysis~\cite{ciomaga2017image,alvarez2017corner}. Another limitation is the pixelation effect~\cite{bachmann2016perception}, where pixel corners do not necessarily correspond to perceived shape corners.  The goal of vectorization is to represent a  raster image by a finite number of parameterized boundary curves and regions with certain rendering rules. The resulting  resolution-free representation, called \textit{vector graphic}, is presented under many variants in the literature~\cite{tian2022survey,dziuba2023image}.

A majority of vectorization methods  produce \textit{piecewise-constant vector graphics}~\cite{he2021finding,he2022silhouette,he2023binary,he2023topology,he2023viva,reddy2021im2vec,li2020differentiable}, where the composing elements are regions filled with constant colors and enclosed by parametric curves such as lines, arcs, and B\'{e}zier curves. These elements are often encoded and organized in a particular format, called Scalable Vector Graphic (SVG)~\cite{ferraiolo2000scalable}. 
Most image viewers and web browsers can instantly read and render\footnote{Rasterization converts vector graphics back to bitmaps, which is the inverse process of vectorization. This rendering is required to display or print a vector graphic. Since this process is generally real-time, it creates an impression of directly manipulating the SVG files. Rasterization follows instantly after modifying the SVG file via well-designed user interfaces.} the contents, and common graphic software such as Adobe Illustrator~\cite{AI} support interactive editing. There are more advanced forms involving mesh generation~\cite{sun2007image} and elliptic partial differential equations (PDEs)~\cite{orzan2008diffusion} for photo-realistic \textit{smooth vector graphics}. 
Typically, the vectorization process involves edge detection or domain segmentation followed by curve fitting for boundaries of homogeneous regions, and these decisions strongly affect the results' visual quality.
In \cite{he2021accurate}, it was proven that binary raster shape vectorization was optimally performed by affine shortening, and  a fast code and online execution of the method is provided~\cite{he2023binary} for binary images. The method~\cite{he2022vectorizing} sketches an extension of the silhouette affine shortening  to general image vectorization relying on color quantization.  A generalized affine shortening is proposed to smooth the set of resulting region boundaries.  The method focuses on the question of moving adequately the curves supporting T-junctions \cite{he2023topology}. 

For color or grayscale images, 
\textit{color quantization} techniques
~\cite{celebi2023forty} are widely applied to find initial domain partitions.  
Well-known methods include median cut~\cite{heckbert1982color},
$k$-means clustering~\cite{lloyd1982least},
Octree~\cite{gervautz1988simple}, self-organizing maps~\cite{chang2005new}, and competitive learning~\cite{celebi2009effective}. 
\begin{figure}
\centering
\begin{tabular}{c@{\hspace{2pt}}c@{\hspace{2pt}}c}
(a)&(b)&(c)\\
\includegraphics[width=0.3\textwidth]{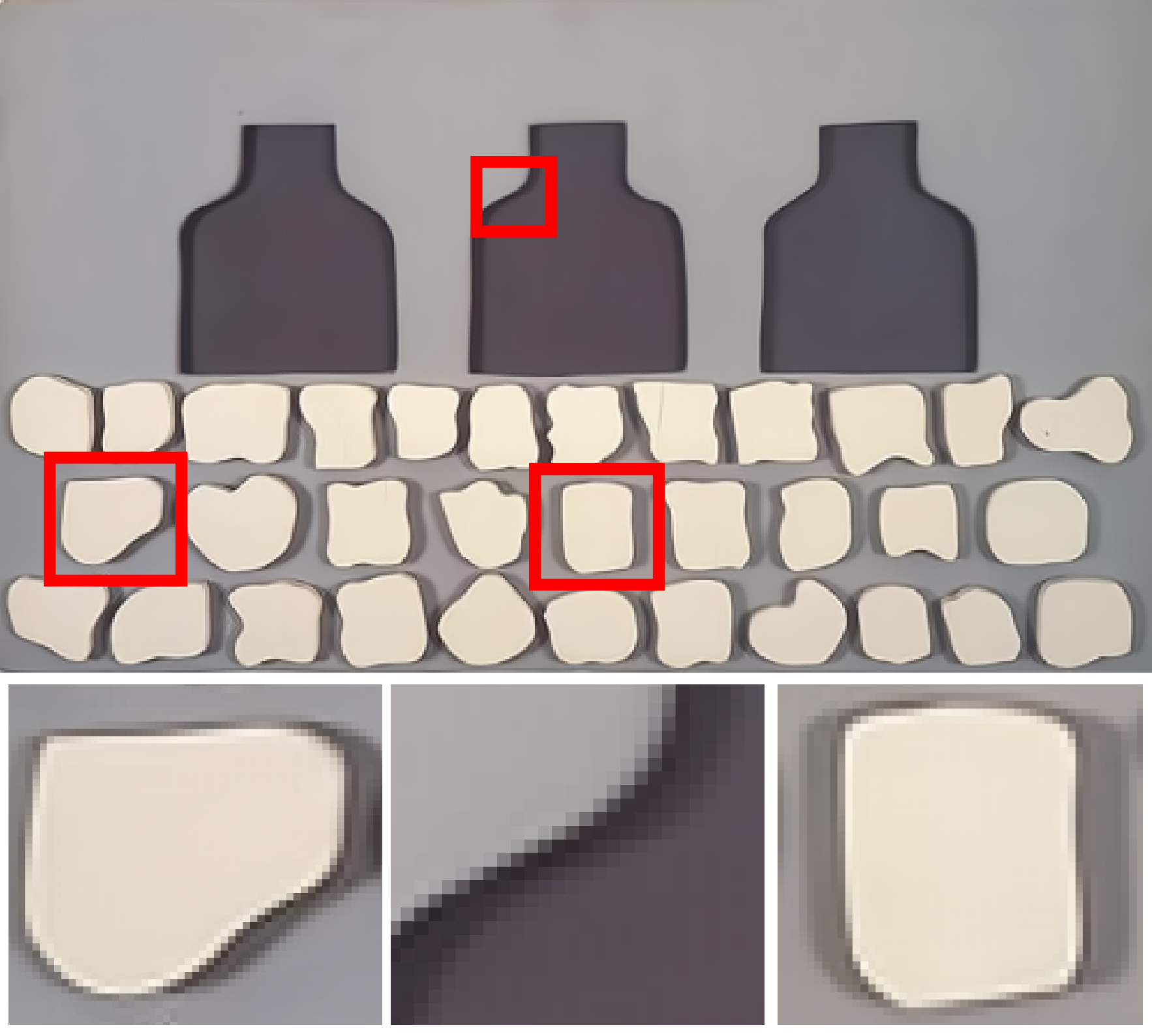}&
\includegraphics[width=0.3\textwidth]{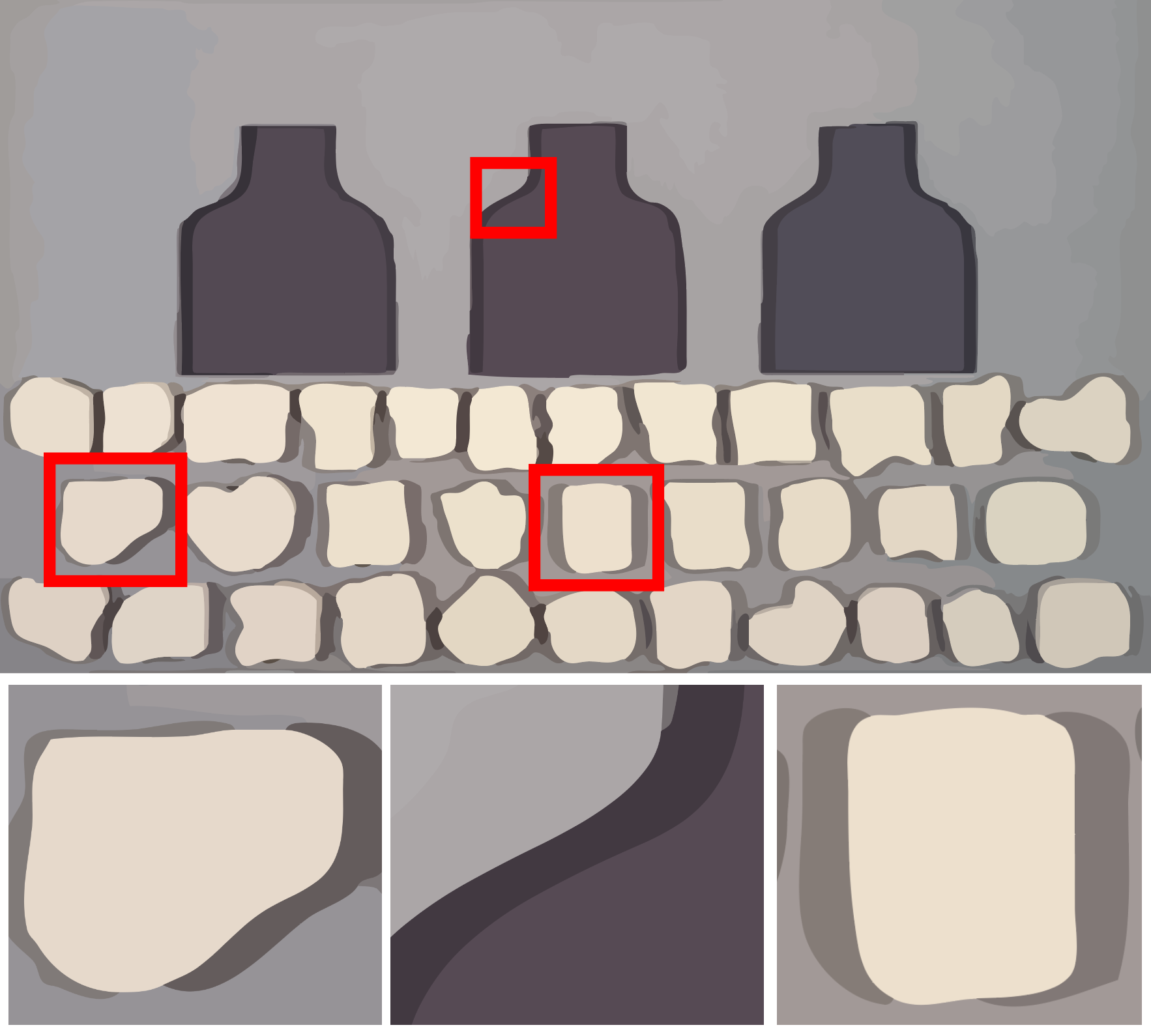}&
\includegraphics[width=0.3\textwidth]{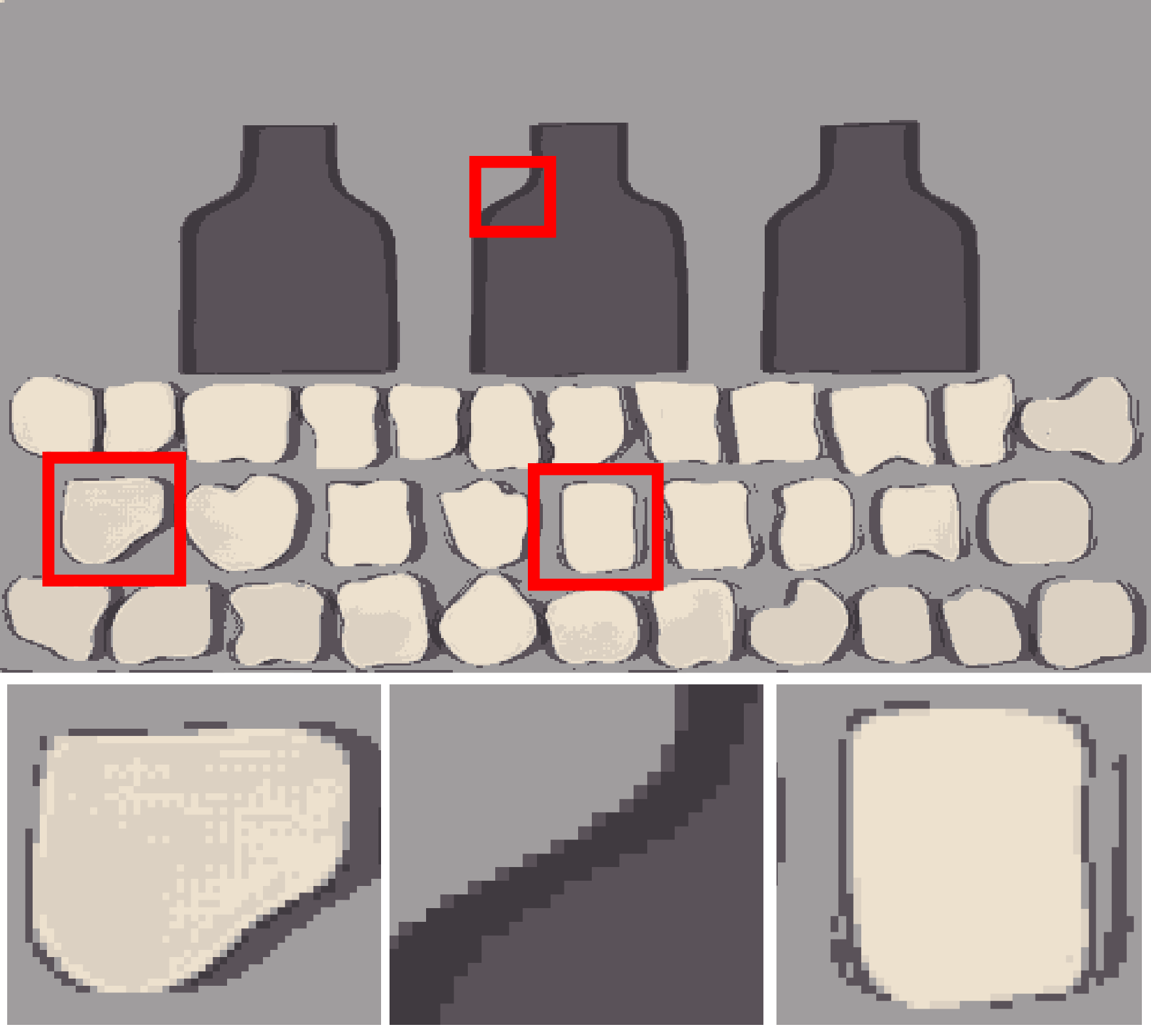}
\end{tabular}
\caption{Number of colors and region complexity. 
(a) Raster image of size $500\times292$ (painting by \textit{Forrest Bess}). (b) The proposed Area region merging~\eqref{eq_area_gain} gives $N=151$ regions, yielding simpler boundaries and flatter regions.  (c) Color quantization by k-means with $k=5$ colors. This results in $2418$ connected components with complicated boundaries.  Region merging (b) reduces the geometric complexity, which is more suitable for efficient vectorization; whereas quantization (c) reduces complexity only in the color space. } 
\label{fig_demo_quant}
\end{figure}
 For example, Figure~\ref{fig_demo_quant} (c) shows a quantization for (a) using $k$-means with $5$ colors, and the connected components are then filled with average colors of the pixels therein. The quantized image contains $2418$ connected components. There are many small regions near the shape boundaries and highly irregular contours  in homogeneous regions.  Dithering~\cite{teuber2011dithering} or half-toning~\cite{gräf2012quadrature} techniques can be used to reduce such contouring artifacts, but they can create even more regions.  Figure~\ref{fig_demo_quant} (b) shows a domain partition based on a region merging method proposed in this paper. It contains only $151$ regions while exhibiting fewer artifacts, thus achieving a highly efficient representation.

A key vectorization criterion is the local regularity of regions and boundaries.  The proximity principle in Gestalt theory~\cite{rock1990legacy} suggests that the similarities in colors, geometry, or textures of nearby regions facilitate grouping unstructured units. There is also evidence showing that this assimilation process eventually contributes to visual perception~\cite{prinzmetal1981principles,torralba2009many}.  When presented with raster images consisting of thousands of squares of constant colors, human observers tend to perceive smooth boundaries or acute angles while ignoring discrete pixel features. The underlying mechanism has been a source of ongoing discussions and research~\cite{dijkstra2019shared,pascucci2023serial}, and it also inspires multiple inventions in computational models that simulate different aspects of visual perception.
Among the various developments~\cite{chan2001active,caselles1997geodesic,jung2007multiphase,bouman1994multiscale,land1971lightness}, one of the most celebrated models is the \textit{Mumford-Shah functional}~\cite{mumford1989optimal}:
\begin{equation}\label{equ:MS}
   \mathcal{E}_f(u,\Gamma)=\int_{\Omega \setminus \Gamma}\|u(\bx)-f(\bx)\|_2^2\,d\bx+\int_{\Omega\setminus \Gamma}\|\nabla u(\bx)\|_2^2\,d\bx+\lambda \int_{\Gamma}\,d\sigma(\bx)\;.
\end{equation}
It models the perceived image $u:\Omega\to\mathbb{R}^d$ ($d=1$ or $3$)  as a regularized approximation for the observation $f:\Omega\to\mathbb{R}^{d}$. In~\eqref{equ:MS},  $\Gamma\subset\mathbb{R}^2$ is the discontinuity set of $u$,  $\lambda>0$ is the regularization parameter,  and $d\sigma$ is the length element. The perceptual simplification is characterized by smoothness of continuous regions in the second term and length-minimization in the third term. Its objective is to partition the image domain into finite regions, each of which ideally corresponds to an object characterized by homogeneous interiors or semantically defined as a whole.

There are natural connections between image segmentation and image vectorization.
Both areas  are motivated by the simplifying process of human visual perception.  Natural objectives 
such as length minimization, contour continuity, and spectral proximity are frequently used in both families of models~\cite{prasad2006vectorized,mumford1989optimal,selinger2003potrace,morel2012variational,dziuba2023image}. 
They emphasize large-scale or intrinsic geometric features of the image while removing  small-scale oscillations caused by pixelation or noise.   Secondly, both focus on detecting critical boundary curves such that the interior details of each enclosed region are deemed minute or deductible. The idea initiated by Elder's studies on edge map~\cite{elder1999edges}, followed by Orzan's development of diffusion curves~\cite{orzan2008diffusion} as well as the further developments~\cite{jeschke2011estimating,finch2011freeform,bezerra2010diffusion} for image vectorization is parallel to the road paved by~\cite{chan2001active,shen2001variational,sapiro2005anisotropic}  for image segmentation; both streams of work contain interesting discussions about edge strength modeling.
On the dual side, both vectorization algorithms such as~\cite{zhang2009vectorizing,favreau2017photo2clipart} and  segmentation methods including~\cite{salman2006image,koepfler1994multiscale,adams1994seeded} exploit a  region-growing strategy. Thirdly, like segmentation, many vectorization algorithms produce partitions of the image domain. Both approaches sometimes appear interchangeable. For example,  Prasad and Skourikhine~\cite{prasad2006vectorized} proposed a segmentation method by modifying their mesh-based vectorization using constrained Delaunay triangulation. Lecot and L\'{e}vy~\cite{lecot2006ardeco} presented a vectorization model by combining Ilyod's formulation of Voronoi partition~\cite{lloyd1982least} with Mumford-Shah's length minimization. 

The primary challenge in recognizing the significance of local regularity in image vectorization lies in the fact that variational segmentation models such as~\eqref{equ:MS} are typically more computationally intensive to address compared to color quantization techniques. Variational models often involve manipulating curves and regions, leading to non-convex and non-smooth optimization problems~\cite{kornprobst2006mathematical}. Various advanced algorithms~\cite{pock2009algorithm, chan2001level} and intricate approximations~\cite{ambrosio1990approximation} have been developed to address this issue. Recent advancements in neural network-based approaches further reduce the complexity by optimizing the shape and color control points of deformable objects which are initially randomly placed~\cite{li2020differentiable,vinker2022clipasso,vinker2023clipascene}. Since these methods allow for transparent regions with overlaps, the outcomes may not always preserve shapes and can be difficult to edit.

In this paper, we investigate a general and interpretable image vectorization method that is built  on \textit{region merging} and \textit{curve smoothing}. 
Starting from each pixel as its own region, we consider several region merging methods associated with classic Mumford-Shah~\cite{mumford1989optimal}, Beaulieu-Goldberg~\cite{beaulieu1989hierarchy}, and two new merging methods based on the scale and area of regions.  We formulate this procedure as modifications on an image-induced \textit{dual graph} which represents the adjacency relations among partition regions;  we also show that such local decisions yield globally consistent partitions.
Then, each region boundary is smoothed by an affine shortening flow, and we analyze a theoretically sufficient condition for topological preservation.  We describe this procedure as a transformation on an image-induced \textit{primal graph}, where image junctions correspond to nodes and boundary curves connecting them correspond to edges.
Both components rely on solid foundations whose behaviors are explicitly controlled via user-friendly parameters. 
We summarize this article's contributions as follows.
\begin{enumerate}
\item{We explore a general formulation of image vectorization using region merging based on image segmentation principles.  }
\item{We introduce partition-induced primal and dual graph structures for efficiently integrating region merging and curve smoothing. }
\item{We discuss the properties of different region merging criteria to show various effects and analyze a sufficient condition for topological preservation during the curve smoothing.}
\item{We are led to a general vectorization method with explainable parameters, and obtain accurate vectorization results that compare favourably to state-of-the-art vectorization methods.}
\end{enumerate}

This paper is organized as follows.  In Section~\ref{sec_proposed_AIV}, we present the general framework of the investigated models and introduce the primal and dual graph structures.  In Section~\ref{sec_merging}, we present the region merging (the dual step) in more depth by exploring and analyzing different region merging criteria.  In Section~\ref{sec_smooth}, we present the details of curve smoothing (the primal step), including the extension to networks of curves and the investigation of topological preservation.  Numerical details are presented in Section~\ref{sec_algorithm}, which is followed by various experimental results in Section \ref{sec_experiment}.  Section \ref{sec::conclude} concludes the paper.

\section{Image Vectorization as a Combination of Region Merging and Curve Smoothing}\label{sec_proposed_AIV}

Let $\Omega\subset\mathbb{R}^2$ be a bounded rectangular domain. An open set in $\Omega$ is called \textit{Caccioppoli} if its essential boundary (or perimeter) has a finite one-dimensional Hausdorff measure (its length) \cite{ambrosio2001connected}. 
We denote by $\mathcal{H}^1$ the one-dimensional Hausdorff measure.
A Caccioppoli partition is a finite set $\mathcal{P}=\{O_1,\dots,O_N\}$ of pairwise disjoint Caccioppoli connected open subsets of $\Omega$  such that the union of their topological closures $\bigcup_{i=1}^N\overline{O_i}$ is $\overline{\Omega}$.
For convenience, we introduce an additional element $O_0$ indexed by $0$ to $\mathcal{P}$, corresponding to the out-of-image-domain region $\mathbb{R}^2\setminus\overline{\Omega}$; thus any partition contains at least two elements.
If $\mathcal{H}^1(\partial O_i\cap \partial O_j)\neq 0$,
for $i,j=1,2,\dots,N$ and $i\neq j$, we say that $O_i$ and $O_j$ are adjacent, and we denote this relation by $O_i\sim O_j$. 
A piecewise constant image $f$ associated with a given partition $\mathcal{P}$ is defined by $f(\bx)= \sum_{i=1}^N c_i\mathbb{I}_{O_i}(\bx)$, where $c_i\in\mathbb{R}^d$ is the constant gray level or color in region $O_i$, with $d=1$ or $3$, and the characteristic function $\mathbb{I}_{O_i}(\bx)=1$ if $\bx\in O_n$ and equals $0$ otherwise.  
We use $\#\mathcal{P}$ to denote the number of regions in $\mathcal{P}$, $N(O_i)$ to denote the number of neighbors of $O_i$, and $|O_i|$ to denote the area of $O_i$.

We define a junction as a point $\bP_k$, $k=1, 2,\dots, K$,  where more than three regions meet. 
For each region $O_i\in\mathcal{P}$, its boundary $\partial O_i$ is a finite union of rectifiable curves without junctions in their respective interiors.   The set $\Gamma$ of all these boundary  curves,  $\cC_m$ for $m=1,2,\dots, M$  has two sorts of elements:  either embedded circles or  embedded  intervals whose end points are junctions.  These notations implicitly depend on the partition $\mathcal{P}$, and we suppress this dependence for simplicity when $\mathcal{P}$ is fixed.

In this paper, we interpret vectorization as an image simplification process that exploits local color regularity through effective region-merging, while ensuring boundary smoothness via geometric invariant curve smoothing.  This leads to using two key operations:
\begin{align}
\textbf{Region merging}\quad&  \text{Cost}(O_i,O_j)< \text{threshold}\;,\quad  O_i,O_j\in\mathcal{P}~\text{with}~O_i\sim O_j \;,\label{region_merging_step}\\
\textbf{Curve smoothing}\quad& \partial_t \Gamma = \mathcal{F}(\Gamma) \;.\label{smooth_step}
\end{align}
In the region merging process~\eqref{region_merging_step}, the cost of merging two adjacent regions is evaluated and compared with a threshold.  The cost of merging models involves color contrast and various shape priors. The merging threshold fixes a desired complexity for the final partition. In the curve smoothing process~\eqref{smooth_step}, the set of boundary curves, $\Gamma$, undergoes an evolution based on a flow $\mathcal{F}$ moving all curves by a geometrically intrinsic equation with junctions fixed. 

\begin{figure}
\centering
\begin{tabular}{ccc}
(a)&(b)&(c)\\
\includegraphics[width=0.25\textwidth]{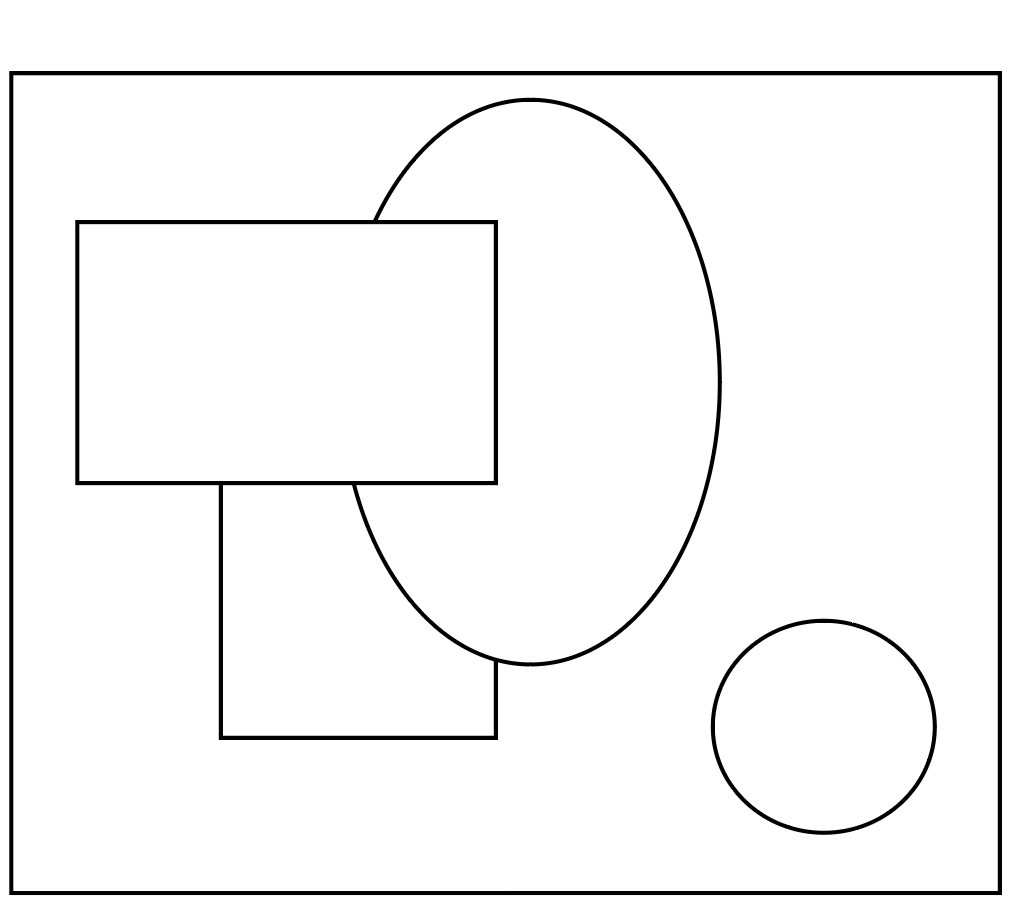}&
\includegraphics[width=0.25\textwidth]{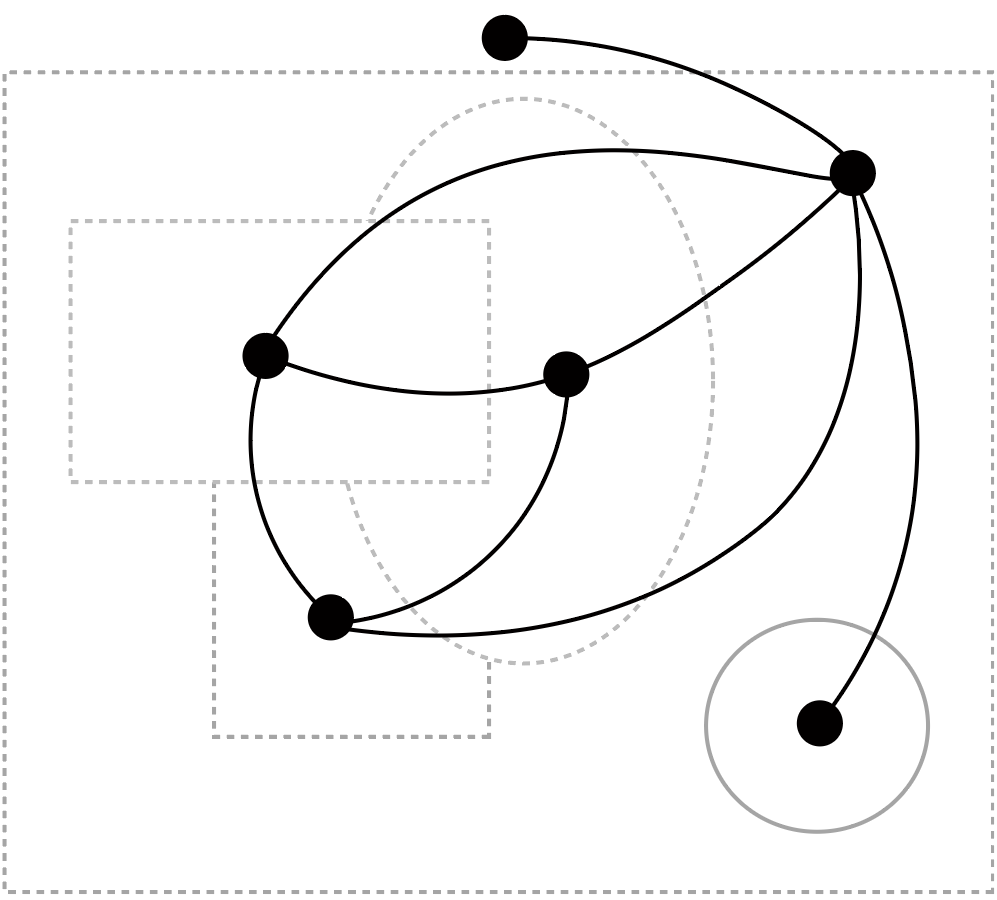}&
\includegraphics[width=0.25\textwidth]{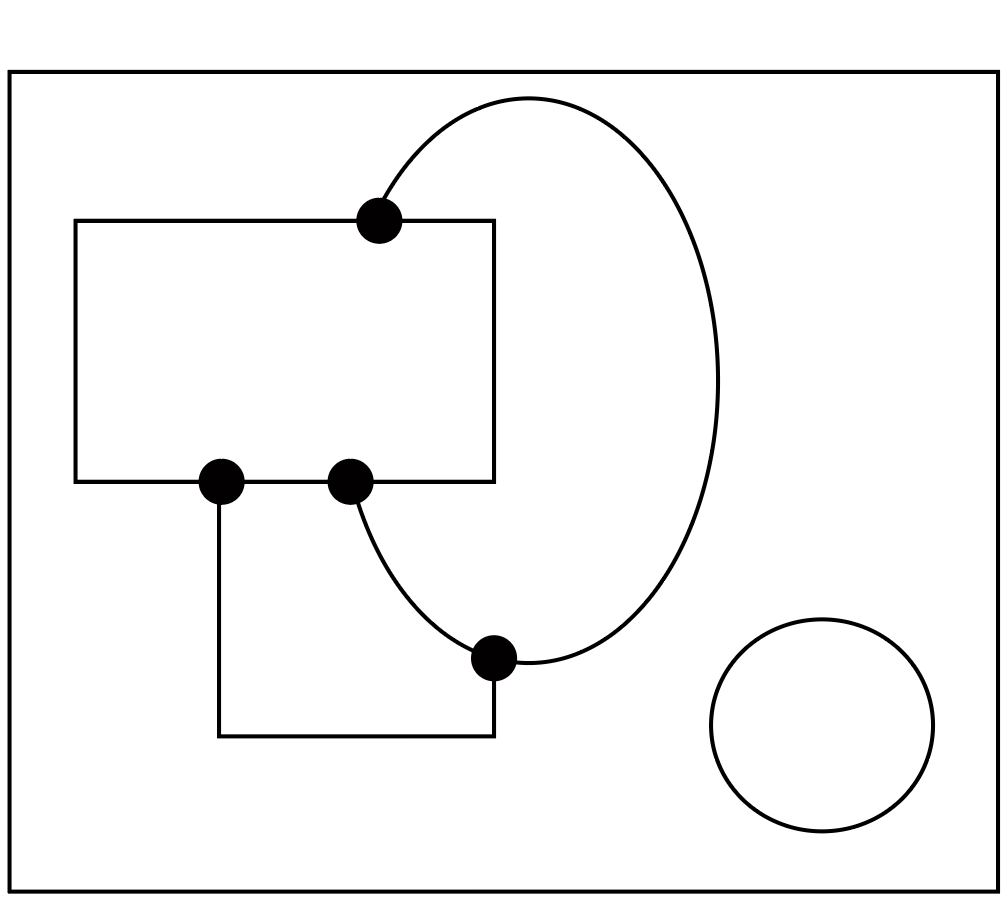}
\end{tabular}
\caption{Primal and dual graph induced from image partition. (a) Region partition  $\mathcal{P}$ of an image $f$. (b) The dual graph represents each region $O_i$ as a node, and two nodes are connected by an edge if the corresponding regions are adjacent. (c) The primal graph represents each junction $\bP_k$ as a node, and two nodes are connected by boundary curves $\cC_m$.  The dual graph reflects the adjacency relations among regions, and the primal graph shows the geometry of the partitioning boundaries.}
\label{fig_dualprimalgraph}
\end{figure}

We formulate the two processes \eqref{region_merging_step} and \eqref{smooth_step} as operations acting on the \textit{dual} and \textit{primal graphs} induced from image domain partitions,  respectively. 
The dual graph encodes the  adjacency relations among regions in the partition, where each node stands for a connected component, and two nodes are connected by an edge if their corresponding regions are adjacent.  The primal graph models the geometry of the partition by denoting each junction point as a node, and two nodes are joined by an edge if they are connected by a boundary curve.  In the primal graph, there may be multiple edges between nodes, and there may be isolated edges incident to no nodes. 
This is illustrated in Figure \ref{fig_dualprimalgraph}. Associated with the region partition $\mathcal{P}$ of the given image $f$ in (a), the dual graph in (b)  represents each region $O_i$ as a node and each adjacency between regions as an edge. The primal graph in  (c) represents each junction $\bP_k$ as a node, and nodes are connected by boundary curves $\cC_m$. Note that some of the boundary curves $\cC_m$ can be closed.

\begin{figure}
\centering
\begin{tabular}{cccc}
(a)&(b)&(c)&(d)\\
\includegraphics[width=0.20\textwidth]{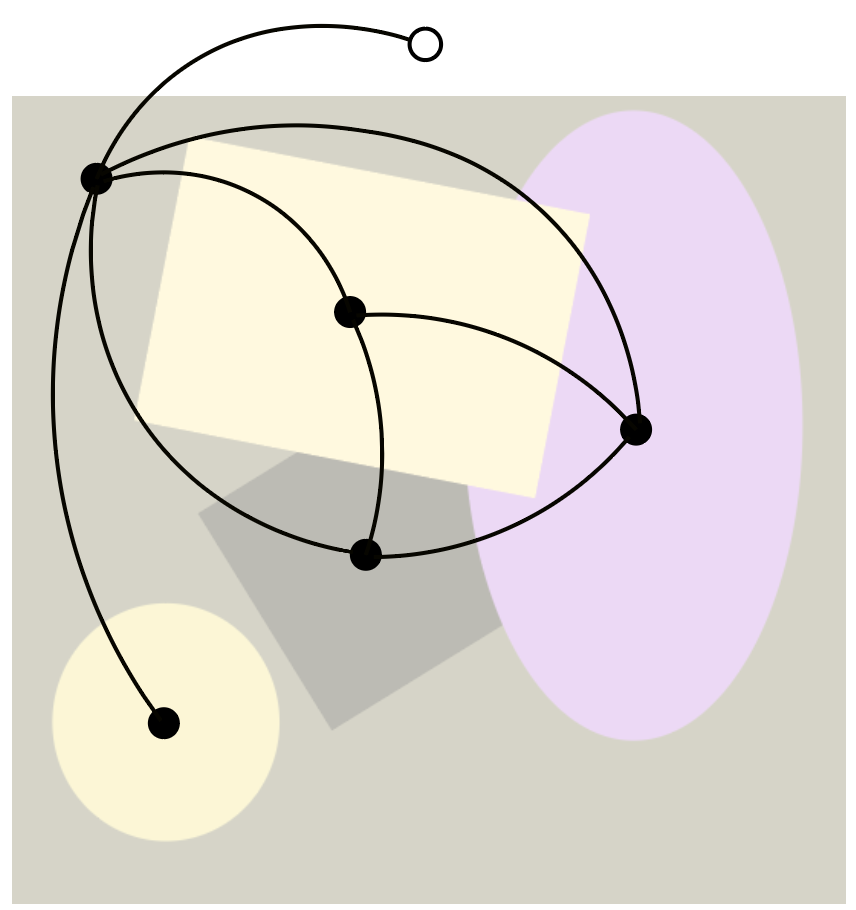}&
\includegraphics[width=0.20\textwidth]{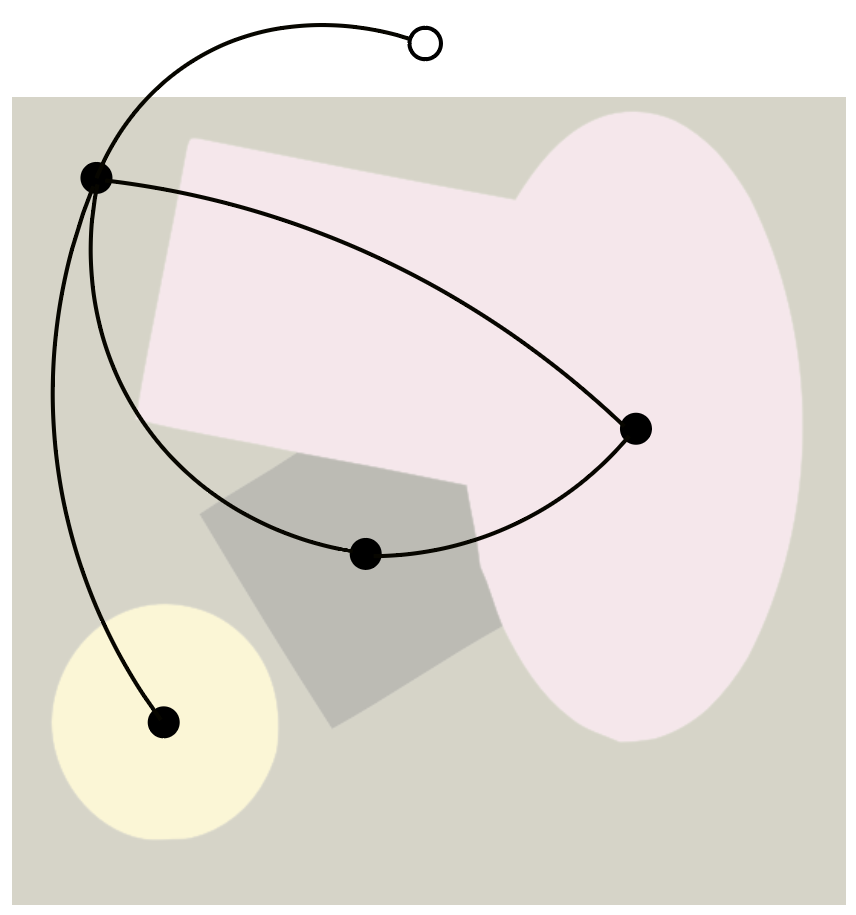}&
\includegraphics[width=0.20\textwidth]{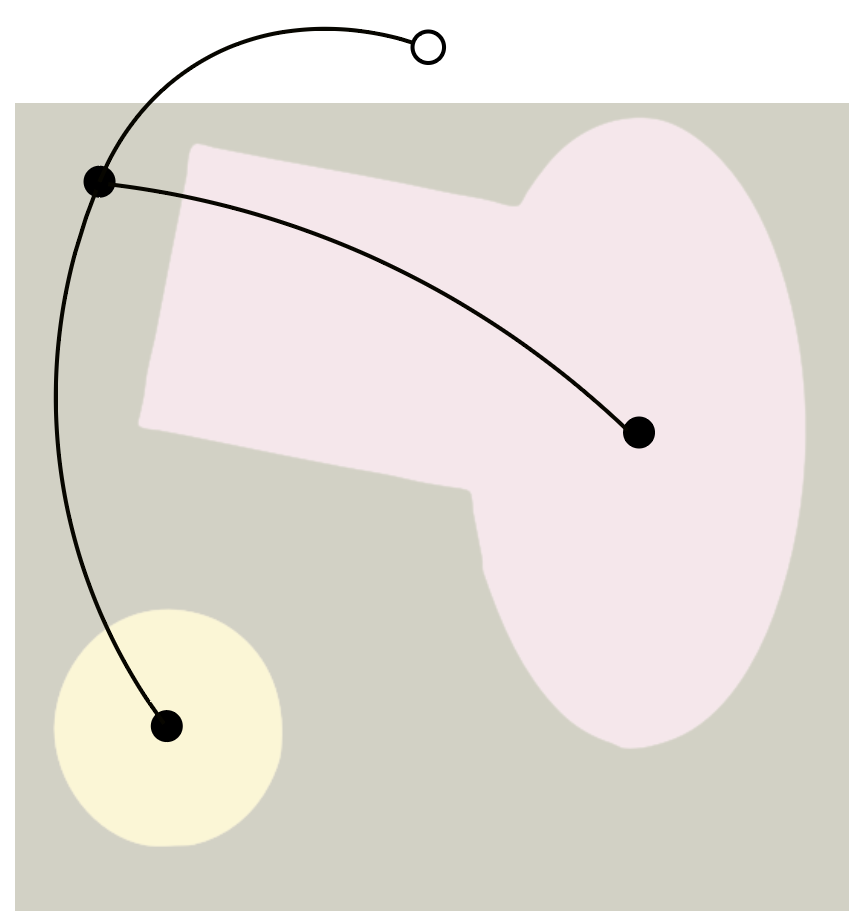}&
\includegraphics[width=0.2\textwidth]{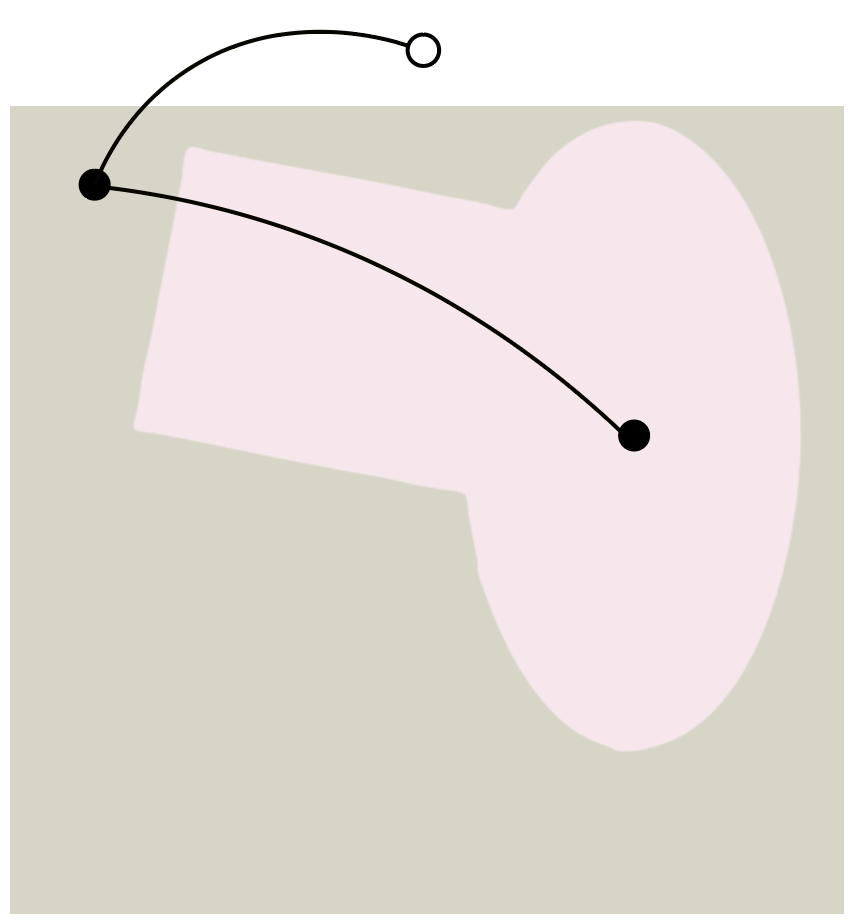}
\end{tabular}
\caption{The dual graph update.  From the given partition in (a), (b) illustrates when the  yellow rectangle and the purple ellipse  are merged.  One node from the dual graph is removed as well as the associated edges. After one primal step, in (c), the dark region is  merged with the background; and after another one, in (d), the yellow disk is merged with the background, yielding only two regions in the image domain. The additional element $O_0$ (the circle node) denotes the out-of-domain region.   
}\label{fig_dual_demo}
\end{figure}

\begin{figure}
\centering
\begin{tabular}{c@{\hspace{2pt}}c@{\hspace{2pt}}c@{\hspace{2pt}}c@{\hspace{2pt}}c}
(a)&(b)&(c)&(d)&(e)\\
\includegraphics[width=0.18\textwidth]{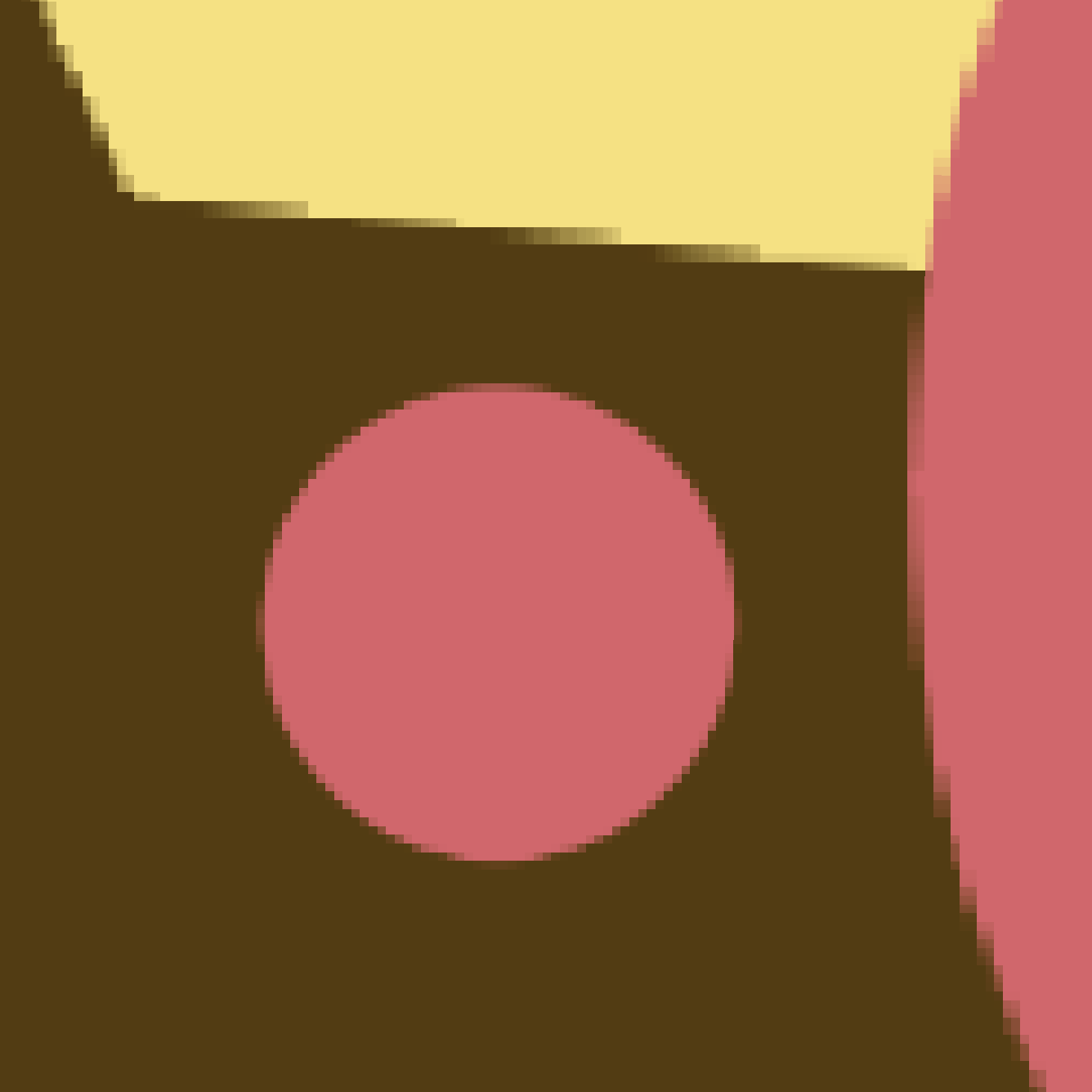}&
\includegraphics[width=0.18\textwidth]{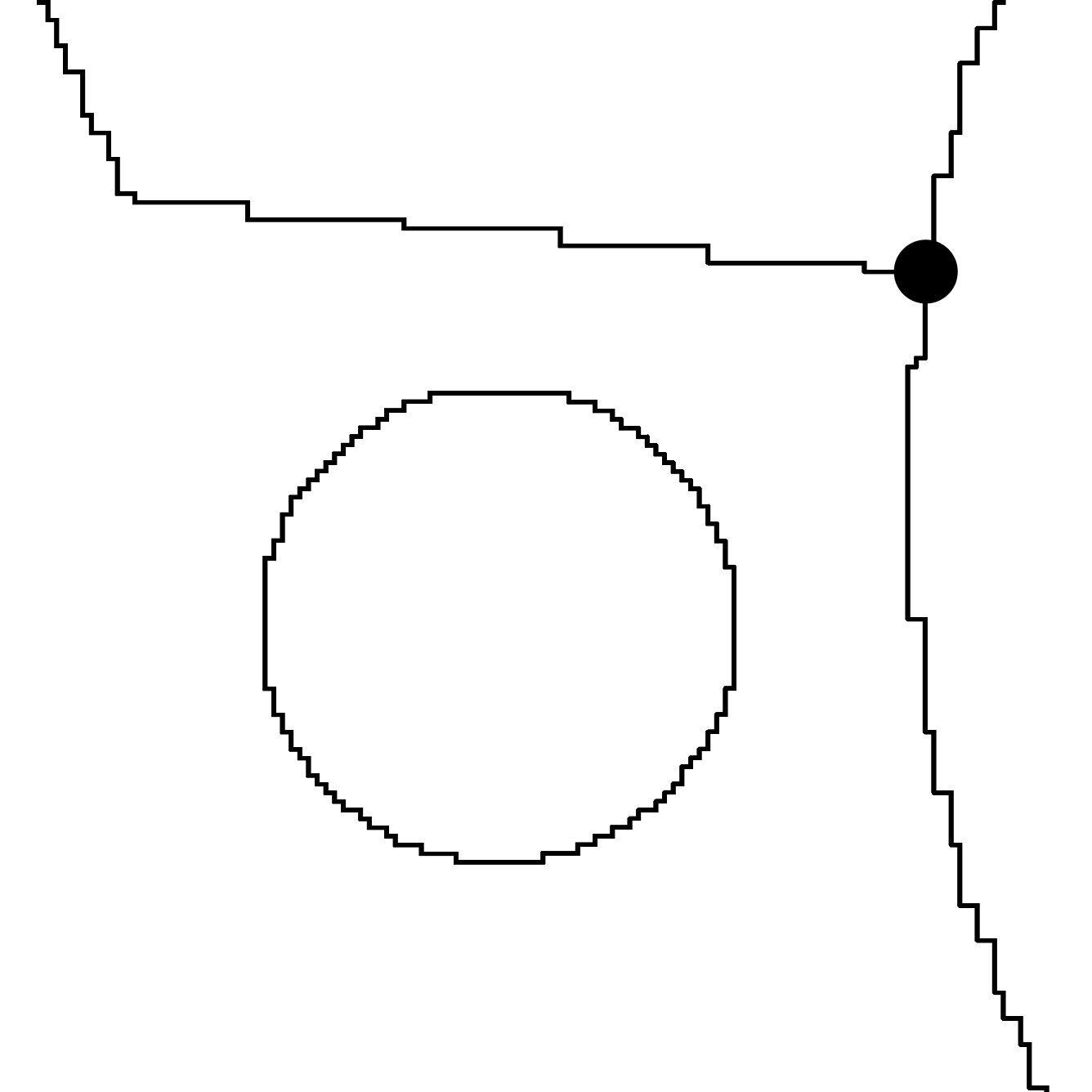}&
\includegraphics[width=0.18\textwidth]{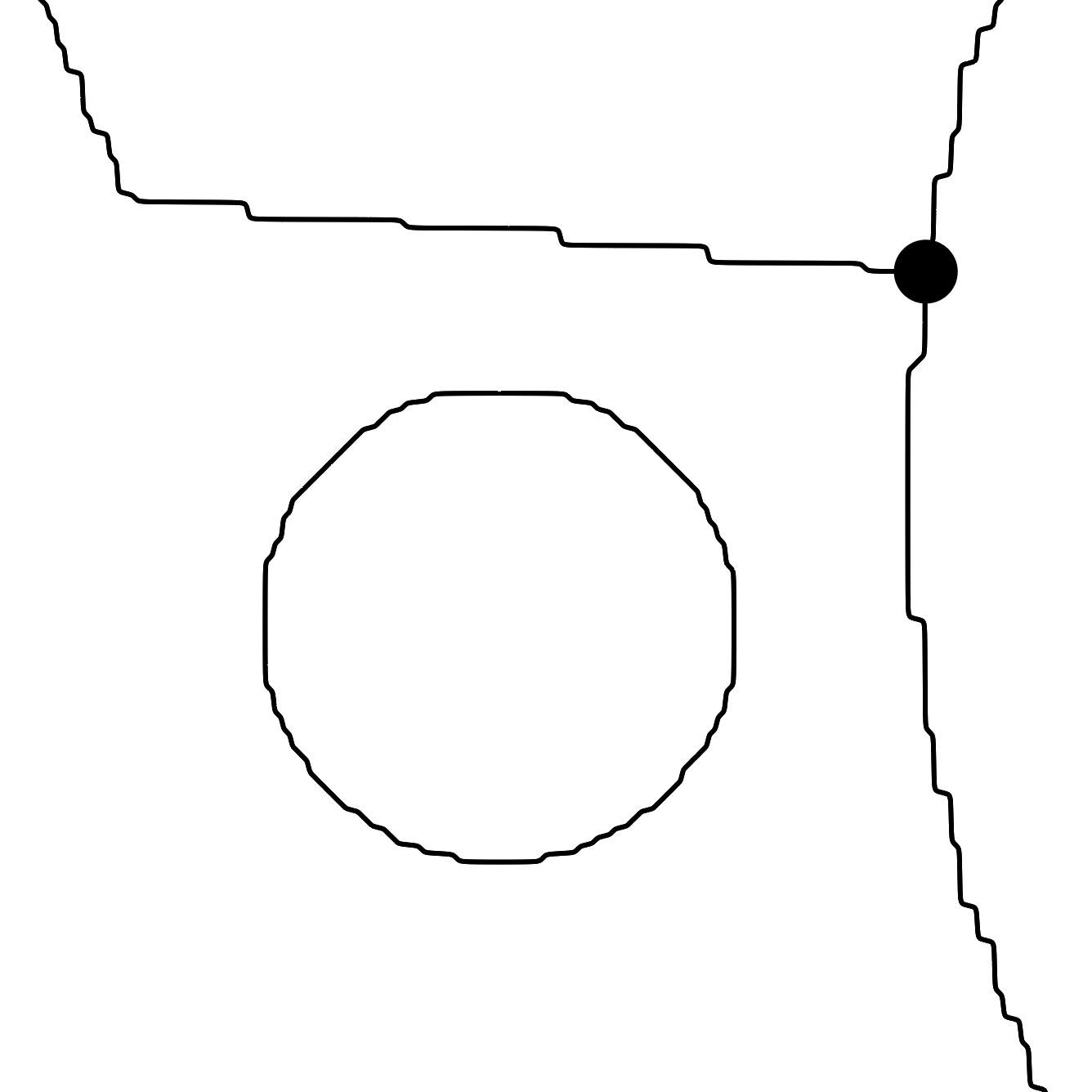}&
\includegraphics[width=0.18\textwidth]{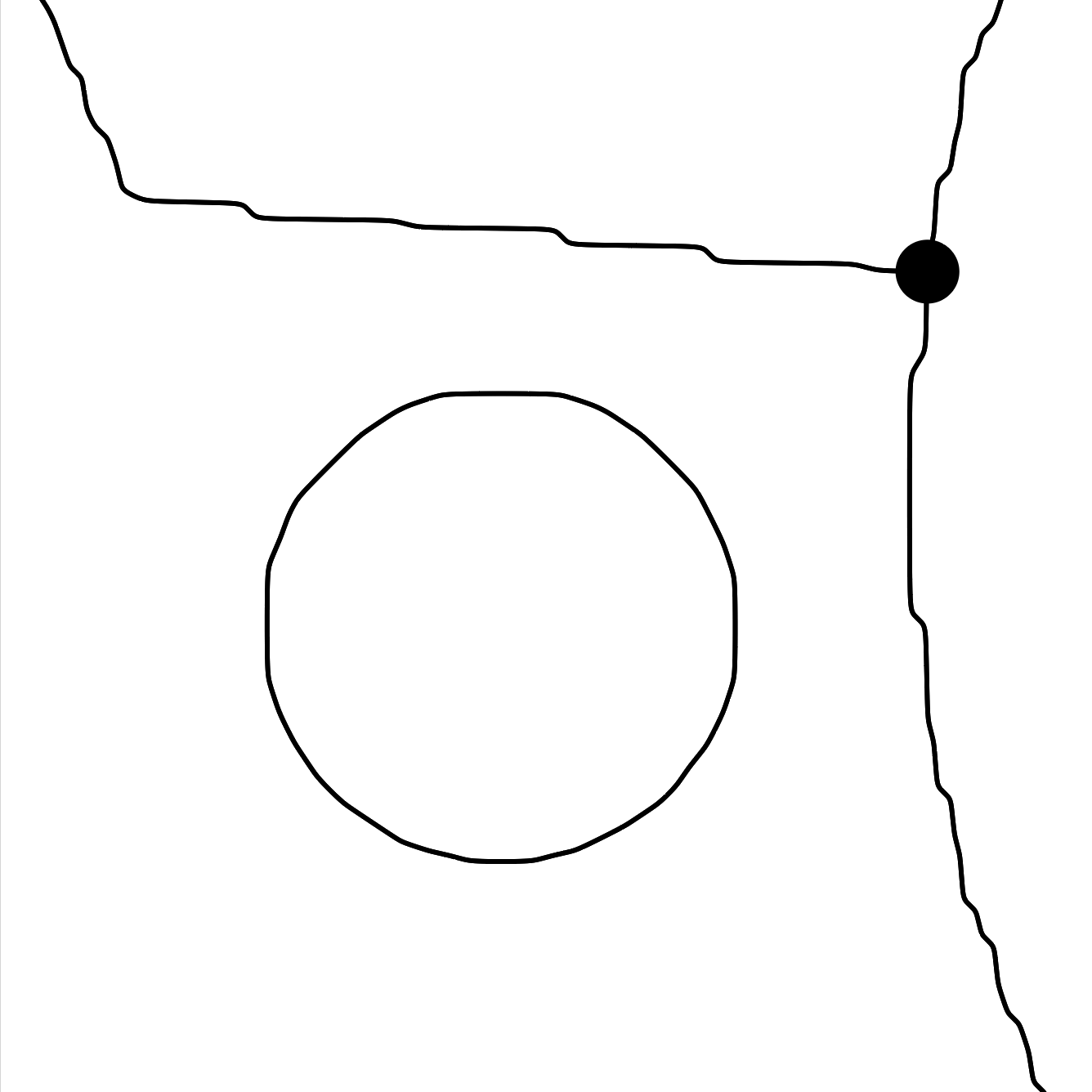}&
\includegraphics[width=0.18\textwidth]{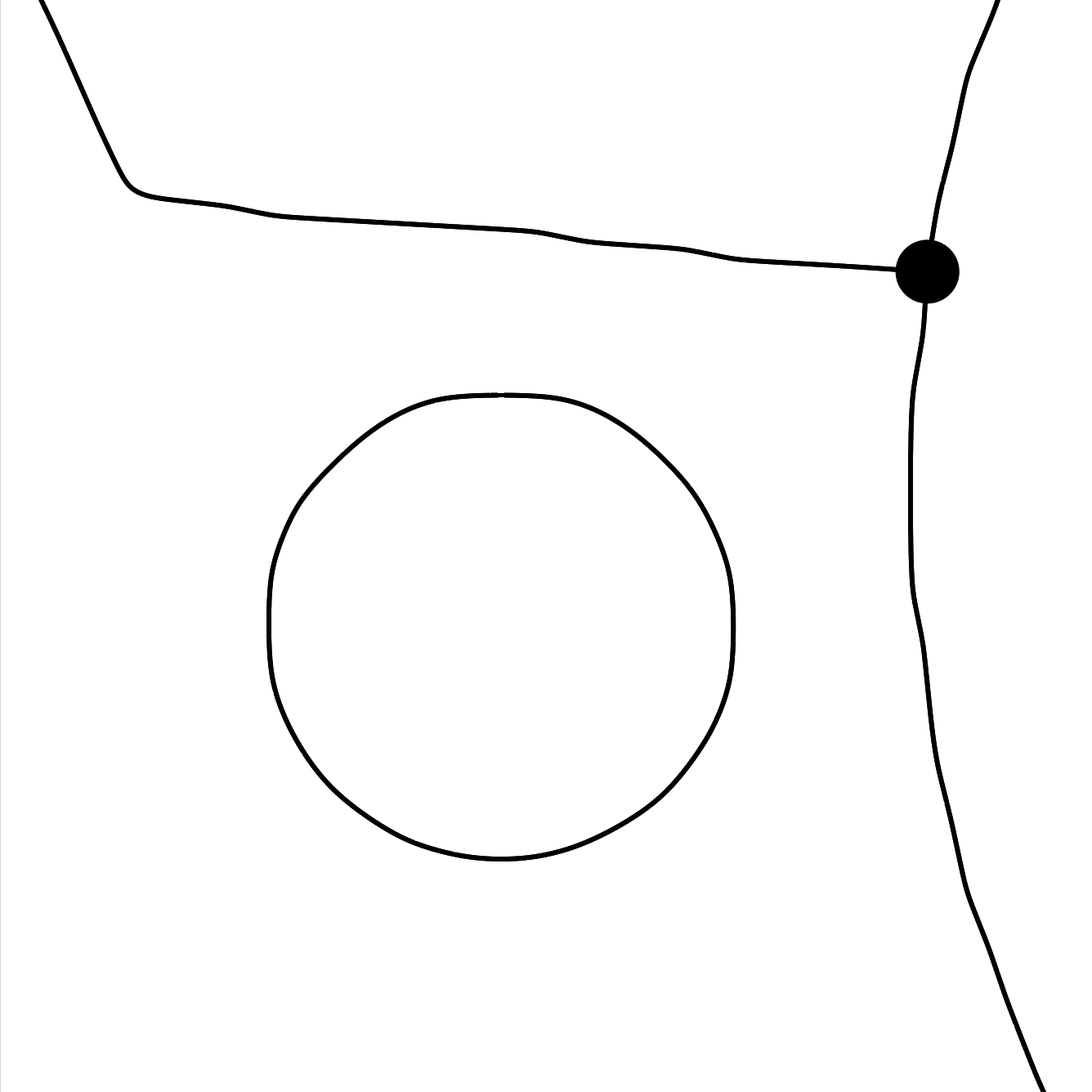}\\
\end{tabular}
\caption{The primal graph update. After dual steps, the raster image in (a) induces a primal graph with pixelated boundaries in (b), i.e., $T=0.0$.  The boundary curves are smoothed by 
the affine shortening flow~\eqref{eq_as_flow} 
with the evolution time (c) $T=0.2$, (d) $T=1.0$, and (e) $T=1.5$. The black dot specifies the junction point, which is fixed during the primal update. }\label{fig_primal_demo}
\end{figure}

With this setting, our proposed generic method alternates   the dual region merging step acting on the dual graph, and the primal curve smoothing  step acting on the primal graph. 
 The iteration starts from an input raster image, where pixels  viewed as square regions form  an initial partition $\mathcal{P}$. The topology and geometry of the image partition is updated by iterating the primal and dual steps:
\begin{itemize}
\item \textbf{Dual step}: 
Given the initial partition or the partition from the previous primal step,   adjacent regions are merged according to~\eqref{region_merging_step}. This step may change the topology of the partition. The details are discussed in Section~\ref{sec_merging}. 
\item\textbf{Primal step}: Given the  partition processed by the dual step,  the boundary curves evolve by the affine shortening flow~\eqref{eq_as_flow} with suitable boundary conditions.  This step reduces pixelation effects while preserving shape corners. The details are in Section~\ref{sec_smooth}.  

\end{itemize}
Finally,  boundary curves $\cC_m$ of the final  partition are approximated by parametric curves and  the color $c_i$ within each region $O_i$ is computed as the mean of the original image colors in $O_i$.  This leads to generating an SVG file enabling real-time rendering. 
  Figure \ref{fig_dual_demo} illustrates the region merging dual step. Starting from the dual graph in (a),  the yellow rectangle is merged  with the purple region in (b), causing the removal of one node and its incident edges.  In (c), the dark region is further merged with the background region. In (d), the yellow disk is merged with the background, leaving two regions in the image domain. The primal process is illustrated in Figure \ref{fig_primal_demo}. From the raster image in (a), we apply the dual update and obtain the primal graph in (b). It corresponds to the initial condition ($T=0.0$) for the affine shortening flow~\eqref{eq_as_flow}. In (c)-(e), we show the effect of the primal step when $T=0.2$, $T=1.0$, and $T=1.5$, respectively. The curve smoothing denoises each boundary curve while keeping the junction location, i.e., the black dot, to prevent topological changes.    More details are presented in Subsection \ref{sec_network_curves}.

In the following sections, we present the details of the major processes.  The dual step of region merging is presented in Section \ref{sec_merging}, and the primal step of curve smoothing in Section \ref{sec_smooth}. Section~\ref{sec_algorithm} contains more implementation details of the proposed algroithm.

\section{The Dual Step of Region Merging}\label{sec_merging}

In image segmentation, one of the most celebrated models is the Mumford-Shah functional \eqref{equ:MS}.  It enforces homogeneity within regions delineated by $\Gamma$ while minimizing the length of $\Gamma$ for denoising the boundary, which fits well with the vectorization we formulate in this paper.  However, the major computational difficulty comes from dealing with curves and regions simultaneously.   An alternative approach for pursuing the minimum of such functional is via region-merging~\cite{koepfler1994multiscale}. 

\textit{Region merging} is a technique used in image processing and computer graphics to form smooth image regions. It  works by  merging adjacent regions that share similar colors or textures. The iterated merging continues until a stopping criterion is met, which is commonly associated with a desired complexity level.  The region merging criteria reflect low-level visual processing privileging contrasted edges and  regions with uniform color~\cite{pavlidis1972segmentation,haralick1980edge,horowitz1976picture,beaulieu1989hierarchy}. 

The dual graph structure is the key to the efficiency of region merging. 
In traditional data clustering methods, including most color quantization algorithms, the process of conducting pairwise comparisons  requires $\mathcal{O}(N^2)$ computations for  $N$ data points. However, by encoding the adjacency relations, each region is only compared to its neighbors. This significantly reduces the number of comparisons to $\mathcal{O}(NR)$, where $R$ denotes the average number of neighbors. Even with an additional cost  of indexing the neighbors, 
the use of adjacency lists and advanced search algorithms~\cite{leiserson1994introduction} makes this indexing overhead negligible for most applications.

To determine the pairs of adjacent regions to be merged, several criteria inspired by visual perception can be formulated.
Brice and Fennema~\cite{brice1970scene} proposed the weakness heuristics for merging condition,
which depends on discernible jumps across the shared boundary.  
Their emphasis on the color jump between neighboring regions closely relates their model to the Kass-Witkin-Terzopoulos model (known as  \text{snake} model)~\cite{kass1988snakes}. The notion of edge strength is the object of numerous works such as~\cite{haralick1980edge}. 
Pavilidis~\cite{pavlidis1972segmentation} proposed the merging condition as a way
to minimize the number of partitioning regions while controlling the approximation error. Beaulieu and Goldberg~\cite{beaulieu1989hierarchy} cast the region merging paradigm as a sequential optimization process and deduced a simple merging condition: merge $O_1$ with $O_2$ if 
\begin{align}
\Delta E(O_1,O_2) \leq  \min(\min_{O_i\sim O_j}(\Delta E(O_i,O_j)), \lambda)\;.\label{merge_BG}
\end{align}
In the  formula above, 
\begin{equation}
\Delta E(O_1,O_2) = \text{Var}(f, O_1\cup O_2)-\text{Var}(f, O_1)-\text{Var}(f, O_2), \quad\text{Var}(f,O)=\int_{O}\|f(\bx)-\overline{f}(O)\|_2^2\,d\bx\;,\label{BG1}
\end{equation} 
with $\overline{f}(O)=|O|^{-1}\int_O f(\bx)\,d\bx$ for $O\subset\Omega$, and $\min_{O_i\sim O_j}$ means selecting  the minimal pair of adjacent regions. 
Notably, the condition~\eqref{merge_BG} is global and gradual. 

The above region merging condition can be derived for several examples of variational models as a discrete  gradient descent. Conversely, some merging conditions lead to schemes approximately minimizing certain functionals~\cite{morel2012variational,rose2010unifying}. 
For example, Koepfler, Lopez, and Morel~\cite{koepfler1994multiscale} considered the reduced piecewise constant MS model~\cite{mumford1989optimal}
\begin{equation}\label{equ:red_MS}
    \mathcal{E}_f(u)=\int_{\Omega\setminus \Gamma}\|u(\bx)-f(\bx)\|_2^2\,d\bx+\lambda \int_{\Gamma}\,d\sigma(\bx),\;
\end{equation}
where the approximating image is constrained to have a piecewise constant color.
They proposed an efficient region merging algorithm to attain a suboptimal segmentation for~\eqref{equ:red_MS} where $\Delta E(O_i,O_j)$ in~\eqref{merge_BG}
 is changed to
\begin{align}
\Delta E(O_1,O_2) = \frac{\text{Var}(f,O_1\cup O_2)- \text{Var}(f,O_1)- \text{Var}(f,O_2)}{\calH^1(\partial O_1\cap \partial O_2)}.\label{eq_energy_reduced}
\end{align}
Formula~\eqref{eq_energy_reduced} involves both local color comparison and geometric regularity, thus it improves on simpler thresholding methods~\cite{pavlidis1972segmentation,morel2012variational}. This algorithm enjoys compactness properties ensuring in particular the elimination of small regions (Section 3.2 in~\cite{koepfler1994multiscale}). The expression~\eqref{eq_energy_reduced} was similarly derived in the context of statistical image segmentation~\cite{zhu1996region,crisp2000fast,tao2003useful}. Other region merging methods may not directly involve  variational formulations, including watershed techniques~\cite{haris1998hybrid},  homograms~\cite{cheng2002color}, and local histograms~\cite{beveridge1989segmenting}.

\subsection{Merging costs and gain functions}
\label{ssec_mergingCriteria}

We propose a unifying formulation to explore various region merging criteria.  We use \eqref{merge_BG} as a generic merging condition, and refer to $\lambda>0$ as a \textit{threshold parameter}.
Given a partition $\mathcal{P}$ of an image $f$, we introduce a generic \textbf{merging cost} as 
\begin{align}
\Delta E(O_i,O_j) = \frac{ \text{Var}(f,O_i\cup O_j)- \text{Var}(f,O_i)- \text{Var}(f,O_j)}{\mathcal{G}(O_i,O_j)},\label{eq_Delta_E}
\end{align}
for every pair of adjacent regions $O_i\sim O_j$ in $\mathcal{P}$. We define the denominator  $\mathcal{G}(O_i,O_j)$ of~\eqref{eq_Delta_E} as the \textbf{gain functional}.  The numerator of~\eqref{eq_Delta_E} is the drop in the quadratic approximation error of the image by a piecewise constant image caused by merging $O_i$ and $O_j$.

Region merging segmentation is a step-wise optimization~\cite{beaulieu1989hierarchy}, and the choice of the merging cost progressively determines the model's behavior.  We consider various region merging criteria and study their effects for image vectorization. We start with two classical approaches: 
\begin{itemize}
\item\textbf{Beaulieu-Goldberg (BG) region merging}. This is the simplest criterion by taking a constant gain functional 
\begin{align}
\mathcal{G}_{\text{BG}}(O_i,O_j)=1\;.
\label{eq_BG_gain}
\end{align}
This leads to the Beaulieu-Goldberg region merging criterion in~\eqref{merge_BG}.
\item \textbf{The Mumford-Shah (MS) region merging.} Generalizing from~\eqref{eq_energy_reduced}, we define the (reduced) Mumford-Shah gain as the length of the shared boundary 
\begin{align}
\mathcal{G}_{\text{MS}}(O_i,O_j)=\calH^1(\partial O_i\cap\partial O_j)\;.\label{eq_MS_gain}
\end{align}
Two adjacent regions are eligible for merging if either they have very similar colors, or their shared boundary is long.  As discussed in~\cite{koepfler1994multiscale}, this gives a minimizing sequence of partitions for the reduced Mumford-Shah~\eqref{equ:red_MS}, and the threshold parameter $\lambda$ corresponds to the regularization parameter. 
\end{itemize}
We also propose  to compare the above classic merging criteria to the following  new ones, which are based on region scale and region area.  
\begin{itemize}
\item \textbf{Scale region merging.}  
The scale of a region $O$ is measured by a ratio between its area and perimeter.  This ratio is  closely related to the Cheeger constant~\cite{leonardi2015overview}, which is also associated with the scale-filtering property of  minimizers of Rudin-Osher-Fatemi (ROF) model~\cite{alter2005characterization}.  
This scale is used in \cite{sandberg2009unsupervised} for weighted length-minimization. It stabilizes the multiphase segmentation processes and helps find larger and more convex regions. Moreover, it is shown to allow more detailed boundaries~\cite{kang2013existence}.

We shall also test the  \textit{Scale region merging} using the following gain: 
\begin{align}
\mathcal{G}_{\text{scale}}(O_i,O_j)=\frac{\calH^1(\partial O_i)}{|O_i|}+\frac{\calH^1(\partial O_j)}{|O_j|}-\frac{\calH^1(\partial O_i)+\calH^1(\partial O_j)-2\calH^1(\partial O_i\cap \partial O_j)}{|O_i\cup O_j|} \;. \label{eq_scale_gain}
\end{align}
The gain function \eqref{eq_MS_gain} is established by comparing the lengths of the discontinuity set before and after the merging. Similarly,  Formula \eqref{eq_scale_gain} can be derived from comparing the scales of the adjacent regions with the scale of their union.
In~\eqref{eq_scale_gain}, a larger value of $\mathcal{G}_{\text{scale}}(O_i,O_j)$ indicates that $O_i\cup O_j$  has a larger scale than either $O_i$ or $O_j$ individually. See Remark~\ref{remark_scale_max} for more details.

\item\textbf{Area region merging.} We also propose the \textit{Area region merging} with a simple gain 
\begin{align}
\mathcal{G}_{\text{area}}(O_i,O_j)=\frac{\max(|O_i|,|O_j|)}{|O_i\cup O_j|}\;.
\label{eq_area_gain}
\end{align}
It computes the proportion of the larger region in a union of two adjacent regions. Notice that it is bounded between $1/2$ and $1$. When two adjacent regions have equal areas, the gain of merging them equals $1/2$; when one of the adjacent regions has a much larger area than the other, the gain is close to $1$. Area region merging thus prefers merging small regions with adjacent large regions, which is adequate to remove small boundary regions caused by antialiasing.
By Proposition~\ref{prop_var_equation} (below), the merging condition associated with~\eqref{eq_area_gain} can be written as
\begin{equation}\label{eq_area_mergingC}
    \min(|O_1|,|O_2|)\cdot\|\overline{f}(O_1)-\overline{f}(O_2)\|_2^2<\lambda \;,
\end{equation}
which generalizes the heuristic criterion in~\cite{he2023viva}. 

\end{itemize}

The following proposition is used to compute the explicit merging condition \eqref{eq_area_mergingC} for the area region merging \eqref{eq_area_gain}.
\begin{proposition}\label{prop_var_equation}For any pairs of distinct regions $O_i$ and $O_j$ of the partition for an image $f:\Omega\to\mathbb{R}^d$, we have
\begin{equation}
    \text{Var}(f,O_i\cup O_j)- \text{Var}(f,O_i)- \text{Var}(f,O_j) = \frac{|O_i|\cdot|O_j|}{|O_i|+|O_j|}\|\overline{f}(O_i)-\overline{f}(O_j)\|_2^2\label{eq_area_merge_var}
\end{equation}
\end{proposition}
\begin{proof}
See Appendix~\ref{proof_var_equation}.
\end{proof}

\begin{figure}
	\centering
	\begin{tabular}{c|c@{\hspace{2pt}}c@{\hspace{2pt}}c@{\hspace{2pt}}c}
		(a) Input ($N=8$) &(b) Area &(c) BG &(d) Scale & (e) MS\\\hline
		\multirow{4}{*}{\raisebox{-2.4cm}{\includegraphics[width=0.18\textwidth]{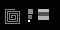}}}&\multicolumn{4}{c}{$N=7$}\\
		&\includegraphics[width=0.18\textwidth]{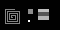}&
		\includegraphics[width=0.18\textwidth]{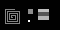}&
		\includegraphics[width=0.18\textwidth]{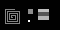}&
  \includegraphics[width=0.18\textwidth]{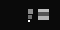}\\
  &\multicolumn{4}{c}{$N=5$}\\
		&\includegraphics[width=0.18\textwidth]{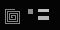}&
		\includegraphics[width=0.18\textwidth]{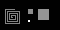}&
		\includegraphics[width=0.18\textwidth]{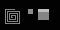}&
  \includegraphics[width=0.18\textwidth]{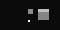}
	\end{tabular}
	\caption{Region merging comparison. (a) The input image contains $8$ regions. We merge regions until there are $N=7$ (first row), and $N=5$ (second row) using (b) Area region merging \eqref{eq_area_gain}, (c) BG region merging \eqref{eq_BG_gain}, (d) Scale region merging~\eqref{eq_scale_gain}, and (e) MS region merging  \eqref{eq_MS_gain}, respectively. Among these methods, Area tends to find larger regions regardless of their perimeters, BG follows a  merging order that is independent of shapes, Scale favors large and  convex region, and MS prefers bright shapes with shorter perimeters.  
 }\label{fig_illustrate1}
\end{figure}

We illustrate one of the key differences among these merging criteria in Figure \ref{fig_illustrate1}.  The image  in (a) shows  a spiral on the left, three squares in the middle, and one square with three horizontal strips on the right.  The spiral consists of 121 pixels with intensity $131$. In the central section, the squares vary in area from top to bottom: the first square has an area of 25 and an intensity value of 140, the second square has an area of 16 with an intensity of 110, and the third square has an area of 4 with an intensity value of 255. On the right side, the square contains three distinct strips: the upper strip has 3 pixels in height with an intensity of 190, the middle strip has a height of 4 pixels with an intensity of 85, and the lower strip also has a height of 4 pixels but with an intensity value of 180.   Figure \ref{fig_illustrate1} (b)-(e) show the results of merging using different criteria until only 7 regions (the first row) and 5 regions (the second row) remain.

In the first row, the MS region merging \eqref{eq_MS_gain} in (e)  privileges shapes with shorter boundaries, thus it eliminates the spiral immediately. Meanwhile, BG, Scale, and Area all decide to merge the dimmer middle square with the background. In the second row, different merging methods exhibit distinct behaviors. Notice that Area merges the bright small square and the middle strip in the right square with the background. BG merges the stripes in the right square to form a uniform region. Scale merges the bright square with the background and the middle strip in the right square with the bottom one. MS merges the central middle dim square with the background and the middle strip with the lower strip in the right square. 

These behaviors are consistent with their respective gains. Area prefers larger and more contrasted shapes, regardless of their perimeters, thus the spirals are preserved while two of the smaller central squares are removed. Since the strips have similar areas, their merging gains are close to $1/2$, which explains why the middle strip merges with the background instead. For BG, the merging decisions solely rely on intensity values, and the resulting images are as close as possible to the  input in average. Scale prefers forming large scale objects, and the spiral remains when $N=5$ due to its intensity; if we lower its intensity, the spiral will be merged with the background. Since the gain grows with the length of shared boundary, MS tends to oversimplify the image by removing elongated shapes, yet preserving bright shapes with relatively short perimeters. We  compare these merging methods with more complicated images in Section~\ref{sec_experiment}.

\vspace{0.2cm}
\begin{remark}
Several merging conditions can be derived from  general variational models of the following form
\begin{align}
\mathcal{E}(\mathcal{P})=\sum_{O\in\mathcal{P}\setminus \{O_0\}}\int_{O}\|f(\bx)-\overline{f}(O)\|_2^2\,d\bx+\lambda\mathcal{R}(\mathcal{P})=\sum_{O\in\mathcal{P}\setminus \{O_0\}} \text{Var}(f,O)+\lambda\mathcal{R}(\mathcal{P})\label{eq_energy}
\end{align} 
where $\mathcal{P}$ is a partition of the image domain  $\Omega$, $f$ is a given image,   $\mathcal{R}(\mathcal{P})$ is a regularization term inducing desired properties of the partition, and $\lambda>0$ is the regularization parameter. 
The simplest criterion of this kind is the Beaulieu-Goldberg criterion as stated in the merging condition \eqref{BG1}, where $\mathcal{R}(\mathcal{P})= \#\mathcal{P}$   simply is the number of regions.  

For minimizing the energy~\eqref{eq_energy}, a greedy approach is to merge pairs of adjacent regions such that there is a significant reduction in the energy after merging. Let $\mathcal{P}_{O_i\cup O_j}$ be the partition obtained after merging $O_i$ with $O_j$, then this merging is permissible if
\begin{align}
\mathcal{E}(\mathcal{P}_{O_i\cup O_j})-\mathcal{E}(\mathcal{P})=\text{Var}(f,O_i\cup O_j)- \text{Var}(f,O_i)- \text{Var}(f,O_j)+\lambda\left(\mathcal{R}(\mathcal{P}_{O_i\cup O_j})-\mathcal{R}(\mathcal{P})\right)\label{eq_func_crit1}
\end{align}
is negative. Since $\text{Var}(f,O_i\cup O_j)- \text{Var}(f,O_i)- \text{Var}(f,O_j)\geq 0$ for any pairs of $O_i$ and $O_j$ and $\lambda>0$ (See Proposition~\ref{prop_var_equation}),~\eqref{eq_func_crit1} is possible only if $\mathcal{R}(\mathcal{P})-\mathcal{R}(\mathcal{P}_{O_i\cup O_j})> 0$. 
In this case, criterion~\eqref{eq_func_crit1} can be written as
\begin{align}
\frac{\text{Var}(f,O_i\cup O_j)- \text{Var}(f,O_i)- \text{Var}(f,O_j) }{\mathcal{R}(\mathcal{P})-\mathcal{R}(\mathcal{P}_{O_i\cup O_j})}<\lambda\;.\label{eq_func_crit2}
\end{align}
Comparing~\eqref{eq_func_crit2} with~\eqref{merge_BG}, we  define the gain  associated with the energy~\eqref{eq_energy} by
$$\mathcal{G}_{\mathcal{E}}(O_i,O_j)=\mathcal{R}(\mathcal{P})-\mathcal{R}(\mathcal{P}_{O_i\cup O_j})\;.$$
It computes the change in the regularization functional before and after merging $O_i$ with $O_j$.   This strategy can be further generalized by considering variational models with different fidelity metrics, such as the mutual information~\cite{crisp2000fast}.

Note that Scale region merging~\eqref{eq_scale_gain} and Area region merging~\eqref{eq_area_gain} do not correspond to any energy functional to be minimized. A necessary condition for such a correspondence is the validity of the association law.  Given three pairwise adjacent regions, $O_1,O_2$ and $O_3$, merging $O_1$ with $O_2$ first, then with  $O_3$ should lead to an identical energy gain as merging $O_2$ with $O_3$ first, then with $O_1$. Neither the Scale region merging  nor the Area region merging   satisfy this condition.
\end{remark}

\begin{remark}\label{remark_scale_max}
 We note that the Scale region merging  gain~\eqref{eq_scale_gain} is different from scale maximization, which corresponds to the following gain function
\begin{align}
\mathcal{G}_{\text{scale-max}}(O_i,O_j)= \frac{|O_i\cup O_j|}{\calH^1(\partial O_i)+\calH^1(\partial O_j)-2\calH^1(\partial O_i\cap \partial O_j)} \;.\label{eq_scale_max}
\end{align}
The  gain functional in~\eqref{eq_scale_max} always prefers merging regions if their union has a large scale, and this is especially true when the scale of one of the adjacent regions is already large.  In contrast, the gain in~\eqref{eq_scale_gain} focuses on the changes. Since merging a small-scale region with a large-scale region may not contribute to a significant increment in the scale,~\eqref{eq_scale_gain} can lead to partitions with more balanced scales.   
\end{remark}

\subsection{Properties of the new region merging methods} 
\label{ssec_dual_analysis}

In this section, we establish analytical properties of Scale region merging \eqref{eq_scale_gain} and Area region merging \eqref{eq_area_gain}.  We find that while the gain functional is established for pairs of neighboring regions, it has an impact on the global properties of the partitions.
Generalizing from~\cite{koepfler1994multiscale},  we say that a partition $\mathcal{P}$ is  \textit{2-normal} with respect to a merging cost~\eqref{eq_Delta_E} if $\Delta E(O_i,O_j)>\lambda$ for any pairs of adjacent regions $O_i,O_j\in\mathcal{P}$. A partition that is 2-normal with respect to one merging cost may not be 2-normal regarding  a different one, e.g., see Figure~\ref{fig_illustrate1}.  For a fixed merging cost, it is possible to have multiple $2$-normal partitions.  In the following analysis, we assume that $f$ is not a constant image.

Firstly, we analyze the Scale region merging with the gain functional~\eqref{eq_scale_gain}.
\begin{proposition}\label{prop_scale_neighbor}[Minimal number of neighbors]
    For any region $O$ in a 2-normal partition for image $f:\Omega \to\mathbb{R}^d$ with respect to the Scale region merging, its number of neighbors $N(O)\in\mathbb{N}$ is bounded from below:
    \begin{equation*}N(O)>\frac{3\lambda|\partial O|}{|\Omega|\cdot|O|\omega_f^2}
    \end{equation*}
    Here $\omega^2_f = \sum_{i=1}^d\left(\sup(f_i)-\inf(f_i)\right)^2$ is the oscillation of $f$ over $\Omega$, and $f_i:\Omega \to\mathbb{R}$ denotes the $i$-th channel of the image.
\end{proposition}
\begin{proof}
See Appendix~\ref{proof_scale_neighbor}
\end{proof}
From this result, we deduce an upper bound on the number of regions.
\begin{theorem}[Maximal number of regions]\label{thm_scale_n_region}
	The total number of regions $\# \mathcal{P}$ of any 2-normal partition $\mathcal{P}$ with respect to the Scale region merging is uniformly bounded from above and we have
	\begin{align}
		\# \mathcal{P}<\frac{48|\Omega|^3\omega_f^4}{C^2\lambda^2}
	\end{align}
 where $C$ is an absolute constant.
\end{theorem}
\begin{proof}
See Appendix~\ref{proof_scale_n_region}
\end{proof}

Secondly, we study the Area region merging with gain functional~\eqref{eq_area_gain}. 

\begin{proposition}[Elimination of small regions]\label{lemma_small_region} For any region $O$ of a $2$-normal partition $\mathcal{P}$ with respect to the Area region merging, its area $|O|$ satisfies
	\begin{align}
		\frac{\lambda}{\omega_f^2}\leq |O|\;,\label{area_lwb}
	\end{align}
	where $\omega^2_f$ is defined in Proposition~\ref{prop_scale_neighbor}.
\end{proposition}
Proposition~\ref{lemma_small_region} is easily derived from the equivalent description~\eqref{eq_area_mergingC} of the Area region merging. It shows the role of $\lambda$ as a threshold parameter. For a given image $f$, greater values of $\lambda$ lead to 2-normal partitions consisting of larger regions. As expected, for a fixed choice of $\lambda$,  images with less color oscillation tend to have larger regions in the  segmentation results.

We also characterize the perimeters of regions in 2-normal partitions by applying the isoperimetric inequality in $\mathbb{R}^2$ together with Proposition~\ref{lemma_small_region}.

\begin{corollary}[Lower bound on perimeter]\label{cor_length} For any region $O$ of a $2$-normal partition $\mathcal{P}$ with respect to the Area region merging, its perimeter $|\partial O|$ satisfies
	\begin{align}
		2\sqrt{\frac{\pi\lambda}{\omega^2_f}}\leq |\partial O|\;.
	\end{align}
\end{corollary}
Moreover, since regions in $\mathcal{P}$ form a partition of the image domain $\Omega$, we immediately have
\begin{corollary}[Maximal number of regions] The total number of regions $\# \mathcal{P}$ of any 2-normal partition $\mathcal{P}$ with respect to the Area region merging is uniformly bounded from above and we have
	\begin{align}
		\# \mathcal{P}\leq \frac{|\Omega|\omega_f^2}{\lambda}\;.\label{n_region_ub}
	\end{align}
\end{corollary}
We discuss some interesting aspects of these results in the following remarks.
\begin{remark}
Thanks to the uniform boundedness of the number of regions and thus the number of edges, by the Arzel\'{a}-Ascoli theorem, we may extract a subsequence from a sequence of 2-normal segmentations that converges to some segmentation in the Hausdorff metric. However, the limit segmentation might have a larger energy than  the inf-limit of the energies of the minimizing sequences. Nevertheless, if we consider segmentations that are unions of pixels, then their number is finite and we can claim that the minimum exists in this class.    
\end{remark}
\begin{remark}
We make a comparison on the upper-bounds for the number of regions in 2-normal partitions using MS, Scale, and Area region merging. In~\cite{koepfler1994multiscale}, the upper-bound for $\#\mathcal{P}$ using MS is proved:
\begin{equation}
\#\mathcal{P}\leq \frac{|\Omega|\omega_f^4}{C'\lambda^2}\;,
\end{equation}
where $C'>0$ is an absolute constant.  For all merging methods,  increasing $\lambda$ or reducing the image contrasts yields fewer regions.  Specifically, for both MS and Scale region merging, since their respective upper-bound depends on the second order of $\lambda^{-1}$, increasing $\lambda$ tends to cause more regions to be merged compared to Area region merging. For smaller images, i.e., smaller $|\Omega|$, we expect that MS and Scale region merging behave similarly. 
\end{remark}

\section{The Curve Smoothing Primal Step}\label{sec_smooth}

Starting from an image partition $\mathcal{P}$, the primal step smooths each boundary curve $\cC_m$ to reduce the pixelation effect of the raster image.  We define an appropriate framework for evolving the discontinuity set of an image partition via curve evolution equations. Focusing on the affine shortening flow~\cite{sapiro1993affine}, we also present a sufficient condition for the maximal evolution time to preserve the topology of primal graph. 
Geometric PDEs for planar curve evolution have high relevance in shape analysis because of their intrinsic relations with subgroups of the planar  projective group~\cite{alvarez1993axioms,olver1993classification,tannenbaum1997invariant} such as the group of similarities or of affine maps.  This means that local geometric features of the evolved curves remain stable under  perspective changes, which is desirable in most image and shape processing tasks. For any  connected Lie group, it is possible to symmetrically derive the corresponding differential invariant flows by solving equations of prolongations of the infinitesimal generator~\cite{olver1993classification}. As shown in~\cite{huisken1990asymptotic,angenent1991formation}, general differential invariant flows can exhibit complex behaviors including developing singularities. However, imposing additional properties such as  a comparison principle and causality leads to second-order PDEs~\cite{alvarez1993axioms} whose solutions constitute scale-space frameworks for image analysis.
An example involving the planar Euclidean group generated by translations and rotations,  is the mean curvature flow~\cite{cao2003geometric}.
The existence and uniqueness of its solution were established by Gage and Hamilton~\cite{gage1986heat} and Grayson~\cite{grayson1987heat}. Some numerical techniques include  finite differences~\cite{catte1995morphological}, finite elements~\cite{dziuk1999discrete}, the level-set method~\cite{osher1988fronts}, and the Merriman-Bence-Osher (MBO) scheme~\cite{merriman1994motion}.

A flow that is symmetric under the special affine group, consisting of translation and transformations by the special linear group $\text{SL}(\mathbb{R}^2)$, is the affine shortening flow~\cite{sapiro1993affine,alvarez1993axioms}
\begin{align}
\frac{\partial \cC}{\partial t} = \kappa^{1/3}\mathbf{N}\;, \label{eq_as_flow}
\end{align}
where $\kappa$ is the curve's  curvature and $\mathbf{N}$  its normal direction (More details in Subsection~\ref{sub_sec_primal_step_detail}). 
The affine shortening flow~\eqref{eq_as_flow} enjoys many desirable properties \cite{angenent1998affine}. Firstly, it effectively eliminates small-scale irregularities while preserving significant curvature features such as corners. Secondly, the curve smoothing process commutes with special affine transformations, guaranteeing that the shape simplification is not affected by general perspective changes.

Some properties and existence theory were discussed and developed in a series of works~\cite{sapiro1993affine,angenent1998affine,alvarez1993axioms}. 
Angenent, Sapiro, and Tannenbaum~\cite{sapiro1993affine} showed that any closed planar curve converges to an elliptical point~\cite{angenent1998affine}, indicating preservation of critical shape features such as corners.
This is different from the asymptotic behavior of the mean curvature flow (which is similar to \eqref{eq_as_flow}, but without the cubic root). 
Alvarez and Morel~\cite{alvarez1997affine} conduct a comparison of corner displacement resulting from the curve shortening flow with that observed in the affine shortening flow. They conclude that the affine shortening flow is more effective for corner detection~\cite{alvarez2017corner}. Similar conclusions were made for image curvature microscopic analysis~\cite{ciomaga2017image} and image vectorization~\cite{he2022silhouette,he2023binary,he2023topology}. Numerically, the affine shortening flow~\eqref{eq_as_flow} can be efficiently addressed by Moisan's geometric scheme~\cite{moisan1998affine,lisani2003theory}, which amounts to finding the envelope of midpoints of chords that enclose small areas.
More general differential schemes invariant to planar projective actions are discussed in~\cite{faugeras1993evolution}.

\subsection{Curve evolution and preservation of topology}
\label{sec_network_curves}

The collection of curves $\Gamma$ and the evolution time $T$ of the affine shortening flow need to satisfy certain conditions for the primal step to be well-defined.  Evolving individual curves in parallel can cause topological changes to the domain partition, leading to inconsistent labels~\cite{he2023topology}. To avoid such issues, we define the geometric conditions satisfied by $\Gamma$ in the following, and we provide a theoretical upper bound on the maximal evolution time to ensure the topological invariance of the primal graph.

\begin{definition}[Network of curves]\label{def_network_curve} A network of curves in a bounded domain $\Omega\subset\mathbb{R}^2$ is a collection of finitely many rectifiable curves $\cC_m:[0,1]\to\overline{\Omega}$, $m=1,2,\dots,M$, satisfying the following conditions
\begin{itemize}
\item Each curve $\cC_m$, $m=1,2,\dots,M$, has no interior self-intersection, namely $\cC_m(s_1)\neq \cC_m(s_2)$ for any $s_1\neq s_2$ and $s_1, s_2\in(0,1)$.
\item Any two distinct curves $\cC_i$, $\cC_j$, $i\neq j$, $i,j=1,2,\dots,M$, do not intersect at their interior points, i.e., $\cC_i(s_1)\neq \cC_j(s_2)$ for any $s_1, s_2\in(0,1)$.
\item Endpoints are not open. For any $m=1,2,\dots,M$, each endpoint of $\cC_i$ is either (i) identical with the other endpoint of $\cC_i$; or (ii) identical with an endpoint of $\cC_j$ for some $j\neq i$; or (iii) belongs to  $\partial\Omega$. 
\item Image boundaries are excluded.  That is, $\cC_m\cap\partial\Omega$ can only have at most two points, for $m=1,2,\dots,M$.
\item All intersection points are junctions $\bP_k$. That is, if $\cC_i$ and $\cC_j$ intersect at $p\in\overline{\Omega}$ for some $i\neq j$, there exists $\cC_k$ distinct from $\cC_i$ and $\cC_j$ such that $p$ is an endpoint of $\cC_k$.
\end{itemize}
\end{definition}

By the following result, we can define the evolution of 
an image domain partition by the evolution of its network of boundary curves.
\begin{proposition}\label{prop_network_curve}
    There is a one-to-one correspondence between a network of rectifiable curves  and a Caccioppoli partition of the image domain. 
\end{proposition}
\begin{proof}

By the definition~\ref{def_network_curve} of network of curves, If $\Gamma=\{\cC_1,\cC_2,\dots,\cC_M\}$ is a network of curves, $\Omega\setminus \bigcup_{n=1}^M\cC_m$ consists of a finite set of connected open planar regions, and the union of their closures is $\overline{\Omega}$. Moreover, since curves in $\Gamma$ have no open ends, the induced open sets do not contain interior boundaries; with an additional out-of-image-domain region,   $\Gamma$ induces a partition of $\Omega$. Conversely, consider a Caccioppoli partition $\mathcal{P}=\{O_1,O_2,\dots,O_N\}$ of $\Omega$. For any $n=1,2,\dots,N$. It is proved in \cite{ambrosio2001connected} that each $\partial O_n$ can be split into a finite set of pieces of curves that are closed or separated by junction points. These curve segments give rise to a network of rectifiable curves. 
\end{proof}

Given an initial collection of curves $\Gamma=\{\cC_1,\dots,\cC_N\}$,  we induce a family of evolving curves 
\begin{equation}\label{eq_gammaT}
\Gamma_t=\{\cC_1(\cdot,t),\cdots,\cC_N(\cdot,t)\}\;,
\end{equation}
where $t\in[0,T]$ represents time in the affine curve evolution \eqref{eq_as_flow}, and $T >0 $ is the maximal evolution time. In this work, we fix all the junction points $\bP_k$ when evolving curves.  
For general curvature-driven evolution of networks of curves with movable junctions, the problem of short-time existence is more delicate and is mainly investigated in the case of a mean curvature flow with non-trivial tangential components~\cite{tortorelli2004motion,magni2013motion}.

\begin{figure}
\centering
\begin{tabular}{cccc}
(a)&(b)&(c)&(d)\\
\includegraphics[width=0.22\textwidth]{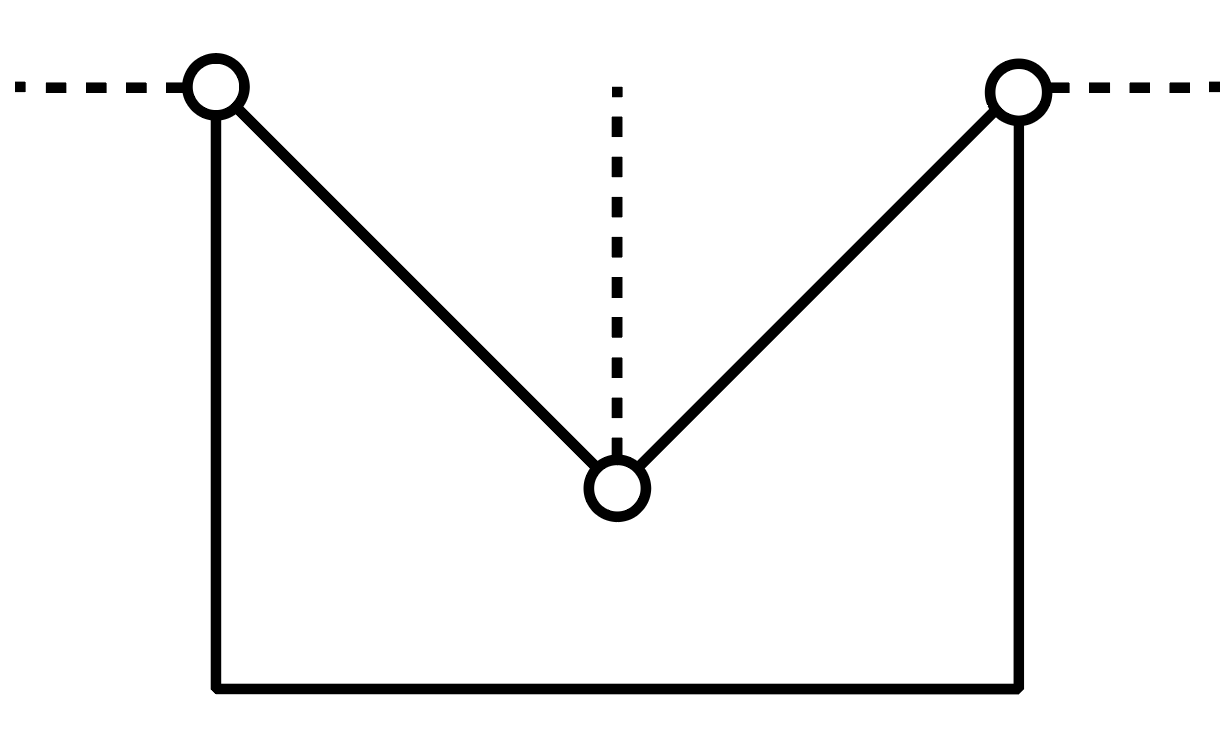}&
\includegraphics[width=0.22\textwidth]{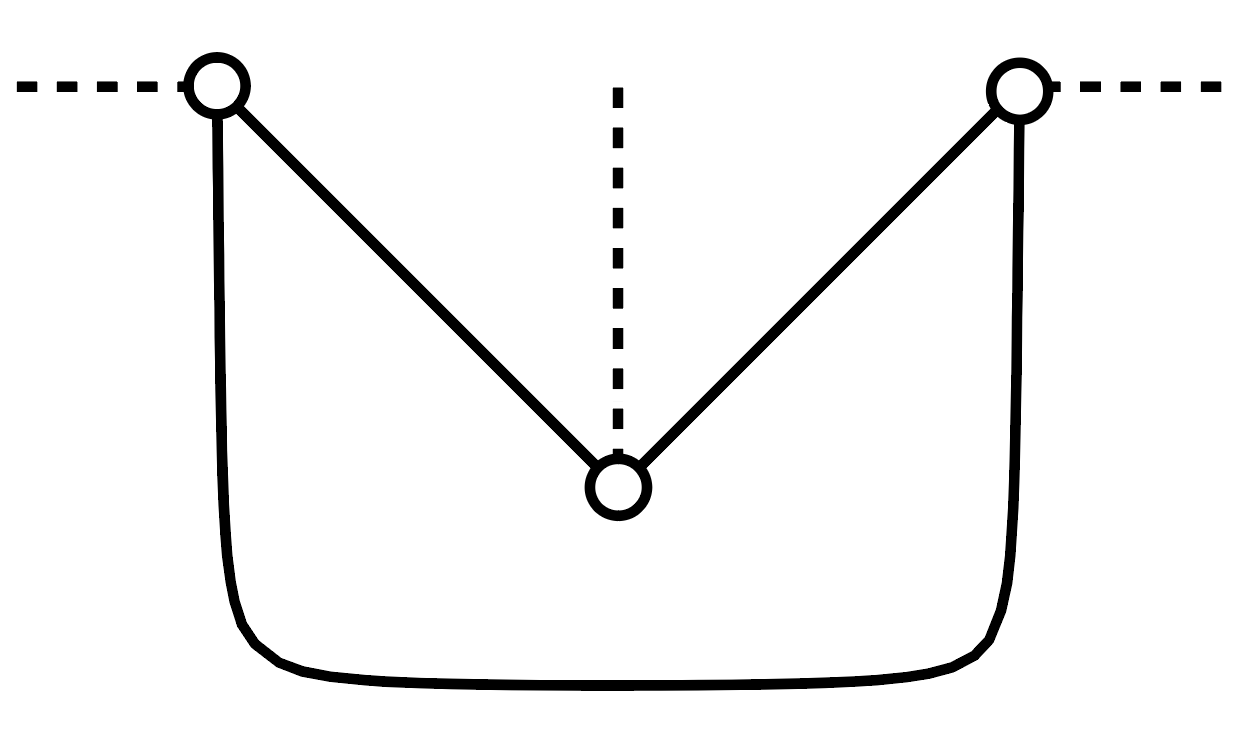}&
\includegraphics[width=0.22\textwidth]{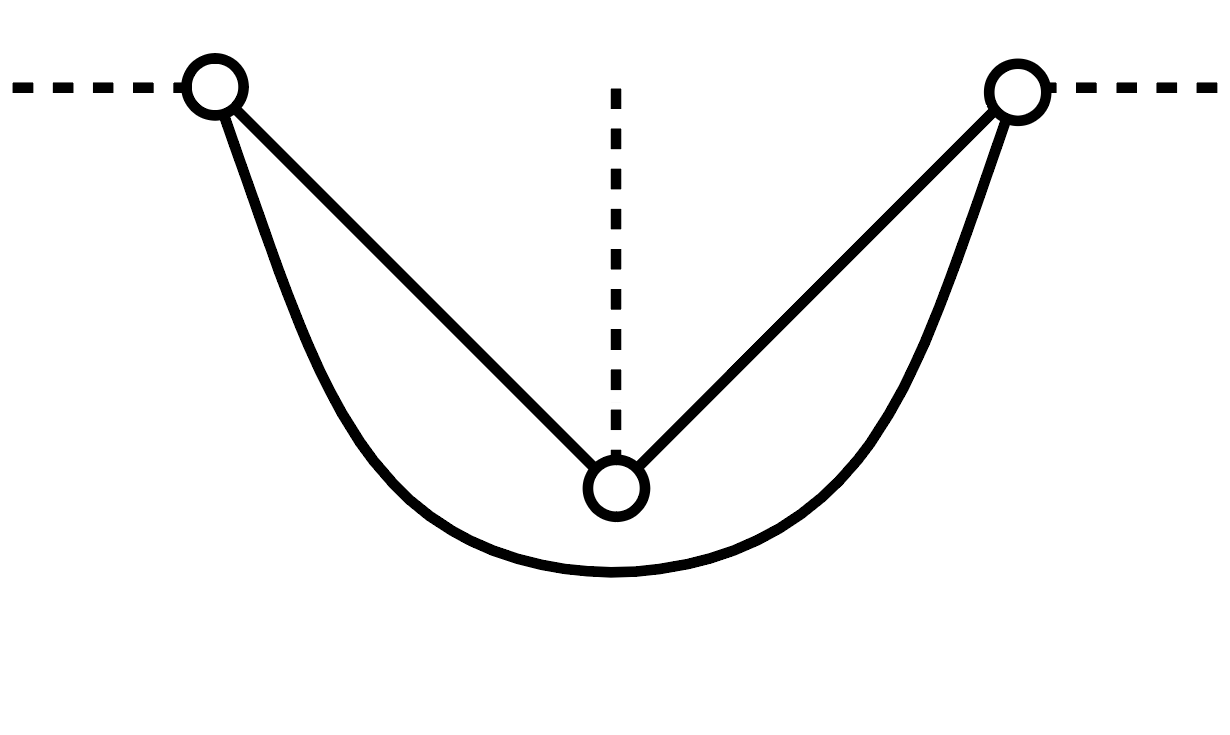}&
\includegraphics[width=0.22\textwidth]{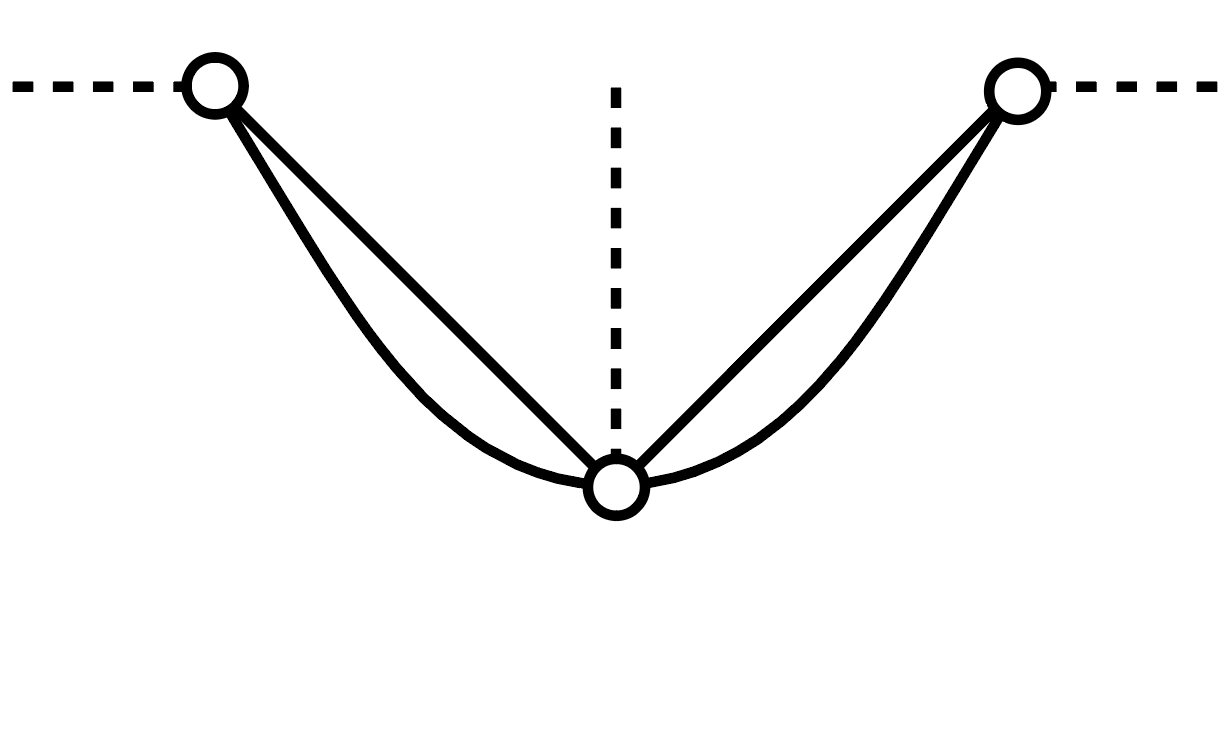}
\end{tabular}
\caption{Illustration of the necessity of controlling the evolution duration. (a) Initial configuration: two straight lines plus a polygonal curve, each of which ends with circular dots representing junction points that are fixed during affine shortening evolution. The dashed lines are the other curves joining at the junction points. Evolution at (b) $t=0.1$ (c) $t=1.0$ and (d) $t=1.45$, when the lower curve segment intersects with the middle junction point, and  the topology of $\Gamma$ changes. The upper-bound for the evolution time in Proposition~\ref{prop_max_time} guarantees that the evolution stops before (d) for general pixel images. }\label{fig_evolve}
\end{figure}

Although the evolution with fixed junctions is simplified, it requires a non-trivial analysis. If we take $T$ too large,
curves with fixed endpoints $\bP_1,\bP_2$ with $\bP_1\neq \bP_2$ will converge to a line segment connecting $\bP_1$ and $\bP_2$; and closed curves can shrink to an ellipse point~\cite{sapiro1993affine}. 
This leads to vanishing of regions or generation of new regions when curve interiors intersect with junctions. Figure~\ref{fig_evolve} illustrates the case.

To avoid such circumstances, the maximal evolution time should be controlled as follows. 

\begin{proposition}\label{prop_max_time} If the evolution time $T$ satisfies
 \begin{align}
 T < \frac{1}{(2\sqrt{2})^{1/3}}\approx 0.7071\;,\label{eq_time_bound2}
  \end{align}
  then the affine shortening flow does not alter the topology of $\Gamma$.
\end{proposition}

\begin{proof}
See Appendix~\ref{proof_max_time}.
\end{proof}

The upper-bound in~\eqref{eq_time_bound2} is pessimistic. Experimental evidence shows that $T\leq 1.0$  suffices to remove pixelation effects while preserving critical shape characteristics.

\subsection{Asynchronous effect in dual-primal iteration}
Thanks to the maximal evolution time derived in Proposition~\ref{prop_max_time}, the primal step does not alter the topology of the dual graph. Moreover, the dual step does not affect the smoothness of the primal graph. This allows  us to proceed the primal and dual step alternatively. Before extending this idea to develop our algorithm in Section~\ref{sec_algorithm}, we discuss an important interplay between the dual and primal step.

\begin{figure}
\centering
\begin{tabular}{ccc}
(a) & (b) & (c)\\
\includegraphics[width=0.30\textwidth]{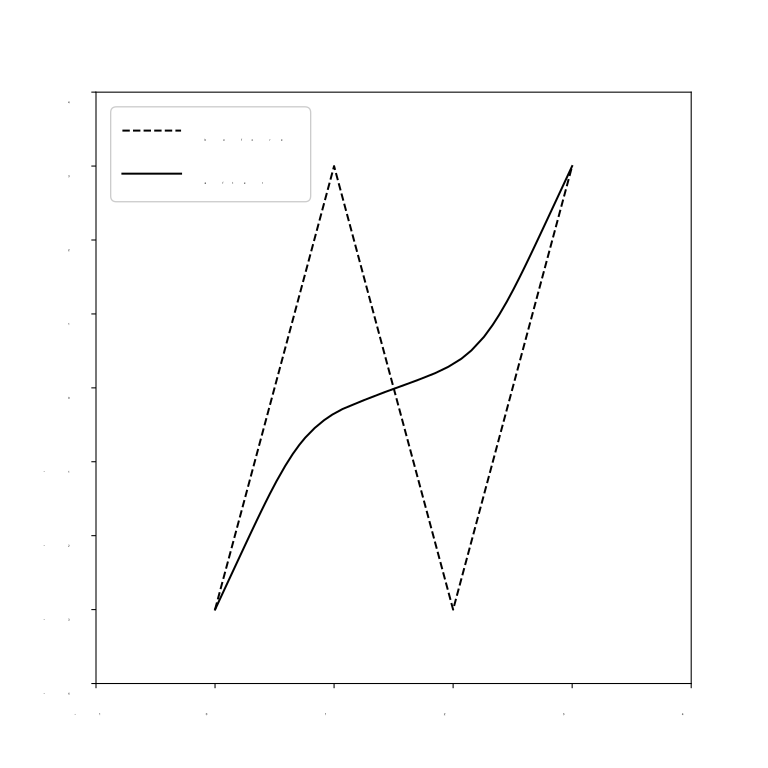}&
\includegraphics[width=0.30\textwidth]{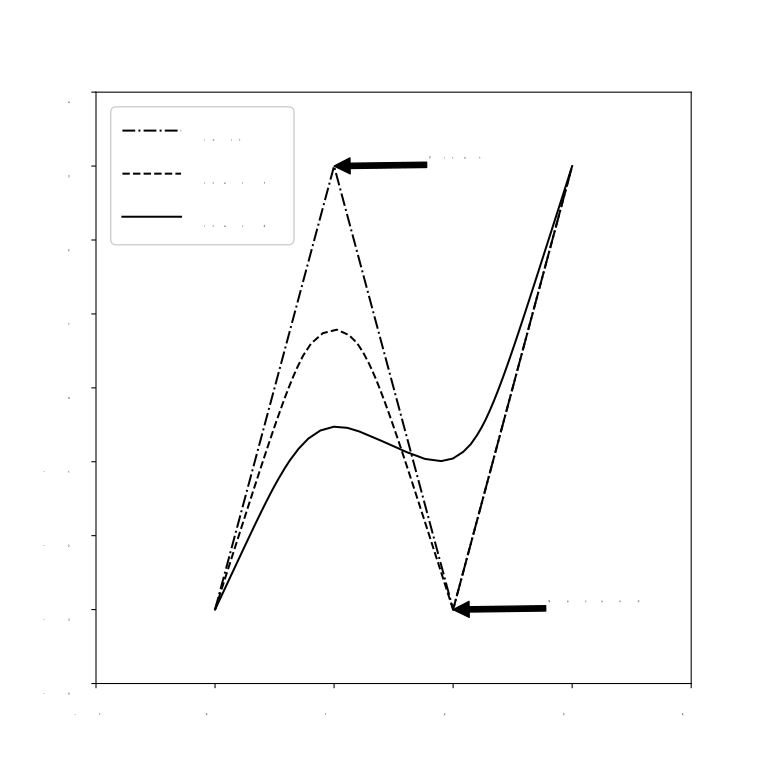}&
\includegraphics[width=0.30\textwidth]{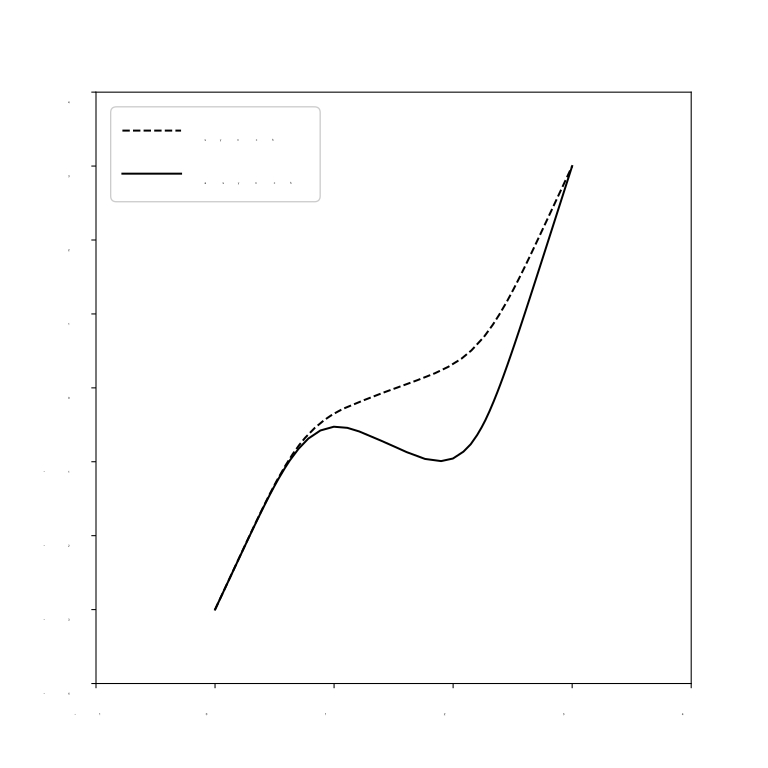}
\end{tabular}
\caption{Asynchronous effect of smoothing an open curve $\cC_m$ during the dual-primal iteration. (a) Assuming the two end points are junction points $\bP_1$ and $\bP_2$, affine smoothing \eqref{eq_as_flow} changes the dotted piecewise-linear curve to the solid curve.  (b) We assume that there are four junctions points,  $\bP_1$ and $\bP_2$ the same as in (a), and two additional junction points $\bP_3$ (labeled First) and  $\bP_4$ (labeled Second).  We experiment that during the first dual iteration, $\bP_3$ was released, then $\bP_4$ was released during the second dual iteration, then the solid curve in (c)  shows the final result.  The dotted curve in (c)  is to be compared with the dotted curve in (a), which shows the asynchronous effect of the dual-primal iteration. }\label{fig_iteration_effecs2}
\end{figure}

In Figure~\ref{fig_iteration_effecs2} (a), the initial curve is given as an open Z-shape curve whose endpoints are junctions, which we label as $\bP_1$ at  $(0,-3)$ and $\bP_2$ at  $(3,3)$.  
In (b), the shape of the initial curve is the same as in (a), but with additional junction points $\bP_3$ (labeled First) at $(1,3)$, and  $\bP_4$ (labeled Second) at $(2,-3)$.  For (a), we apply  the affine smoothing till $T=1.0$, while for (b), we apply the affine smoothing twice, each evolving with  $T=0.5$.  In addition, in (b) we assume that during the first dual iteration the junction point $\bP_3$ is removed, and during the second dual iteration the junction point $\bP_4$ is removed.   In (b), notice that the segment with nodes $(0,-3),(1,3)$ and $(2,-3)$ undergoes longer evolution by the affine shortening flow compared to the other part of the curve.  
The results in (c) show the comparison between the final curve in (a) (dotted curve), and the final curve from (b) (solid curve). In particular, both curves have the same $\bP_1$ and $\bP_2$ as primal graph nodes, but the dotted curve is smoother than the solid one.  

This shows the \textit{asynchronous effect} of the interplay between dual and primal step. It helps to keep details while smoothing the curves.  This approach aligns  well with the property of Area, BG, and Scale region merging methods which  preserve well regional details, as discussed in Subsection \ref{ssec_mergingCriteria}. 
We note that, for the closed curves containing no junction points, these two schemes with $T=1$ or $T=0.5$ twice lead to identical results. This is consistent with the semigroup property~\cite{cao2003geometric} of the affine shortening flow applied to a single closed curve.

\section{Proposed Algorithm and Implementation Details}\label{sec_algorithm}
In this section, we  describe our proposed algorithm for accurate image vectorization with more details.

\subsection{Dual step: Region merging on the dual graph}\label{sec_dual_step}

Consider a partition  $\mathcal{P}^{(0)}=\{O_0^{(0)},O^{(0)}_1,O^{(0)}_2,\dots,O^{(0)}_{N_0}\}$ with $N_0\geq 1$. It can be the initial partition whose elements correspond to single pixels or a partition obtained from a previous primal step.  Our strategy is to construct a sequence of at most $N_0+1$  partitions $\mathcal{P}^{(0)},\mathcal{P}^{(1)},\mathcal{P}^{(2)},\dots$ by iteratively merging adjacent regions.

For $k=0,1,\dots$, let $\mathcal{P}^{(k)}$ be the partition after the $k$-th iteration. Fix a threshold parameter $\lambda>0$. For each pair of adjacent regions, say $O_i^{(k)}, O_j^{(k)}$ in $\mathcal{P}^{(k)}$, $i< j$, $i,j\in\{1,2,\dots, N_k\}$, we conduct a region merging 
as discussed in Section~\ref{sec_merging}. We modify $\mathcal{P}^{(k)}$ by merging $O_i^{(k)}$ with $O_j^{(k)}$ if their merging cost is below $\lambda$ and also the lowest among all adjacent pairs, i.e., 
\begin{equation}
   \Delta E(O_i^{(k)},O_j^{(k)})\leq \min(\min_{O_s\sim O_l} (\Delta E(O_s^{(k)},O_l^{(k)}), \lambda)\label{eq_merging_condition}
\end{equation} 
 The merging operation means removing $O_i^{(k)}$ and $O_j^{(k)}$ from the list $\mathcal{P}^{(k)}$ and inserting the interior of $\overline{O}_i^{(k)}\cup \overline{O}_i^{(k)}$, denoted by $O_{ij}^{(k)}$ as the new region. By this, the new list of regions $$\mathcal{P}^{(k+1)} = \{O_0^{(k)},O^{(k)}_1,\dots,O^{(k)}_{i-1},O_{ij}^{(k)},O_{i+1}^{(k)},\dots,O_{j-1}^{(k)},O_{j+1}^{(k)},\dots,O_{N_k}^{(k)}\}$$ still defines a partition for $\Omega$.

 Besides the obvious upper bound, $N_0$, for the number of merging operations, the value of $\lambda$ also determines a natural terminating condition. The merging process stops whenever the cost of merging every pair of  adjacent regions exceeds $\lambda$, i.e., the corresponding partition is $2$-normal (See Section~\ref{ssec_dual_analysis}).  Hence, $\lambda$ can be treated as a parameter that controls the number of regions in the final partition. 
This relation between $\lambda$ and the number of regions is implicit and hard to handle. In Section~\ref{sec_terminating}, we  introduce a strategy for a convenient control of the partition complexity.

\subsection{Primal Step: Curve smoothing on the primal graph}\label{sub_sec_primal_step_detail}

The evolutionary family of curves $\Gamma_t$ in \eqref{eq_gammaT} is defined by evolving each curve $\cC_m:[0,1]\times [0,T)\to\overline{\Omega}$ by the affine shortening flow 
\begin{align}
&\frac{\partial \cC_m}{\partial t}(s,t)=\kappa_m(s,t)^{1/3}\mathbf{N}_m(s,t)\;,~0<s<L_m(t), \; 0<t<T\label{eq_flow}\\
&\cC_m(s,0) = \cC_m(s)\;, 0\leq s \leq L_m(0)\label{eq_init}
\end{align}
for $m=1,\dots,M$.  Here $L_m(t)$ is the arc-length, $\kappa_m(s,t)=\|\mathbf{k}_m(s,t)\|$ is the curvature, and $\mathbf{N}_m(s,t)=\mathbf{k}_m(s,t)/\|\mathbf{k}_m(s,t)\|$ with $\mathbf{k}_m(s,t) = \frac{\partial^2 \mathcal{C}_m(s,t)}{\partial s^2}$ is the normal direction of the curve at $\cC_m(s,t)$ pointing towards the center of curvature. 
Depending on the locations of endpoints, $\cC_m(0)$ and $\cC_m(L_m(0))$, we have the following two cases.
 
 \begin{itemize}
 \item Case I: If $\cC_m(0)=\cC_m(L_m(0))$ is not a junction point nor on $\partial\Omega$, then we enforce a periodic boundary condition:
\begin{align}
\cC_m(0,t)=\cC_m(L_m(t),t)\;,\quad t\in [0,T]\;, \label{eq_per_bound}
\end{align}
so that closed curves remain closed. By~\cite{angenent1998affine}, any smooth embedding of unit circle $\mathbb{S}^1$ in $\mathbb{R}^2$ remains embedded by the affine shortening flow. Hence, the longtime existence of the solution is guaranteed.
\item Case II: If $\cC_m(0)\neq \cC_m(L_m(0))$ or $\cC_m(0)= \cC_m(L_m(0))=\bP_k$ for $\bP_k$ being a junction point or on $\partial\Omega$, we use a Dirichlet boundary condition:
\begin{align}
\cC_m(0,t)=\cC_m(0,0)~\text{and}~\cC_m(L_m(t),t)=\cC_m(L_m(0),0)\;,\quad t\in [0,T]\;, \label{eq_Dir_bound}
\end{align}
so that if a curve intersects with other curves or with the boundary of the image domain, the intersections remain fixed during the evolution.  By the arc-length evolution formula proved in Lemma 4 of~\cite{sapiro1993affine}, the distance comparison principle (Theorem 2.1 in~\cite{huisken1998distance}) established for the mean curvature flow 
can be generalized to the affine shortening flow applying on smoothly embedded intervals, which can lead to the desired long-time existence and regularity of the solution. 
 \end{itemize}
Both scenarios can be consistently and efficiently handled by Moisan's  scheme~\cite{moisan1998affine} with expected behaviors. 
We thus focus on the numerical aspects, and refer interested readers to~\cite{huisken1990asymptotic,huisken1998distance,sapiro1993affine,angenent1998affine} for more theoretical discussions related to curve shortening flow.

\subsection{Algorithm and model parameters}\label{sec_algo_code}

\begin{algorithm}[t!]
	\caption{Proposed Algorithm for Accurate Image Vectorization}\label{alg_pseudo}
 \begin{algorithmic}
        \REQUIRE{Input image $f:\Omega\to\mathbb{R}^d$ ($d=1$ or $3$) with $N$ pixels ($N\geq 2$); a cost of merging associated with some gain functional $\Delta E(\cdot,\cdot)$;  upper-bound on the number of regions $N^*$ with $1\leq N^*< N$; smoothness parameter $T^*>0$;  approximation threshold $\tau>0$; and number of iterations for alternating the dual and primal step $I_{\max}$ with $1\leq I_{\max} \leq N-N^*$.}

  \STATE Set $\mathcal{P}^{(0)}$ as the initial partition where each region is a single pixel;

    \STATE{Set  $J=\lceil(N-N^*)/I_{\max}\rceil$ and $\lambda^* = 0$;}
    
    \FOR{$k=1,\dots,I_{\max}$}
            
            \STATE{Create a priority queue $Q_0^k$ storing all pairs of adjacent regions in $\mathcal{P}^{(k-1)}$ sorted by the respective costs of merging in an increasing order;}

            Set $\mathcal{T}^k_0= \mathcal{P}^{(k-1)}$;
            
            \FOR{$j=1,\dots,J$} 
            
                \STATE{(\textit{Dual Step}) Take 
            $$(O,O',\Delta E(O,O'))=\texttt{Pop}(Q_{j-1}^k)$$
            Here $\texttt{Pop}$ outputs a pair of adjacent regions giving the lowest cost $\Delta E(O,O')$ and removes this triplet from the queue;}

            \STATE{Set $\lambda^* \gets \max\{\Delta E(O,O'),\lambda^*\}$;}

               \STATE{Remove $O,O'$ from $\mathcal{T}_{j-1}^k$, insert the interior of $\overline{O}\cup \overline{O'}$ into $\mathcal{T}_{j-1}^k$, and take the resulting partition as $\mathcal{T}_{j}^k$;}
            \ENDFOR

            \STATE{Take $\mathcal{P}^{(k-1/2)}=\mathcal{T}^{k}_{J}$;}

             \STATE{(\textit{Primal Step}) Evolve $\mathcal{P}^{(k-1/2)}$ by the affine shortening flow~\eqref{eq_as_flow} for a duration of $T^*/I_{\max}$. Set the resulting partition as $\mathcal{P}^{(k)}$;}

        \ENDFOR

        \STATE{(\textit{Refine Step}) Apply the dual step until all adjacent regions have costs of merging greater than $\lambda^*$;}

        \STATE{Apply the primal step with $\Delta t =0.01$ and obtain a partition denoted by $\mathcal{P}^*$;}

        \STATE{Approximate the boundary curves of $\mathcal{P}^*$ by  piece-wise B\'{e}zier curves with error below $\tau$;}
            
            \STATE{Fill each connected component with the mean image color;}
            \STATE Produce the corresponding SVG file\footnotemark;
		\RETURN{A vector graphic approximating the raster input $f$.}
\end{algorithmic}
\end{algorithm}
\footnotetext{Details on SVG formats can be found in \href{https://developer.mozilla.org/en-US/docs/Web/SVG/Tutorial}{https://developer.mozilla.org/en-US/docs/Web/SVG/Tutorial}. There are also many libraries and packages that facilitate the creation and editing of SVM files, e.g., \href{https://github.com/sammycage/lunasvg}{\texttt{lunasvg}} for \texttt{C++} and \href{https://pypi.org/project/svglib/}{\texttt{svglib}} for \texttt{Python}.}

Algorithm~\ref{alg_pseudo} presents the pseudo-code of our proposed scheme. Given a raster image $f:\Omega\to\mathbb{R}^d$ ($d=1$ or $3$) with $N$ pixels and a merging condition associated with a chosen gain functional, our method requires two model parameters: 
\begin{itemize}
\item Region number ($1\leq N^*< N$): This parameter specifies an approximation for the number of regions in the output. The actual number can be different from $N^*$. See the discussions in Section~\ref{sec_terminating}. 
\item Boundary smoothing time ($T^*>0$): This parameter specifies the desired level of smoothness of the boundary curves of  the resulting partition.
\end{itemize}
and two algorithmic parameters:
\begin{itemize}
\item The error threshold for approximation  ($\tau> 0$): The maximal Hausdorff distance allowable for the piecewise B\'{e}zier curves approximating the smoothed segmentation boundaries.
\item The number ($I_{\max}\geq 1$) of alternations of the dual and primal step.
\end{itemize}

In each dual step, we merge $\lceil(N^*-N_0)/I_{\max}\rceil$ pairs of regions. We use the priority queue~\cite{ronngren1997comparative} to store pairs of adjacent regions with their associated costs of merging. The pair with the lowest cost can be retrieved efficiently by the \texttt{Pop} operation with complexity $\cO(\log n)$, where $n$ denotes the size of the queue.  
In each primal step, we use the output from the dual step as the initial condition and evolve it according to the affine shortening flow~\eqref{eq_as_flow} for a short time $T^*/I_{\max}$ by the Moisan's scheme~\cite{moisan1998affine}. The number of merged pairs and the evolution time for each dual and primal step respectively are determined such that the total number of merged pairs is around $(N^*-N_0)$ and the total evolution time equals $T^*$.  

After $I_{\max}$ times of alternating the dual and primal steps, we employ the post-processing described in Section~\ref{sec_terminating}.  We keep merging adjacent regions in the partition until all the resulting pairs of adjacent regions have costs of merging greater than $\lambda^*$. To suppress pixelation effects from the additional mergings, we smooth the boundary curves by the affine shortening flow with a small duration $\Delta t$. We fix $\Delta t = 0.01$ in this work.

\subsection{The terminating condition and the final number of regions}\label{sec_terminating}

As discussed in Section~\ref{sec_dual_step}, the value of $\lambda$ implicitly controls the number of regions in the final partition. The maximal number of mergings $K_{\max}$ can be implicitly specified via $\lambda$. For a fixed $\lambda$, the merging  stops if for some $k_\lambda\in\mathbb{Z}$, $\Delta E_{i,j}^{k_\lambda}> \lambda$ holds
for all pairs of adjacent regions $O_i^{(k_\lambda)}, O_j^{(k_\lambda)}$ in $\mathcal{P}^{(k_\lambda)}$, and we have
\begin{align}
K_{\max} = k_\lambda\;.\label{stop_implicit}
\end{align}
Alternatively, we can fix $\lambda=+\infty$  and set a targeted number of regions $N^*\leq N_0$, and let
\begin{align}
K_{\max} = N_0-N^* \;.\label{stop_explicit}
\end{align}

An advantage of the implicit  stopping criterion~\eqref{stop_implicit} over the explicit one~\eqref{stop_explicit} is its uniformity.  All  regions of the resulting partition satisfy the desired geometric property or relations with their respective neighbors specified by the  gain functional. Such an appropriate $\lambda^*$ is strongly influenced by the image contents.  Typically, for images with rich details, $\lambda^*$ should be small; and for images dominated by large homogeneous patches, $\lambda^*$ should be large. This is analogous to the scale-selection property of the MS model~\cite{koepfler1994multiscale,mumford1989optimal}. 
 Yet, depending on the choice of gain functional, the range of $\lambda$ can be very different. For instance, the parameter $\lambda$ corresponding to the Mumford-Shah criterion~\eqref{eq_MS_gain} has the same dimension as length; whereas the parameter $\lambda$ corresponding to the area criterion~\eqref{eq_area_gain} has the same dimension as area. On the other hand, \eqref{stop_explicit} compared to~\eqref{stop_implicit} is more convenient. It offers a direct control of the \textit{a priori} complexity of the resulted partition.

To combine the benefits of both stopping criteria, we allow the user to specify the approximate number of regions $N^*$, and we define $\lambda^*=0$. We apply the region merging process for $N_0-N^*$ times. Every time we merge a pair of regions $O$ and $O'$, we update $\lambda^*$ by the maximum of the merging cost and the current value for $\lambda^*$, i.e., 
\begin{align}
\lambda^* \gets \max\{\Delta E(O,O'),\lambda^*\}\;.
\end{align}
After $N_0-N^*$ merging rounds, we continue iterating the merging procedure until the implicit stopping criterion holds for $\lambda^*$. By doing so, we ensure that merging any pair of adjacent regions of the resulting partition would invoke greater costs.  Consequently, the number of regions in the resulting partition can be smaller than $N^*$.

\section{Numerical Experiments}\label{sec_experiment}

In this experimental section, we discuss and compare the properties of  vectorization algorithms. For the proposed vectorization method, we fix  $T^*=1.0$ for the affine shortening stopping time, $\tau=0.5$ for the   approximation error threshold, and $I_{\max}=3$ for the number of iterations. We tested the algorithms on images sampled from the following datasets.
\begin{itemize}
\item \textbf{SAVOIAS}~\cite{SaraeeJaBe18}:  A  dataset of more than 1400 images with sizes ranging from below $200\times 300\times 3$ to above $6000\times 4000\times 3$. Moreover, all images are labeled as seven categories: Scenes, Advertisements, Visualization and infographics, Objects, Interior design, Art, and Suprematism.  This dataset was introduced to analyze the visual complexity of images~\cite{SaraeeJaBe20}.
\item \textbf{DECOR}\footnote{\href{https://www.kaggle.com/datasets/olgabelitskaya/traditional-decor-patterns}{https://www.kaggle.com/datasets/olgabelitskaya/traditional-decor-patterns}}: A collection of $485$ color images ($150\times 150\times 3$) of traditional decor patterns, which contain rich details and complex shapes. 
\end{itemize}

To reduce noise and artifacts due to JPEG compression, all the input raster images are preprocessed by the  FFDNet~\cite{zhang2018ffdnet}, which is a CNN-based discriminative image denoising technique. This method requires an estimated noise level ($0\sim 255$) which we fix to $10$ for all clean images. The goal is to alter minimally the shapes present in the images, while removing small noise and JPEG artifacts that should not be kept in any vector representation.  
A typical way to evaluate the accuracy is to use PSNR which compares pixel values.  Since vector graphics are not stored in pixels, 
it is common to compare vectorization results after converting them back to pixel images via rasterization. Note that rasterization is a non-trivial process, and different implementations can produce images with varying levels of quality. In this paper we consistently use \texttt{Inkscape}\footnote{\href{https://inkscape.org}{https://inkscape.org}} for converting  vector graphics to raster images in all experiments for the PSNR computation. 
\begin{figure}
\centering
\begin{tabular}{c@{\hspace{2pt}}c@{\hspace{2pt}}c}
(a) Zoom$\times 1$&(b) Zoom$\times 3$&(c) Zoom$\times 9$\\
\includegraphics[width=0.2\textwidth]{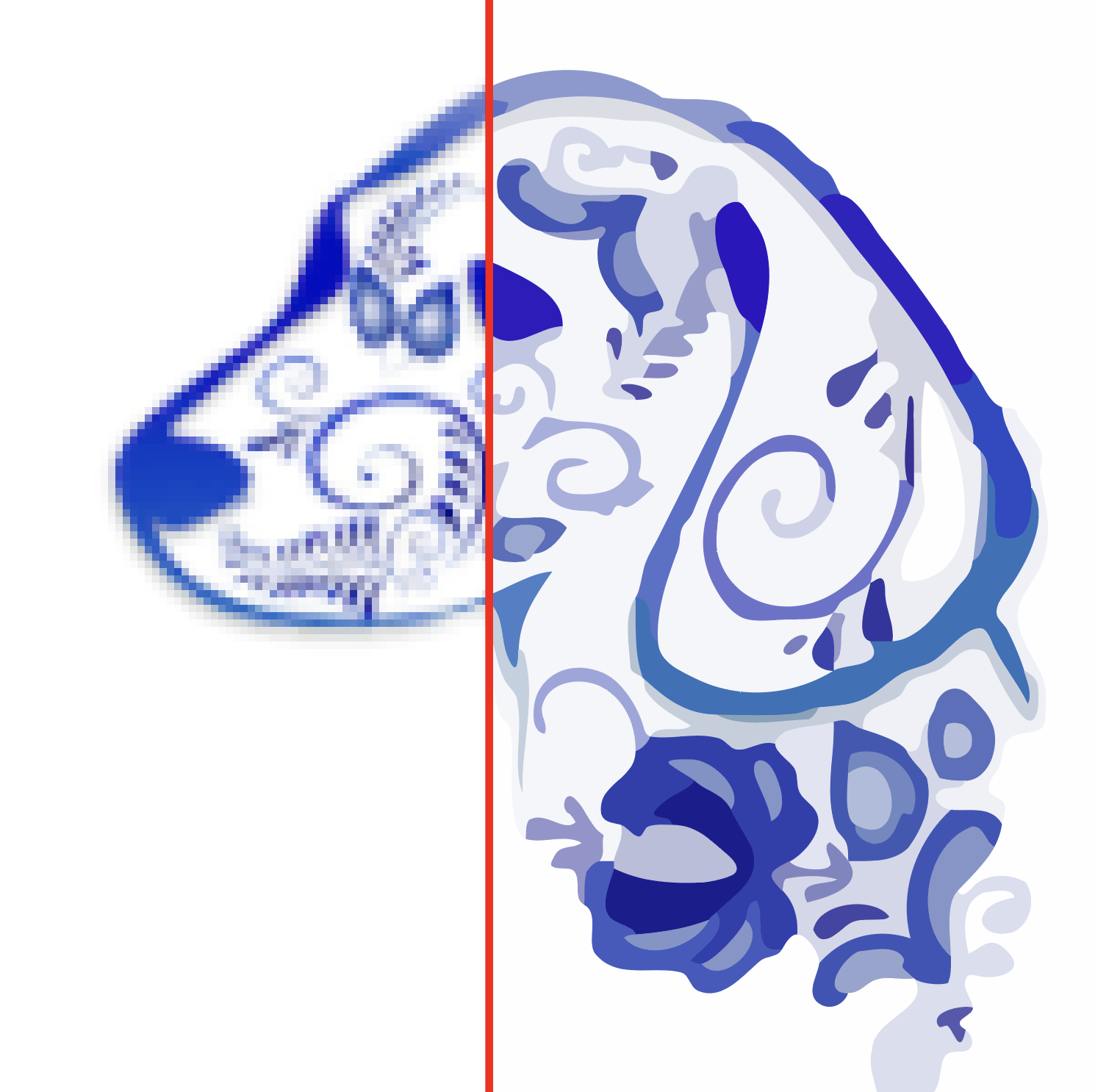}&
\includegraphics[width=0.2\textwidth]{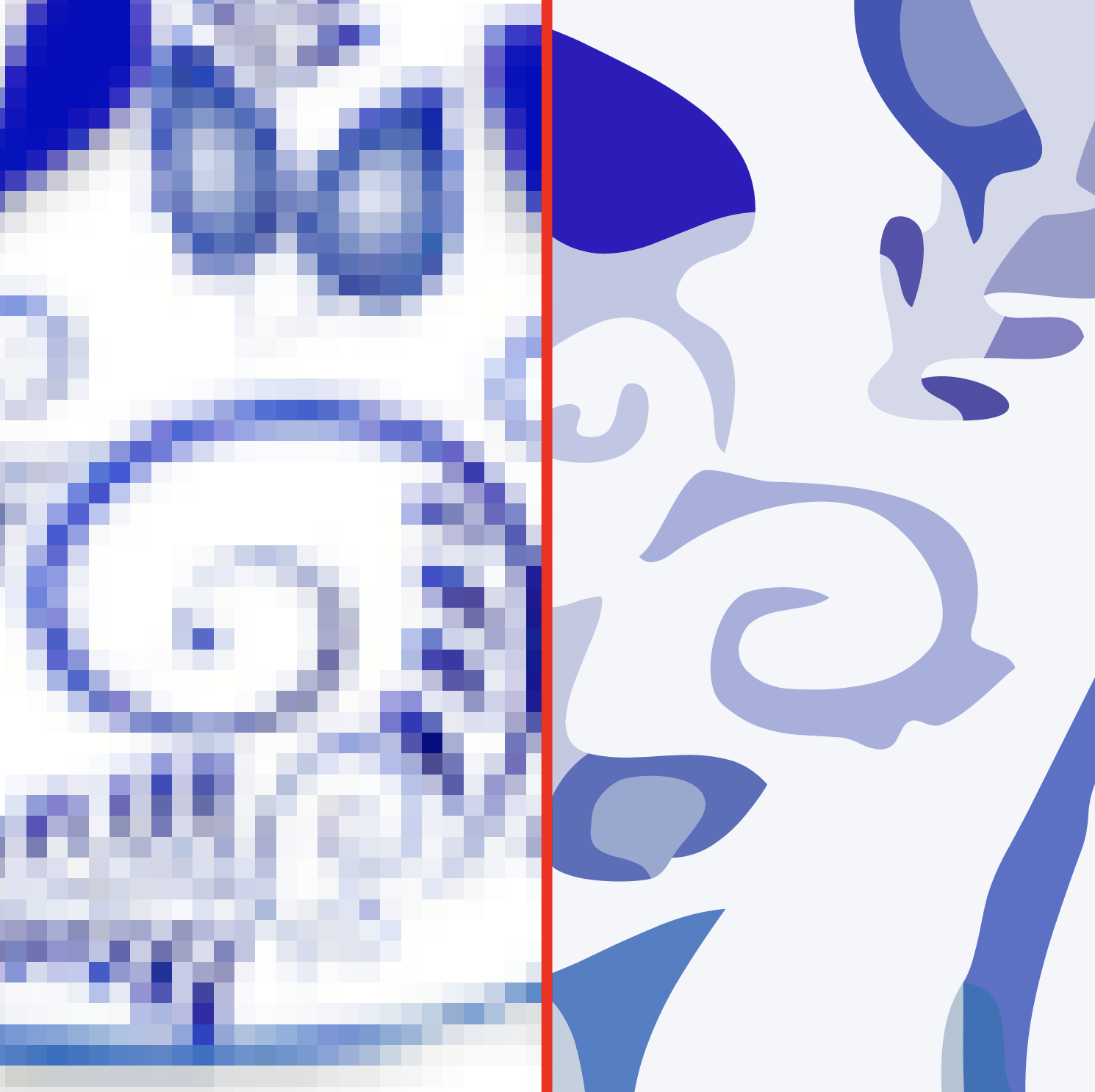}&
\includegraphics[width=0.2\textwidth]{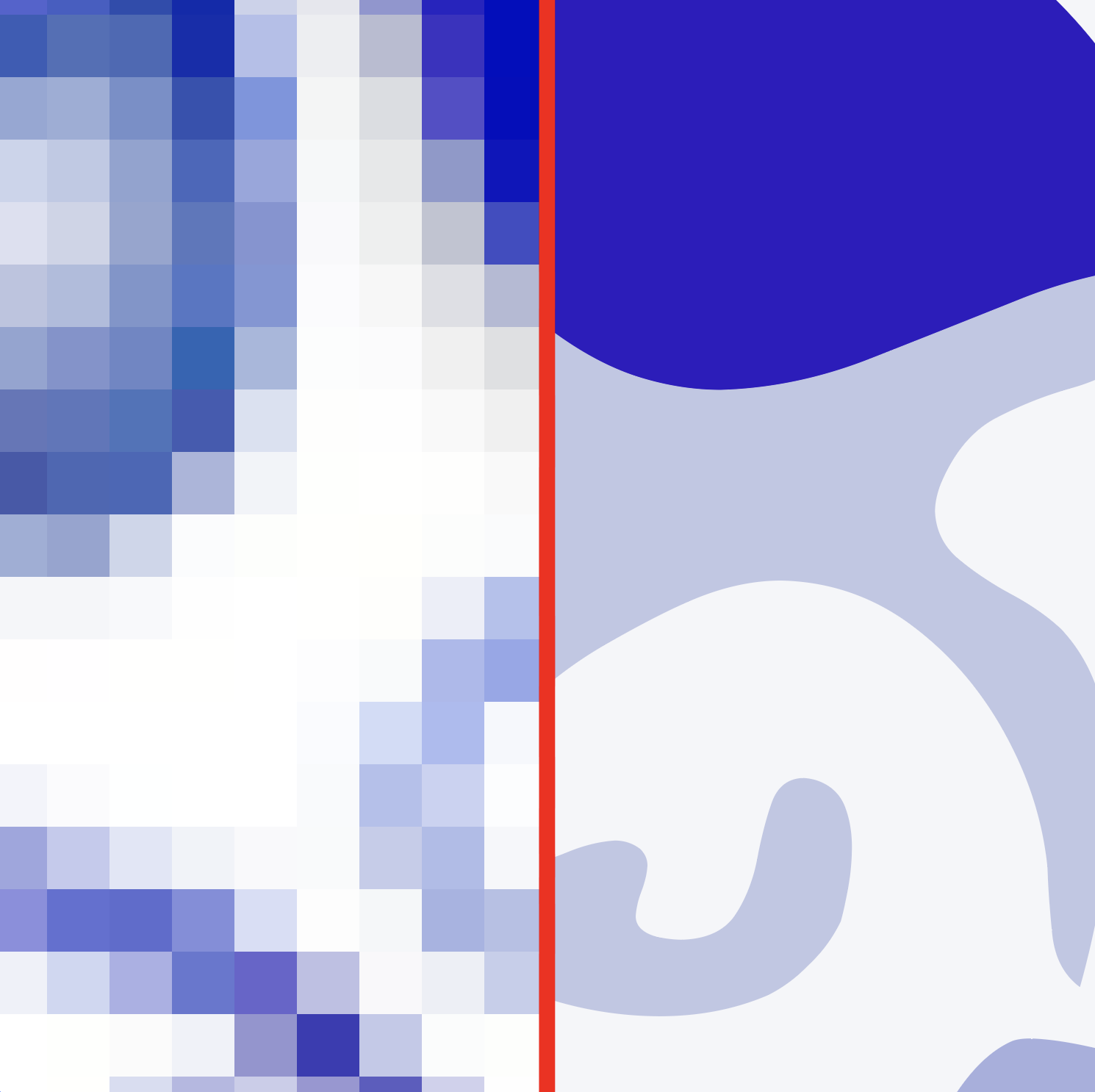}\\
\includegraphics[width=0.3\textwidth]{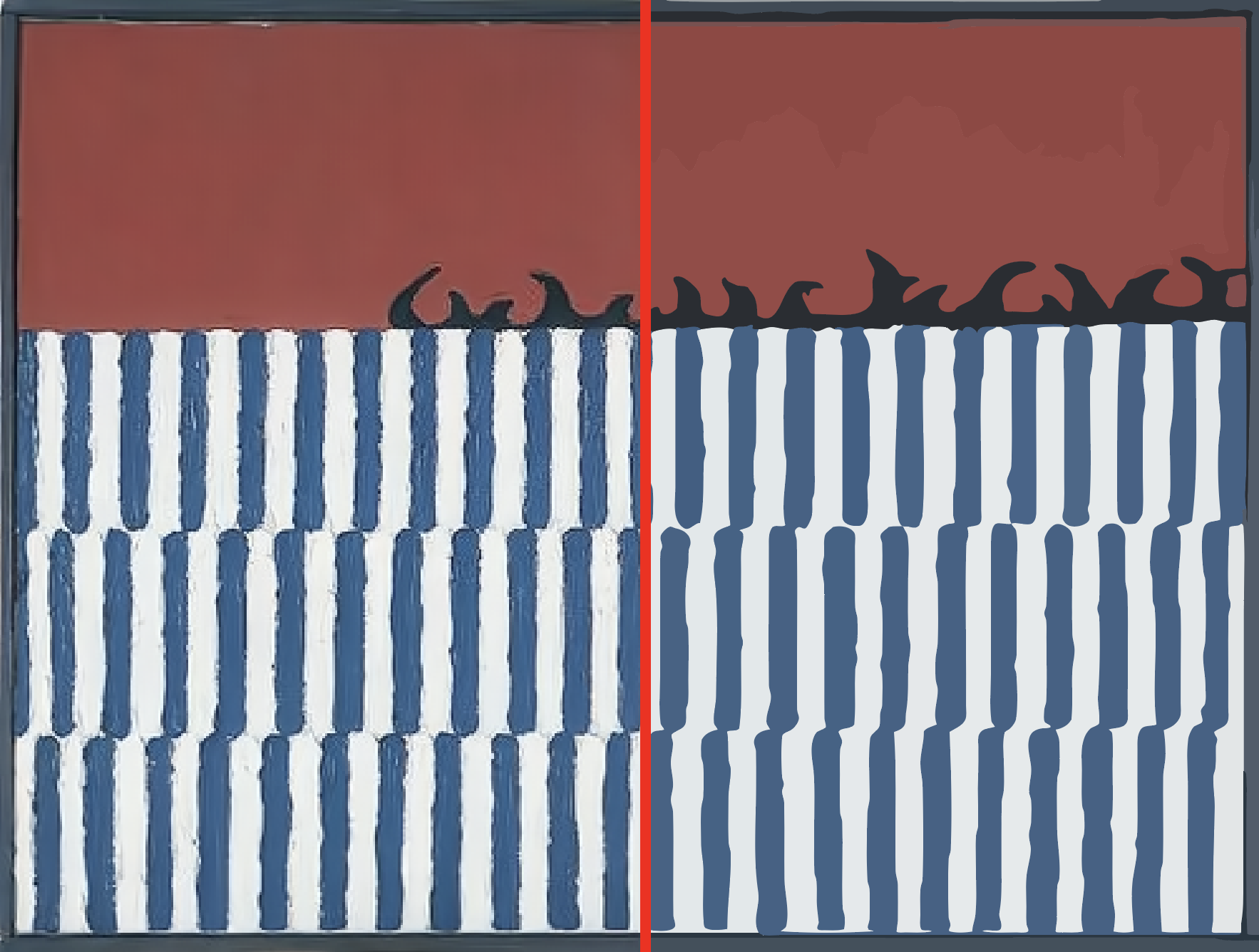}&
\includegraphics[width=0.3\textwidth]{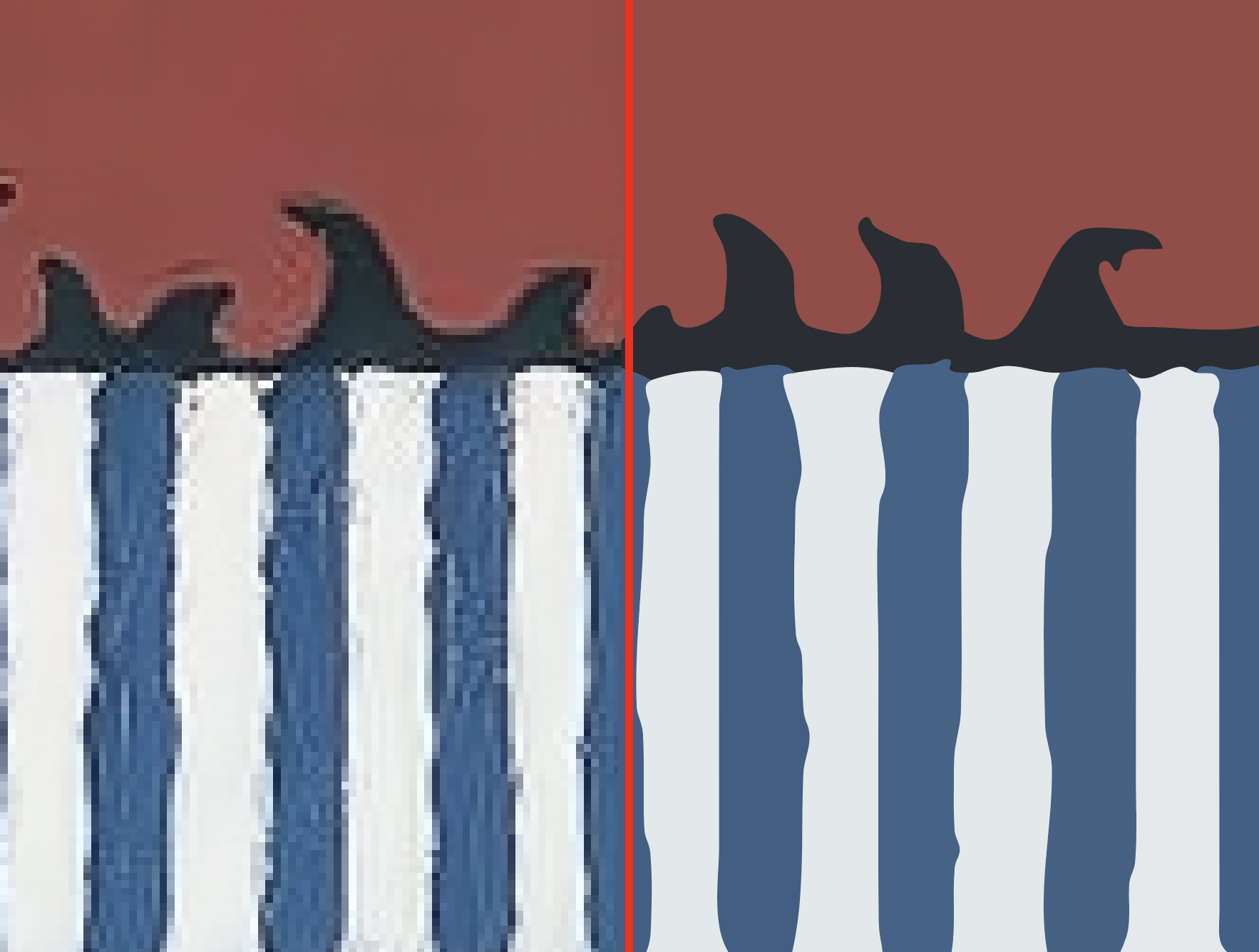}&
\includegraphics[width=0.3\textwidth]{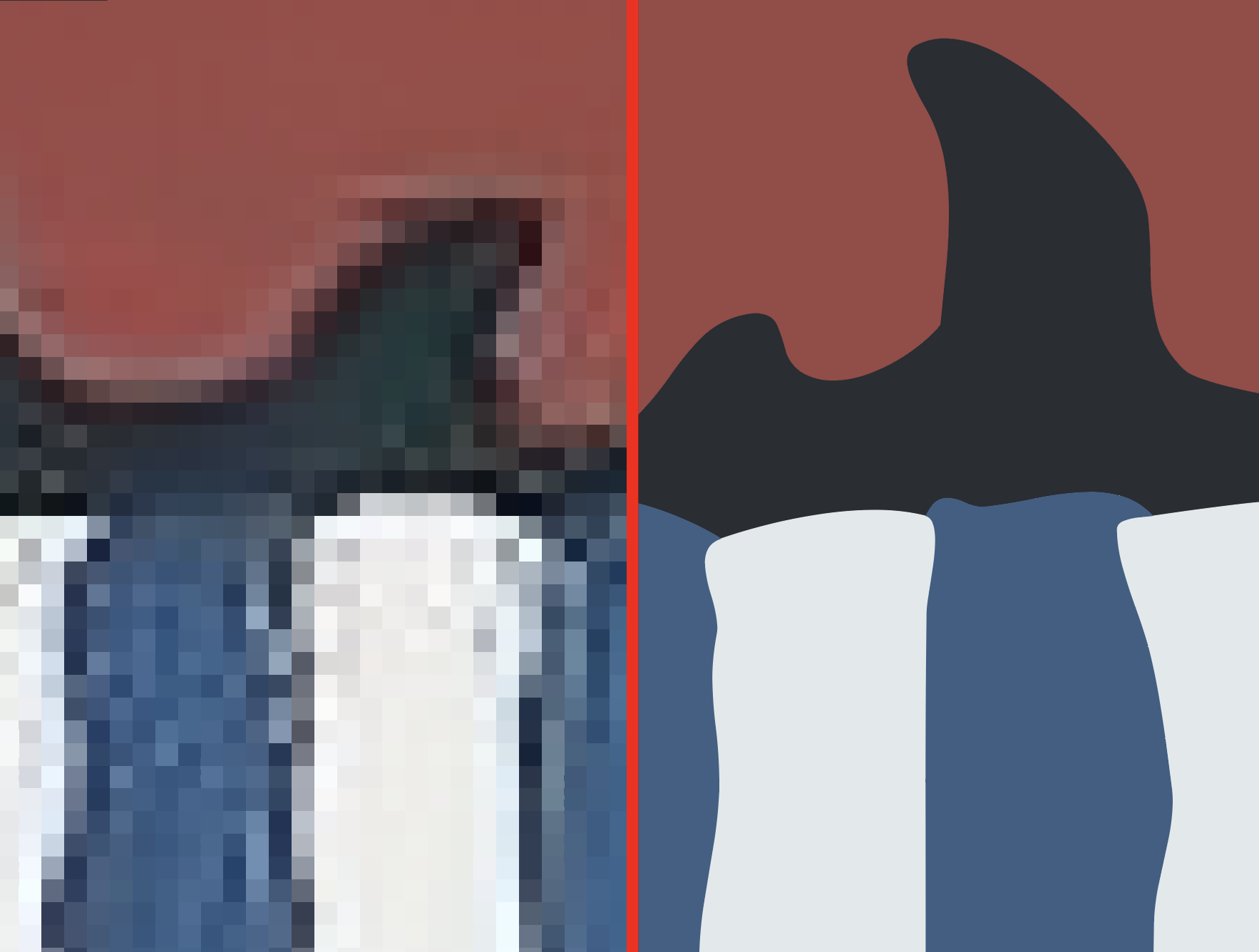}\\
\includegraphics[width=0.3\textwidth]{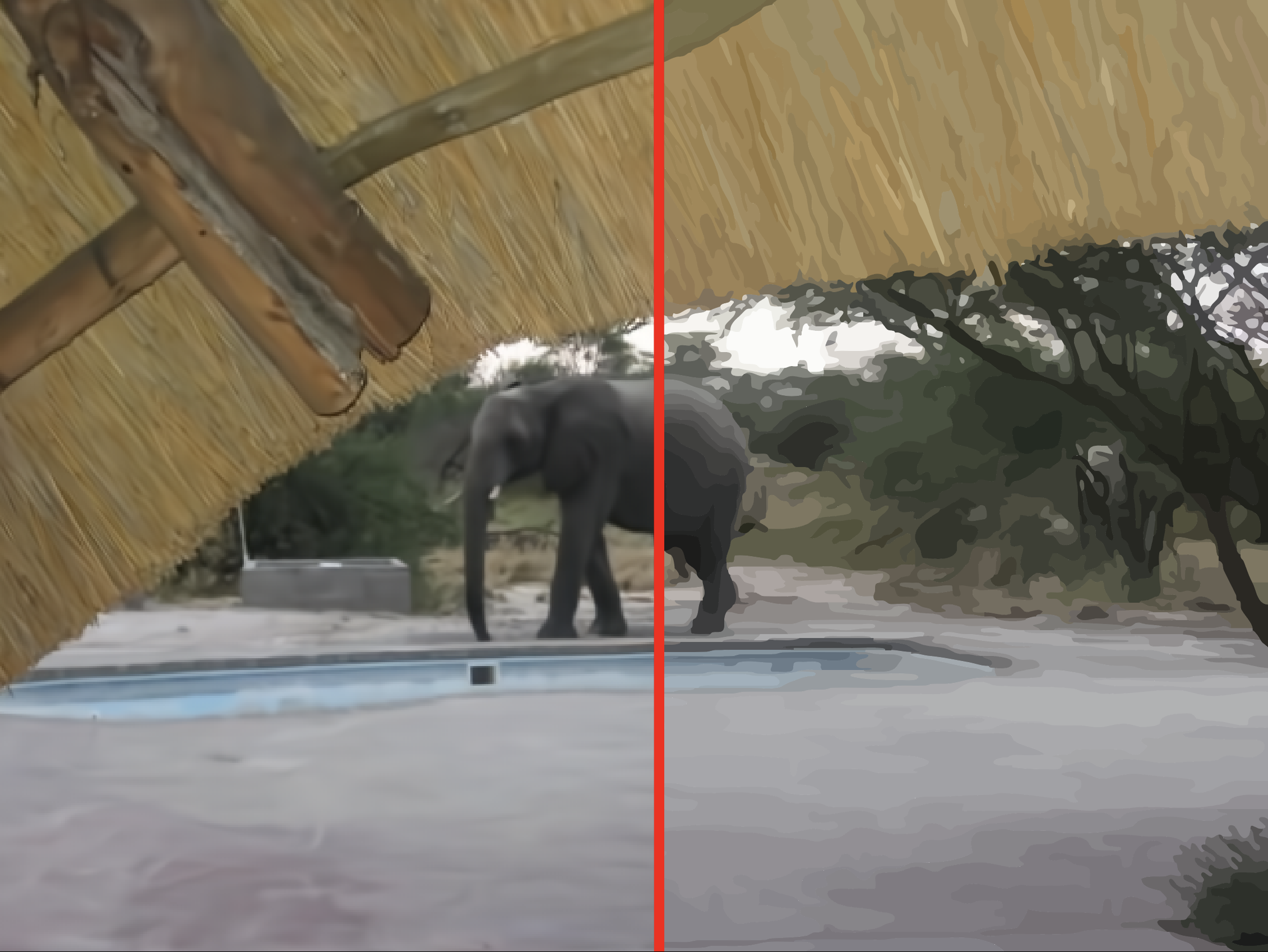}&
\includegraphics[width=0.3\textwidth]{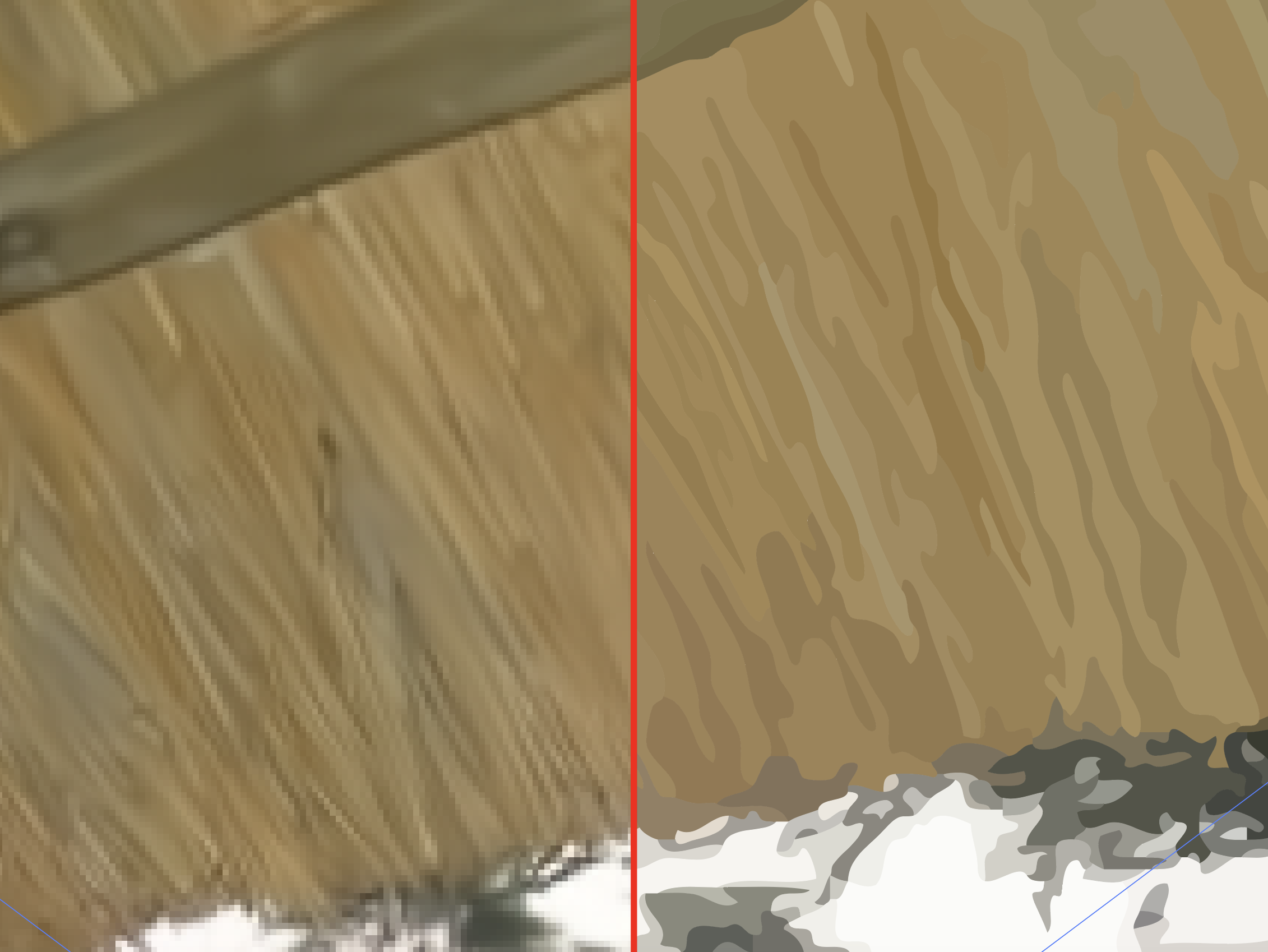}&
\includegraphics[width=0.3\textwidth]{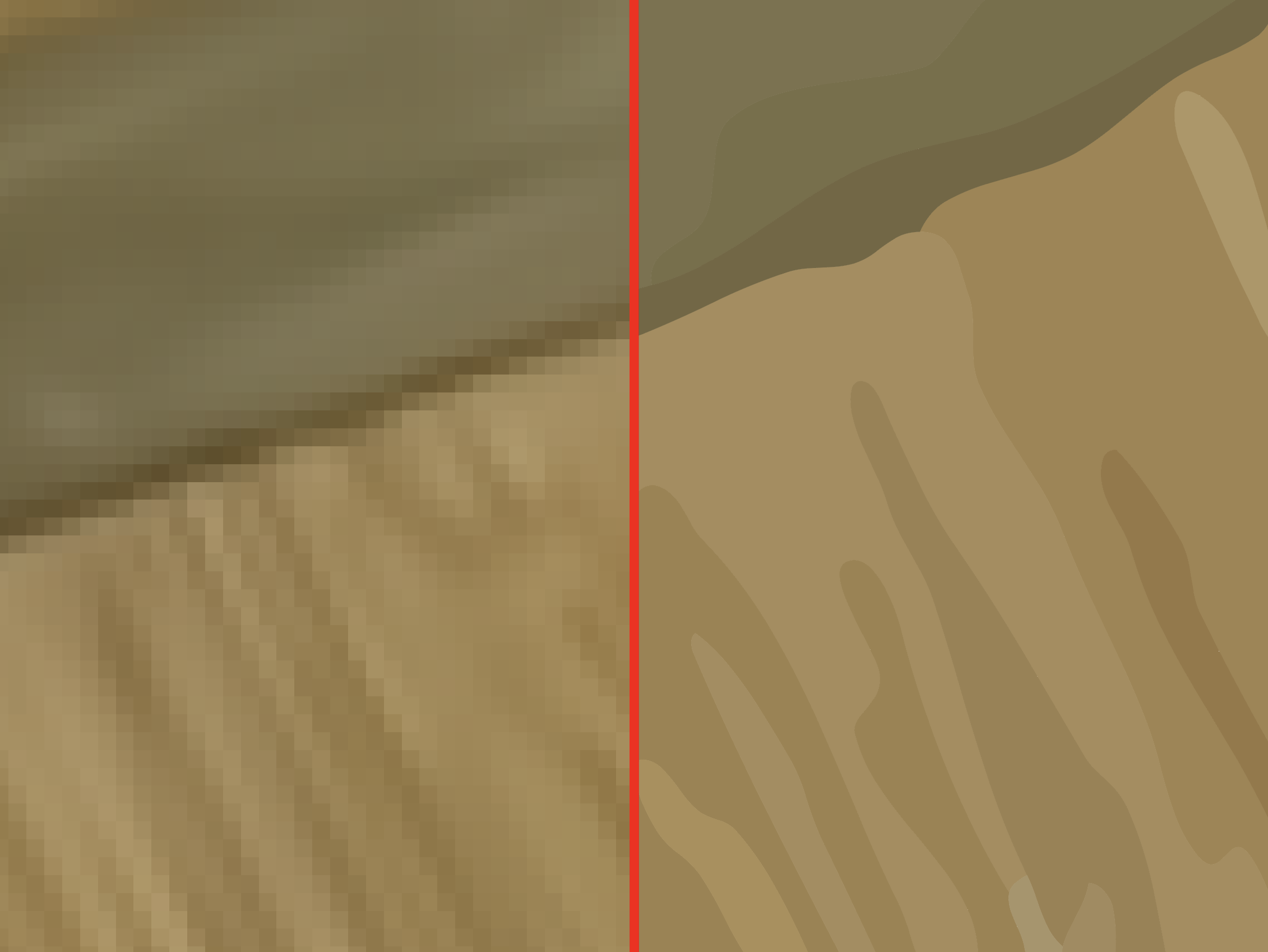}
\end{tabular}
\caption{General performances of the Area region merging~\eqref{eq_area_gain}. The image in the first row has size $150\times 150$, and the vectorized result contains $N=88$ regions (Execution Time (ET) $=2.18$ sec); the image in the second row has size $500\times 378$, and the vectorized result contains $N=50$ regions (ET $=3.72$ sec); the image in the third row has size $640\times 480$, and the vectorized result contains $N=667$ regions (ET $=7.83$ sec). In each image, the left panel shows half the raster input, and the right panel shows  the vectorized representation of the second half.}\label{fig_general}
\end{figure}
\subsection{General performances on vectorization}
We first illustrate the performances of the proposed vectorization algorithm. We use the Area merging gain~\eqref{eq_area_gain} and test it on images of different styles and sizes. We shall compare different merging criteria in the following subsections. 

Figure~\ref{fig_general} shows the results. In the first row, we test on a small image of size $150\times 150$. When zoomed with a factor of $3$ and $9$ respectively, the raster image exhibits strong  pixelation artifacts, as shown in the left panels. In each right panel, we show the vectorized results containing $88$ regions in the corresponding scales. Compared to the raster images, the vectorized graphics can be magnified arbitrarily while the quality remains unchanged. In the second row, we show an image with size $500\times 378$ and our vectorized representation contains only $N=50$ regions. We observe that excessive color variations are present due to antialiasing, especially when the raster image is zoomed by a factor of $9$; our vectorization algorithm effectively removes such undesirable features.  We also note that most blue vertical stripes are connected at corners, thus $N$ is smaller than the number of stripes. In the third row, the image has size $640\times 480$ and the vectorized result contains $N=667$ regions.  For photographs, the number of regions is usually increased due to complex textures and gradient. In its original scale, we see that the proposed algorithm produces a piecewise constant vector graphic with flexible regions closely approximating the input. The raster image gets blurred when we zoom in, whereas the vector graphic remains clear and the regions' boundaries identify with objects' contours and highlight important textures. The Execution Time (ET) for the proposed vectorization is acceptable for practical applications. Using the default number of iterations ($I_{\max}=3$), the ET for the small image in the first row is measured at 2.18 seconds; for the image in the second row, it is 3.72 seconds; and for the third row image, it reaches 7.83 seconds.  It is observed that the ET increases as the size of the image grows and as it contains more intricate details.

\newcommand{\mysize}{0.12}
\subsection{Region merging  criteria and their general behaviors}\label{sec_exp_general}
\begin{figure}
\centering
\begin{tabular}{c@{\hspace{2pt}}c@{\hspace{2pt}}c@{\hspace{2pt}}c@{\hspace{2pt}}c@{\hspace{2pt}}c}
(a)&(b)&(c)&(d)&(e)&(f)\\
\raisebox{1.6cm}{\multirow{3}{*}{
\includegraphics[width=0.295\textwidth]{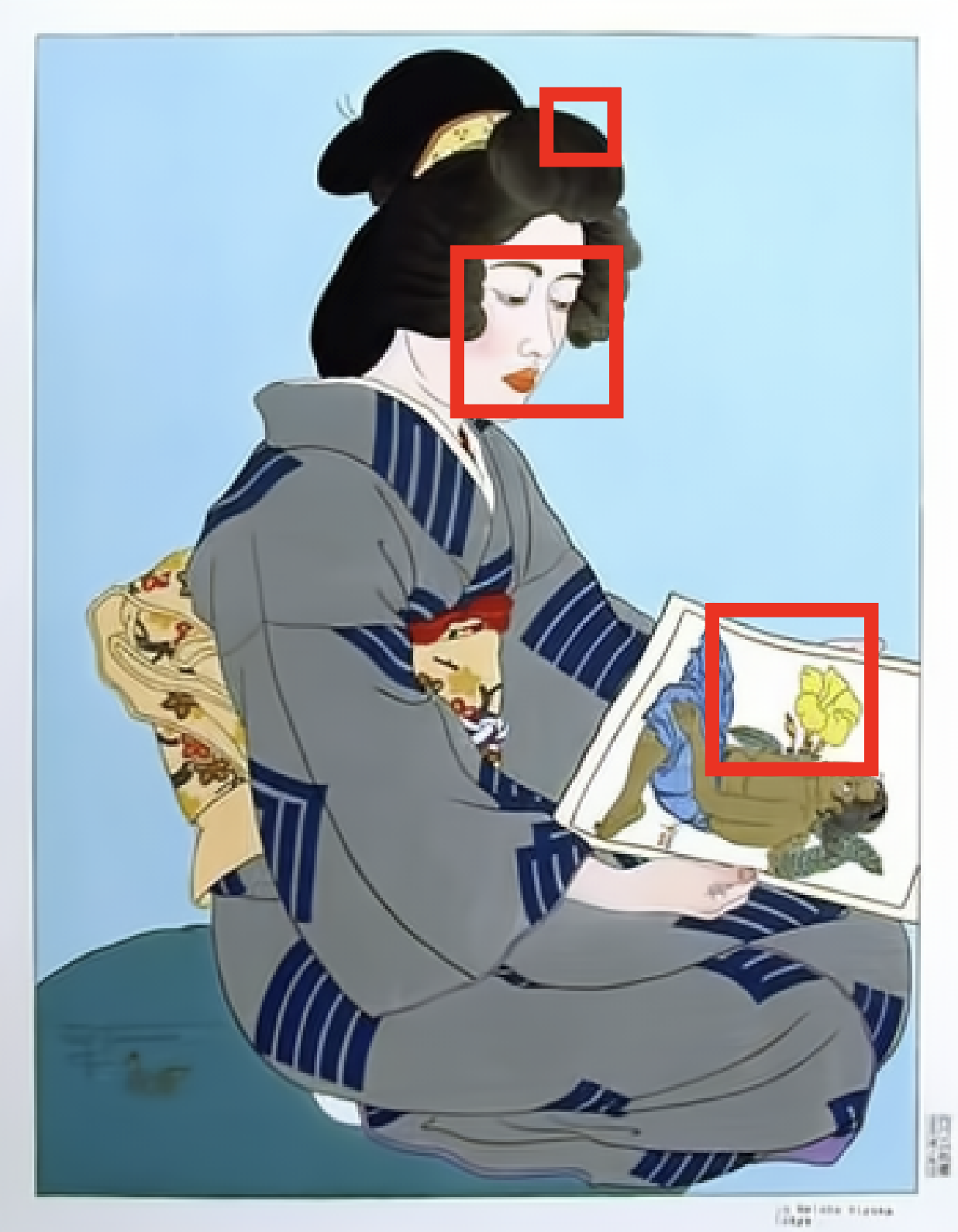}}}&\includegraphics[width=\mysize\textwidth]{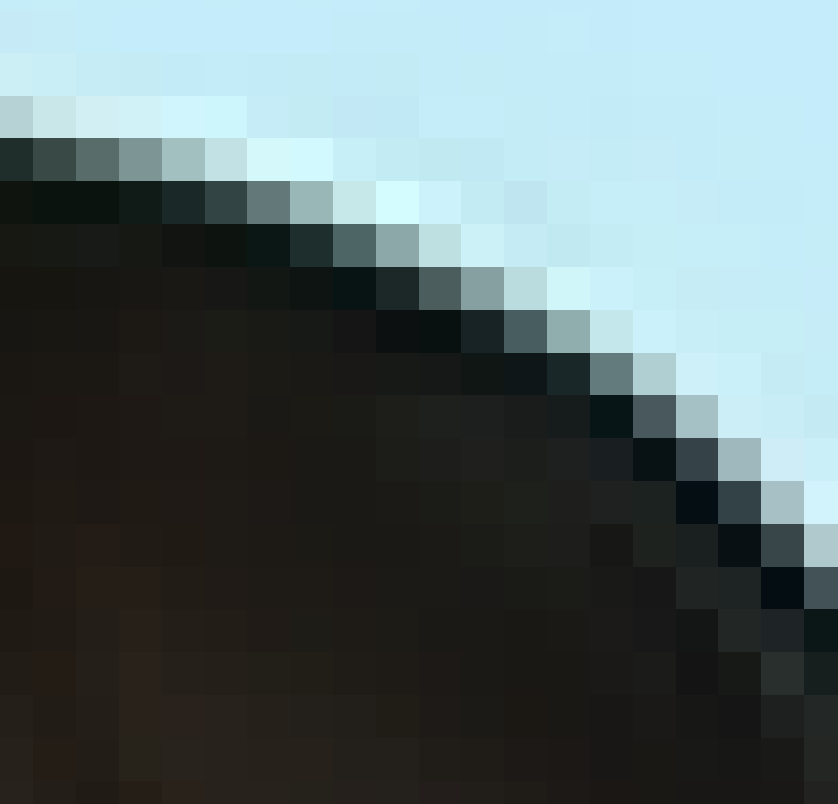}&\includegraphics[width=\mysize\textwidth]{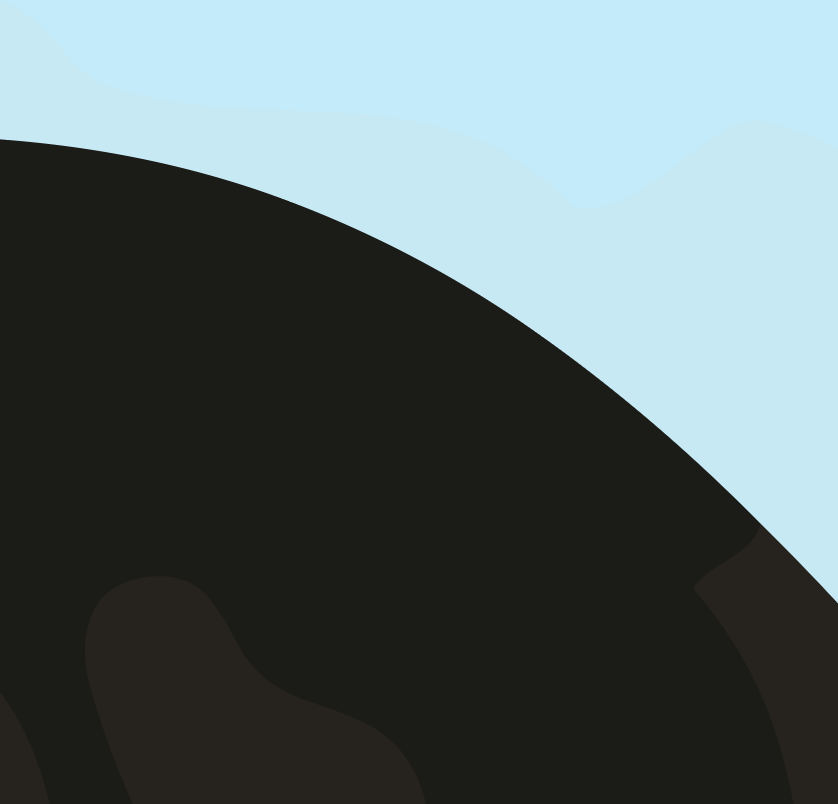}&
\includegraphics[width=\mysize\textwidth]{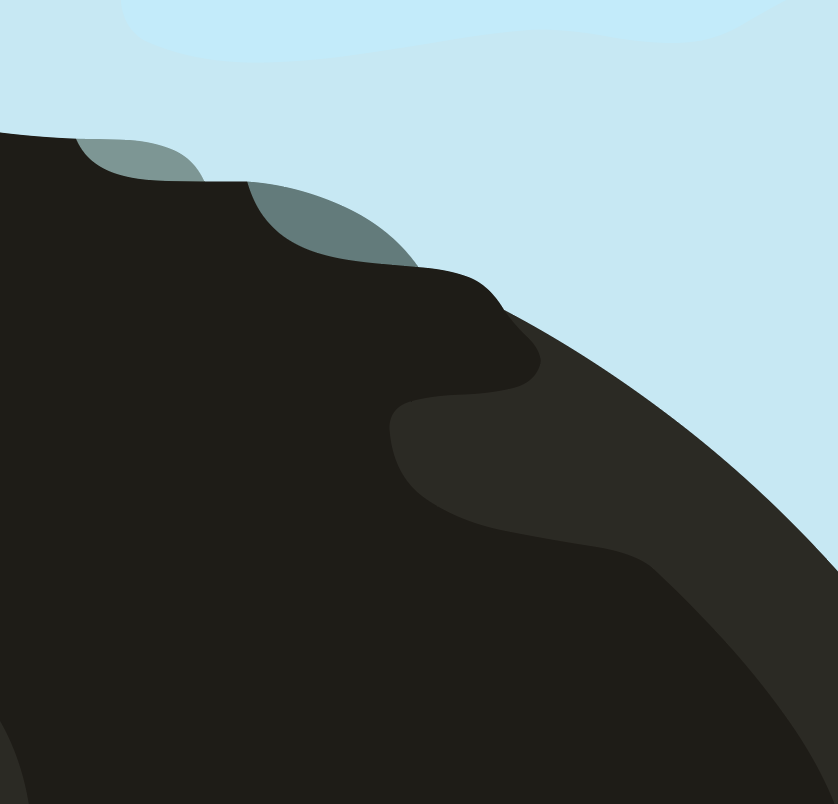}&
\includegraphics[width=\mysize\textwidth]{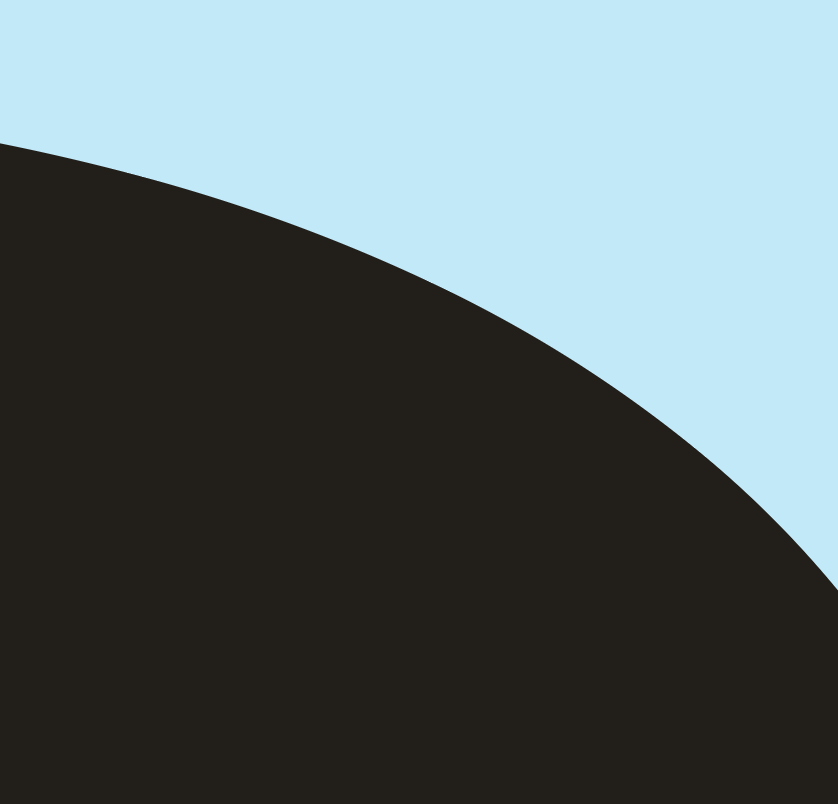}&
\includegraphics[width=\mysize\textwidth]{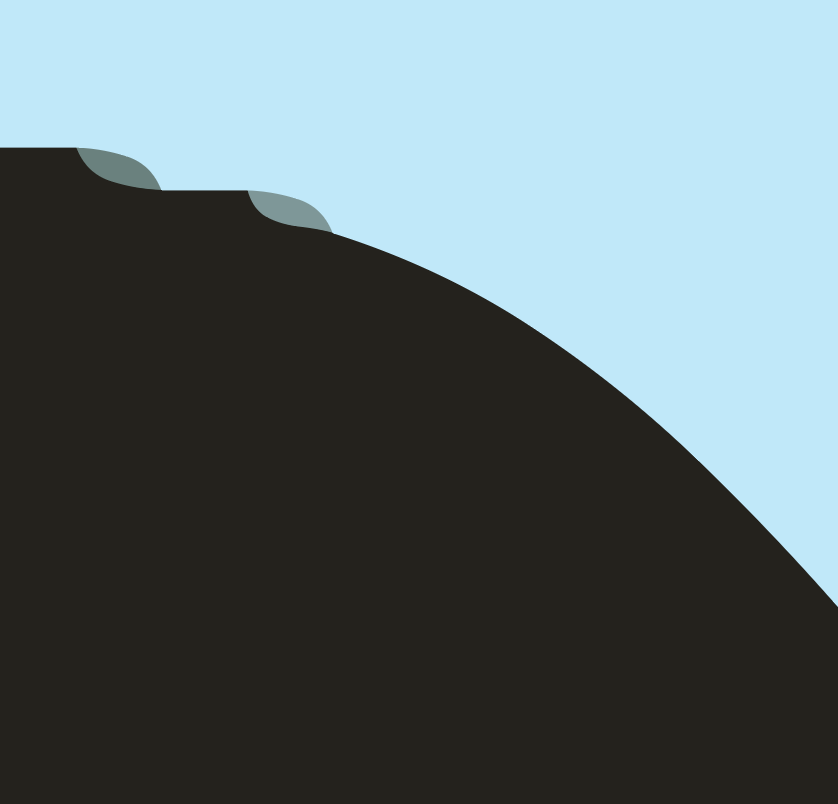}\\
&\includegraphics[width=\mysize\textwidth]{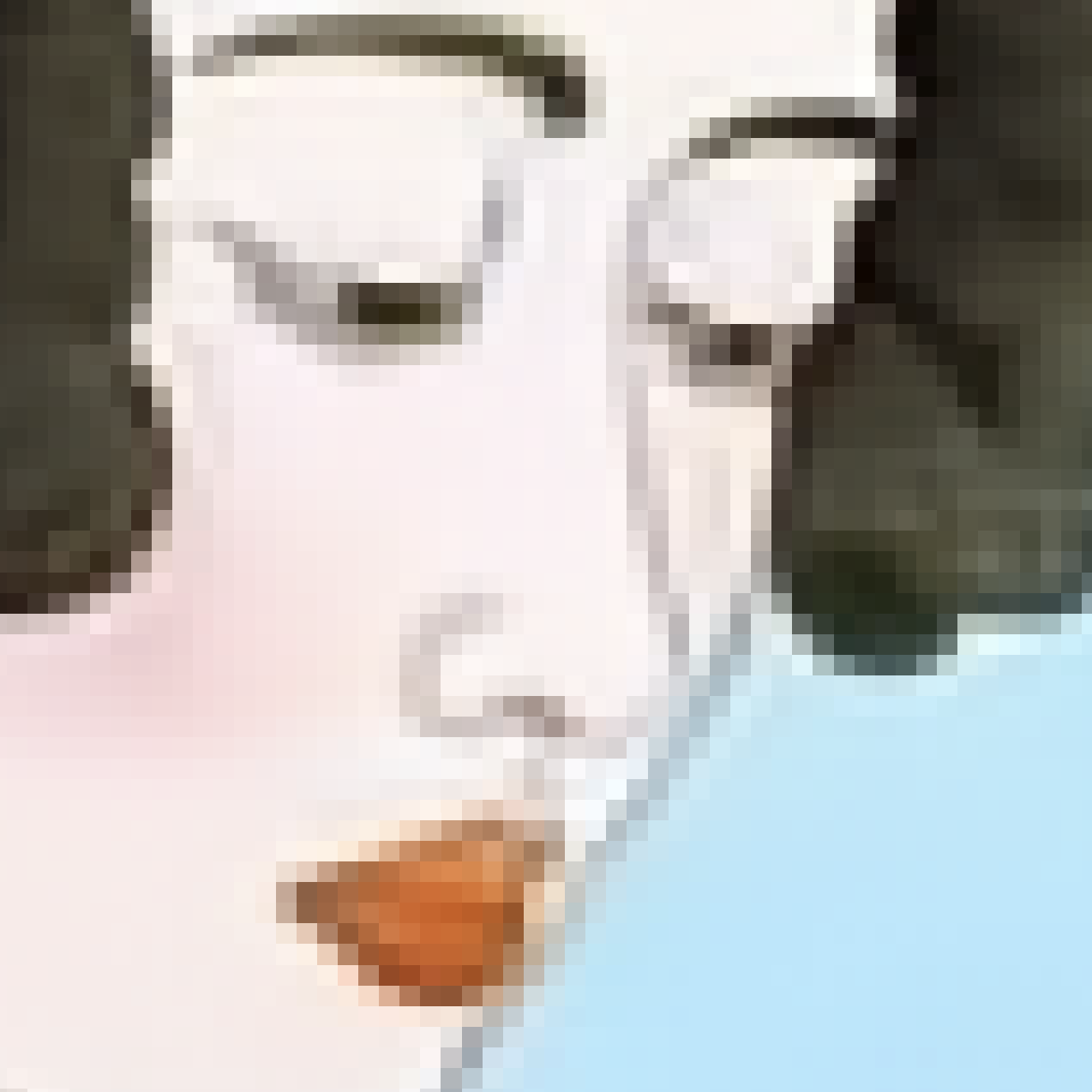}&\includegraphics[width=\mysize\textwidth]{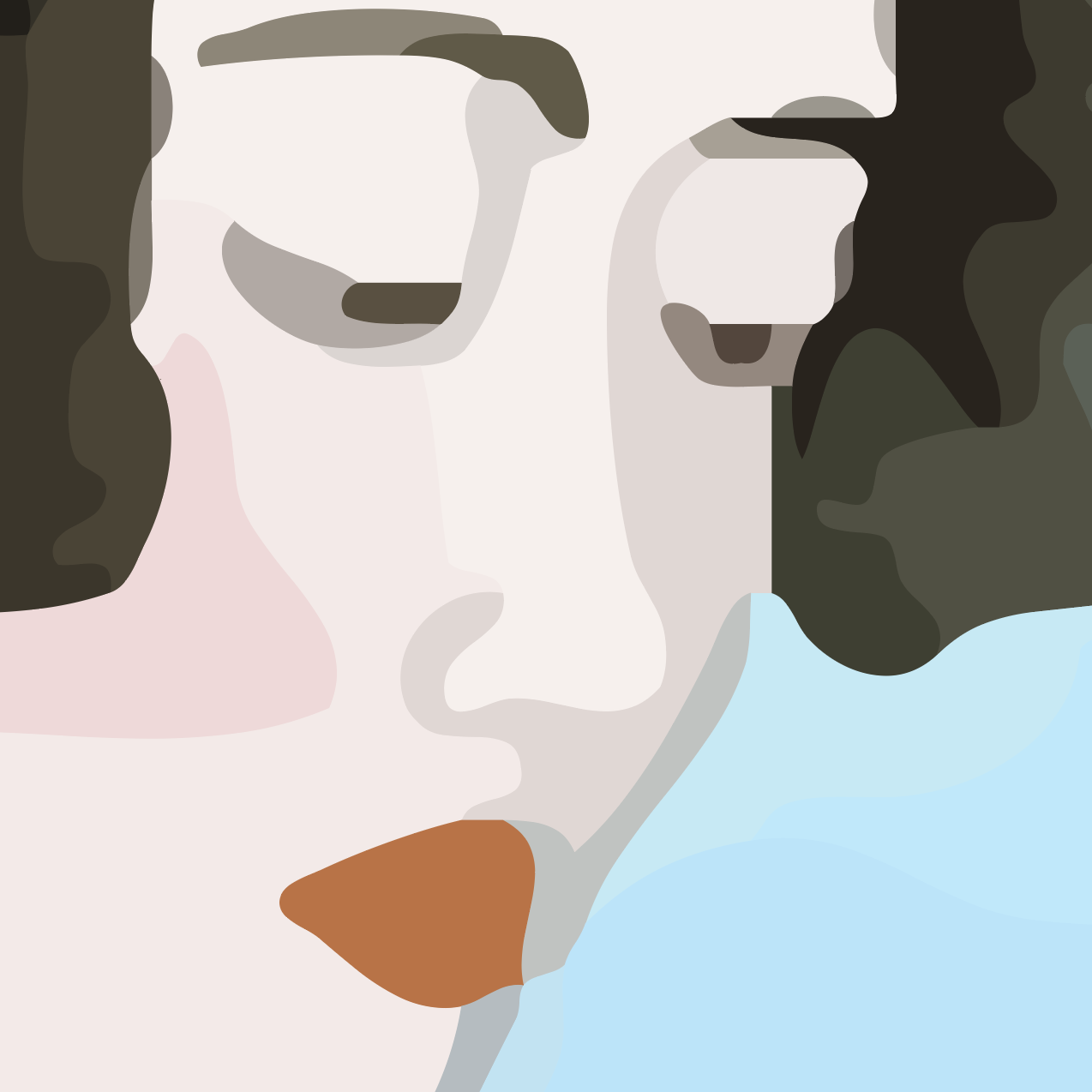}&
\includegraphics[width=\mysize\textwidth]{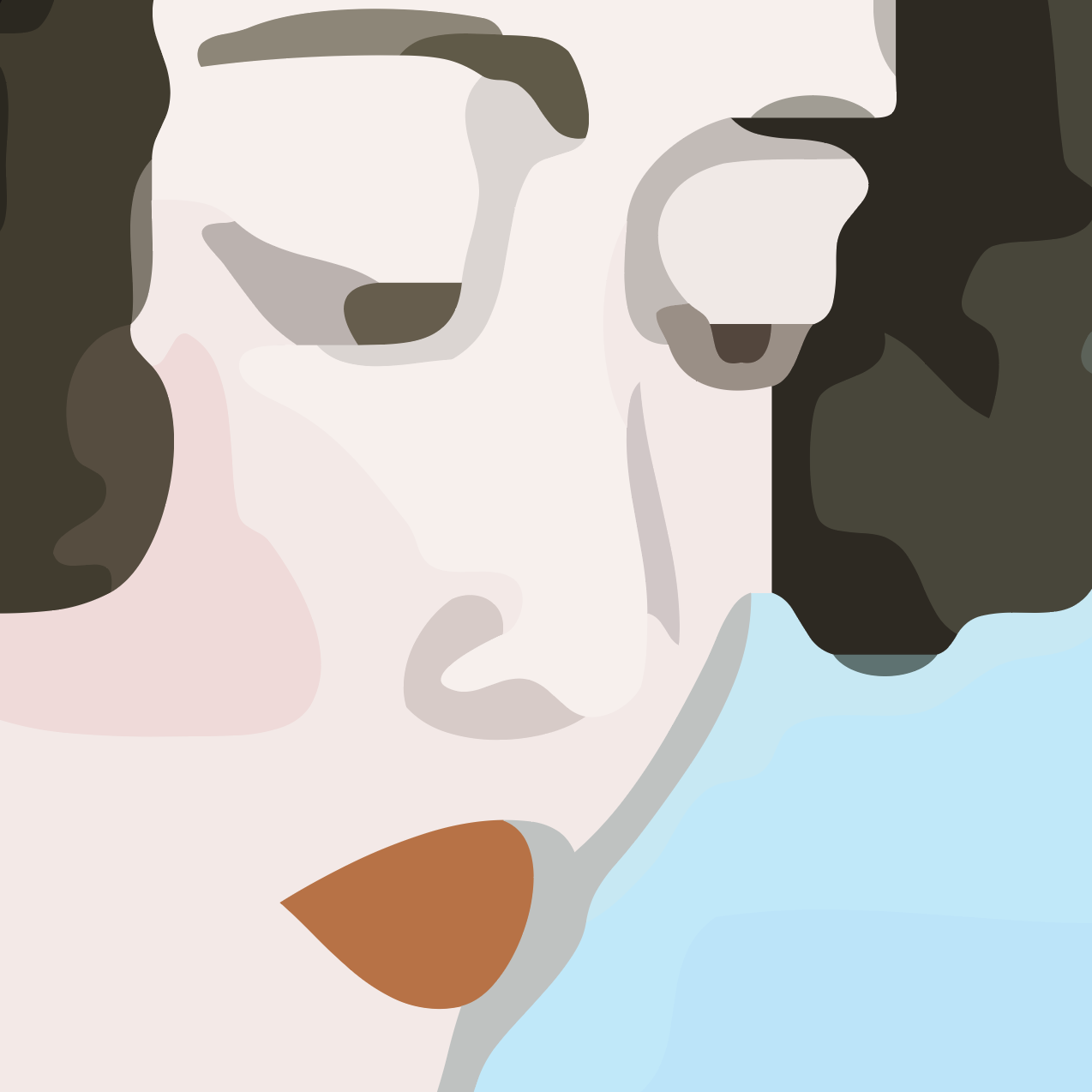}&
\includegraphics[width=\mysize\textwidth]{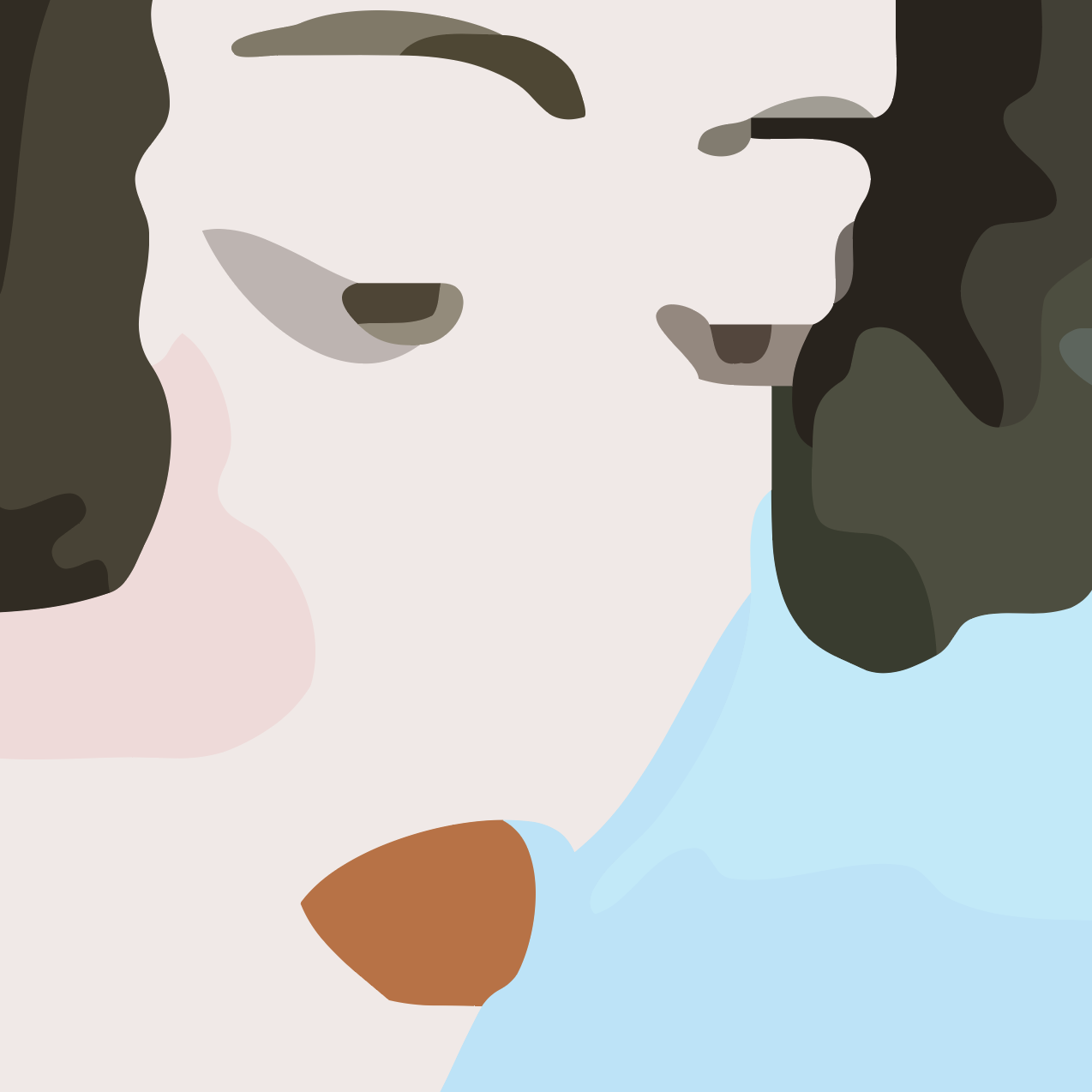}&
\includegraphics[width=\mysize\textwidth]{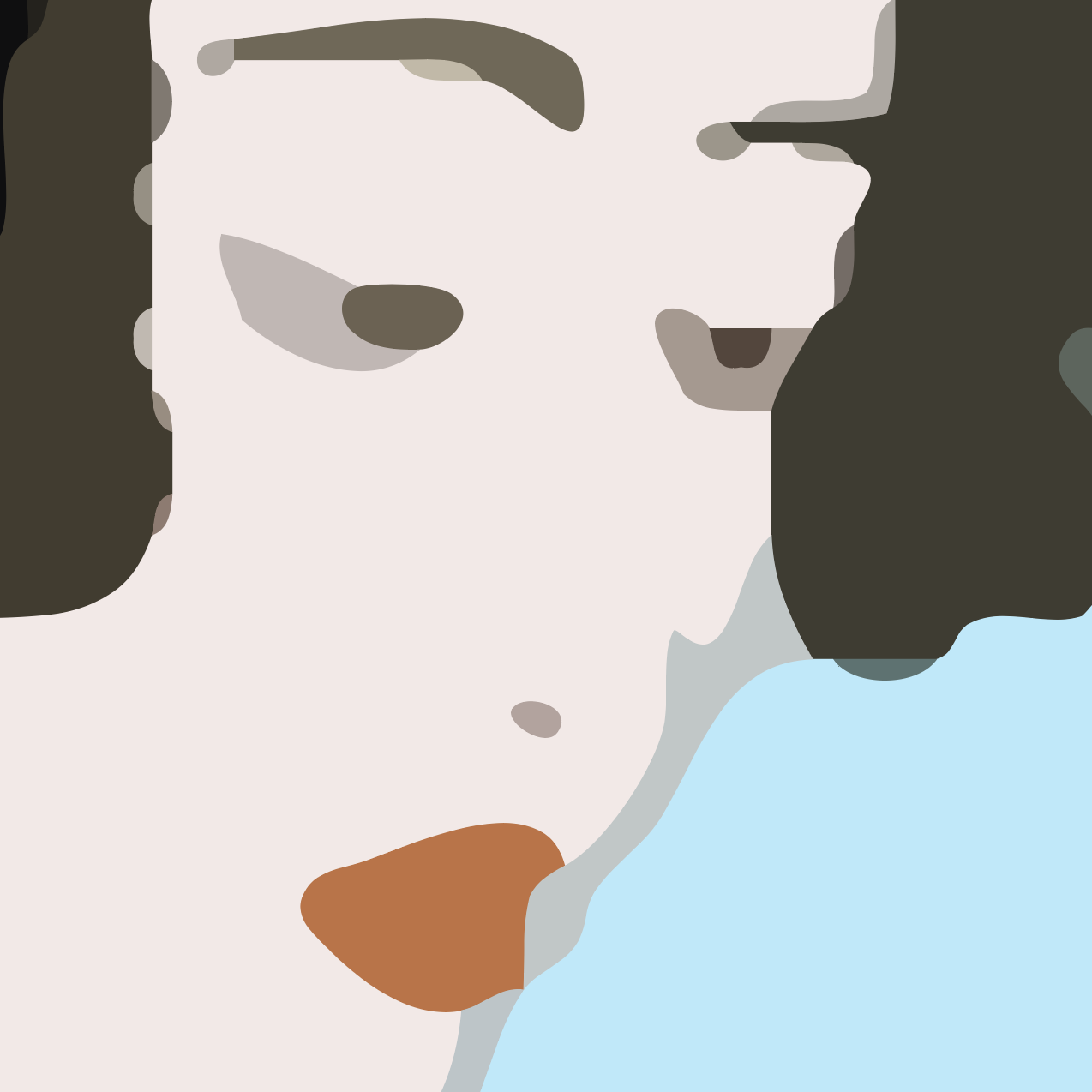}\\
&\includegraphics[width=\mysize\textwidth]{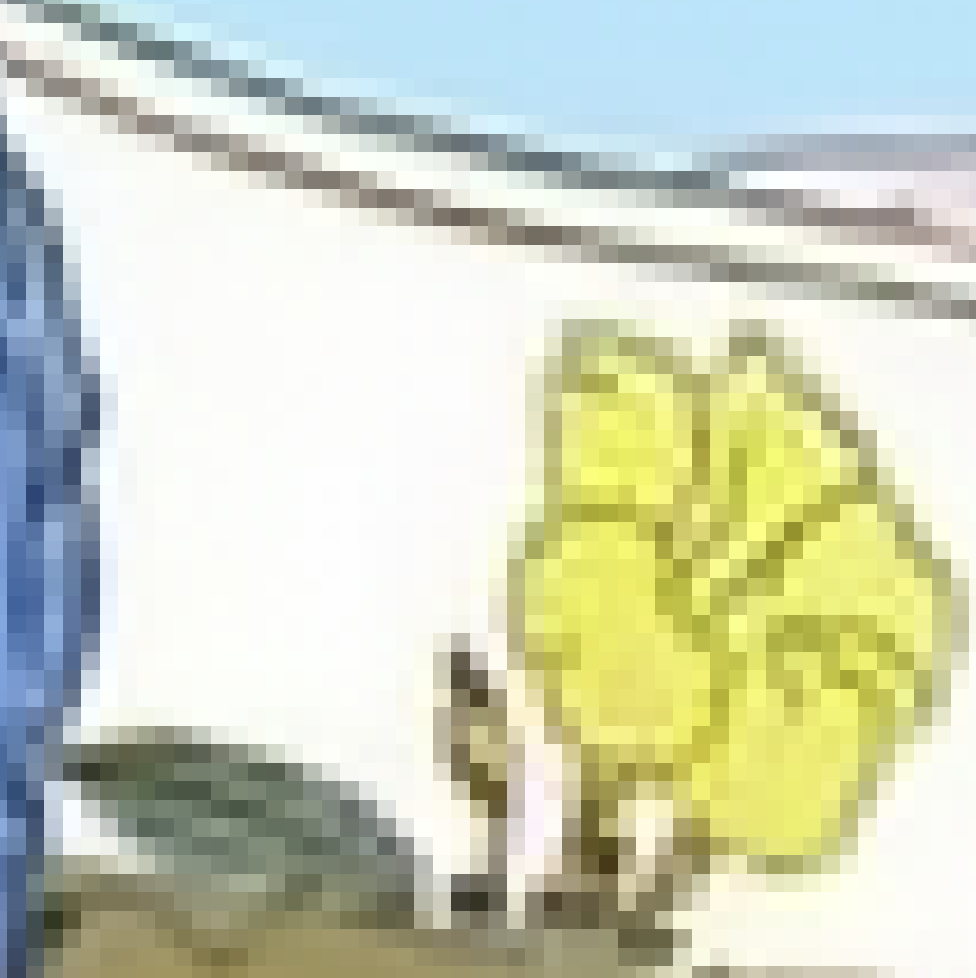}&\includegraphics[width=\mysize\textwidth]{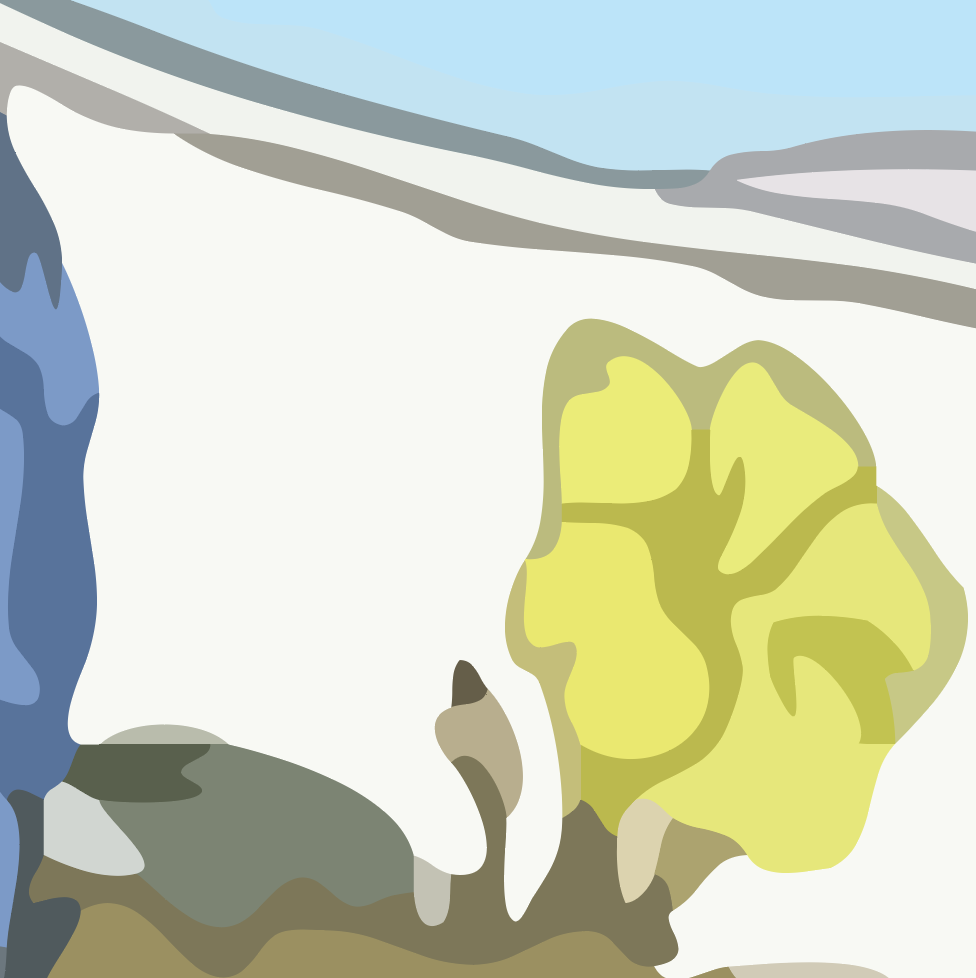}&
\includegraphics[width=\mysize\textwidth]{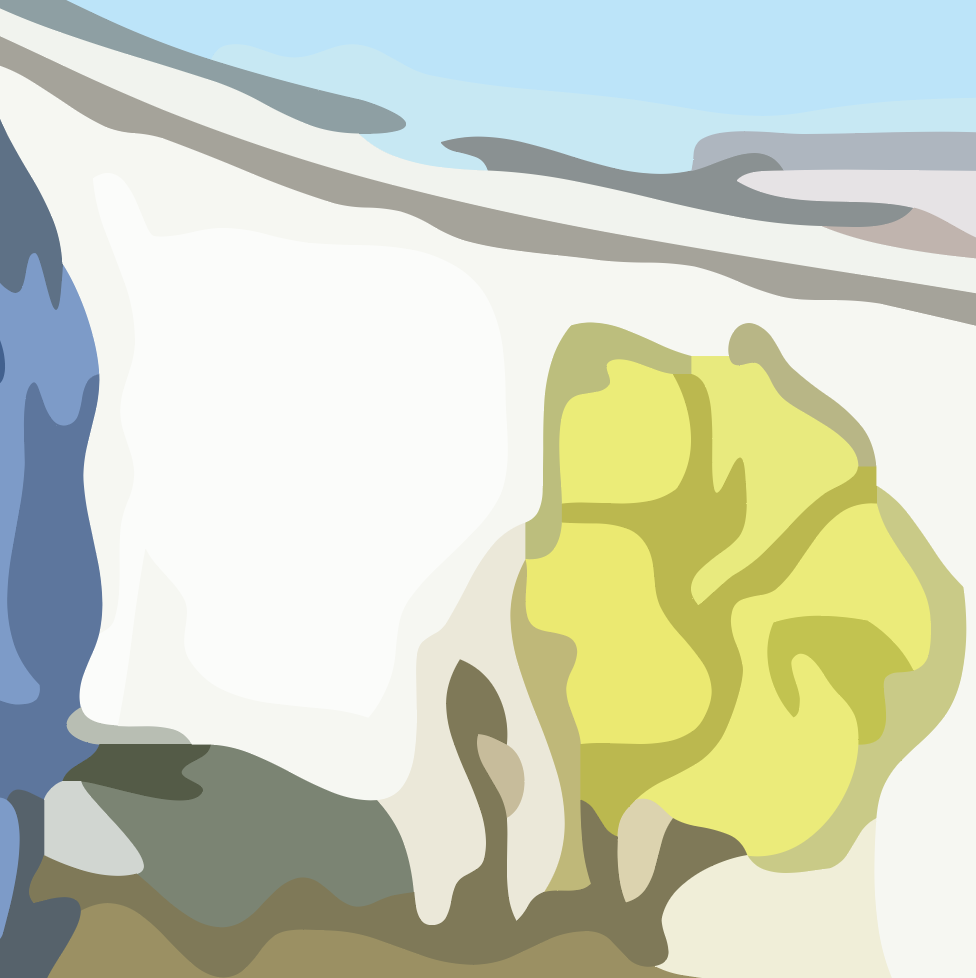}&
\includegraphics[width=\mysize\textwidth]{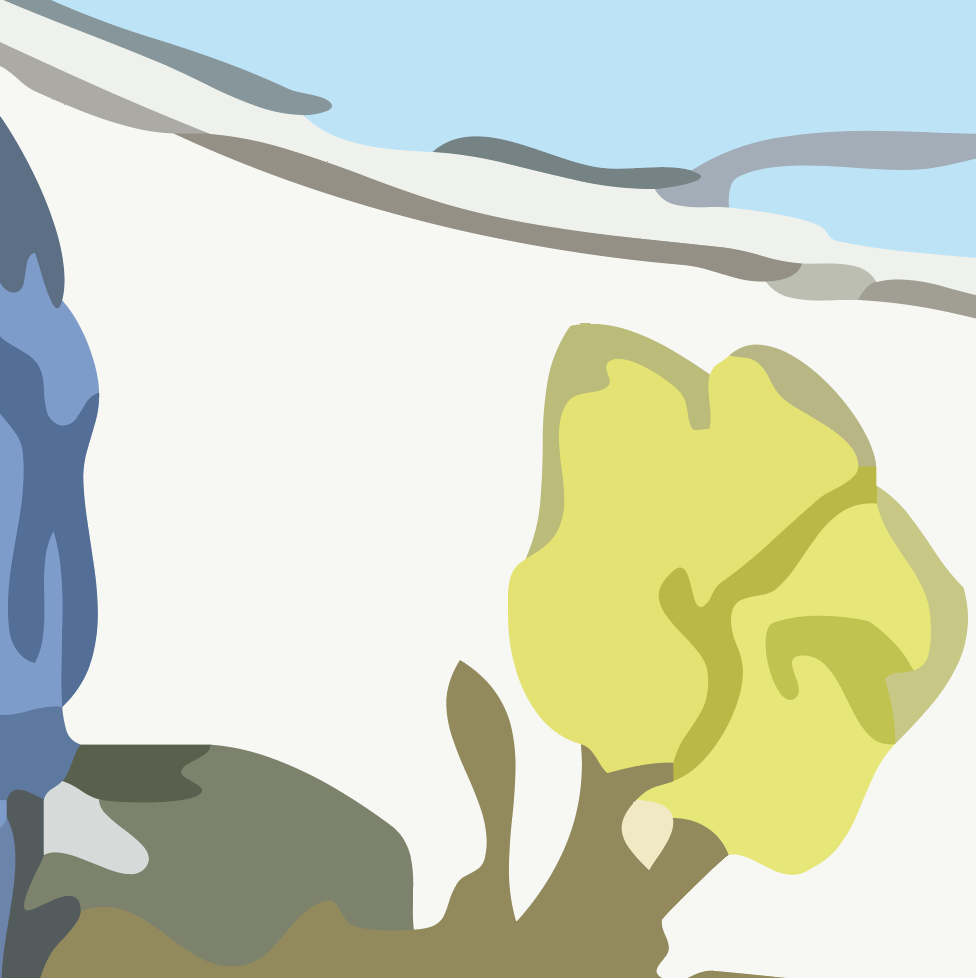}&
\includegraphics[width=\mysize\textwidth]{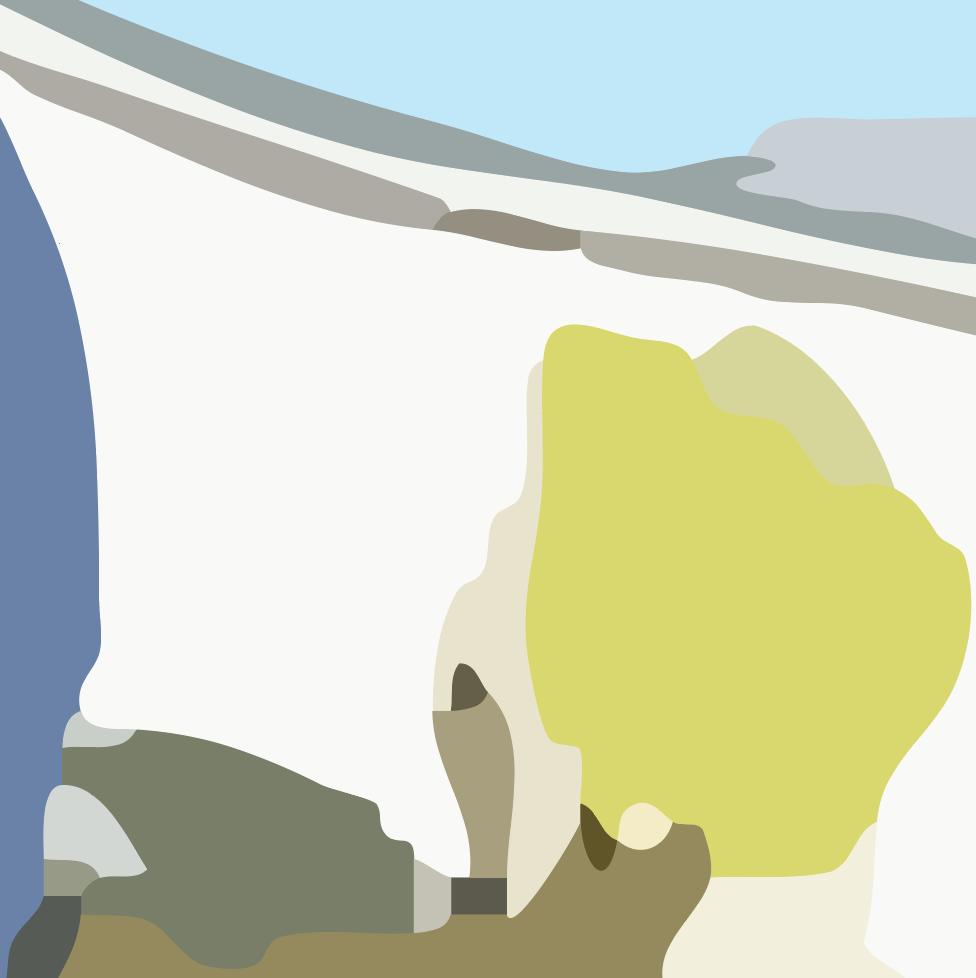}\\
\end{tabular}
\caption{Effects of different merging gains when applied to a painting by Paul Jacoulet. (a) Input image with red boxes for zooming. (b) Zoomed input. The right panel shows zoomed results ($N^*=500$) by (c) Area~\eqref{eq_area_gain}, (d) BG~\eqref{eq_BG_gain}, (e) Scale~\eqref{eq_scale_gain}, and (f) MS~\eqref{eq_MS_gain}, respectively. MS eliminates shapes with elongated contours and yields oversimplified representations. BG retains image details solely based on their contrasts and can violate the geometry of the objects. Scale keeps more details  compared to MS, but still omits thin shapes.  Area preserves most critical features while effectively avoiding small blobs near boundaries due to antialiasing. }\label{fig_gain_compare1}
\end{figure}

\begin{figure}
\centering
\begin{tabular}{c@{\hspace{2pt}}c@{\hspace{2pt}}c@{\hspace{2pt}}c}
\hline
\multicolumn{4}{c}{Input}\\
\includegraphics[width=0.38\textwidth]{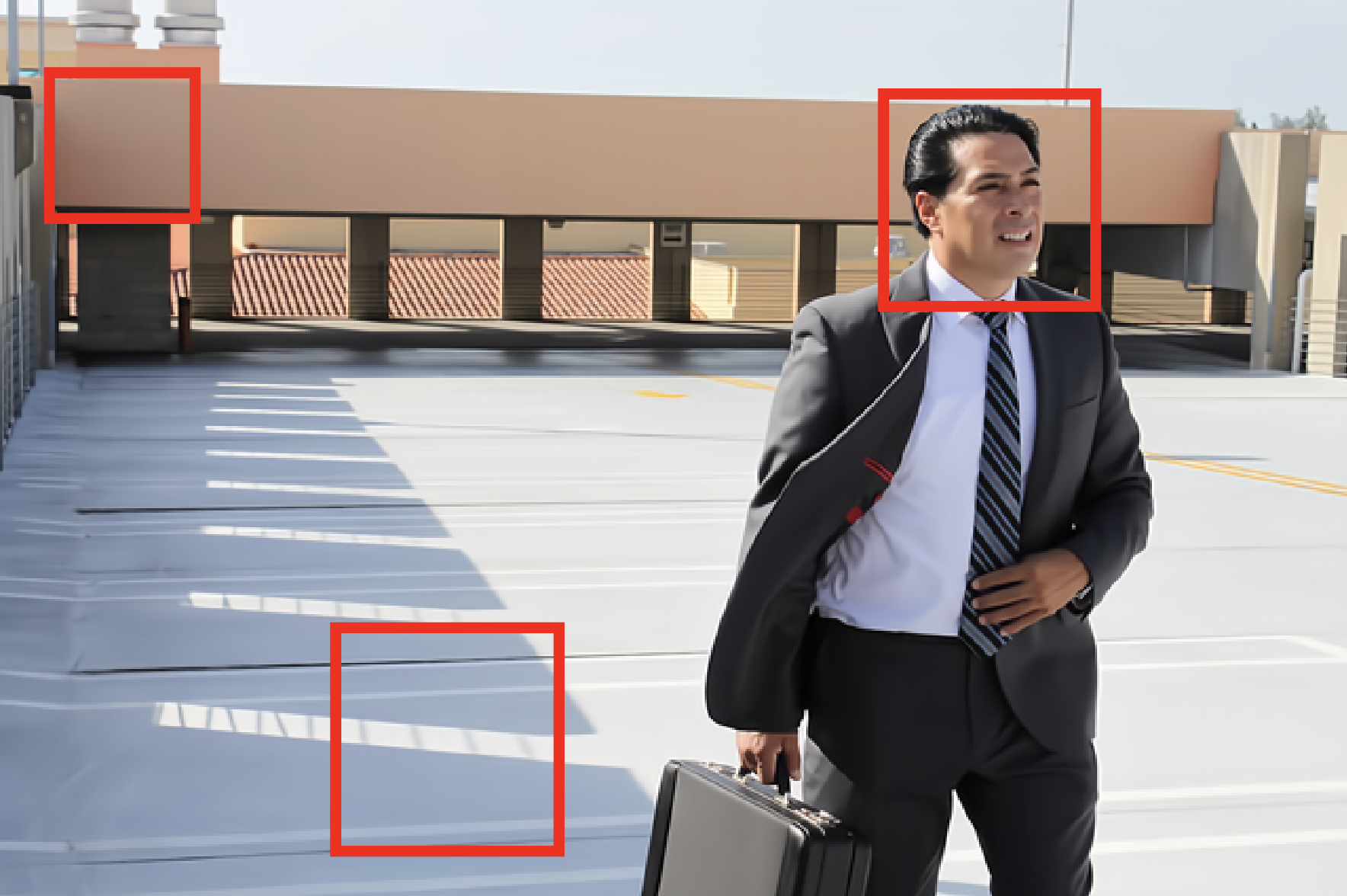}&
\raisebox{0.55cm}{\includegraphics[width=0.19\textwidth]{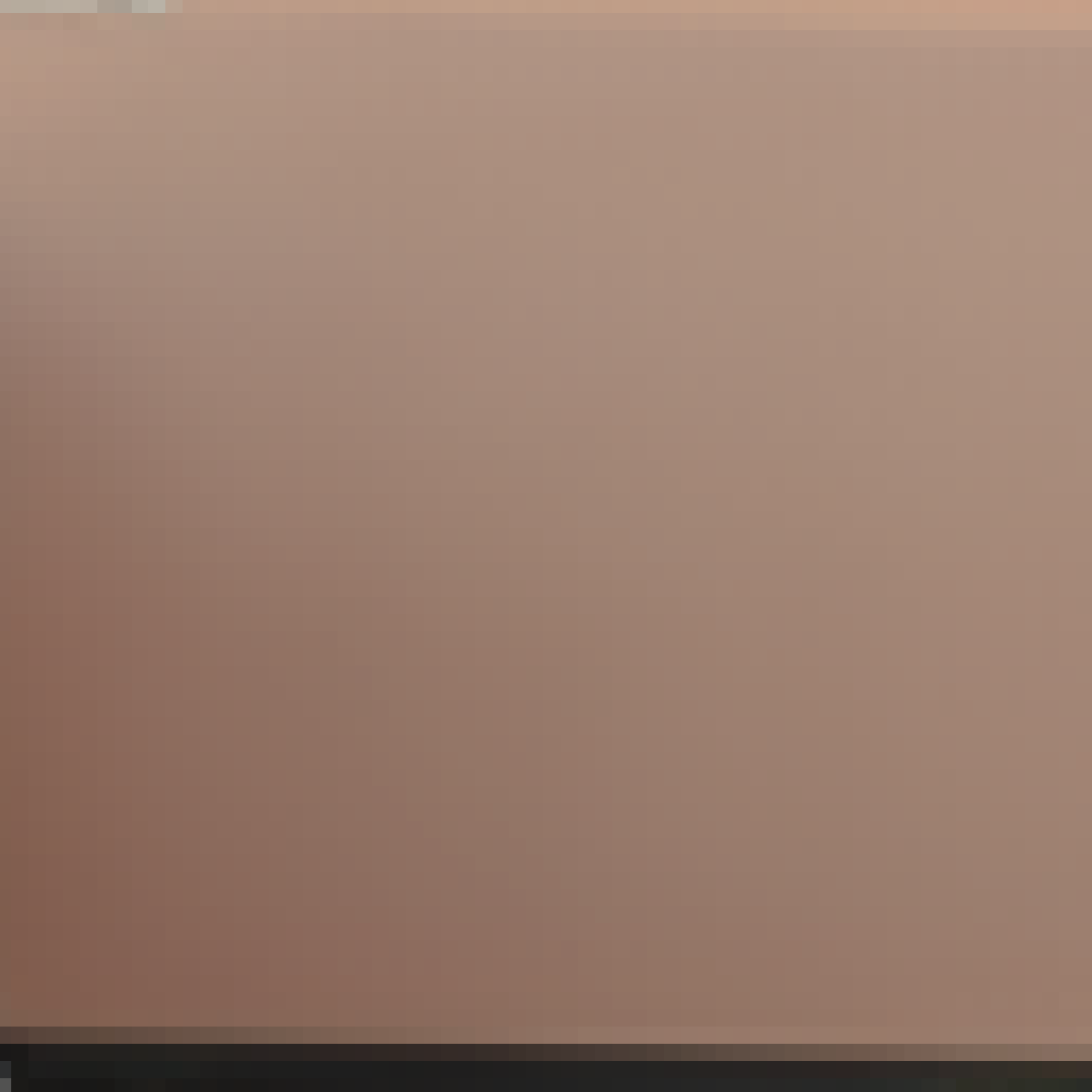}}&
\raisebox{0.55cm}{\includegraphics[width=0.19\textwidth]{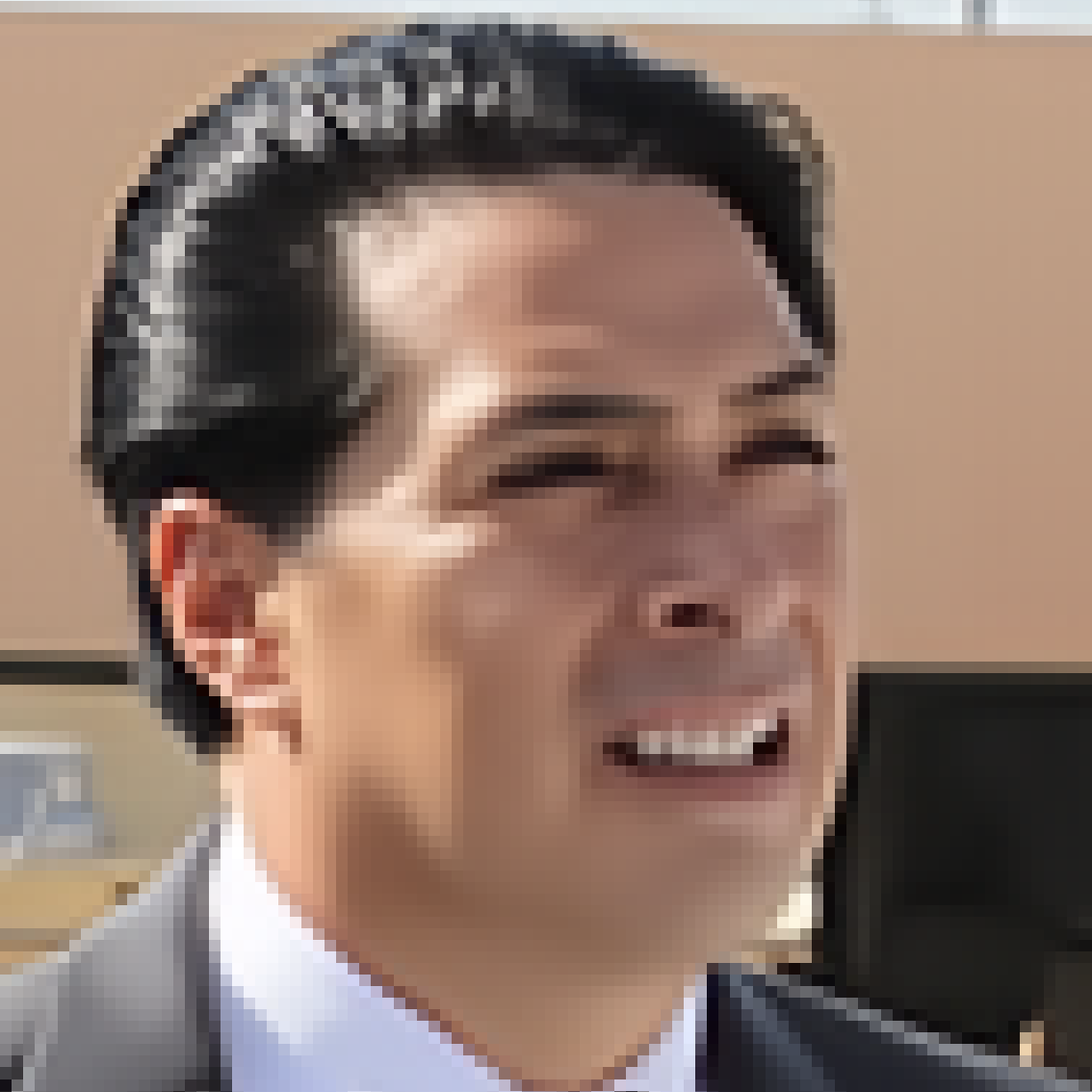}}&
\raisebox{0.55cm}{\includegraphics[width=0.191\textwidth]{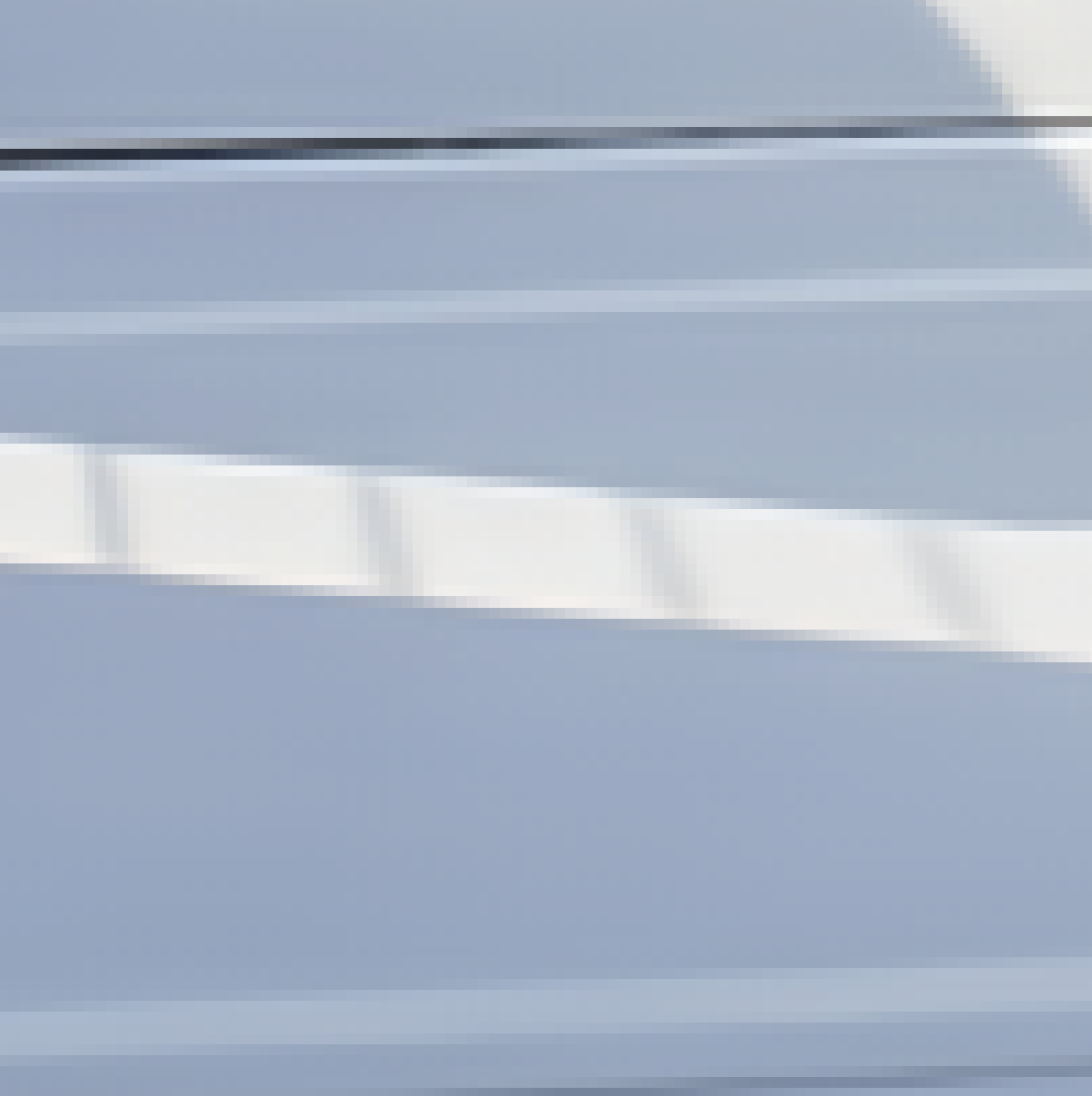}}\\\hline
\end{tabular}
\begin{tabular}{cccc}
\multicolumn{4}{c}{Vectorized results}\\
(a)&(b)&(c)&(d)\\
\includegraphics[width=0.19\textwidth]{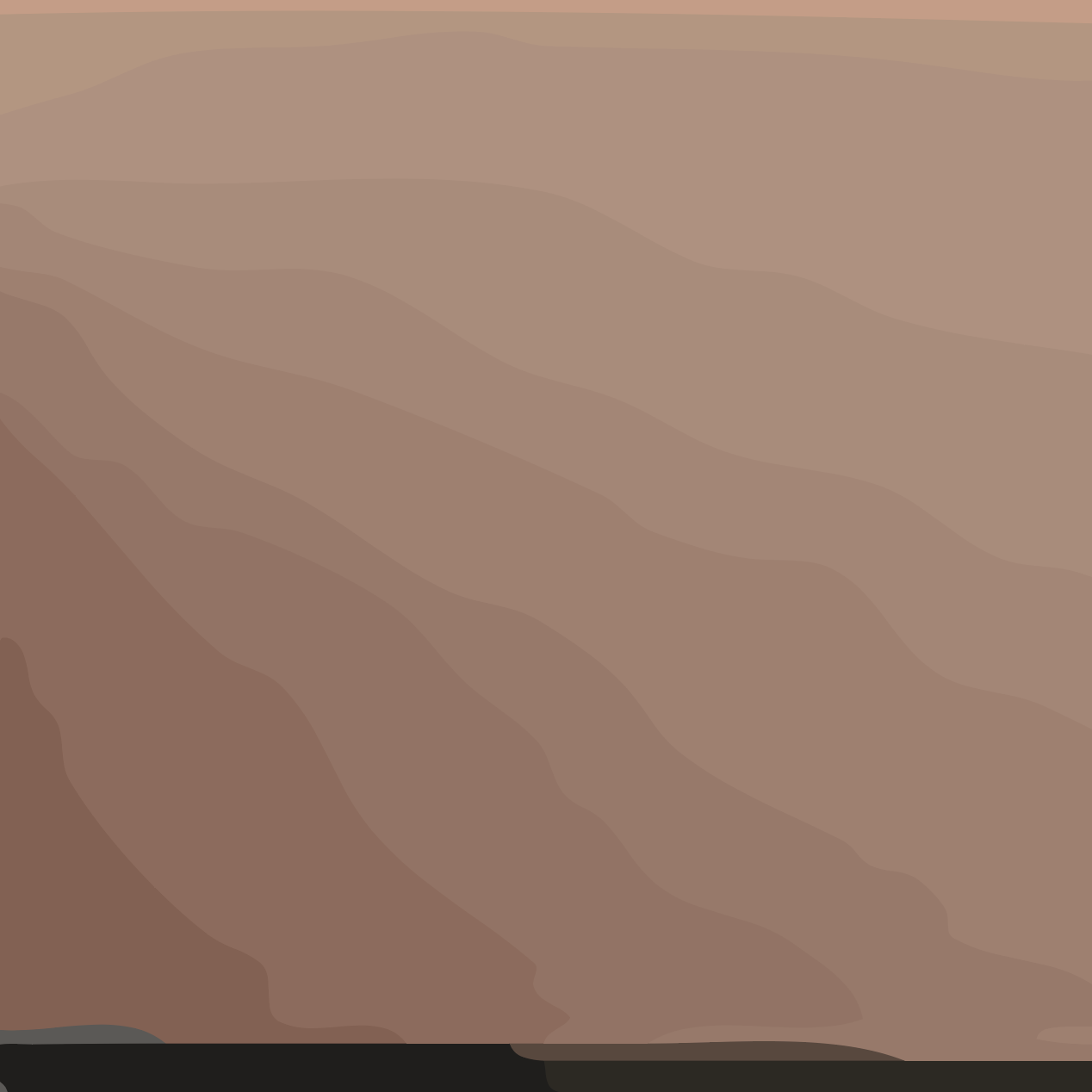}&
\includegraphics[width=0.19\textwidth]{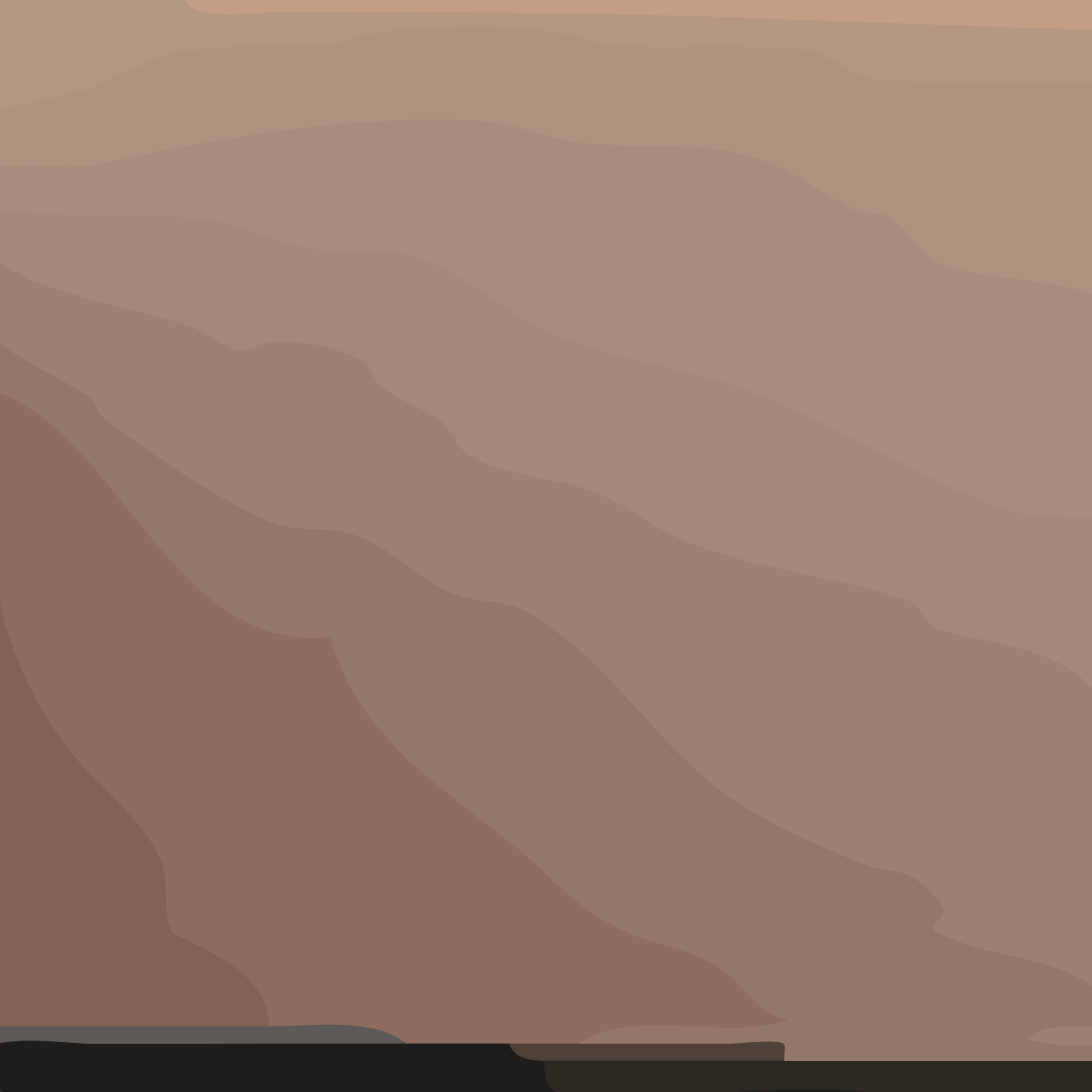}&
\includegraphics[width=0.19\textwidth]{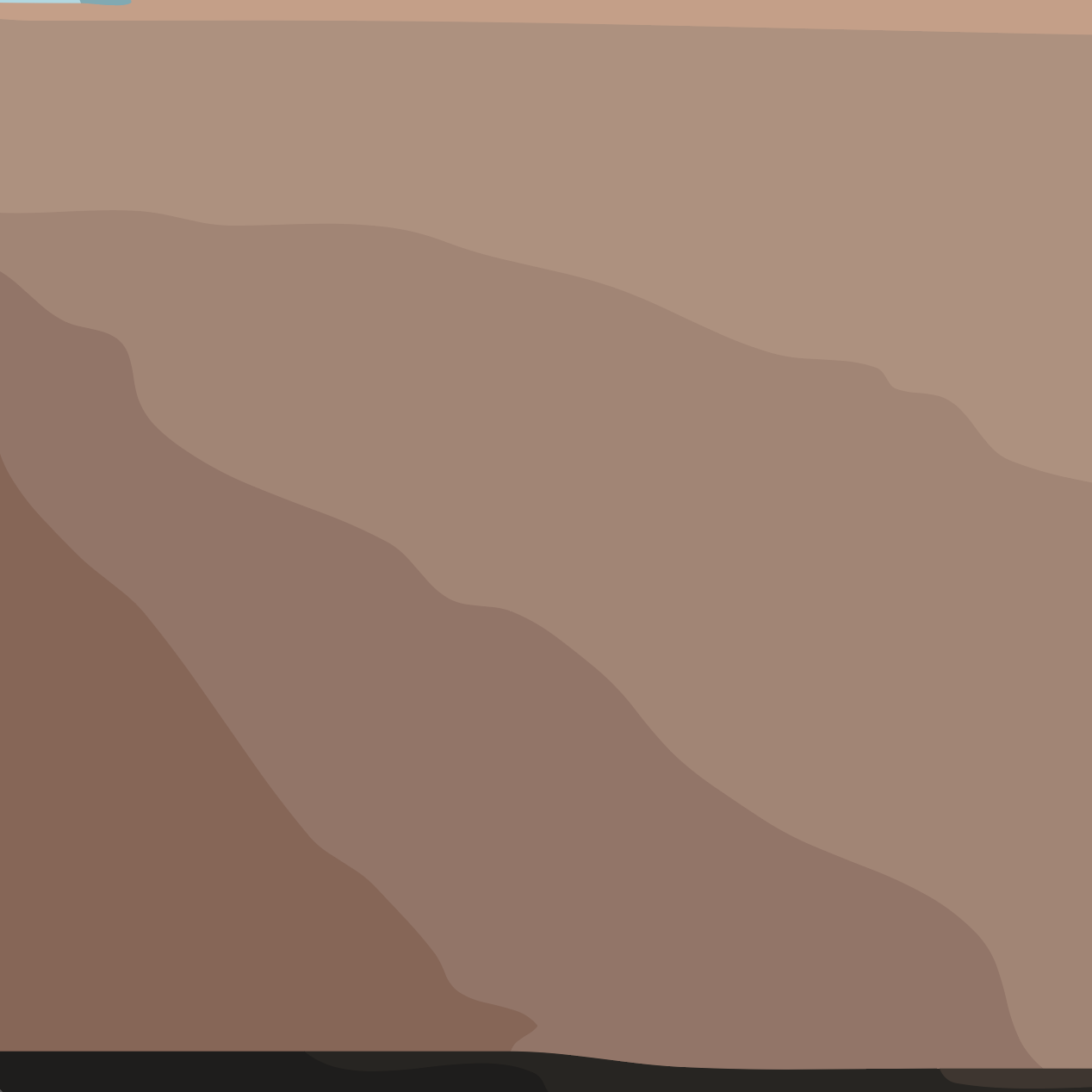}&
\includegraphics[width=0.19\textwidth]{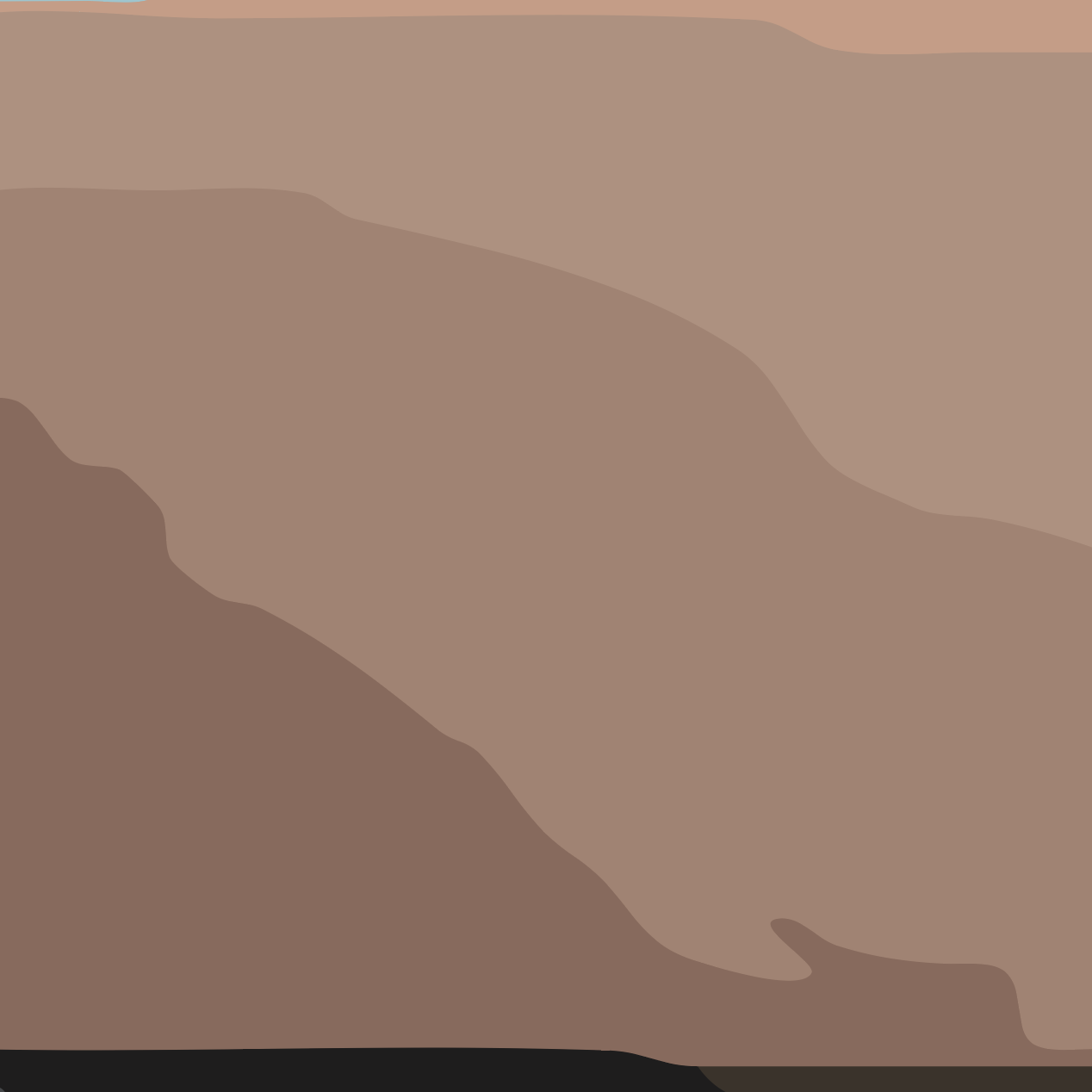}\\
\includegraphics[width=0.19\textwidth]{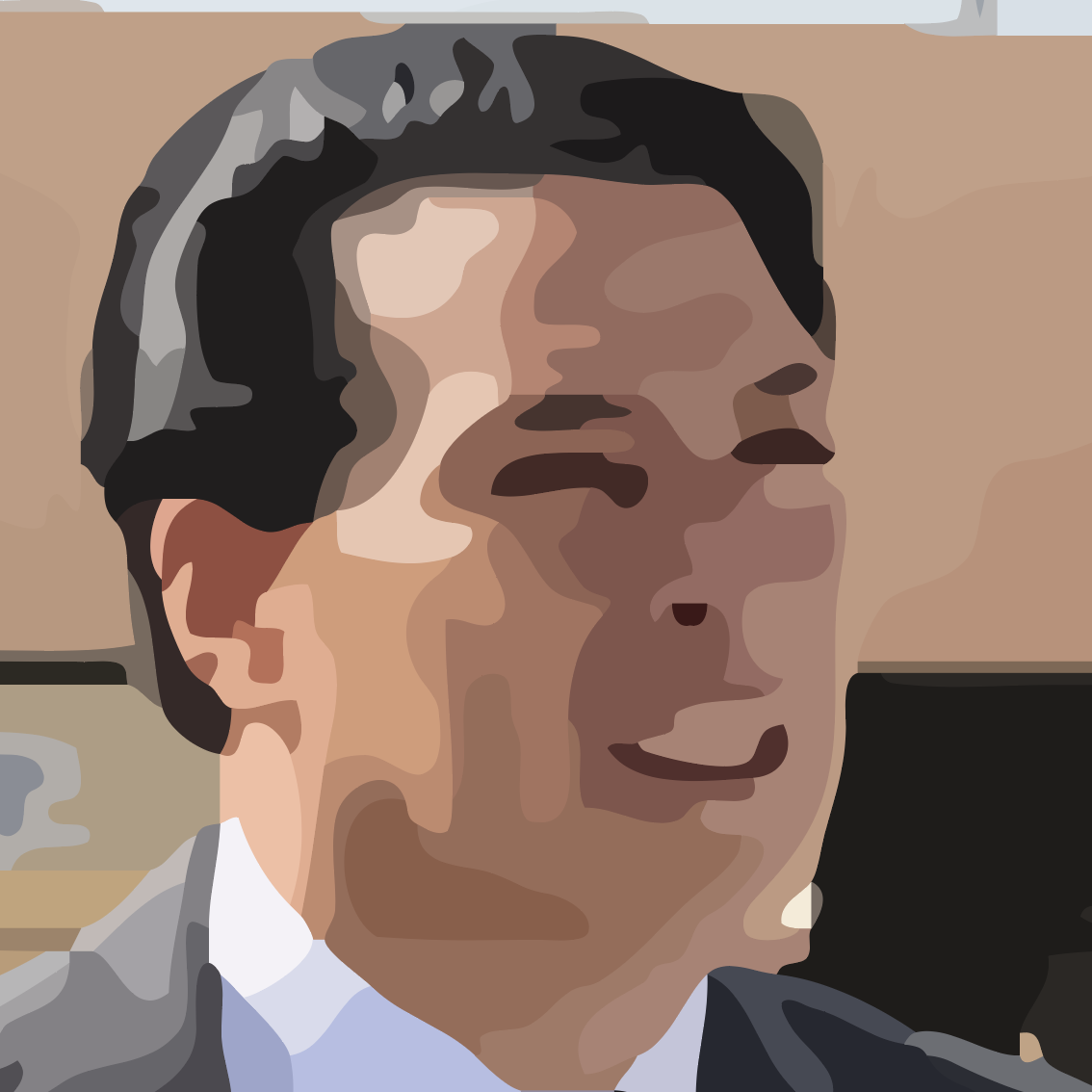}&
\includegraphics[width=0.19\textwidth]{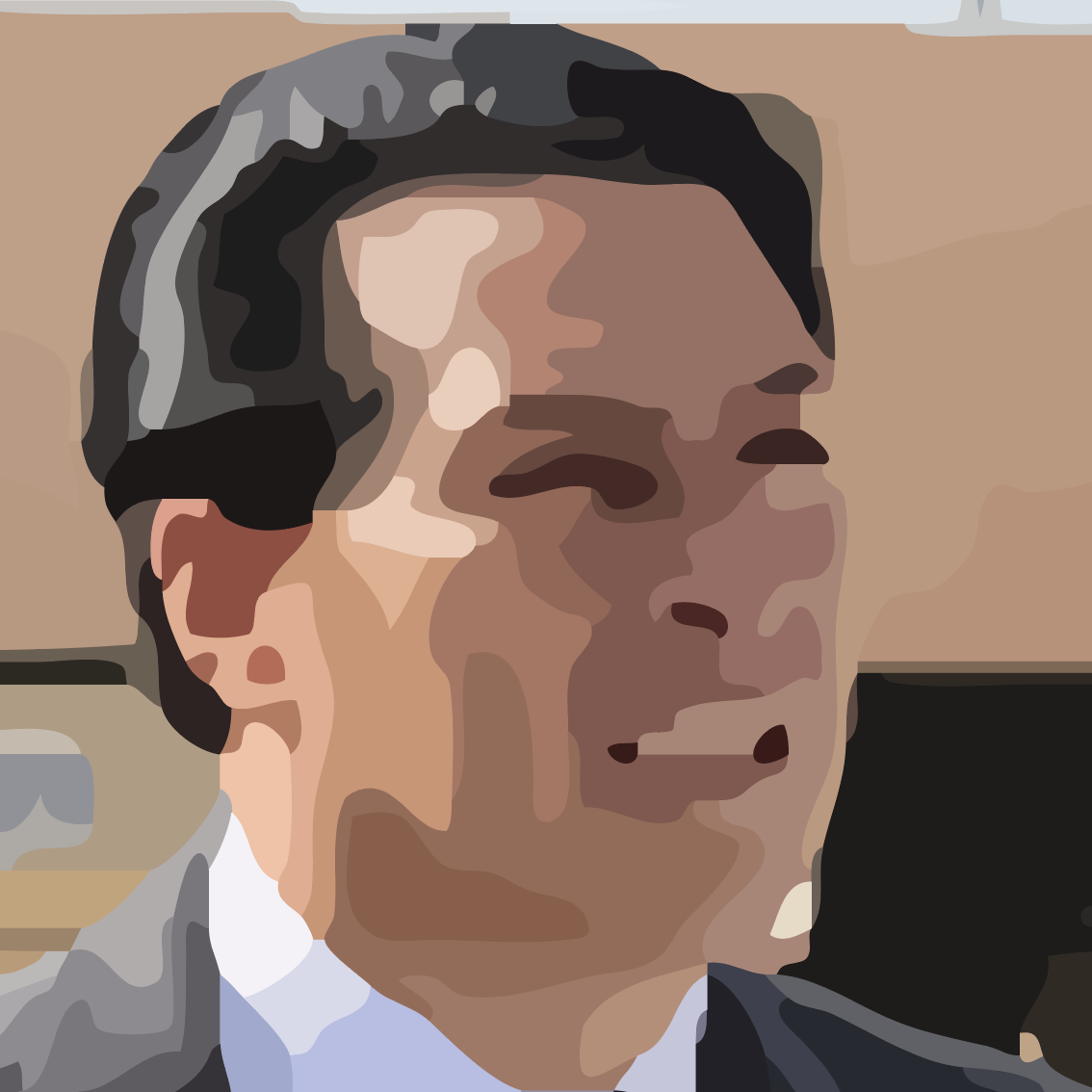}&
\includegraphics[width=0.19\textwidth]{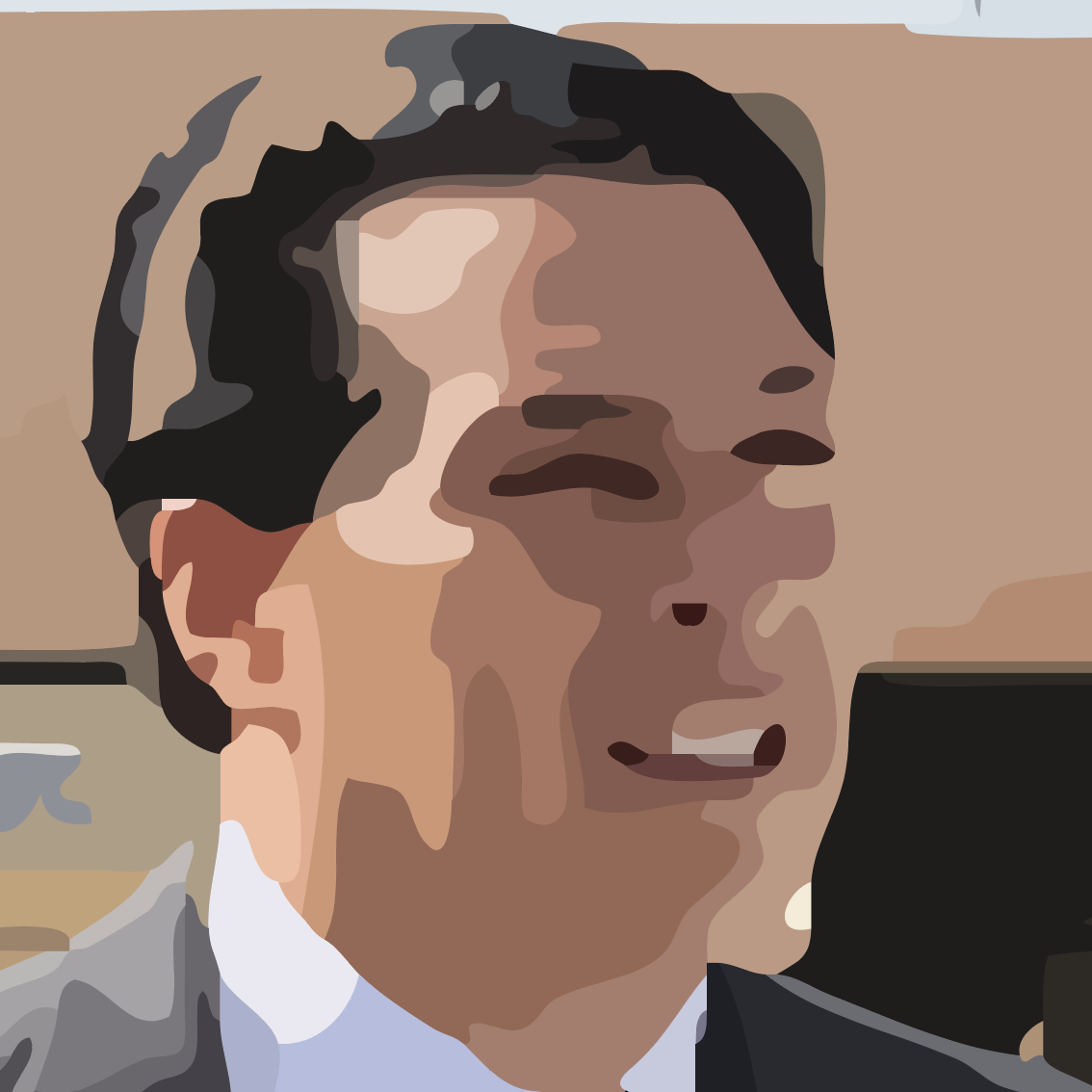}&
\includegraphics[width=0.19\textwidth]{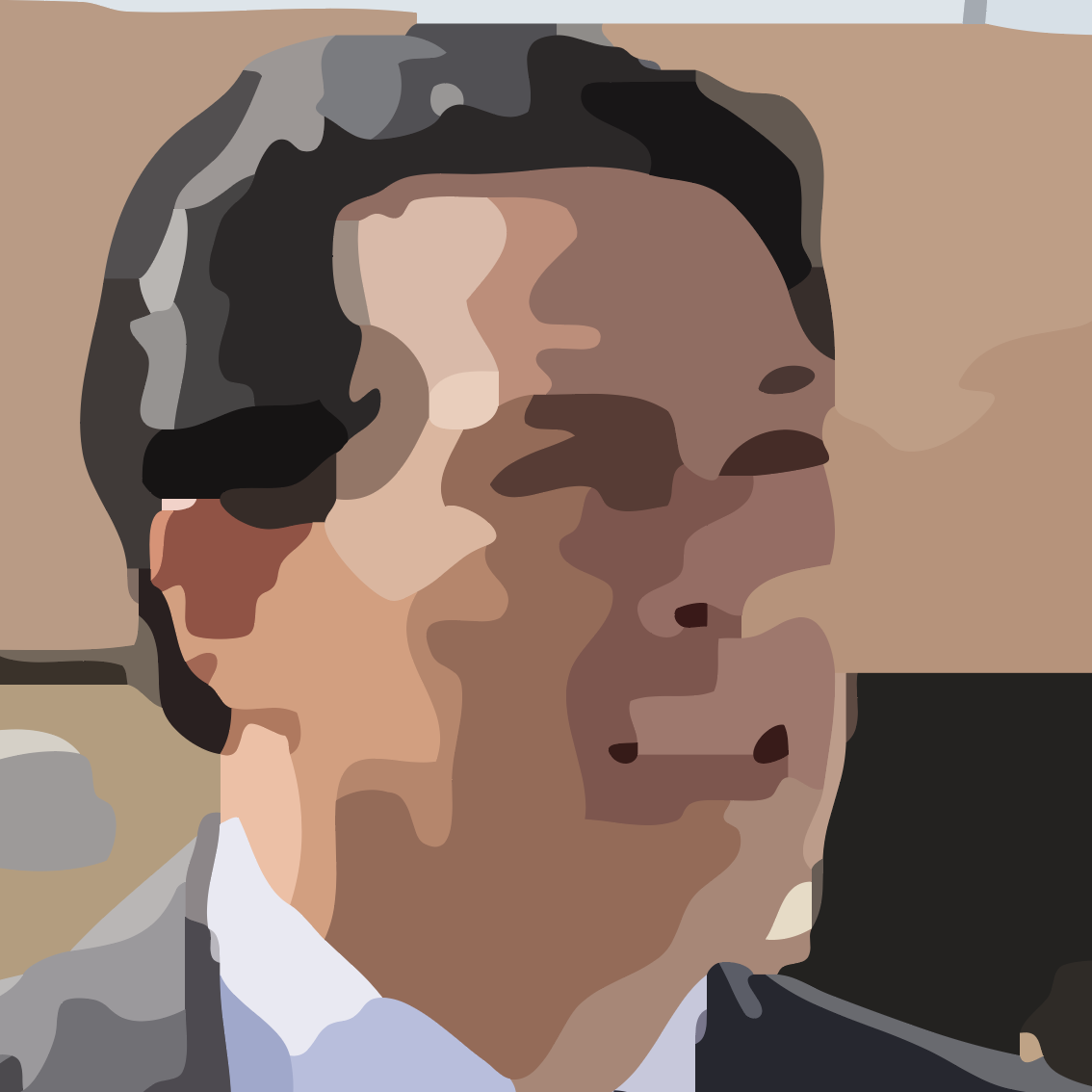}\\
\includegraphics[width=0.19\textwidth]{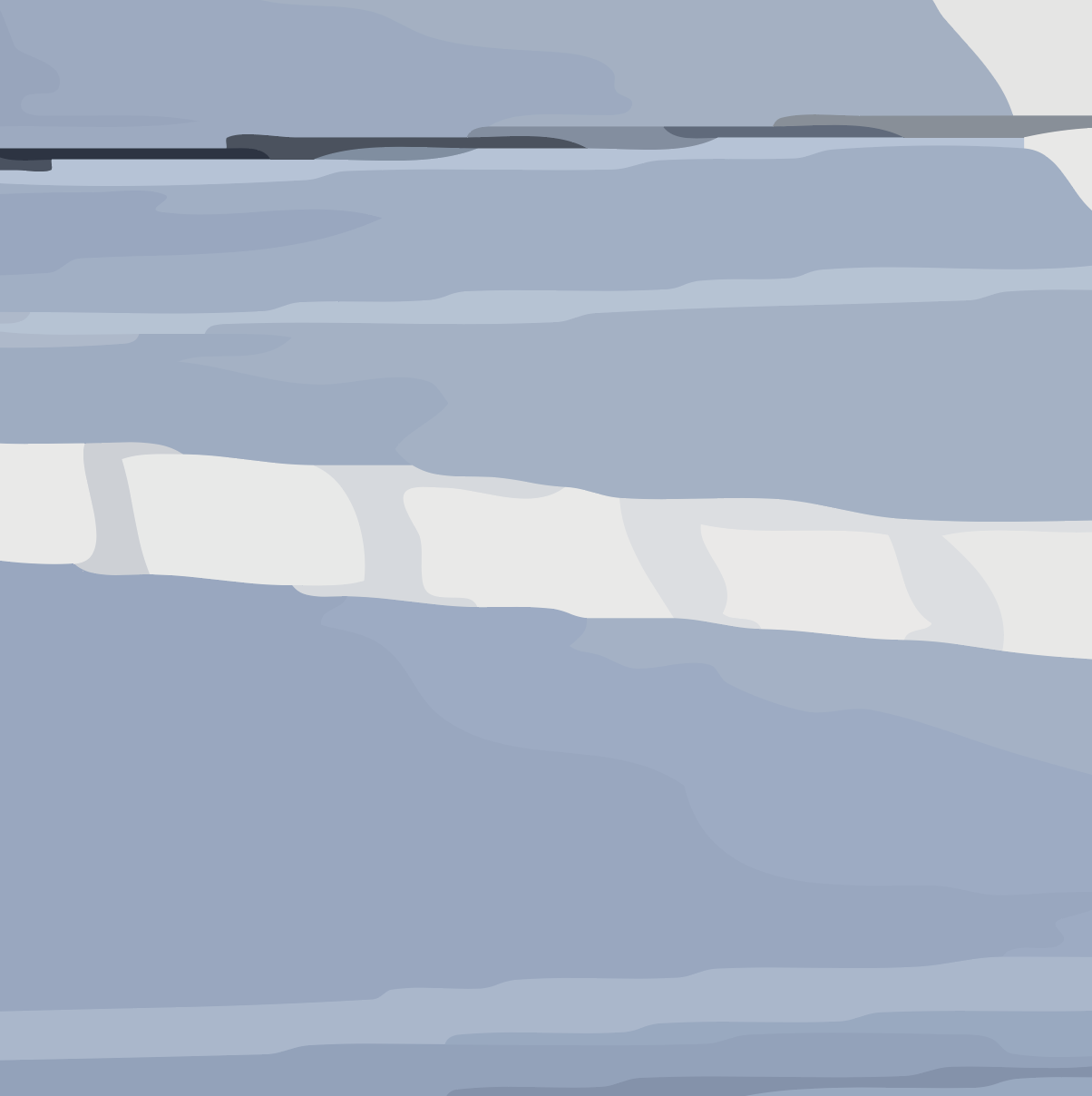}&
\includegraphics[width=0.19\textwidth]{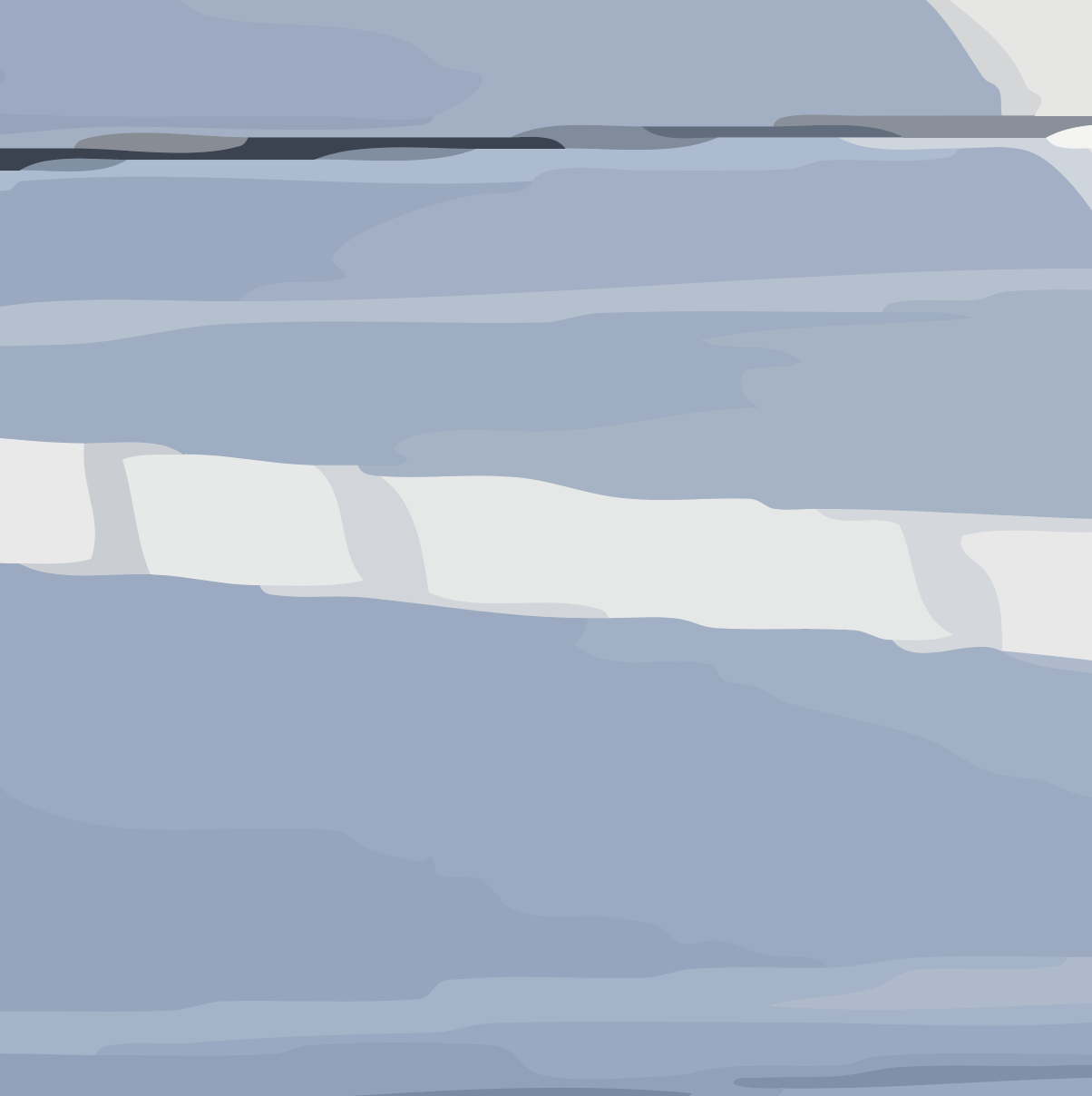}&
\includegraphics[width=0.19\textwidth]{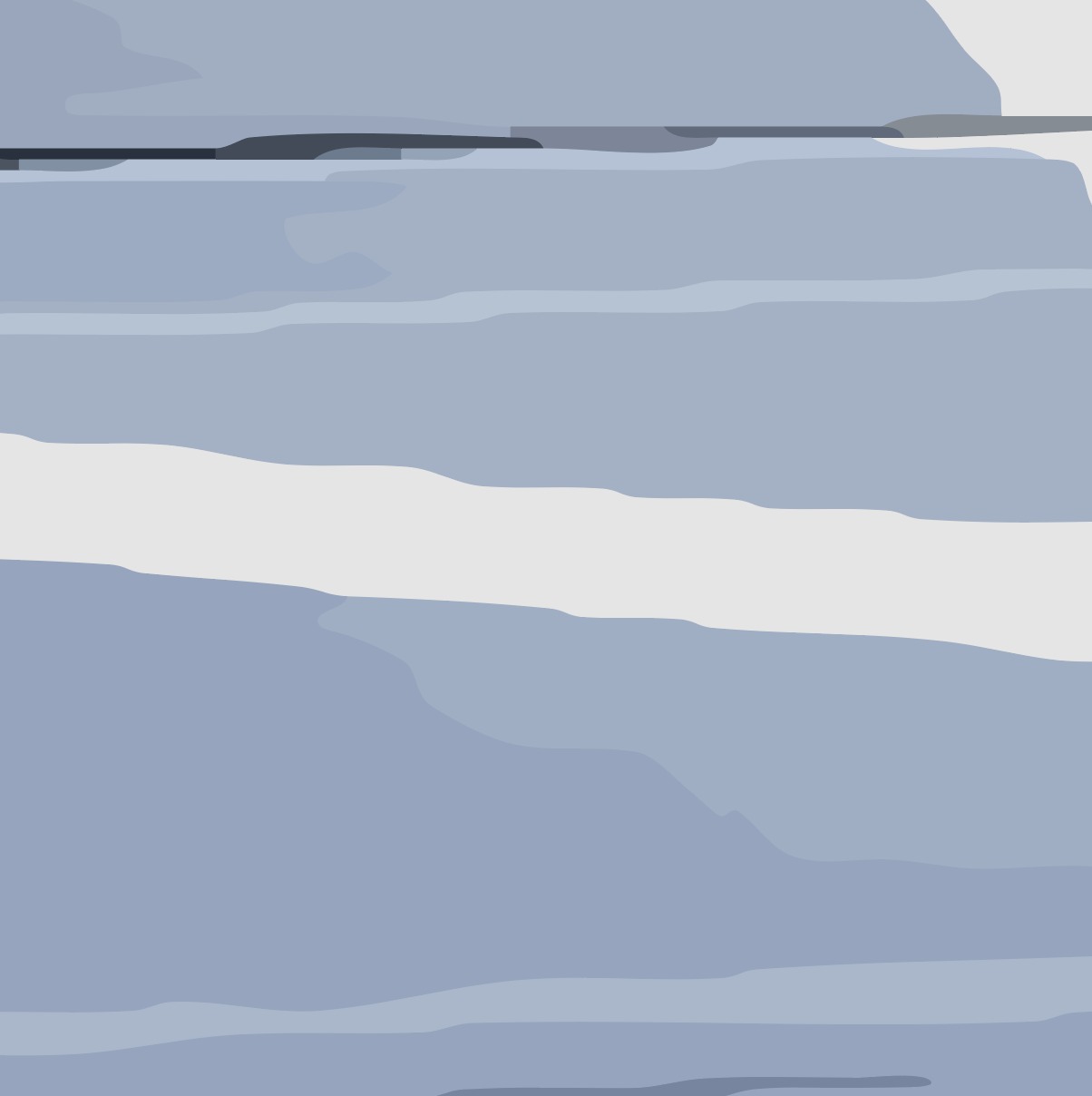}&
\includegraphics[width=0.19\textwidth]{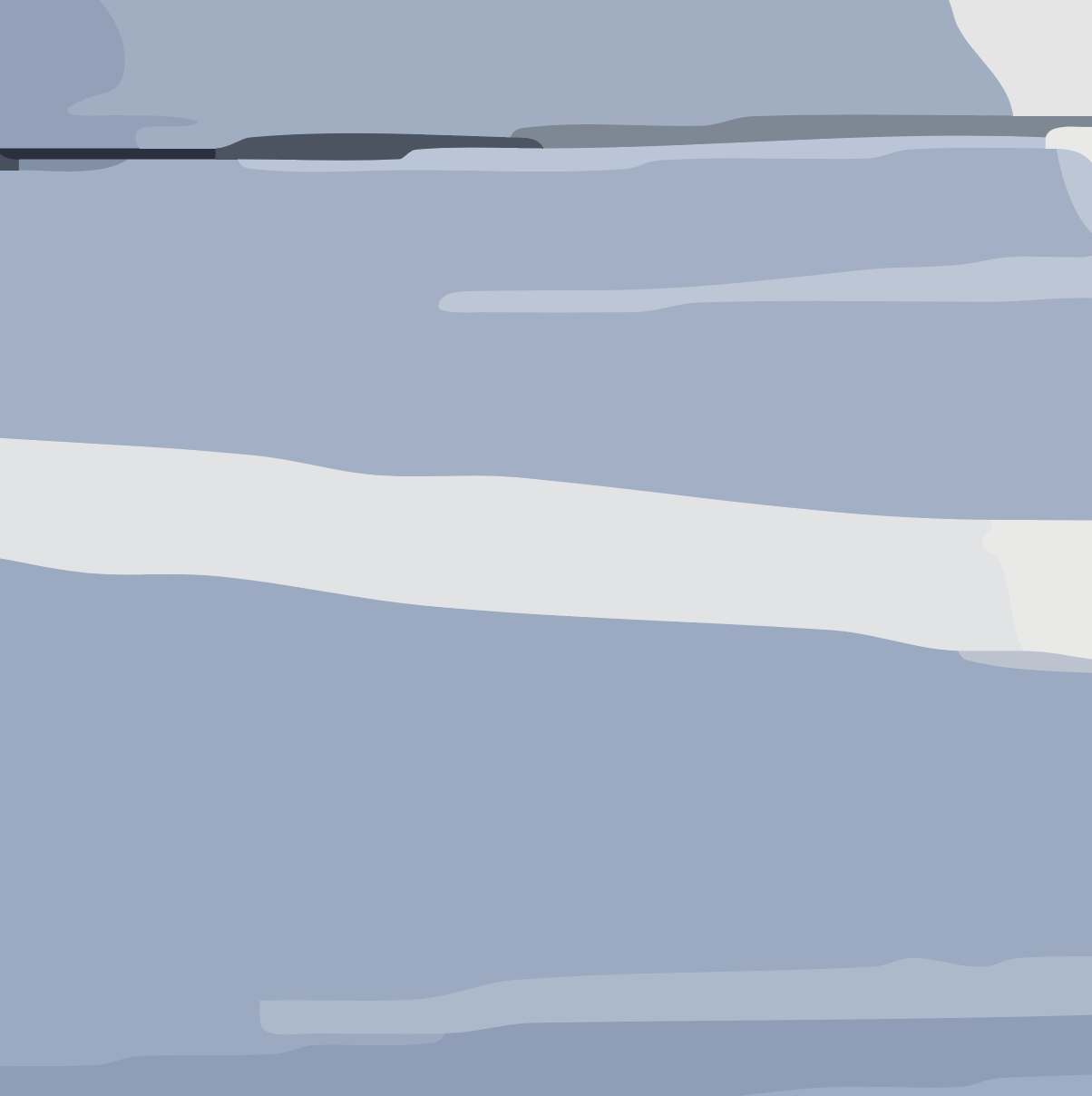}
\end{tabular}
\caption{Effects of merging gains when applied to a  picture. The first row shows the input and the zoom-ins in the red boxes. In the panel below, we show the zoomed-in vectorized results ($N^*=1000$) by (a) Area~\eqref{eq_area_gain}, (b) BG~\eqref{eq_BG_gain}, (c) Scale~\eqref{eq_scale_gain}, and (d) MS~\eqref{eq_MS_gain}. Area accurately approximates the smooth gradient, reconstructs the facial features, and successfully identifies the shadows of the bars.}\label{fig_gain_compare1_3}
\end{figure}

\begin{figure}
\centering
\begin{tabular}{cc}
(a)&(b)\\
\includegraphics[width=0.4\textwidth]{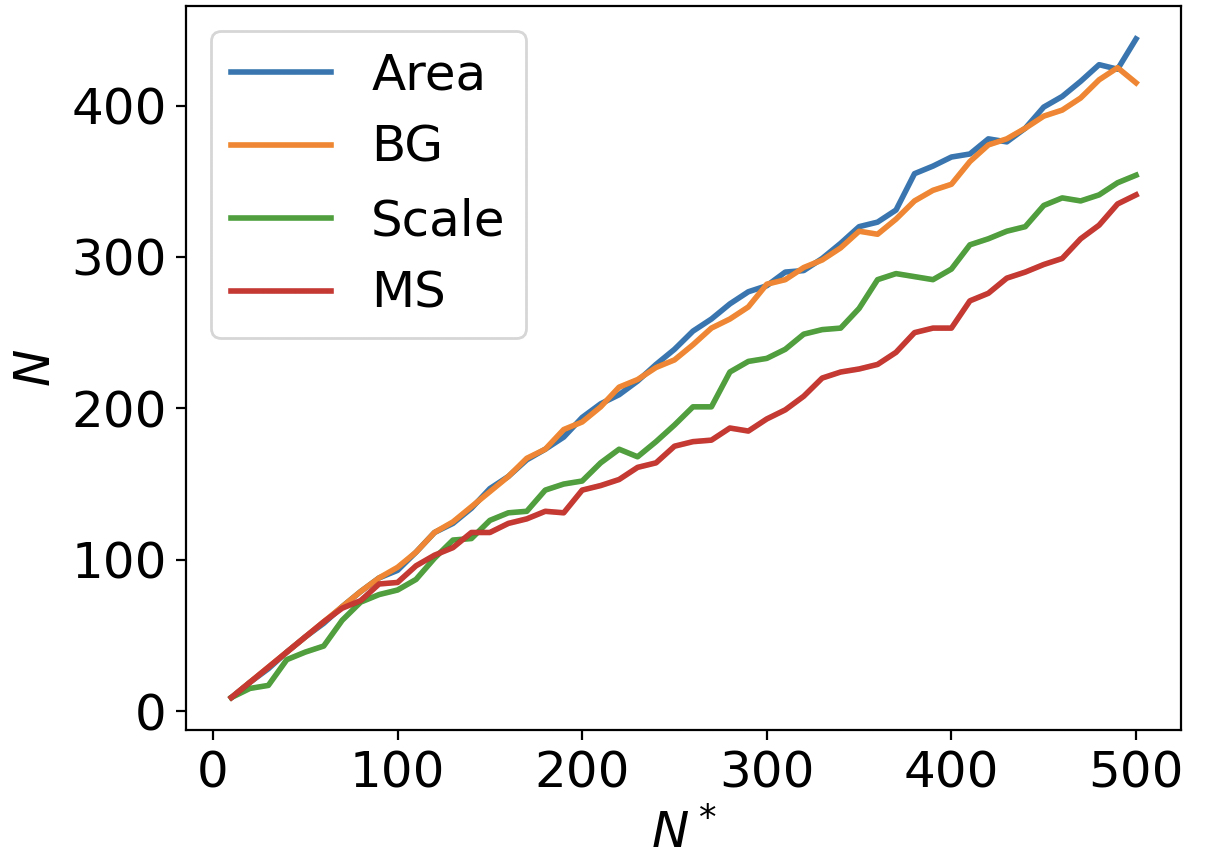}&
\includegraphics[width=0.4\textwidth]{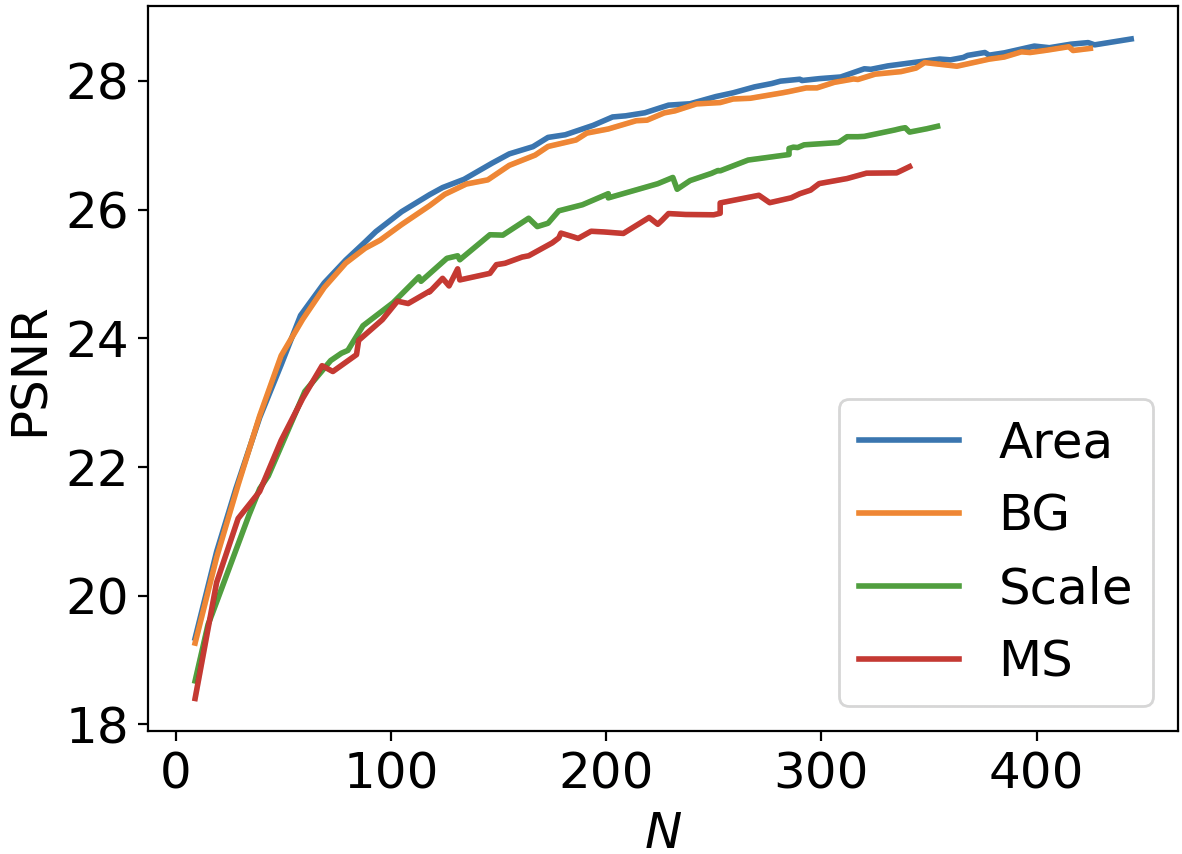}
\end{tabular}
\caption{Quantitative comparison among merging gains. Figure~\ref{fig_gain_compare1} (a) shows the input image. (a) The complexity parameter ($N^*$) versus the number of regions in the resulting vector graphic ($N$). (b) The number of regions ($N$) versus the PSNR between the rasterized vector graphic and the input image. Both Area and BG can yield more regions than Scale or MS when the same $N^*$ is used. Meanwhile, with the same number of regions, vectorized results by Area and BG have better quality than Scale and MS, thus achieving higher efficiency in the representations.  }\label{fig_gain_compare2}
\end{figure}

In this section, we compare the behaviors of the proposed general scheme using different merging gains: MS region merging~\eqref{eq_MS_gain}, BG region merging~\eqref{eq_BG_gain}, Scale region merging~\eqref{eq_scale_gain}, and Area region merging~\eqref{eq_area_gain}. 

In Figure \ref{fig_gain_compare1} (a), we show the rasterized \textit{La Geisha Kiyoka} by Paul Jacoulet (1953) with size $350\times 450$. The regions in the red boxes are zoomed in the column (b), where boundaries show severe pixelation effects. Setting $N^*=500$, we show the zoomed vector graphics obtained by Area, BG, Scale, and MS in column (c)-(f), respectively.  Observe that both Area and BG preserve more important image features compared to Scale and BG. Strongly influenced by the length minimization, MS tends to overlook shapes with elongated contours, such as the strokes delineating the nose and the flower. Additionally, it does not effectively remove small artifacts along objects' boundaries due to antialiasing.   While Scale successfully eliminates these boundary artifacts, it does not manage to recover intricate details. In contrast, BG retains a greater amount of image detail but struggles to produce clean boundaries.  In this example, Area excels at both preserving the image features and removing the blobs along the boundaries, thus yielding the most faithful and efficient representation.

We also test our merging methods on photos. In Figure~\ref{fig_gain_compare1_3}, we focus on comparing different merging methods in three particular regions, and the results are listed in columns (a)-(d) for Area, BG, Scale, and MS. The first region contains smooth transition. Both Scale and MS exhibits strong contouring artifacts; BG shows less severe staircasing; and Area yields the most visually satisfying reconstruction. The second region contains a face, where Area and BG perform similarly while Scale and MS ignore many critical features. The third region contains the shadow of bars with relatively low contrast. Scale and MS fail to keep these recognizable patterns; BG finds it hard to retain all the bars; and Area successfully preserves such details with the highest accuracy.

As explained in Section~\ref{sec_terminating}, the resulting vector graphics may have fewer than $N^*$ regions. In the example shown in Figure~\ref{fig_gain_compare1}, there are 444 regions by Area, 415 regions by BG, 356 regions  by Scale, and 331 regions by MS, respectively. By plotting the numbers of regions ($N$) corresponding to varying levels of complexity ($N^*$) in Figure~\ref{fig_gain_compare2} (a), we see that it is generally the case that MS yields representations with the fewest regions, Scale produces slightly more, whereas both BG and Area render the most regions. However, a smaller $N$ does not necessarily correlate with a less accurate representation. Figure~\ref{fig_gain_compare2} (b) shows the relation between $N$ and the PSNR of the rasterized vector graphic for different merging methods. For all criteria, increasing $N$ generally yields higher PSNR. When $N$ is fixed, we observe that Area and BG always achieve higher accuracy than Scale and MS. To reach the highest PSNR value by MS or Scale ($N\approx 350$), both Area and BG use only $N\approx 170$ regions. The flexibility about the regional contours allows Area and BG quickly adapt to complex shapes in the input image, while the excessive complexity  contributes to a minor increment in the PSNR. We investigate the discrepancy of number of regions in Appendix~\ref{sec_discrepancy} with more quantitative details.

\subsection{Comparison between region merging and color quantization}\label{sec_num_quant}

\begin{figure}
\centering
\begin{tabular}{cc}
(a)&(b)\\
\includegraphics[width=0.4\textwidth]{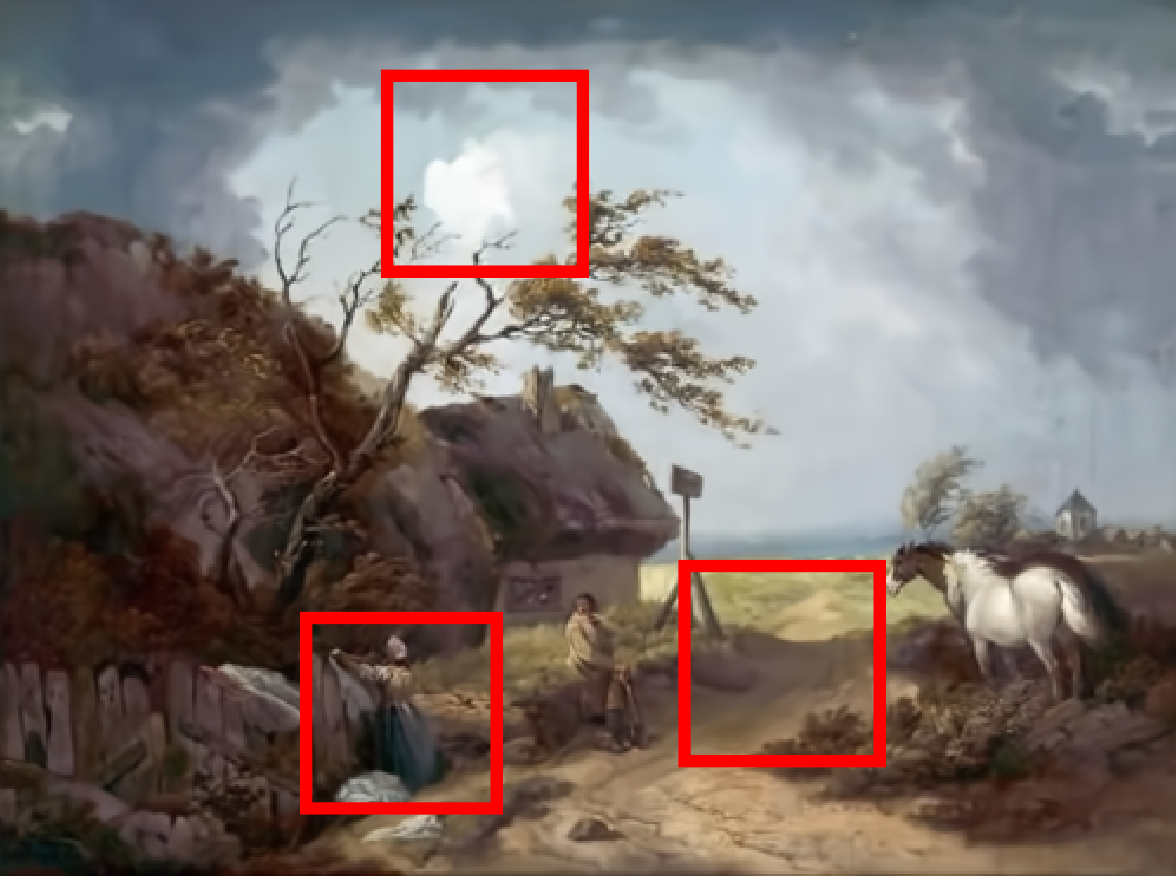}&
\includegraphics[width=0.48\textwidth]{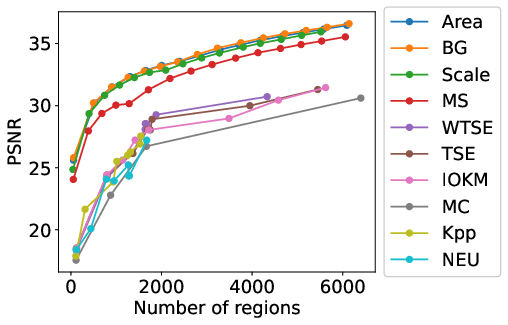}
\end{tabular}
\begin{tabular}{c@{\hspace{2pt}}c@{\hspace{2pt}}c}
(c) Area, $N=519$  & (d) BG, $N=497$ & (e) Scale, $N=743$ \\
\includegraphics[width=0.3\textwidth]{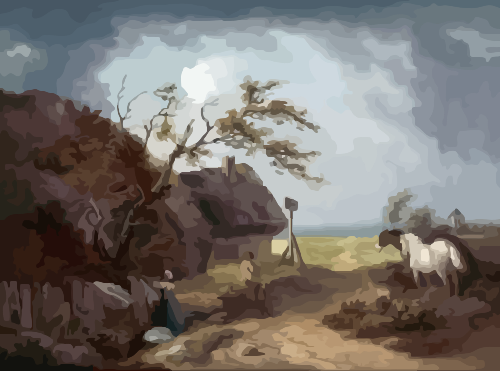}&
\includegraphics[width=0.3\textwidth]{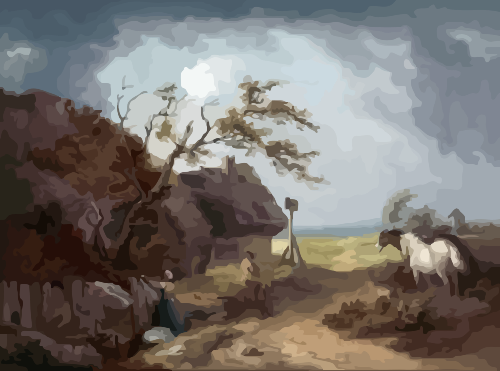}&
\includegraphics[width=0.3\textwidth]{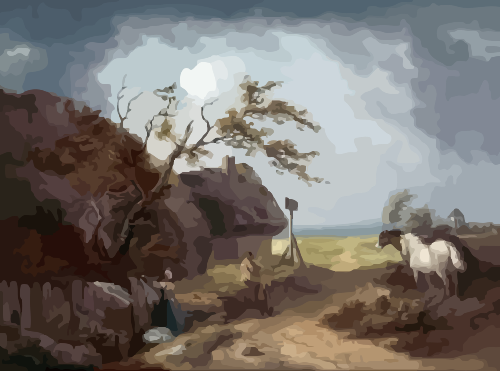}\\
(f) MS, $N=683$ & (g) WTSE, $N=1638$ & (h) IOKM, $N=1417$\\
\includegraphics[width=0.3\textwidth]{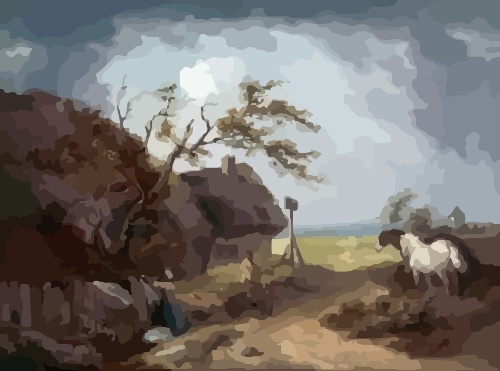}&
\includegraphics[width=0.3\textwidth]{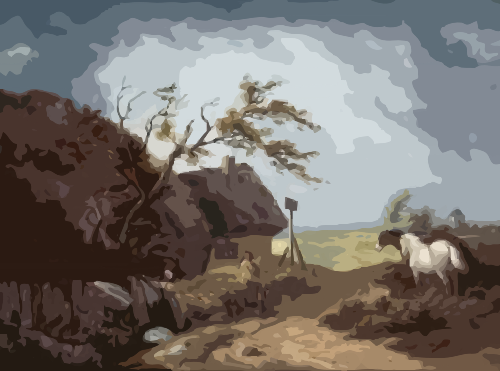}&
\includegraphics[width=0.3\textwidth]{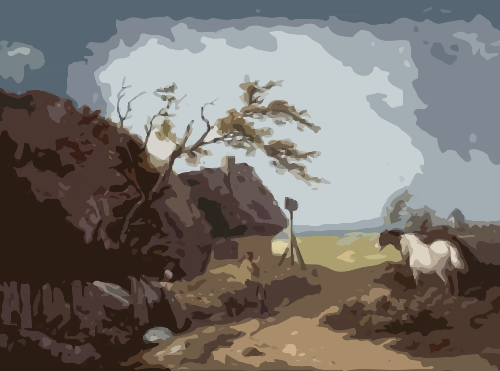}\\
(i) MC, $N=1665$& (j) Kpp, $N=1012$ & (k) NEU, $N=954$\\
\includegraphics[width=0.3\textwidth]{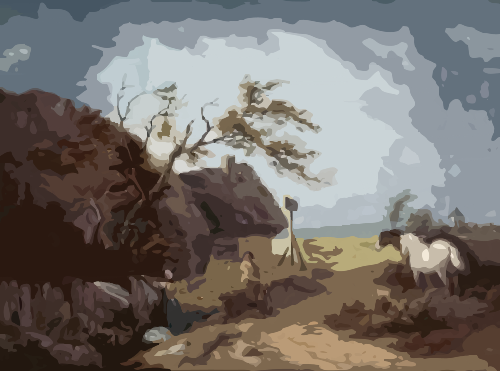}&
\includegraphics[width=0.3\textwidth]{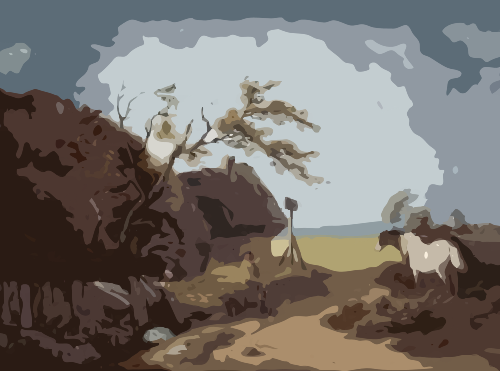}&
\includegraphics[width=0.3\textwidth]{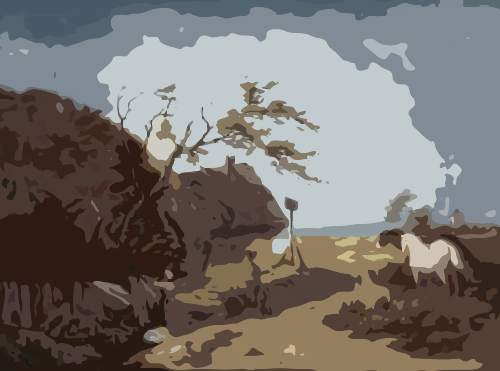}\\
\end{tabular}
\caption{(a) Raster input image. (b) PSNR for all methods with increasing number of regions. (c) Proposed method with Area region merging~\eqref{eq_area_gain}; (d) BG region merging~\eqref{eq_BG_gain}; (e) Scale region merging~\eqref{eq_scale_gain} and (f) MS region merging~\eqref{eq_MS_gain}. Quantization-based vectorization using (g) WTSE~\cite{orchard1991color}; (h) IOKM~\cite{abernathy2022incremental}; (i) MC~\cite{heckbert1982color}; (j) Kpp~\cite{arthur2007k}; and (k) NEU~\cite{dekker1994kohonen}.  Region merging based results are closer  to (a) with smaller number of regions. Zoom-ins of the boxed regions in (a) are compared in Figure~\ref{fig_quant_compare_zoom}.
}\label{fig_quant_compare}
\end{figure}

In this section, we compare the effectiveness of region merging compared with that of color quantization for image vectorization. Figure~\ref{fig_quant_compare} (a) shows an input raster image for this set of experiments. For vectorization based on color quantization, we test multiple benchmark algorithms including Median Cut (MC)~\cite{heckbert1982color}, K means++ (Kpp)~\cite{arthur2007k},  Incremental Online K-Means (IOKM)~\cite{abernathy2022incremental},  Weighted TSE (WTSE)~\cite{orchard1991color}, and quantization based on self-organizing Kohonen neural network (NEU)~\cite{dekker1994kohonen}. These algorithms are efficient and commonly integrated into various software. We refer the readers to~\cite{brun2017color} for a systematic review on related techniques. Using each of these algorithms, we quantize the input with $C$ colors, where $C$ varies from $2$ to $16$. These  results respectively induce image partitions, and we color each connected domain with the mean colors of the pixels therein, leading to raster images with no fewer than $C$ colors. Then we vectorize the resulting   color quantized images  by applying the  same curve smoothing primal step to their discontinuity sets with $T^*=1$, the default smoothness parameter. The compared region merging methods are MS~\eqref{eq_MS_gain}, Scale~\eqref{eq_scale_gain}, and Area~\eqref{eq_area_gain} with  their default parameters. 

Figure~\ref{fig_quant_compare} shows the results. We observe that using similar region numbers, the vectorized results based on region merging are more faithful to the original image than those based on quantization techniques. The results in (c)-(f), produced by Area, BG, Scale, and MS region merging  respectively, preserve both transitional details in relatively smooth regions such as the sky and the ground, as well as structural details in low-contrast areas such as the woman's dress in the bottom left and the grass textures on the roof.  Noticeably,  WTSE in (g) adopts a  merging strategy  in the color space weighted by the image gradient information. This explains why its result is closer to our proposed scheme. Compared to the other quantization methods,  WTSE also shows weaker contouring artifacts. Results in (h)-(k) rely on data clustering techniques applied to color quantization, and they exhibit stronger contouring artifacts in smooth regions. Moreover, they preserve fewer details than the merging methods in (c)-(g).

\begin{figure}
\centering
\begin{tabular}{c@{\hspace{2pt}}c@{\hspace{2pt}}c@{\hspace{2pt}}c@{\hspace{2pt}}c}
(a) Given input & (b) Area & (c) BG & (d) Scale & (e) MS \\
\includegraphics[width=0.18\textwidth]{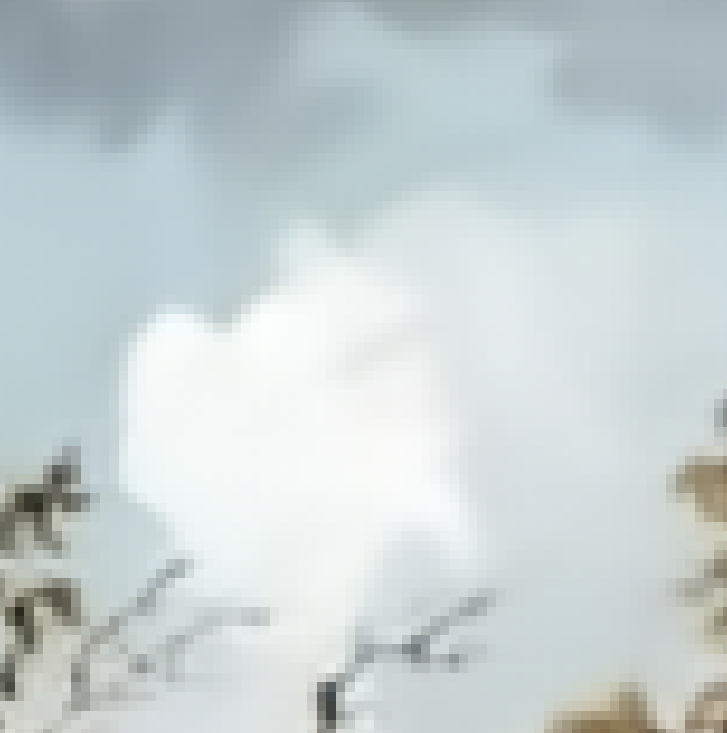}&
\includegraphics[width=0.18\textwidth]{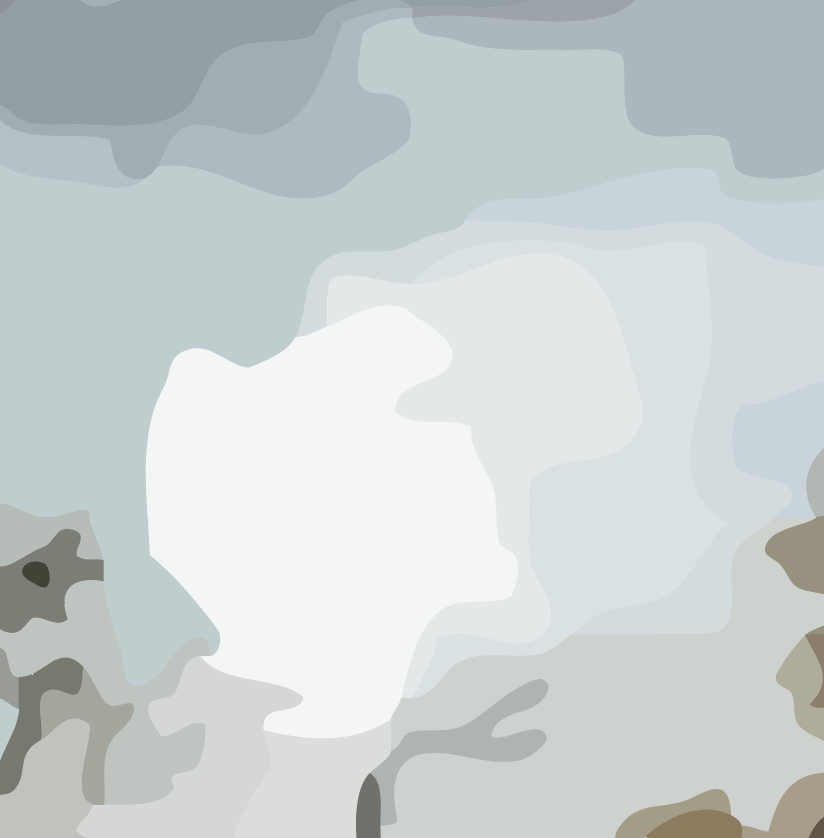}&
\includegraphics[width=0.18\textwidth]{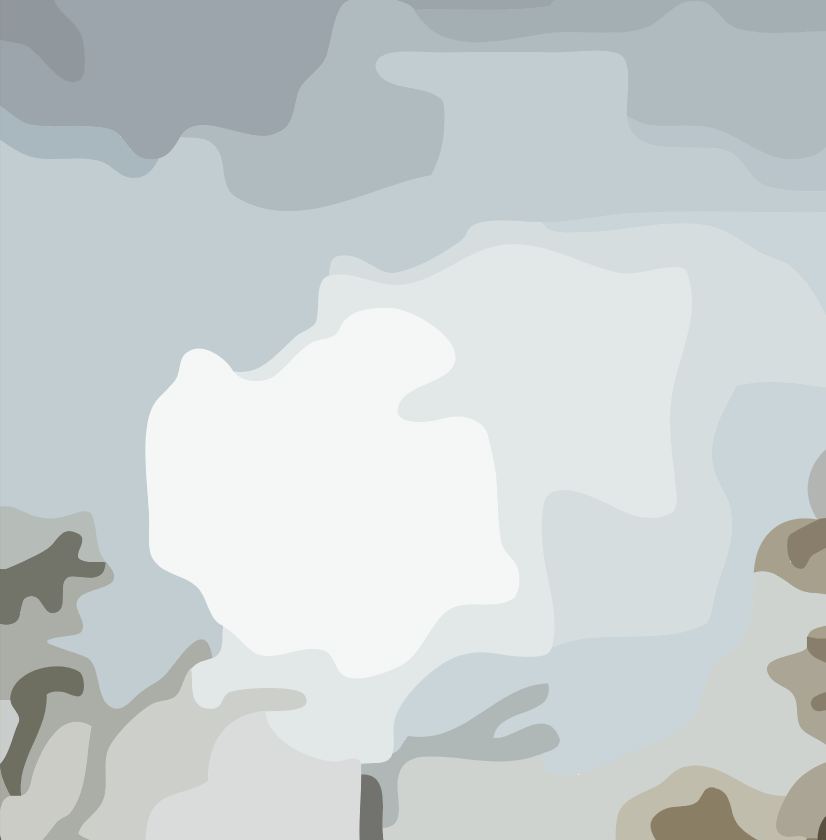}&
\includegraphics[width=0.18\textwidth]{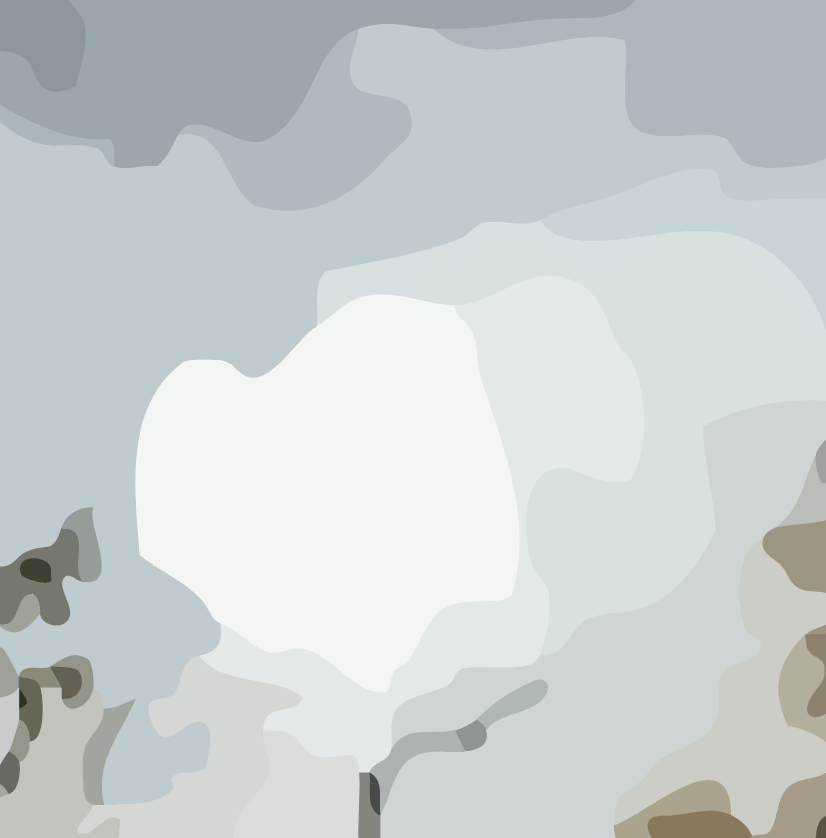}&
\includegraphics[width=0.18\textwidth]{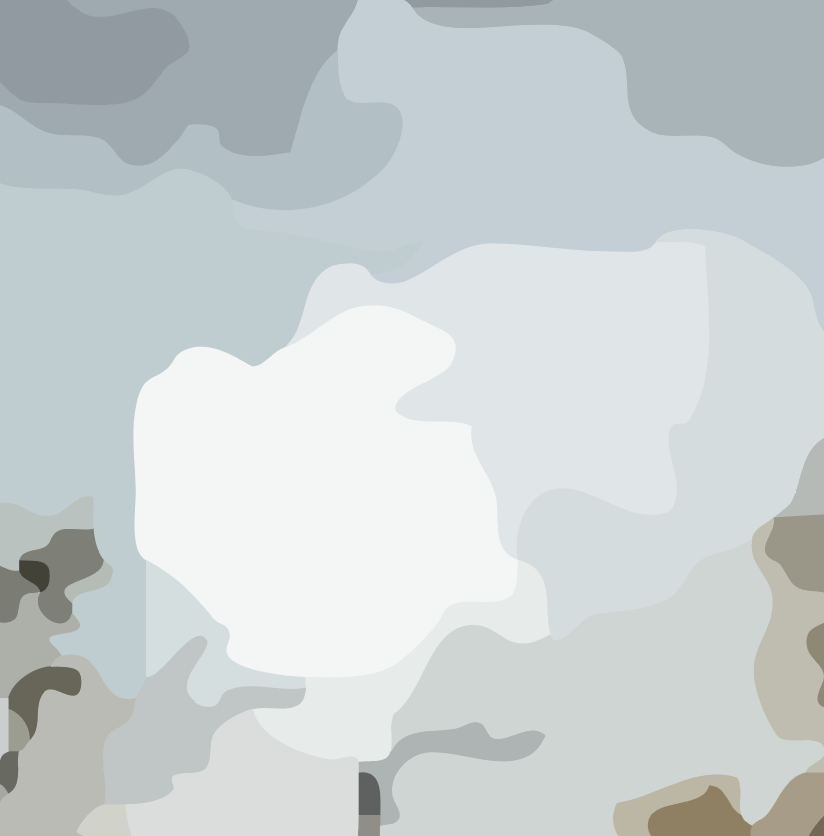}\\
(f) WTSE & (g) IOKM & (h) MC & (i) Kpp & (j) NEU \\
\includegraphics[width=0.18\textwidth]{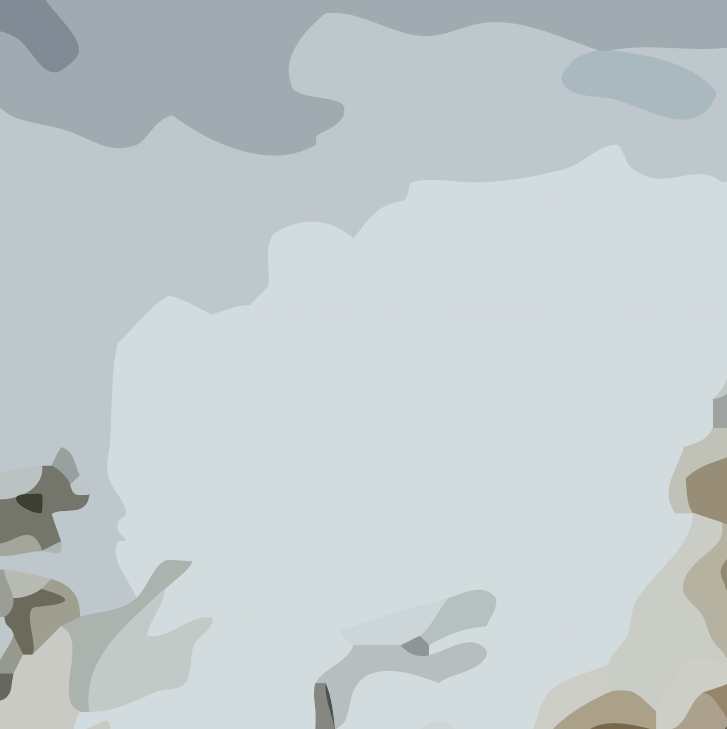}&
\includegraphics[width=0.18\textwidth]{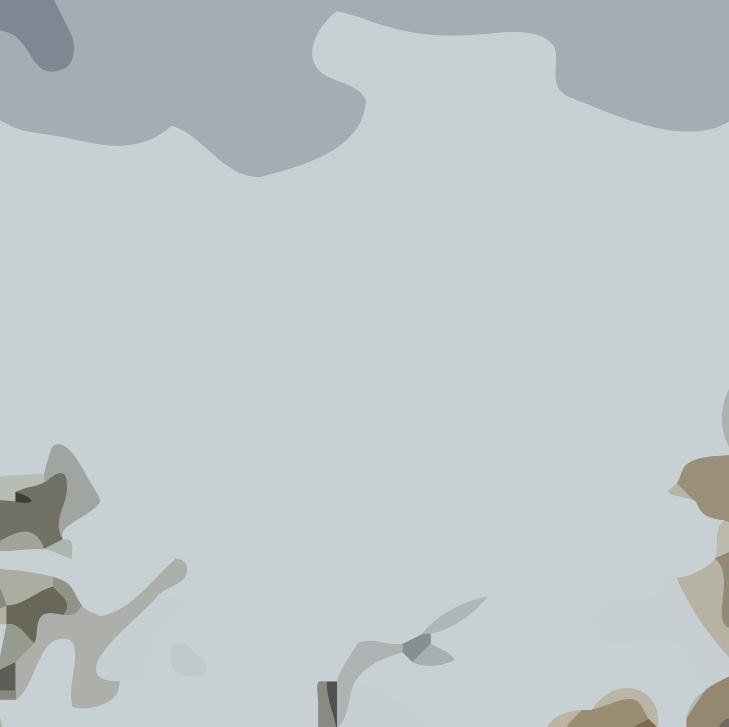}&
\includegraphics[width=0.18\textwidth]{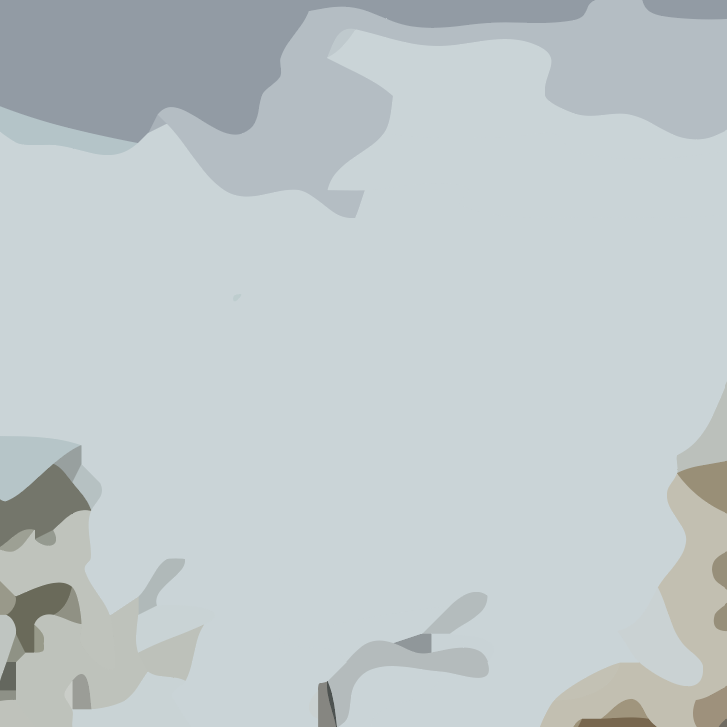}&
\includegraphics[width=0.18\textwidth]{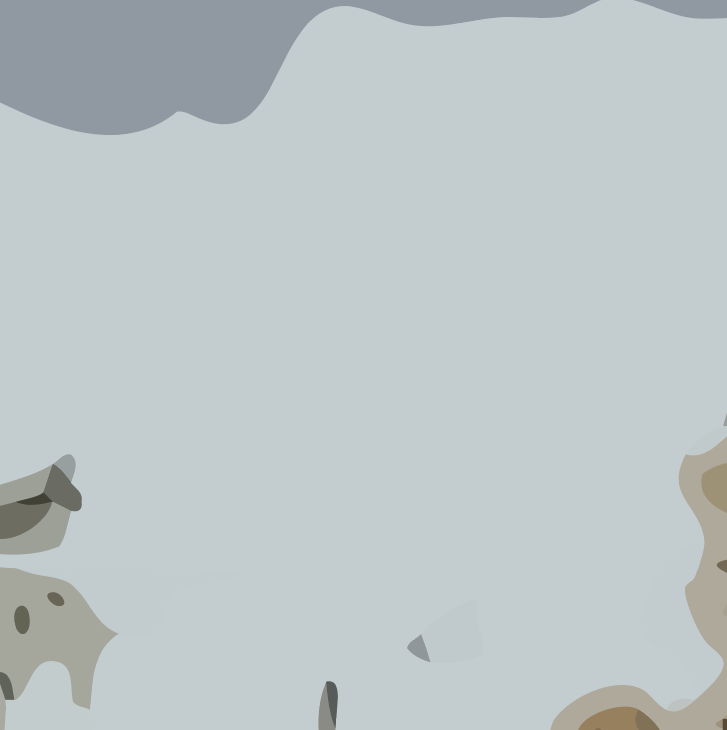}&
\includegraphics[width=0.18\textwidth]{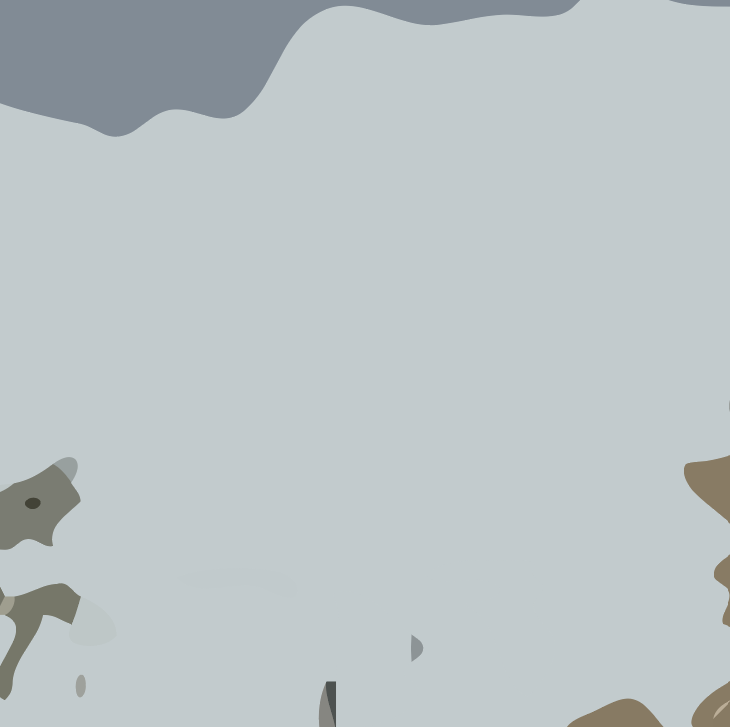}\\ \\
(a) Given input & (b) Area & (c) BG & (d) Scale & (e) MS \\
\includegraphics[width=0.18\textwidth]{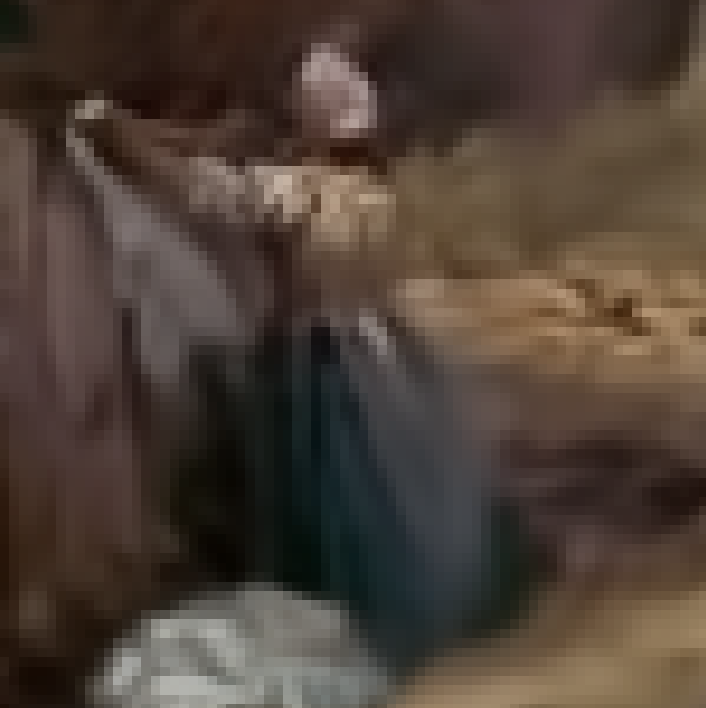}&
\includegraphics[width=0.18\textwidth]{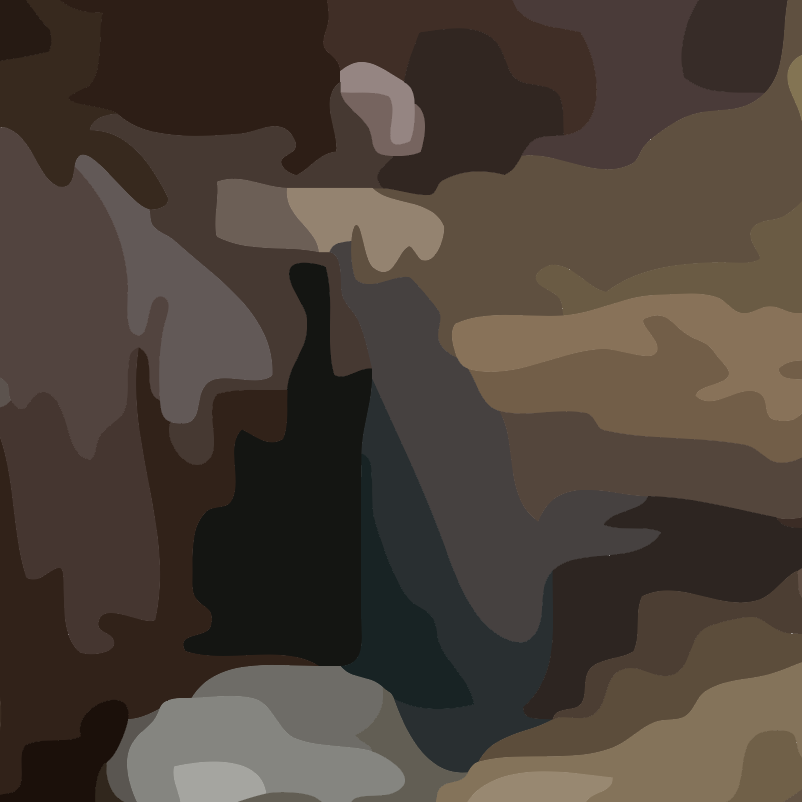}&
\includegraphics[width=0.18\textwidth]{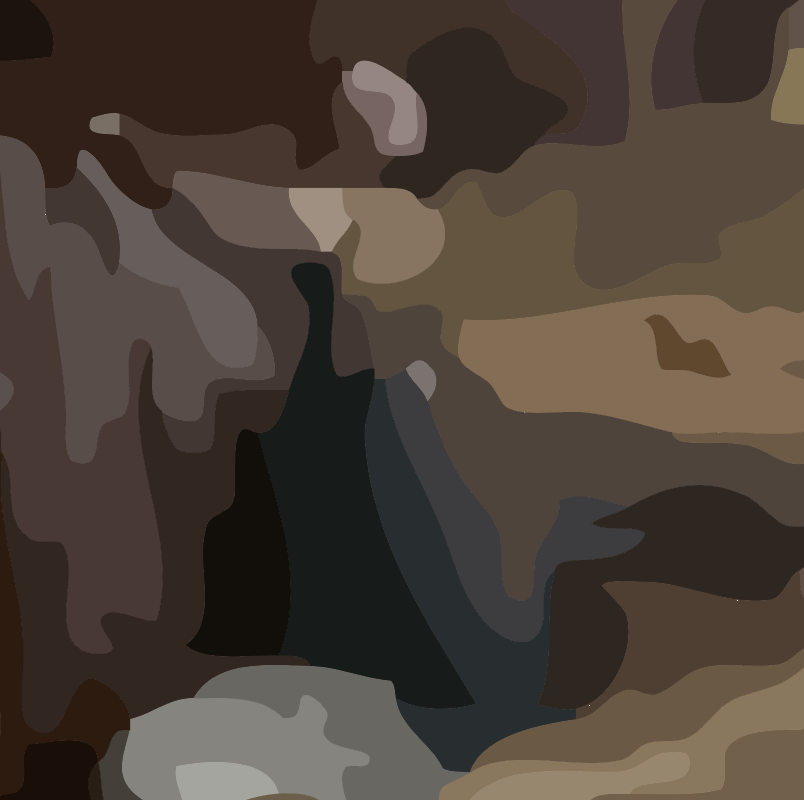}&
\includegraphics[width=0.18\textwidth]{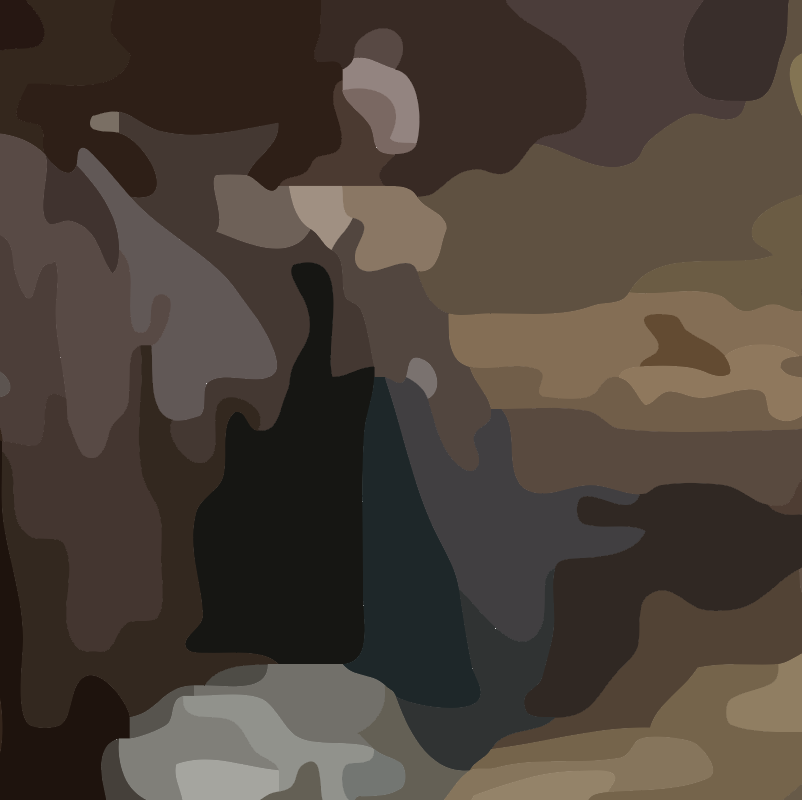}&
\includegraphics[width=0.18\textwidth]{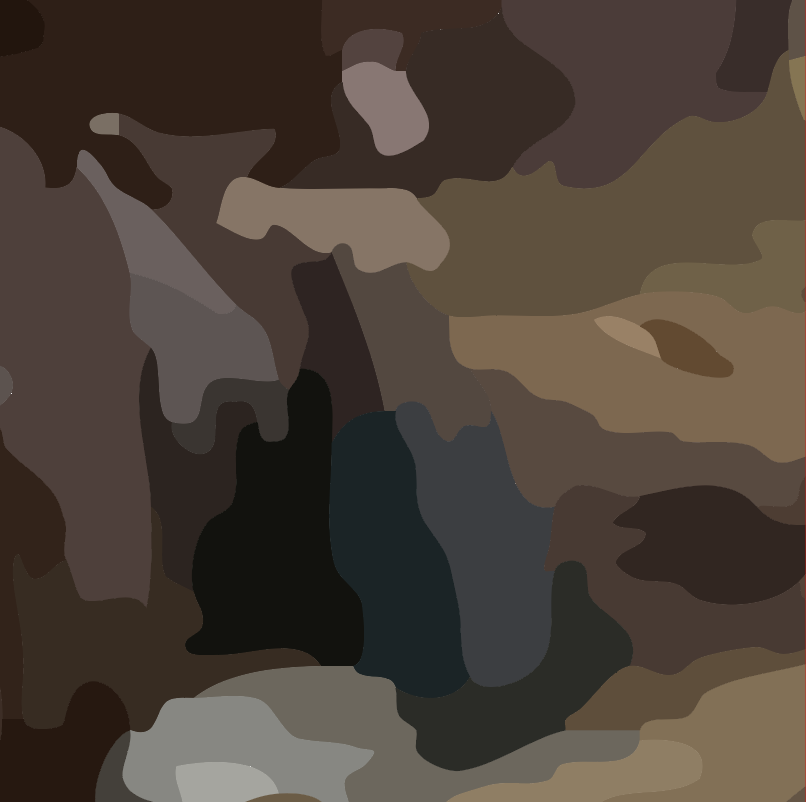}\\
(f) WTSE & (g) IOKM & (h) MC & (i) Kpp & (j) NEU \\
\includegraphics[width=0.18\textwidth]{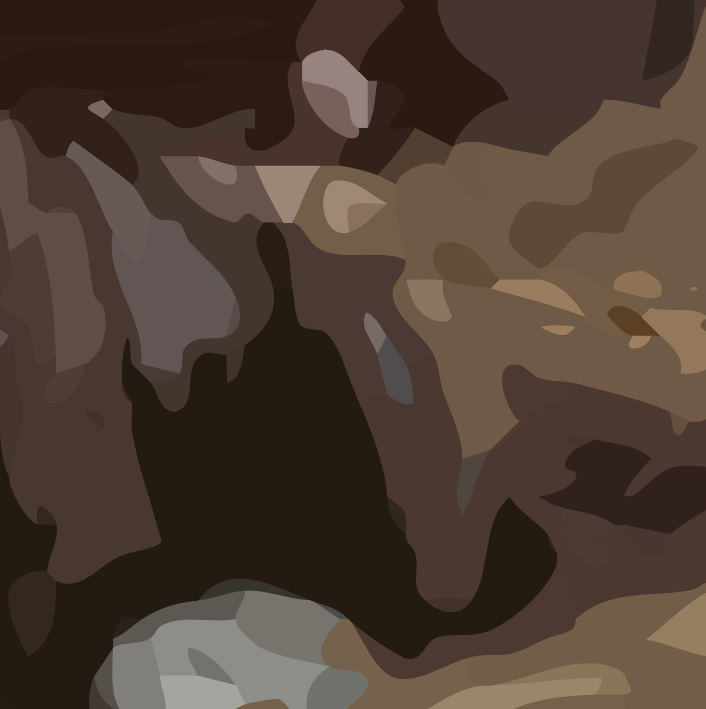}&
\includegraphics[width=0.18\textwidth]{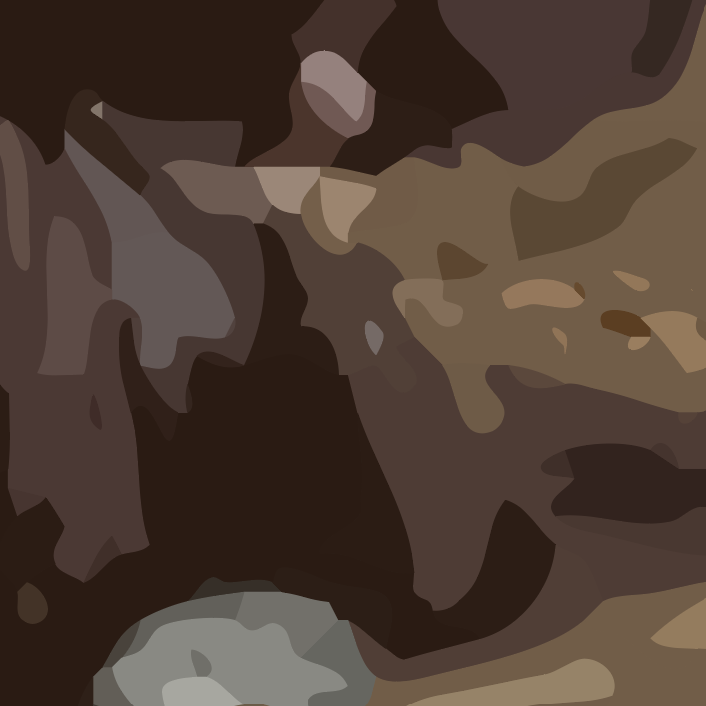}&
\includegraphics[width=0.18\textwidth]{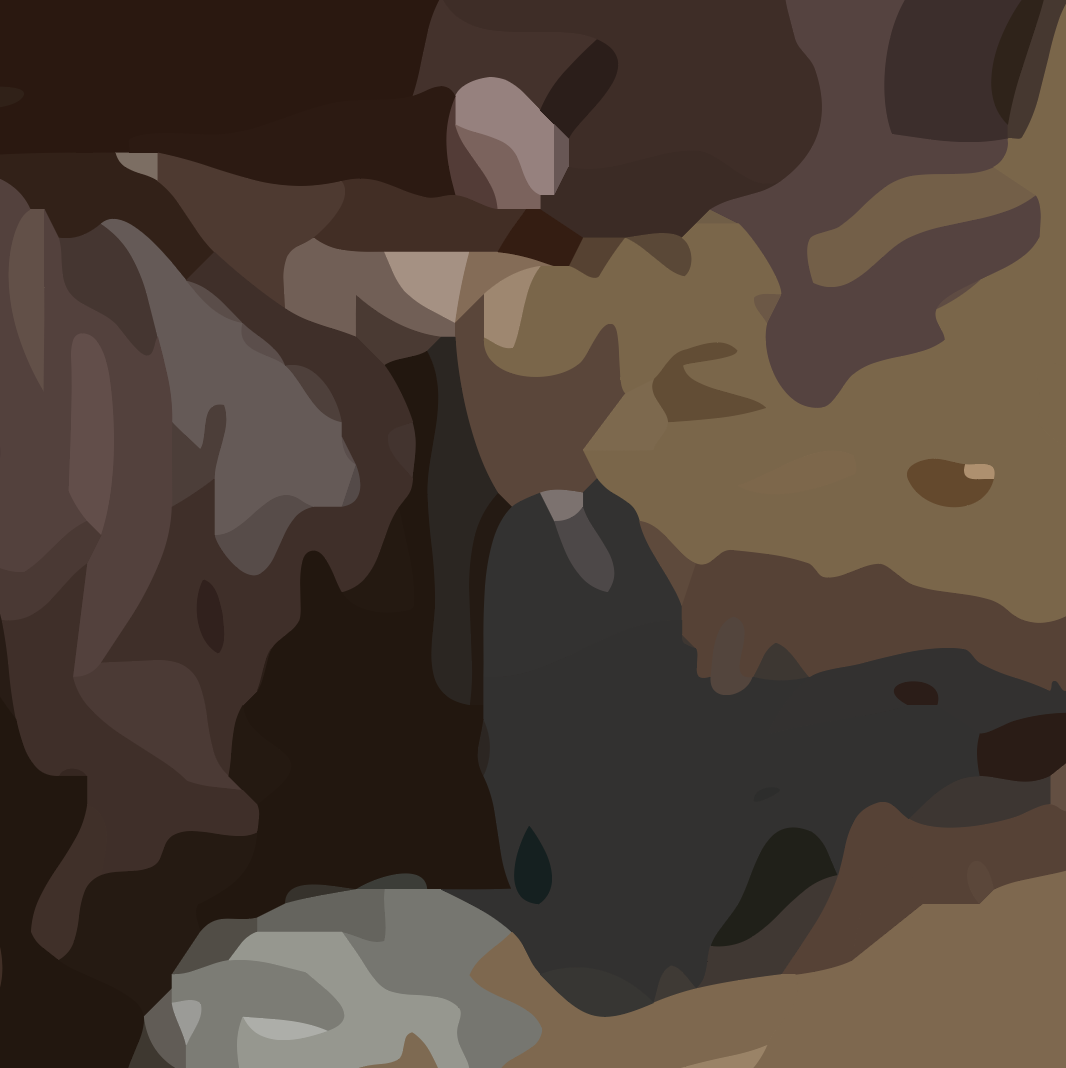}&
\includegraphics[width=0.18\textwidth]{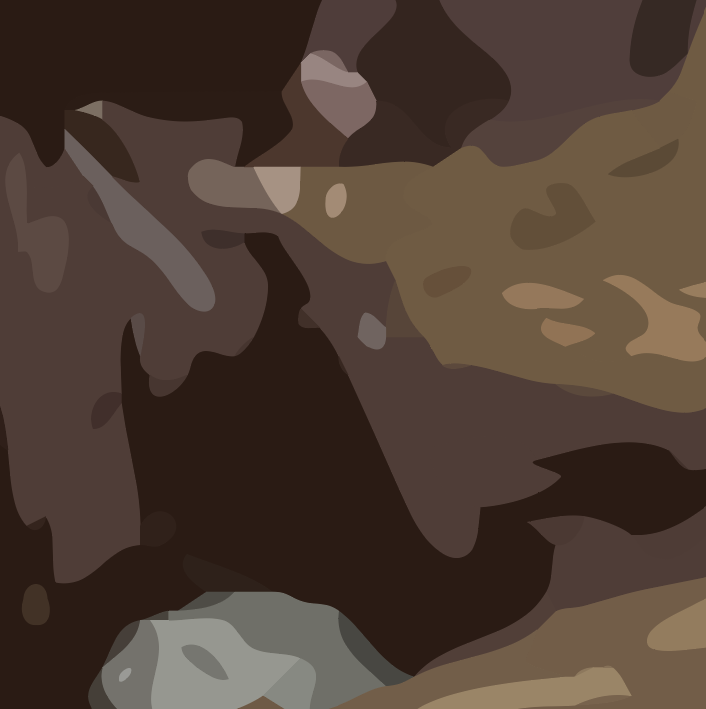}&
\includegraphics[width=0.18\textwidth]{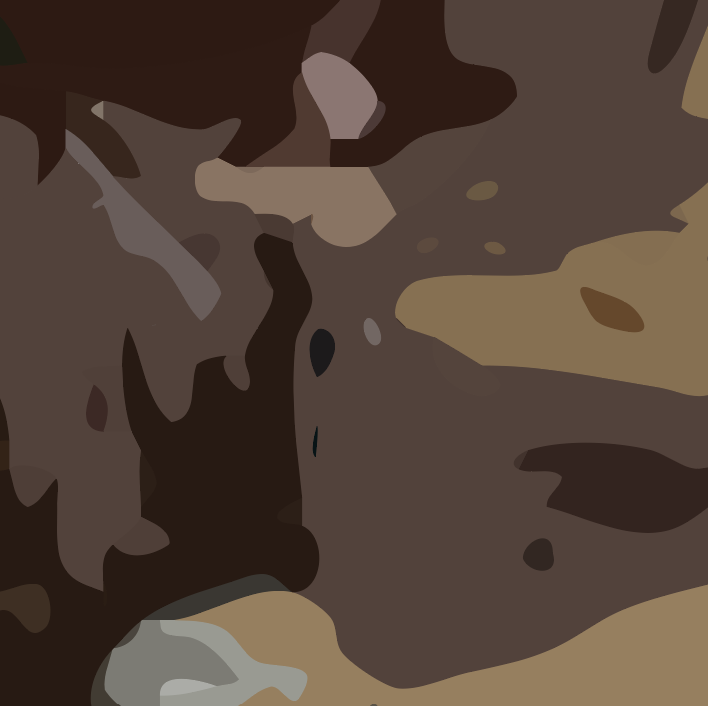}\\ \\
(a) Given input & (b) Area & (c) BG & (d) Scale & (e) MS \\
\includegraphics[width=0.18\textwidth]{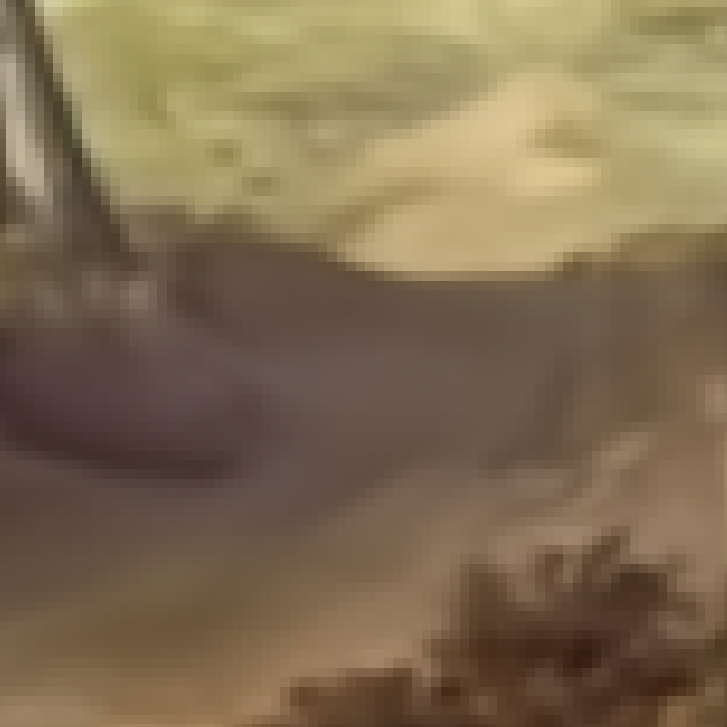}&
\includegraphics[width=0.18\textwidth]{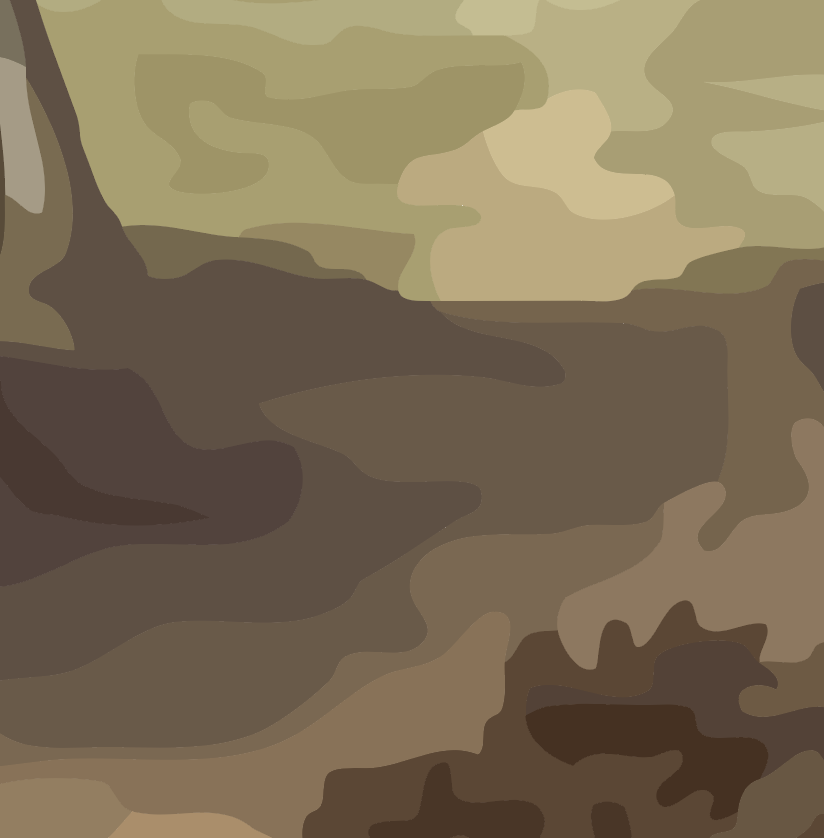}&
\includegraphics[width=0.18\textwidth]{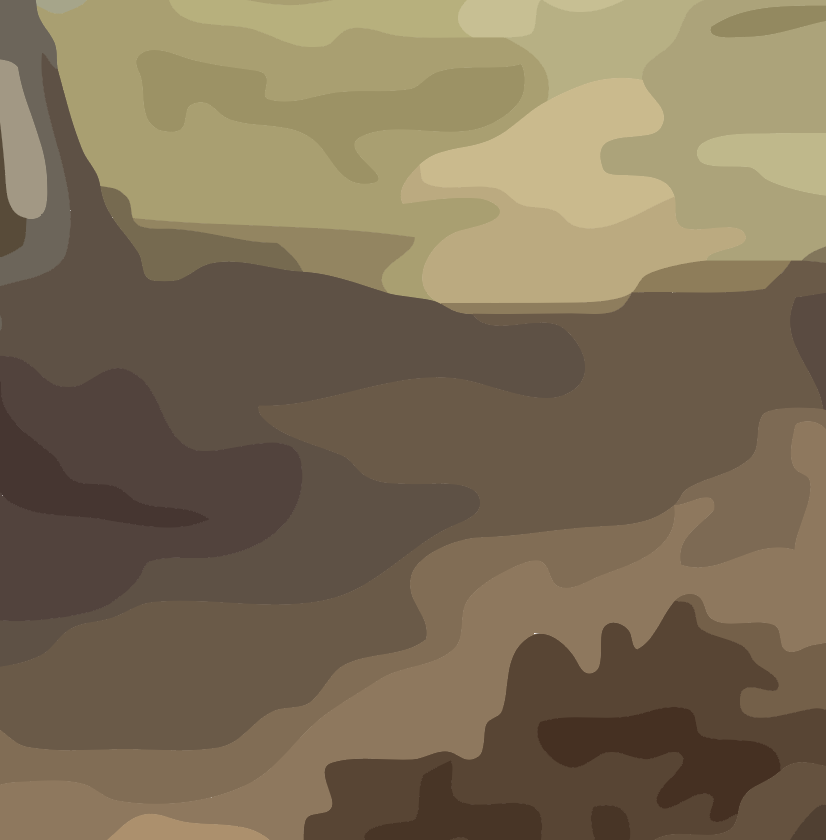}&
\includegraphics[width=0.18\textwidth]{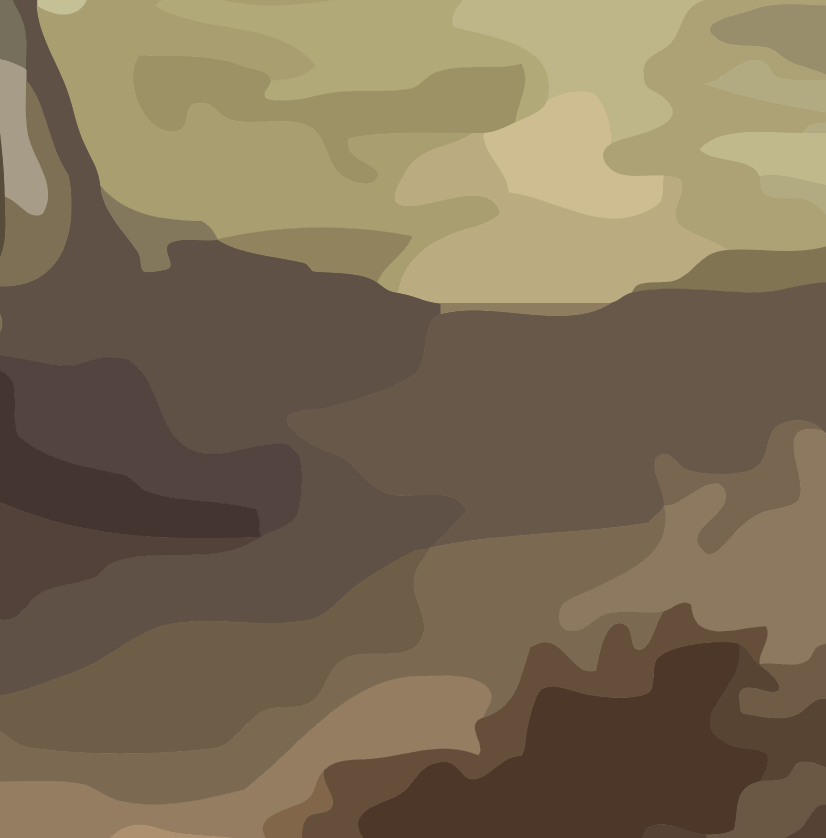}&
\includegraphics[width=0.18\textwidth]{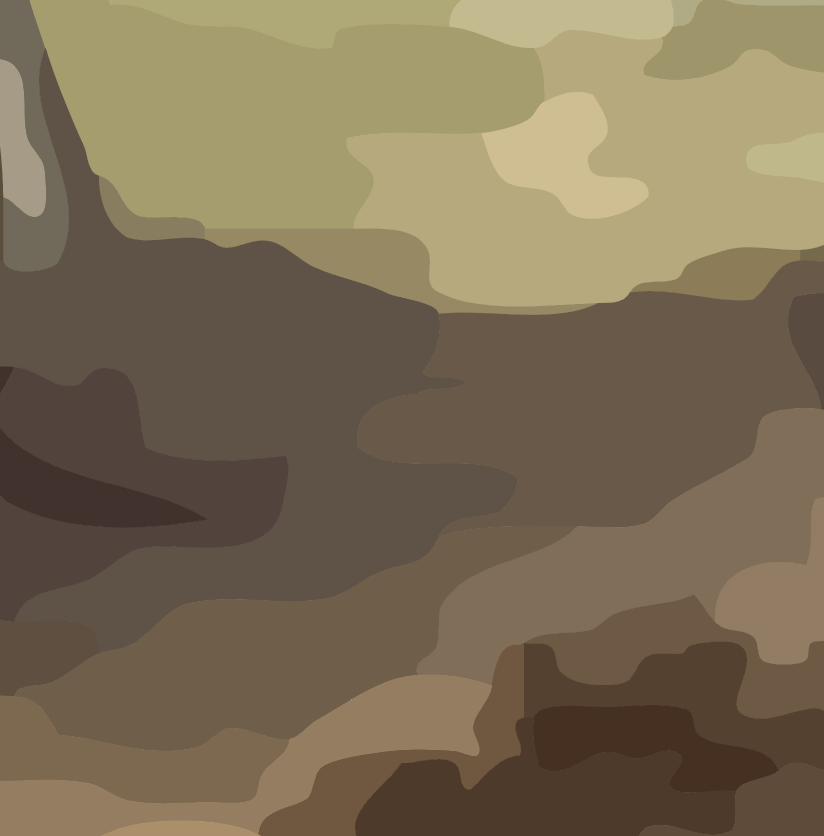}\\
(f) WTSE & (g) IOKM & (h) MC & (i) Kpp & (j) NEU \\
\includegraphics[width=0.18\textwidth]{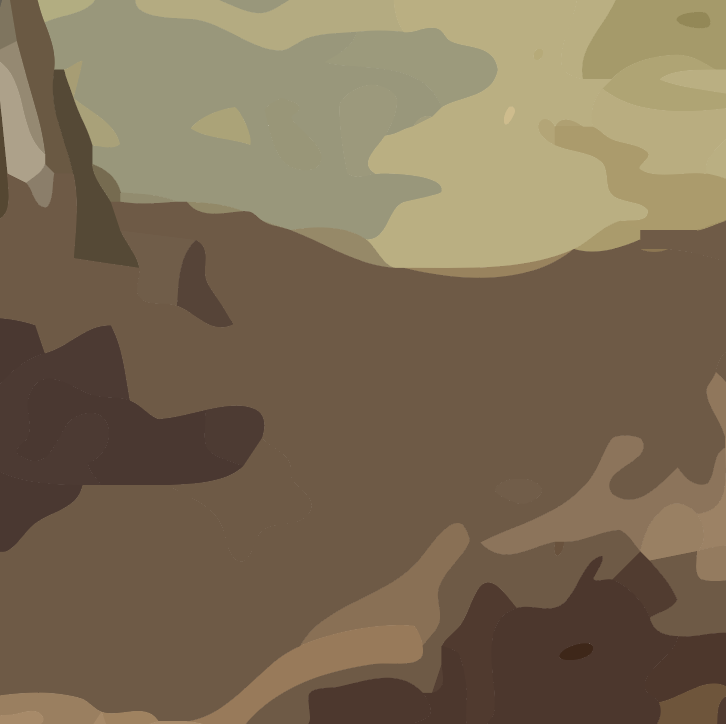}&
\includegraphics[width=0.18\textwidth]{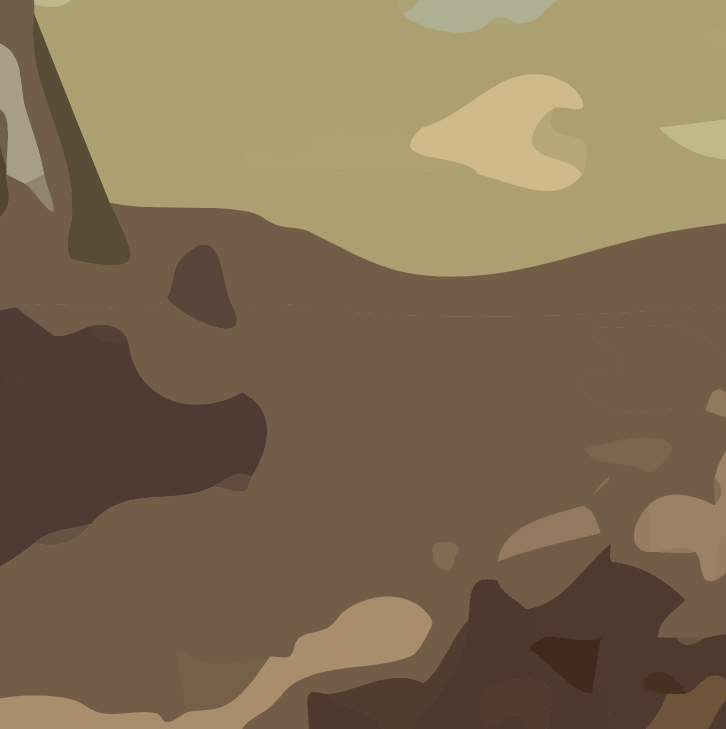}&
\includegraphics[width=0.18\textwidth]{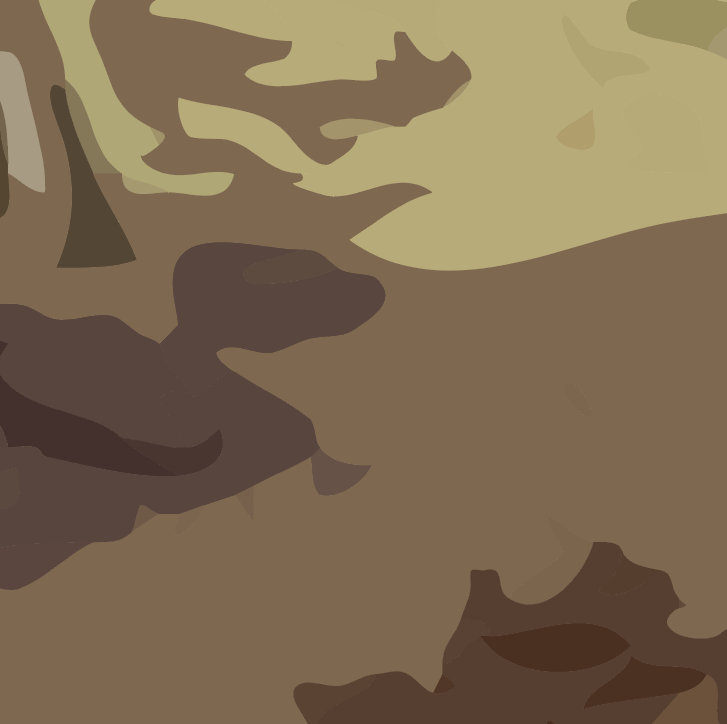}&
\includegraphics[width=0.18\textwidth]{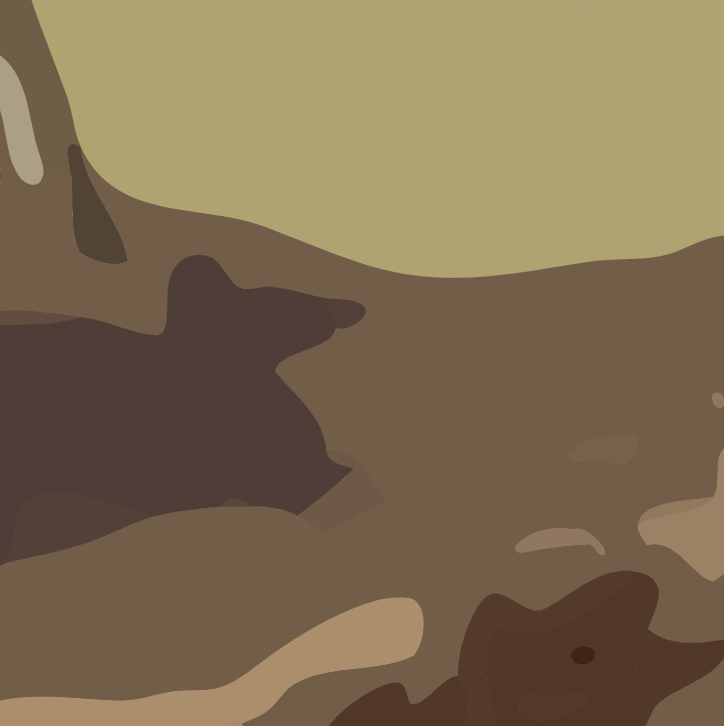}&
\includegraphics[width=0.18\textwidth]{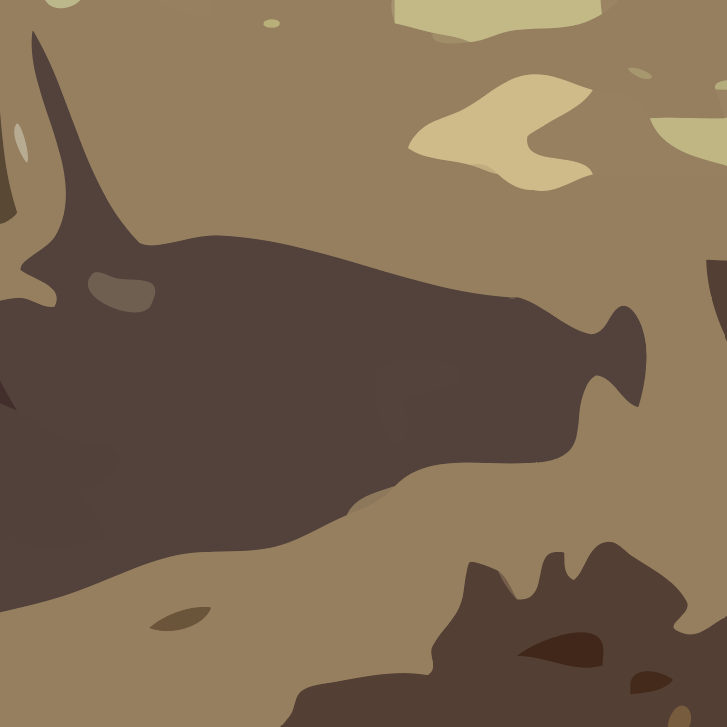}
\end{tabular}
\caption{Zoom-in of red boxes in Figure~\ref{fig_quant_compare}.  In the first group, Area, BG, Scale and MS region mergings show the cloud more accurately compared to all other color quantization based methods.  
In the second group, Area, BG, Scale, MS and MC approximate the skirt color better than other methods.
In the third group, Area, BG, Scale, and MS give  results that are closer to the input.
}\label{fig_quant_compare_zoom}
\end{figure}

In Figure~\ref{fig_quant_compare_zoom}, we compare the results with more details in the zoom-in regions boxed in (a). In the first group, we notice that the cloud, which is a perceivable feature of the input image,  is successfully reconstructed by the proposed scheme using the different region merging criteria. It is merged with the large background region by the other methods. In the second group, most of the quantization-based algorithms fail to recognize the dark blue dress, except for MC. The  region-merging  method not only identifies the dress, but also distinguishes the grass and mud in the background, which MC mistakenly merges with the shoulder of the woman. In the third group, the contouring artifacts of quantization-based results are accentuated. We find that the region-merging-based approaches and MC are better at recovering fine structure in low-contrast regions. However,  MC merges regions with wrong colors, causing low fidelity levels. Among the region merging methods, Area performs  the best in terms of  detail preservation and smooth region rendering.

Quantitatively, we also compare the different approaches in terms of the PSNR and the complexity of the representation measured by the number of regions in the vector graphics. As shown in Figure~\ref{fig_quant_compare}~(b), all the proposed region-merging-based methods outperform the other color-quantization-based results. More specifically, when represented by roughly the same number of regions, region merging  methods yield a closer approximation of  the raster input. Note that the regions produced by color quantization are filled with their respective mean colors; thus our comparison reveals the lack of compatibility of the partitions generated by data-clustering in color spaces and the benefit of the proposed region-merging scheme for the task of image vectorization.

\subsection{Comparison with state-of-the-art software}

In this section, we compare our methods with   state-of-the-art software including Adobe Illustrator 2020 (AI)\footnote{\href{https://www.adobe.com/products/illustrator.html}{https://www.adobe.com/products/illustrator.html}}, Vector Magic (VM)\footnote{\href{https://vectormagic.com/}{https://vectormagic.com/}}, Vectorizer (VZ)\footnote{\href{https://www.vectorizer.io/}{https://www.vectorizer.io/}}, Vectorizer AI (VAI)\footnote{\href{https://vectorizer.ai/}{https://vectorizer.ai/}}, Inkscape (IS)\footnote{\href{https://inkscape.org}{https://inkscape.org}},  a network-based vectorization method proposed by Li et al. (DiffVG)~\cite{li2020differentiable}, and TOP~\cite{he2023topology} based on color quantization and local boundary surgery. These programs accept raster images of any size and content. In particular, VM allows for low, medium, and high levels of details; VZ allows for min, low, medium, high, very high, maximum, and ultra; for AI, VAI, and IS, the results can be fine-tuned via multiple continuous or discrete parameters. We choose the respective default settings with non-overlapping option in the experiments. For TOP, we apply the Kpp for quantization with $32$ colors and use the default parameters.  For DiffVG, we use the vectorized result from AI as the initialization and run the optimization under the default setting.  

\begin{figure}
\centering
\begin{tabular}{c@{\hspace{2pt}}c@{\hspace{2pt}}c@{\hspace{2pt}}c@{\hspace{2pt}}c}
\hline
Input&Area&BG&Scale&MS\\
\includegraphics[width=0.18\textwidth]{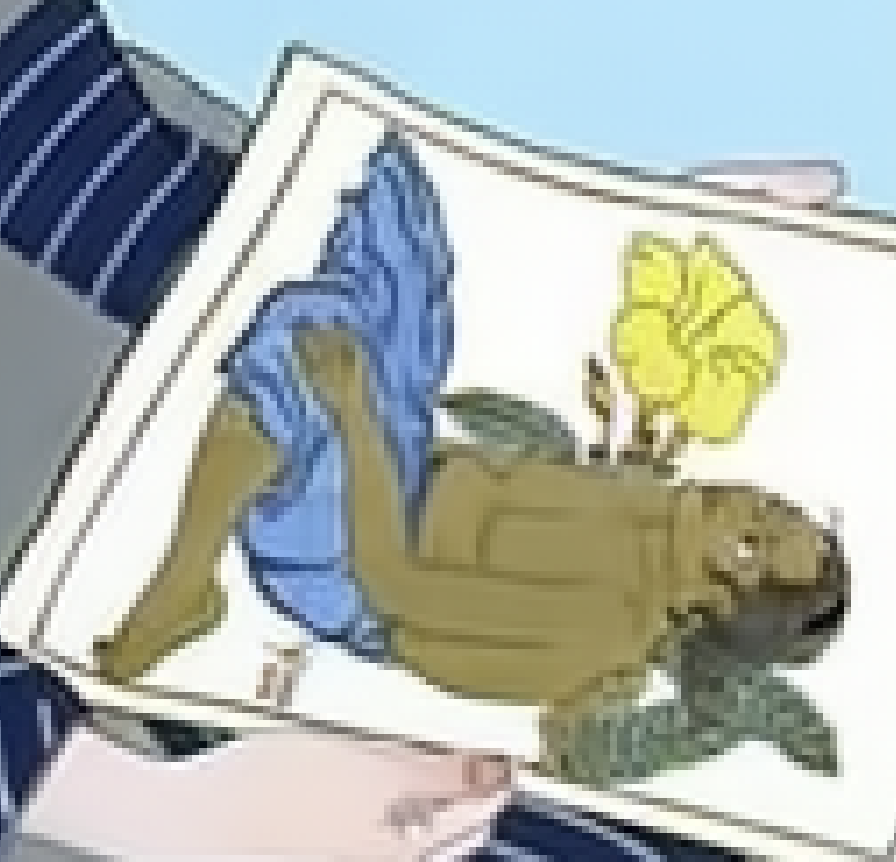}&
\includegraphics[width=0.18\textwidth]{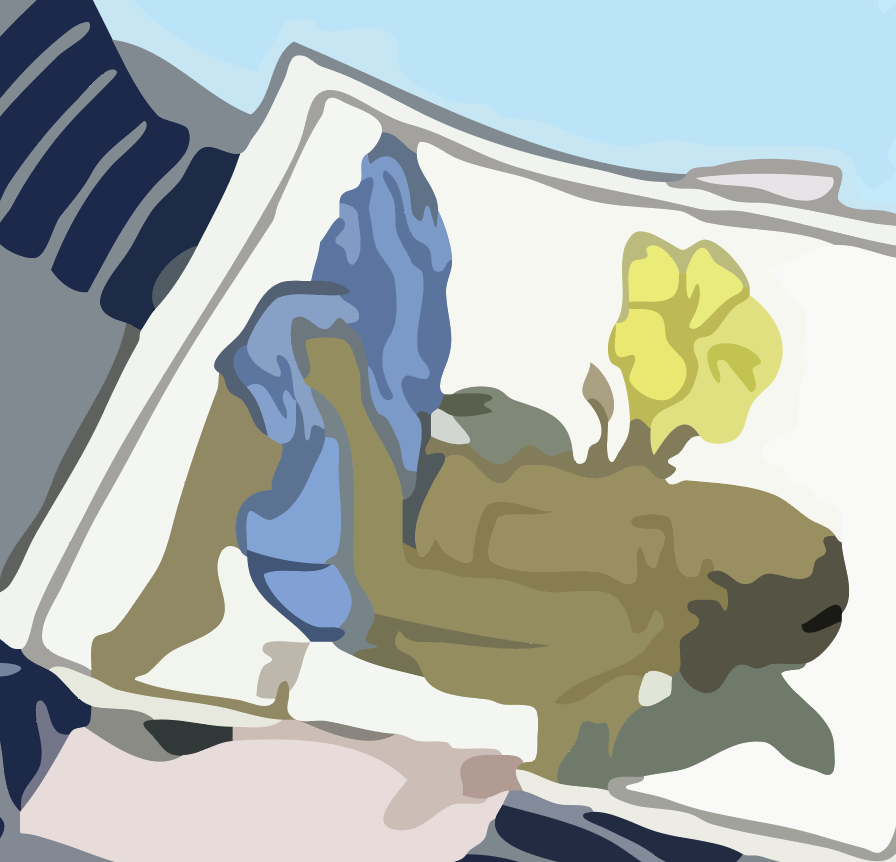}&
\includegraphics[width=0.18\textwidth]{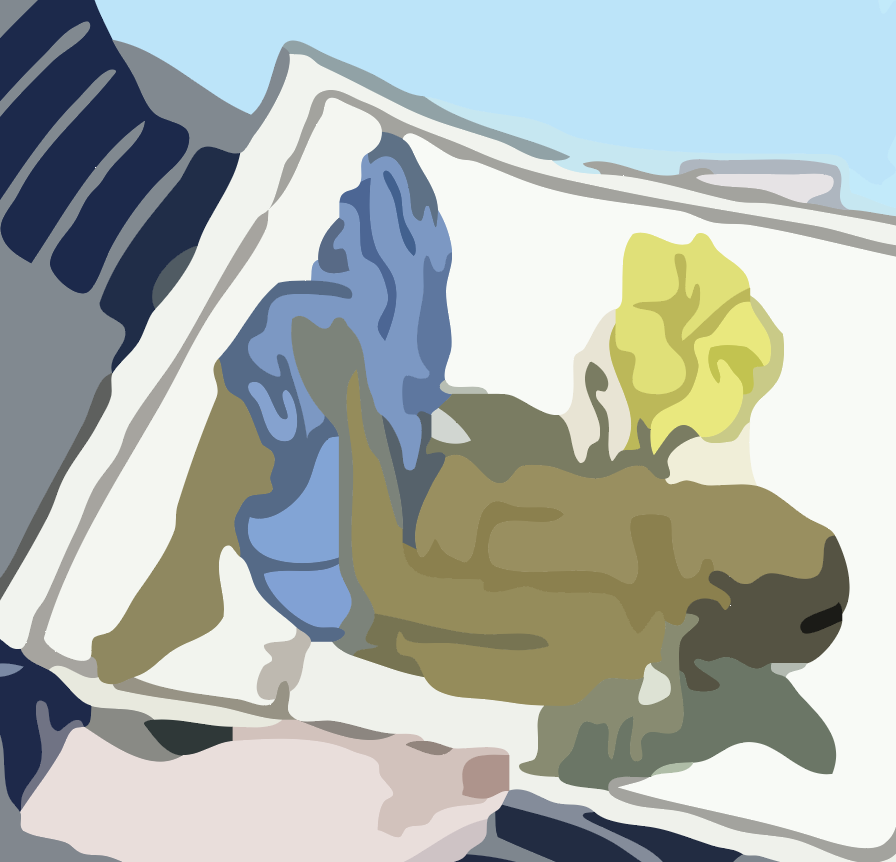}&
\includegraphics[width=0.18\textwidth]{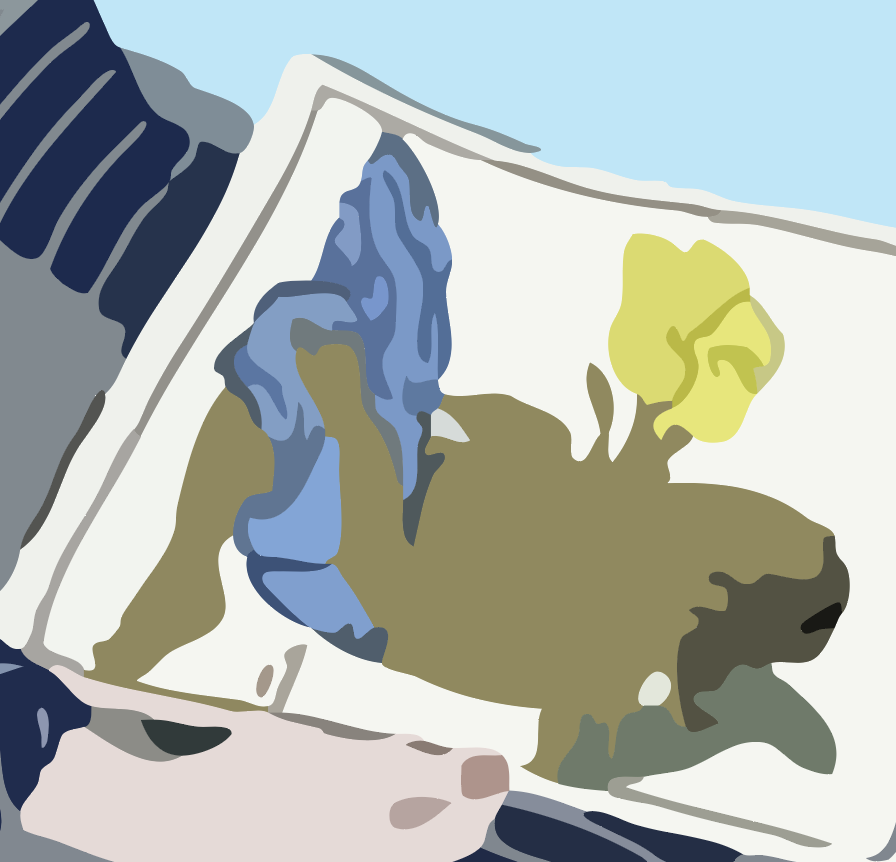}&
\includegraphics[width=0.18\textwidth]{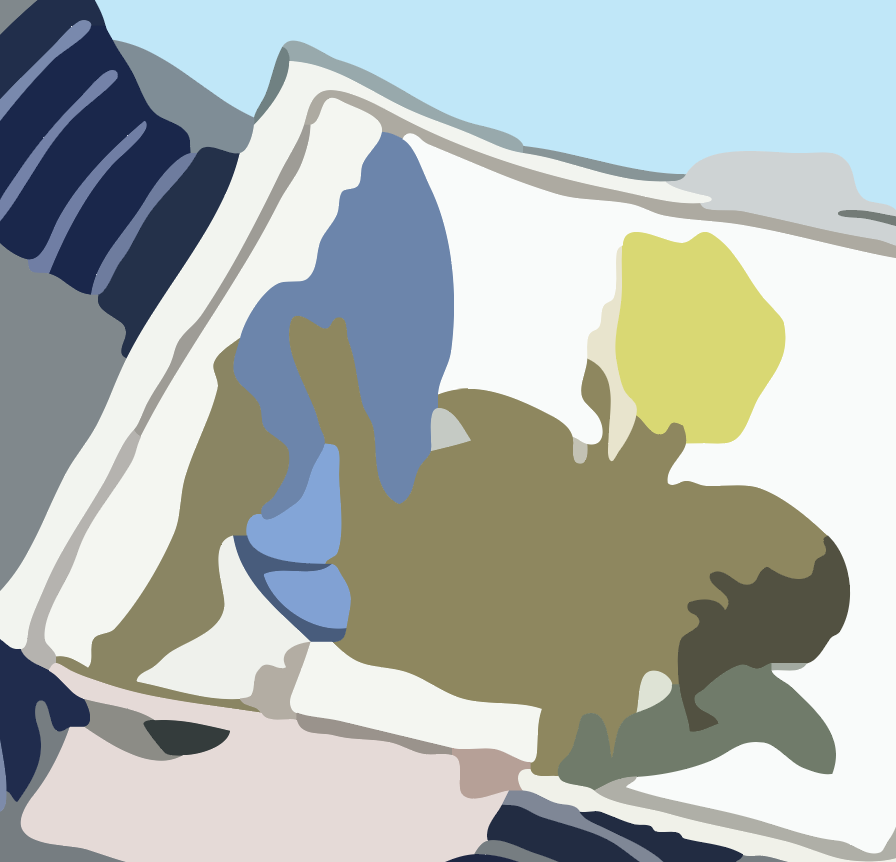}\\
&$N=320$&$N=317$&$N=266$&$N=227$\\\hline
VM High& VM Medium&VM Low&AI&VAI\\
\includegraphics[width=0.18\textwidth]{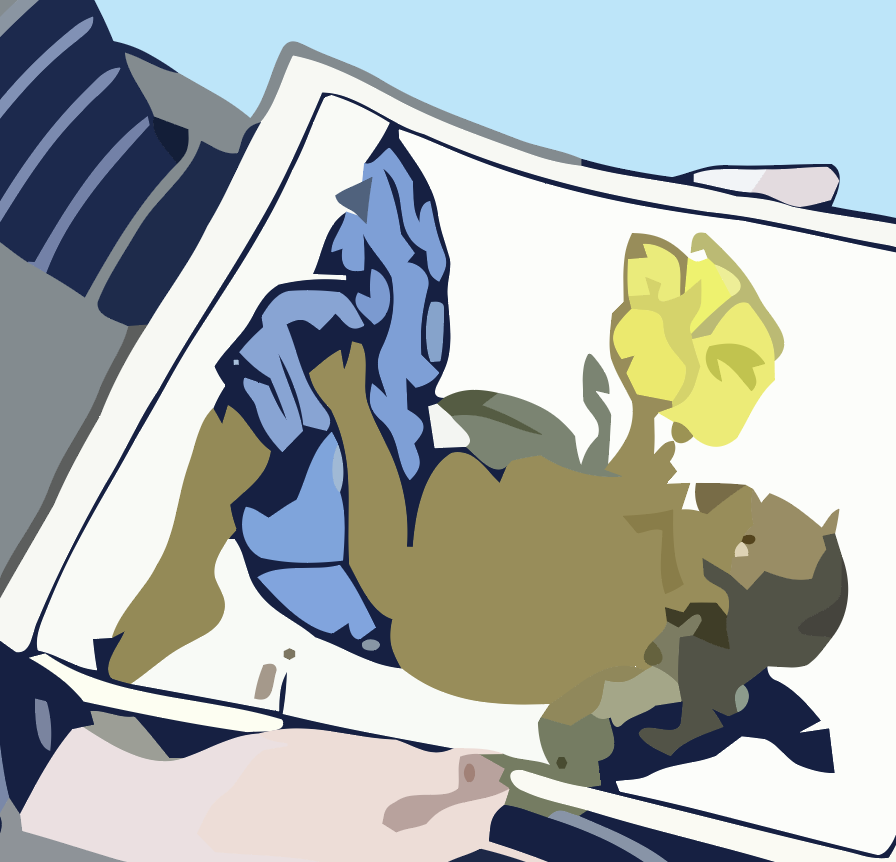}&
\includegraphics[width=0.18\textwidth]{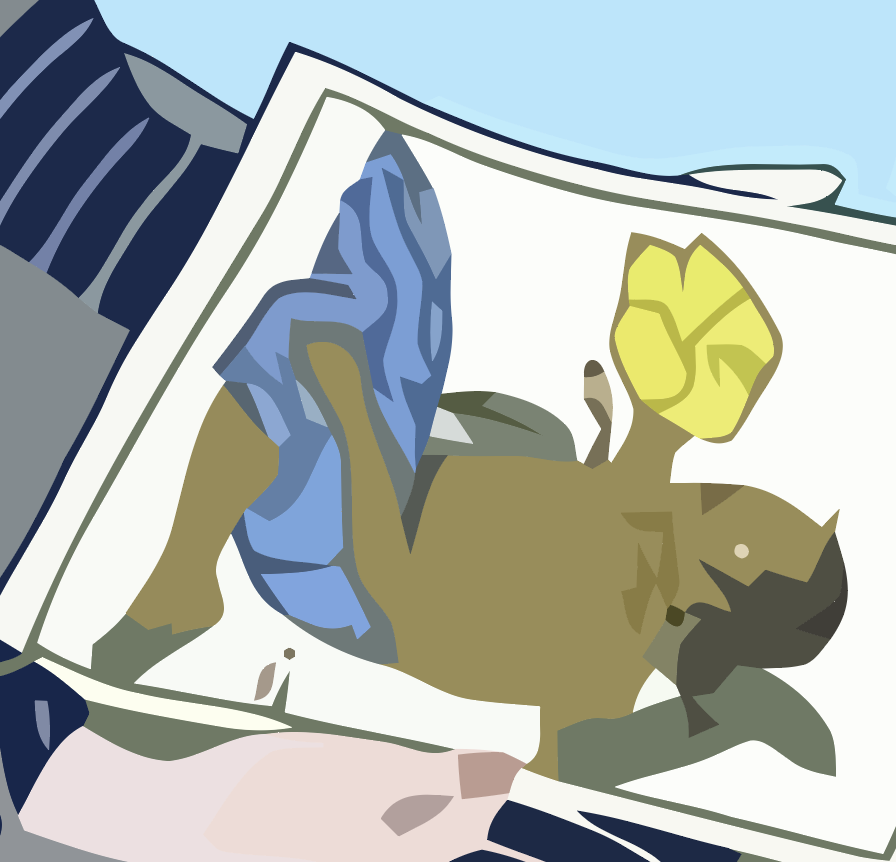}&
\includegraphics[width=0.18\textwidth]{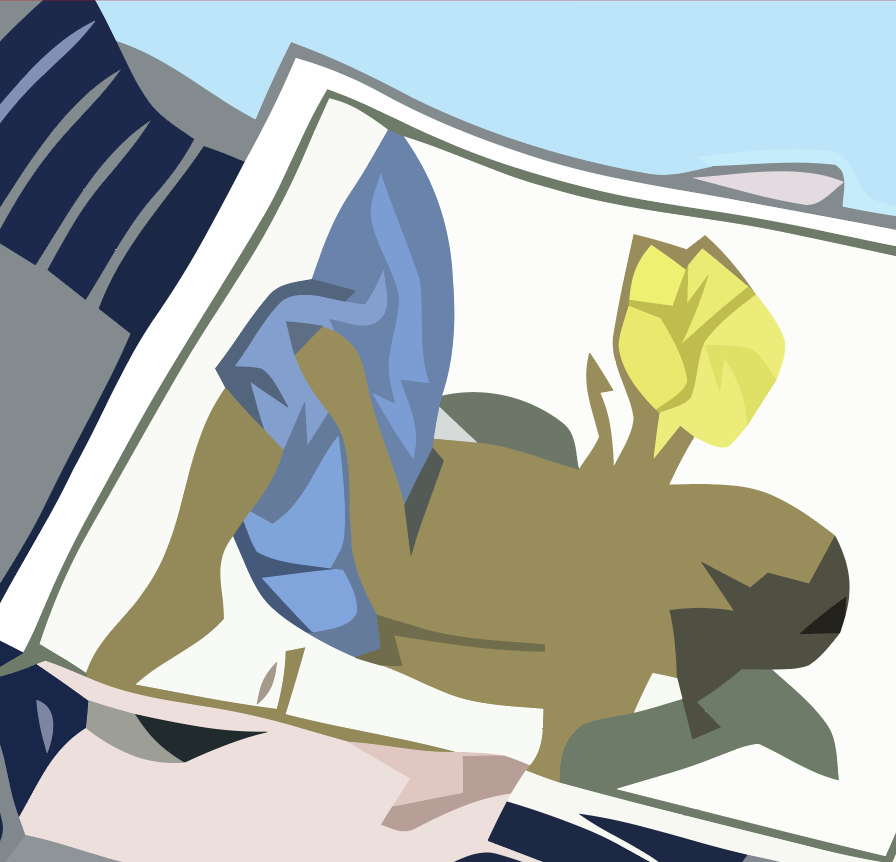}&
\includegraphics[width=0.18\textwidth]{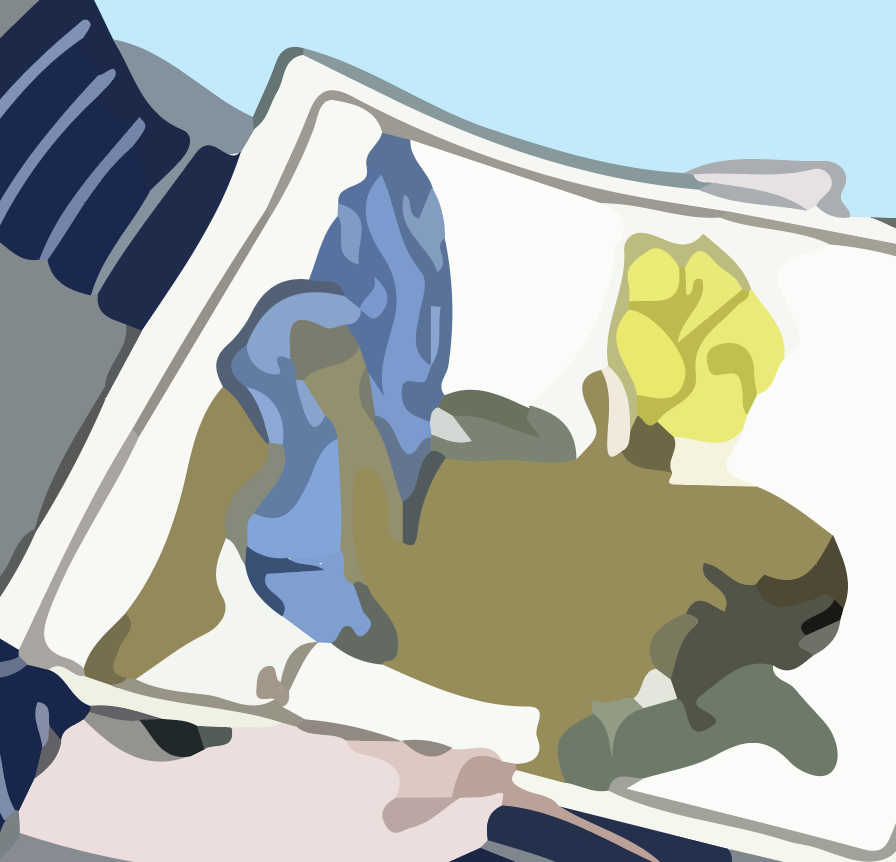}&
\includegraphics[width=0.18\textwidth]{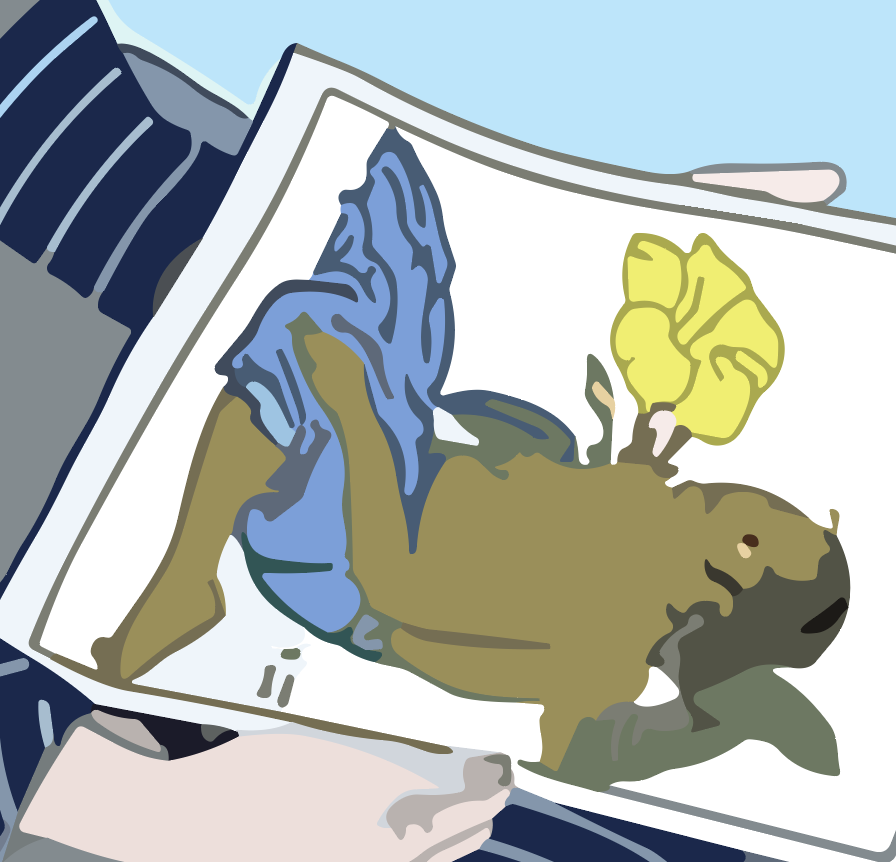}\\
$N=286$&$N=243$&$N=184$&$N=362$&$N=1304$\\\hline
VZ ultra& VZ very high&VZ high&VZ med&VZ low\\
\includegraphics[width=0.18\textwidth]{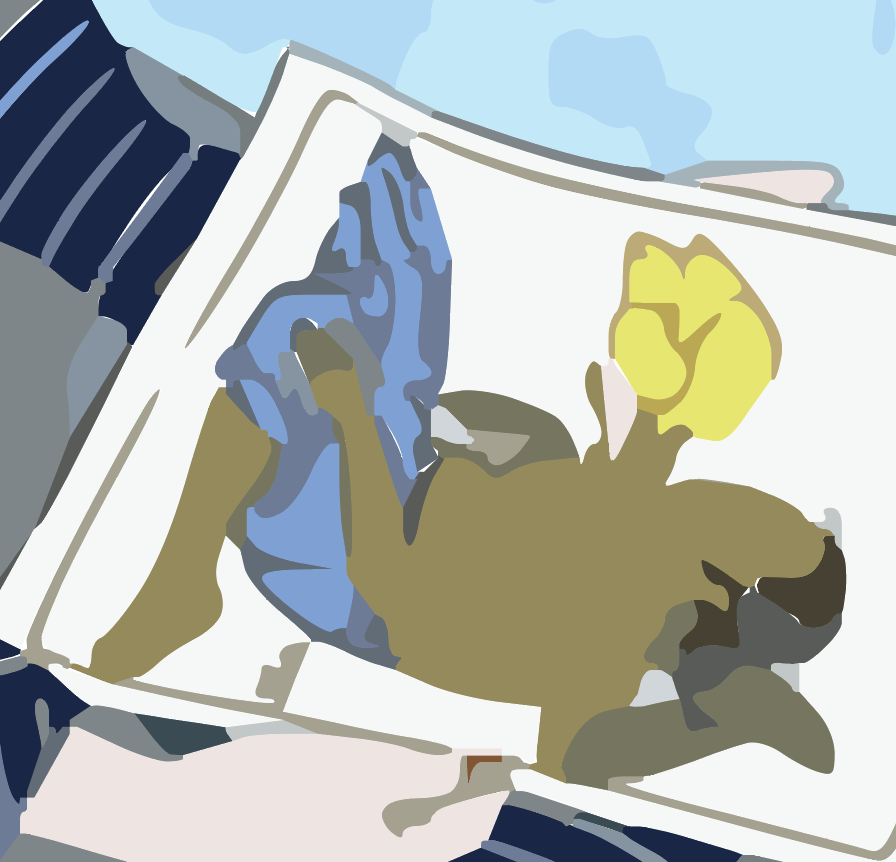}&
\includegraphics[width=0.18\textwidth]{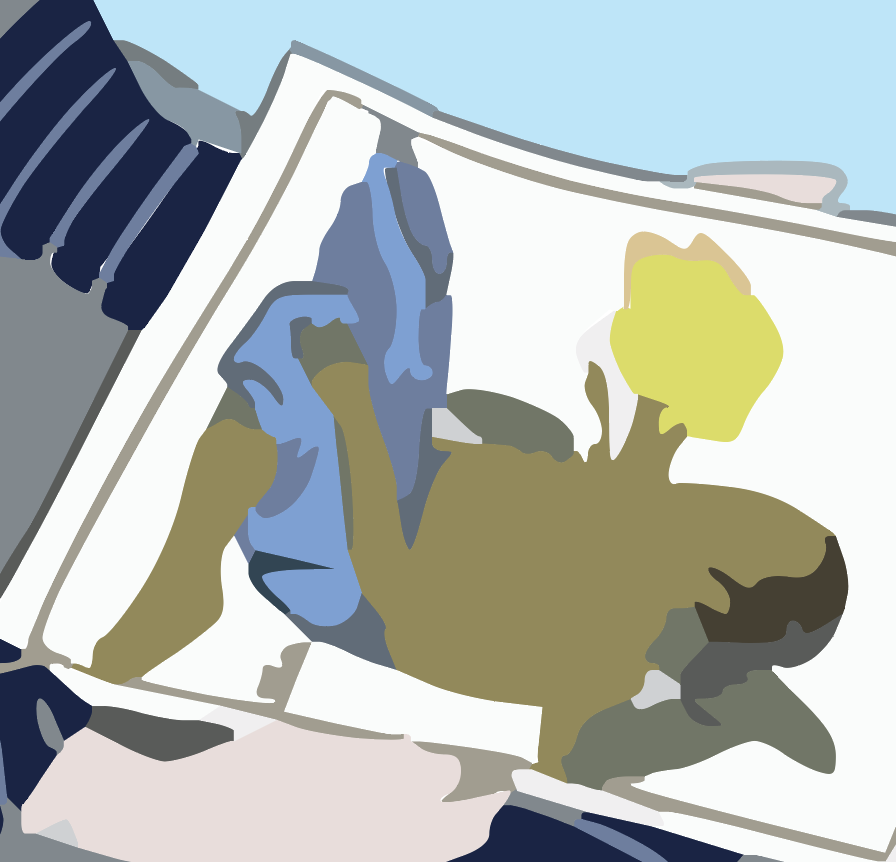}&
\includegraphics[width=0.18\textwidth]{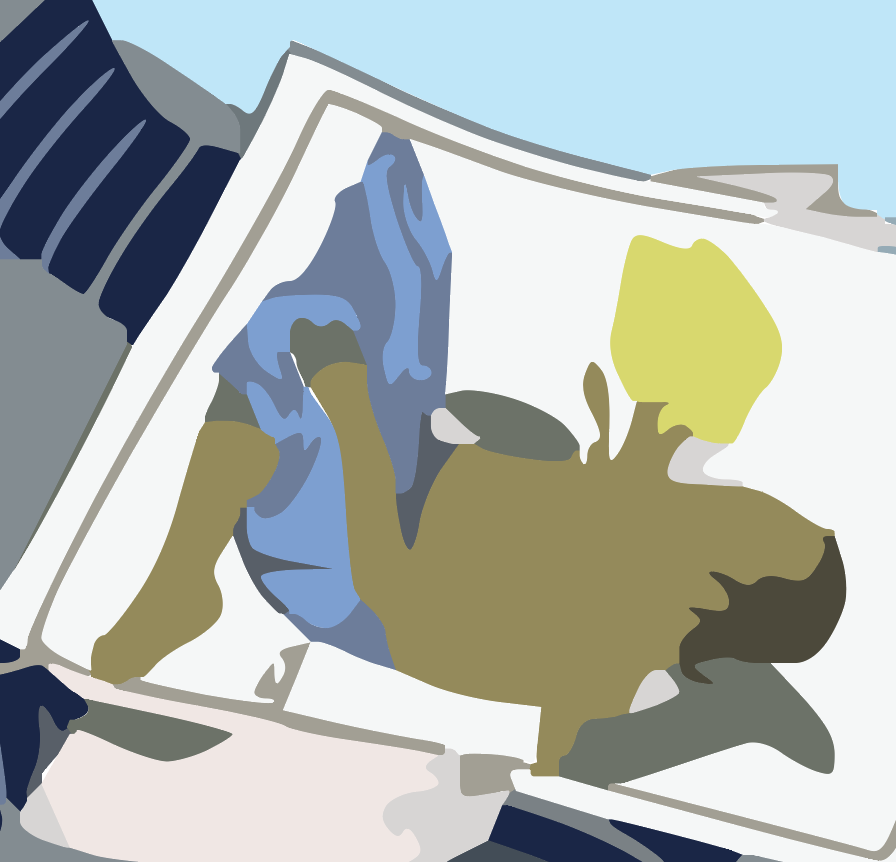}&
\includegraphics[width=0.18\textwidth]{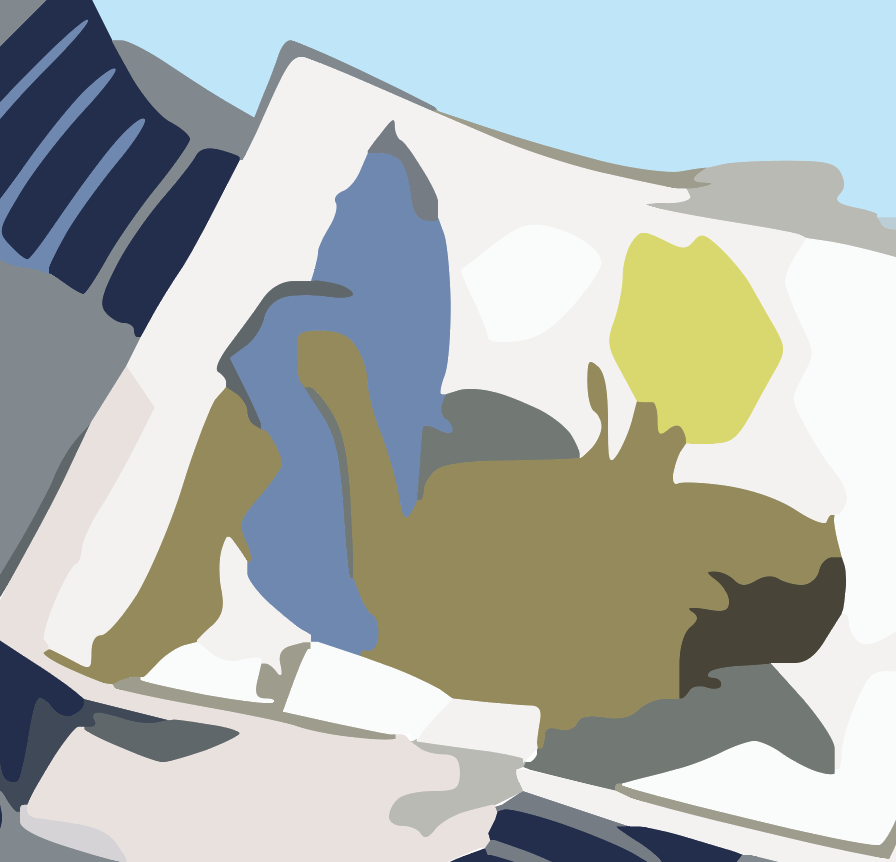}&
\includegraphics[width=0.18\textwidth]{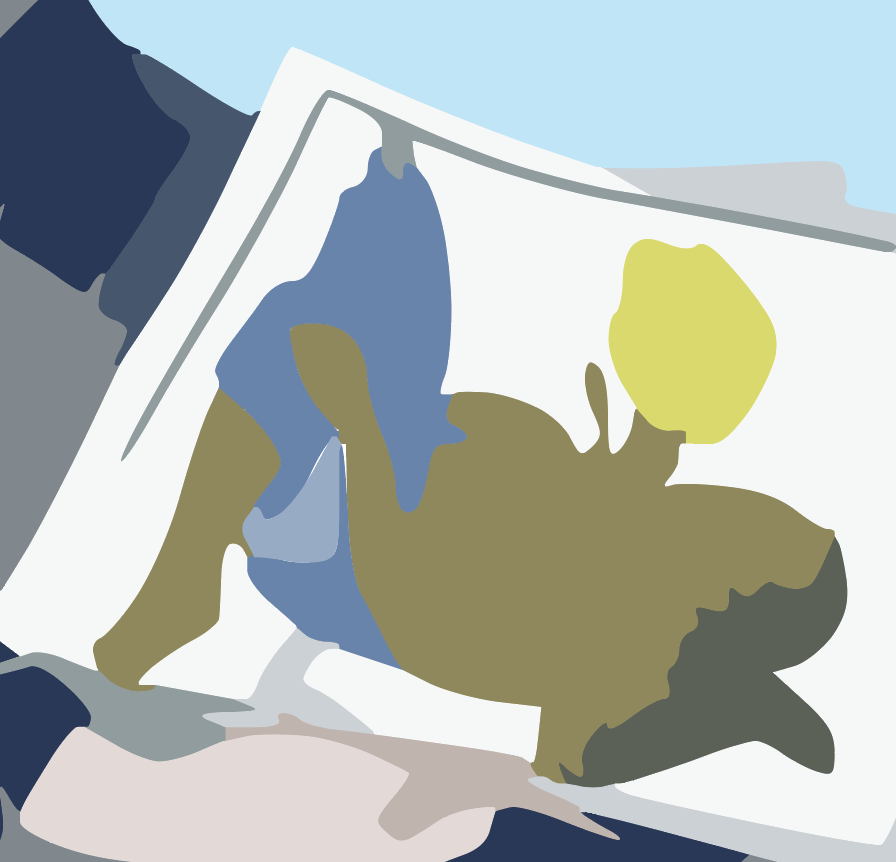}\\
$N=471$&$N=313$&$N=226$&$N=154$&$N=106$\\\hline
VZ min& DiffVG&TOP& IS& VZ max\\
\includegraphics[width=0.18\textwidth]{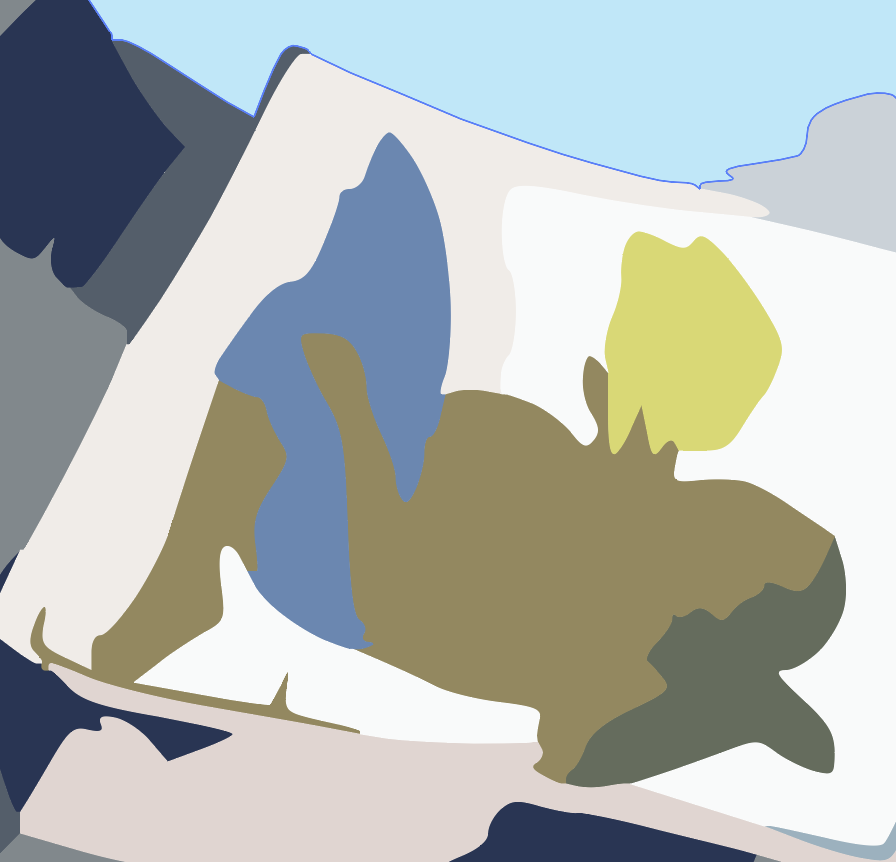}&
\includegraphics[width=0.18\textwidth]{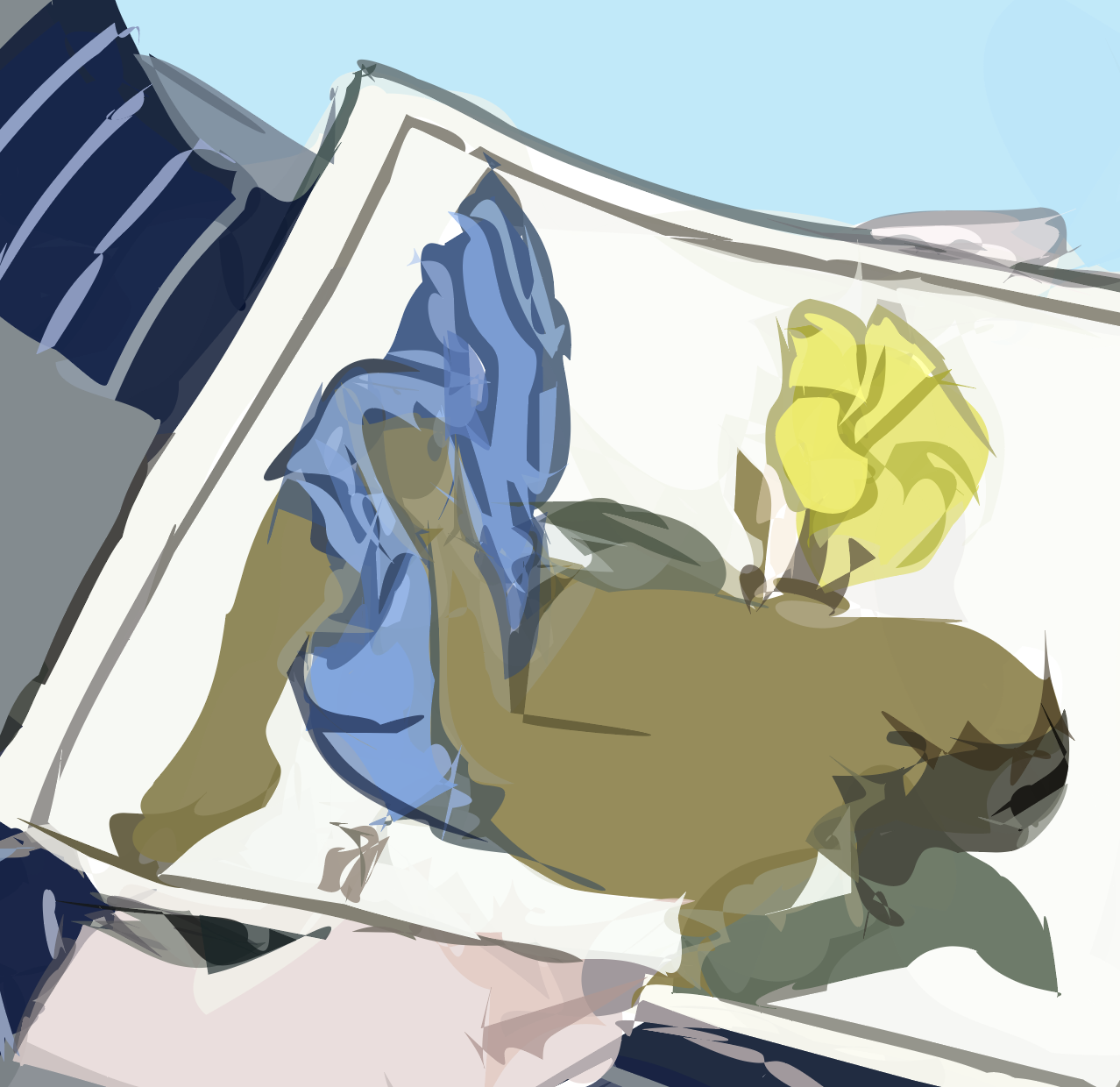}&
\includegraphics[width=0.18\textwidth]{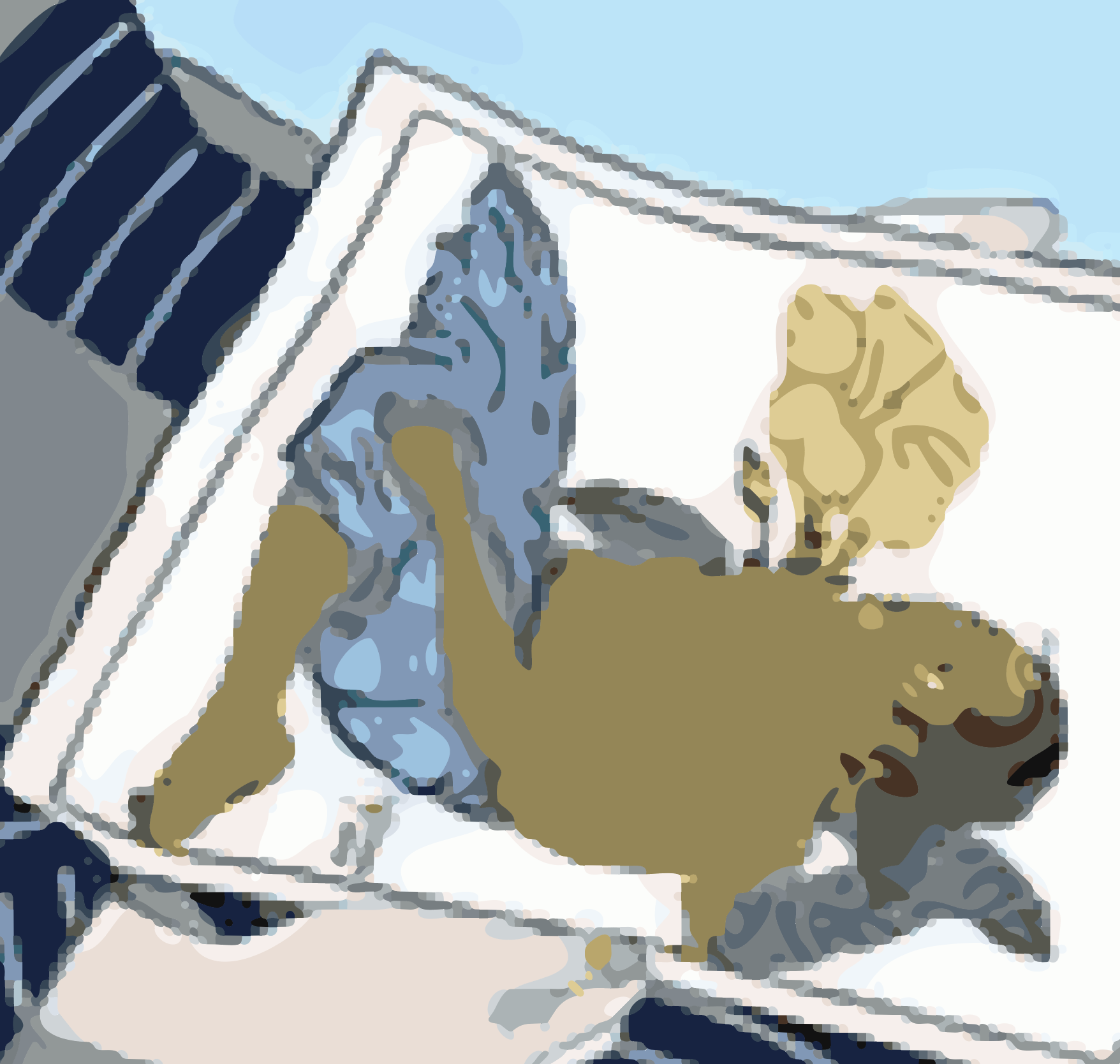}&
\includegraphics[width=0.18\textwidth]{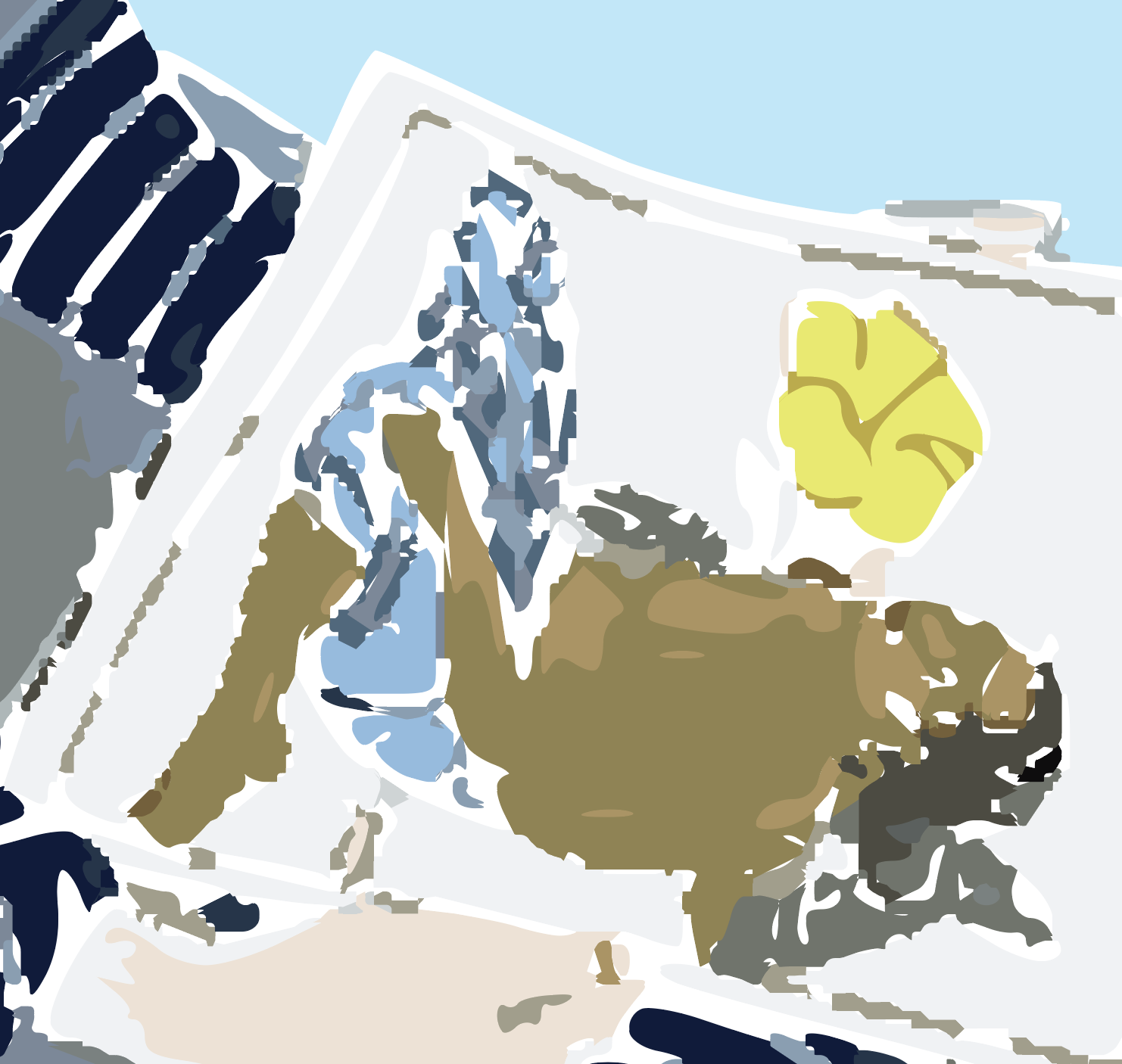}&\includegraphics[width=0.178\textwidth]{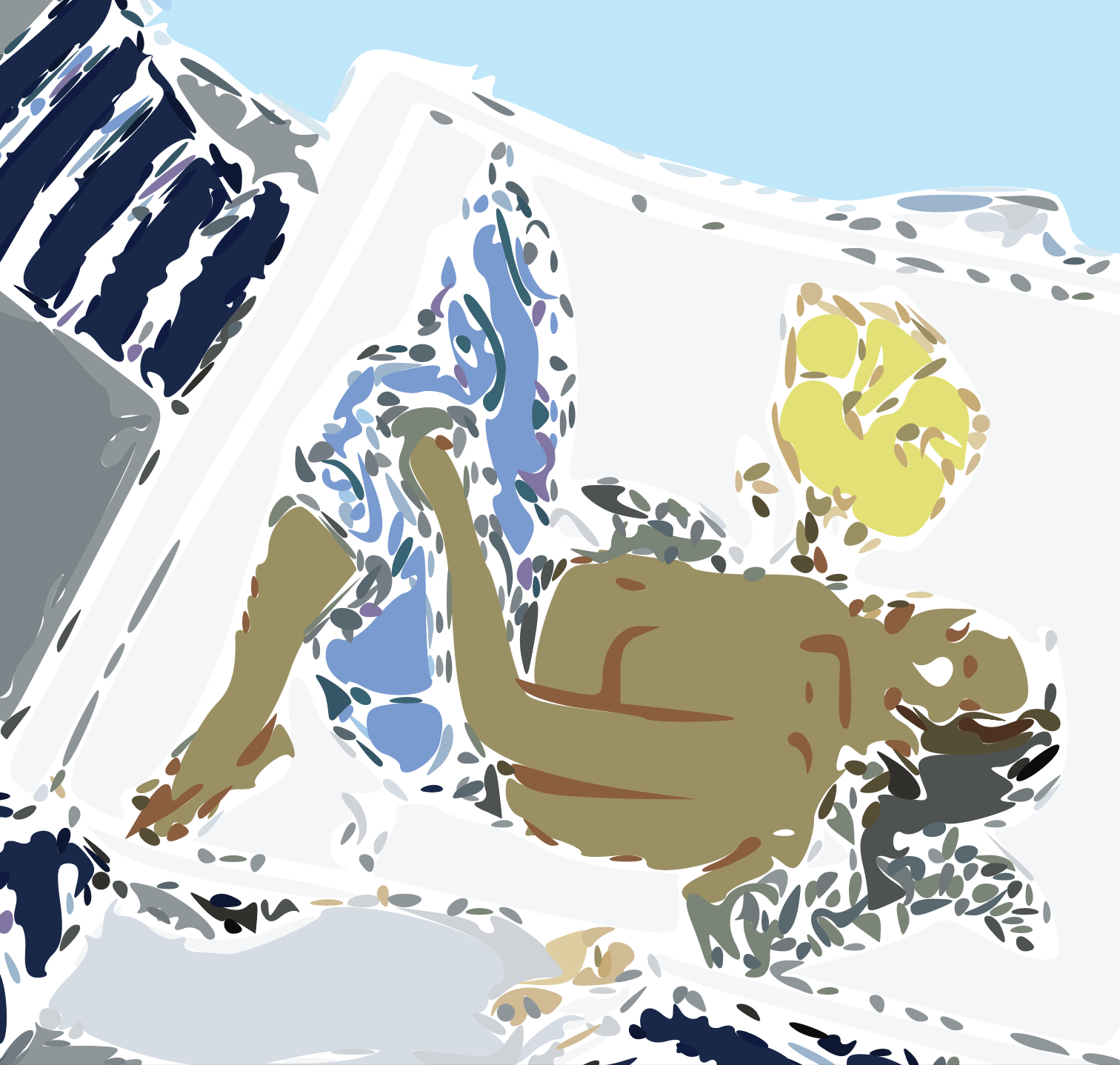}\\
$N=71$&$N=362$&$N=6565$&$N=1554$&$N=2539$
\end{tabular}
\caption{Visual comparison with state-of-the-art software in a zoomed-in region of Figure~\ref{fig_gain_compare1} (a). The results of the region merging methods are produced by setting $N^*=350$. For the other methods, default parameters are used. For them the region number is not a controllable parameter. In this region, BG behaves similarly as Area. The number of regions contained in the whole vector graphic ($N$) is reported below each figure. }\label{fig_SOTA_1_zoom}
\end{figure}

\begin{figure}
\centering
\begin{tabular}{c@{\hspace{2pt}}c@{\hspace{2pt}}c@{\hspace{2pt}}c@{\hspace{2pt}}c}
(a)&(b)&(c)&(d)&(e)\\
\includegraphics[width=0.18\textwidth]{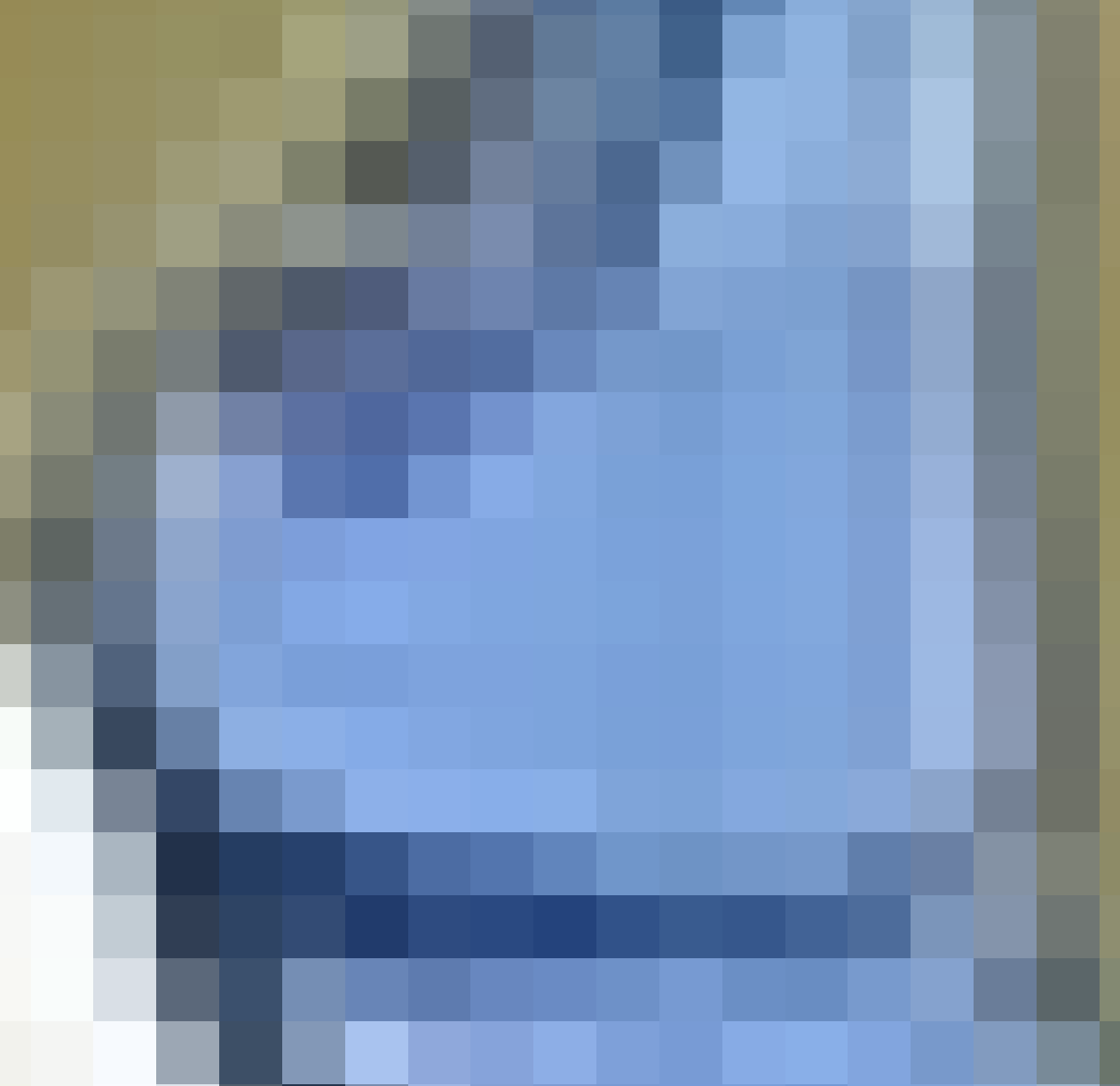}&
\includegraphics[width=0.18\textwidth]{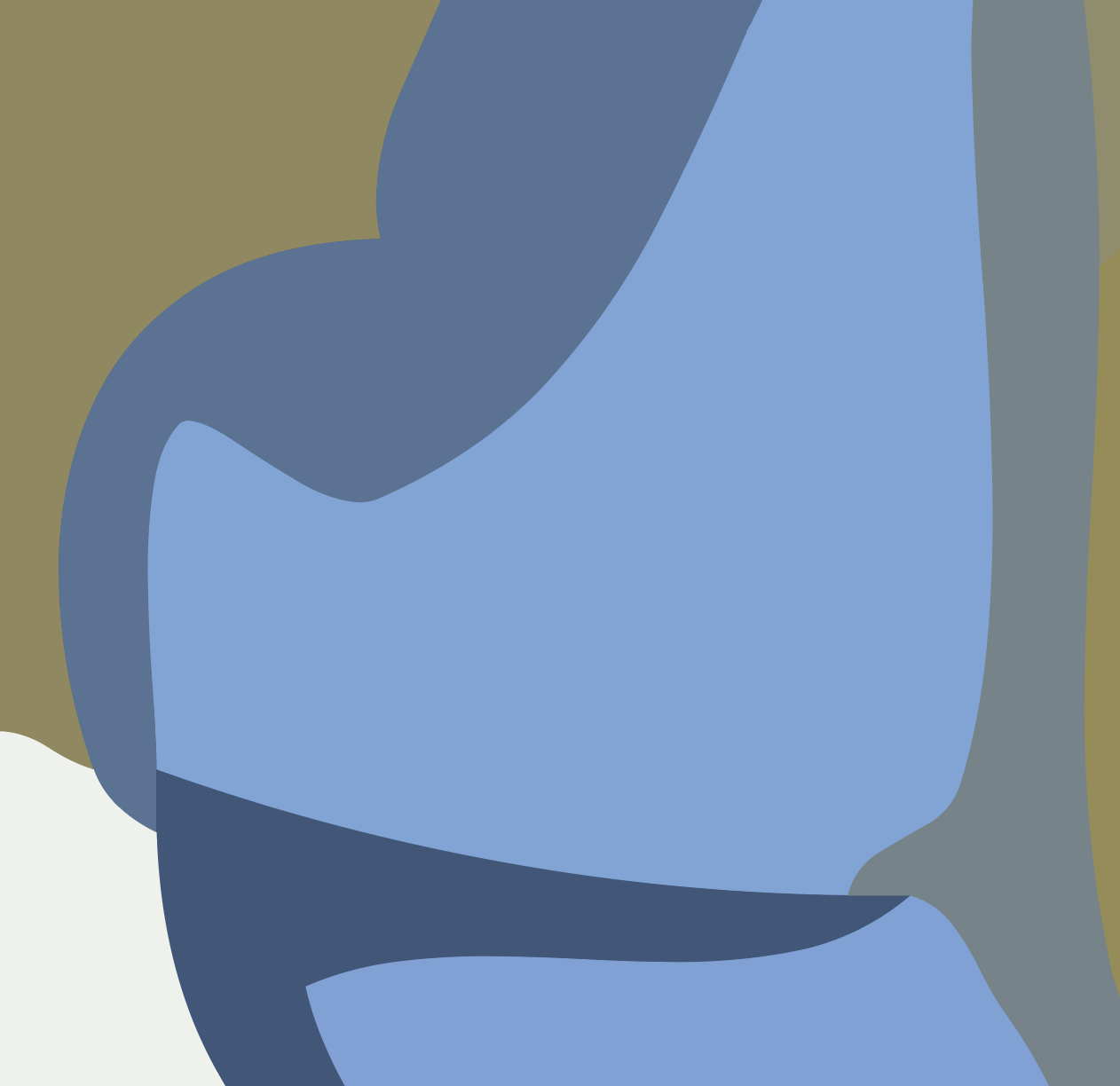}&
\includegraphics[width=0.18\textwidth]{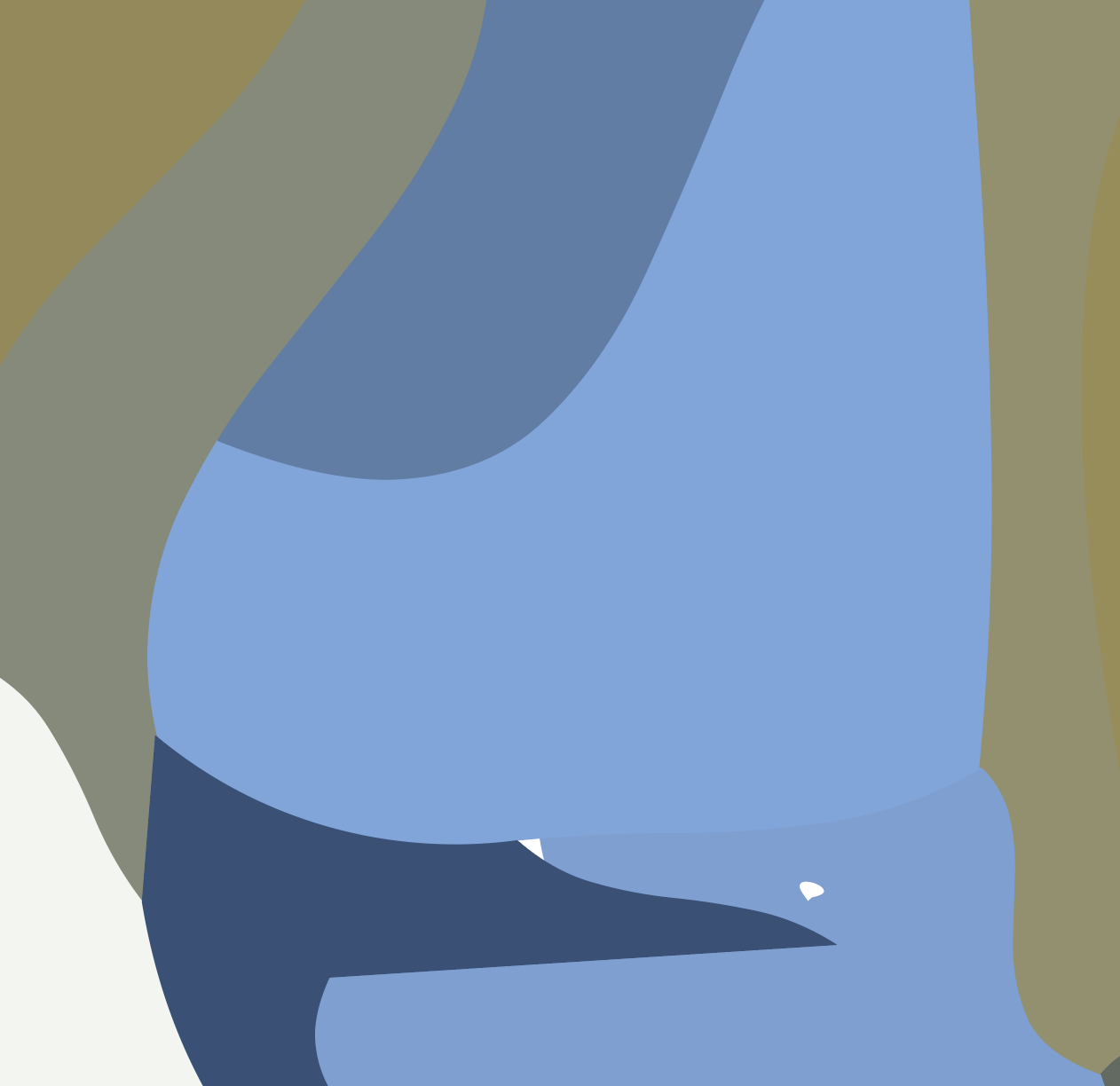}&
\includegraphics[width=0.18\textwidth]{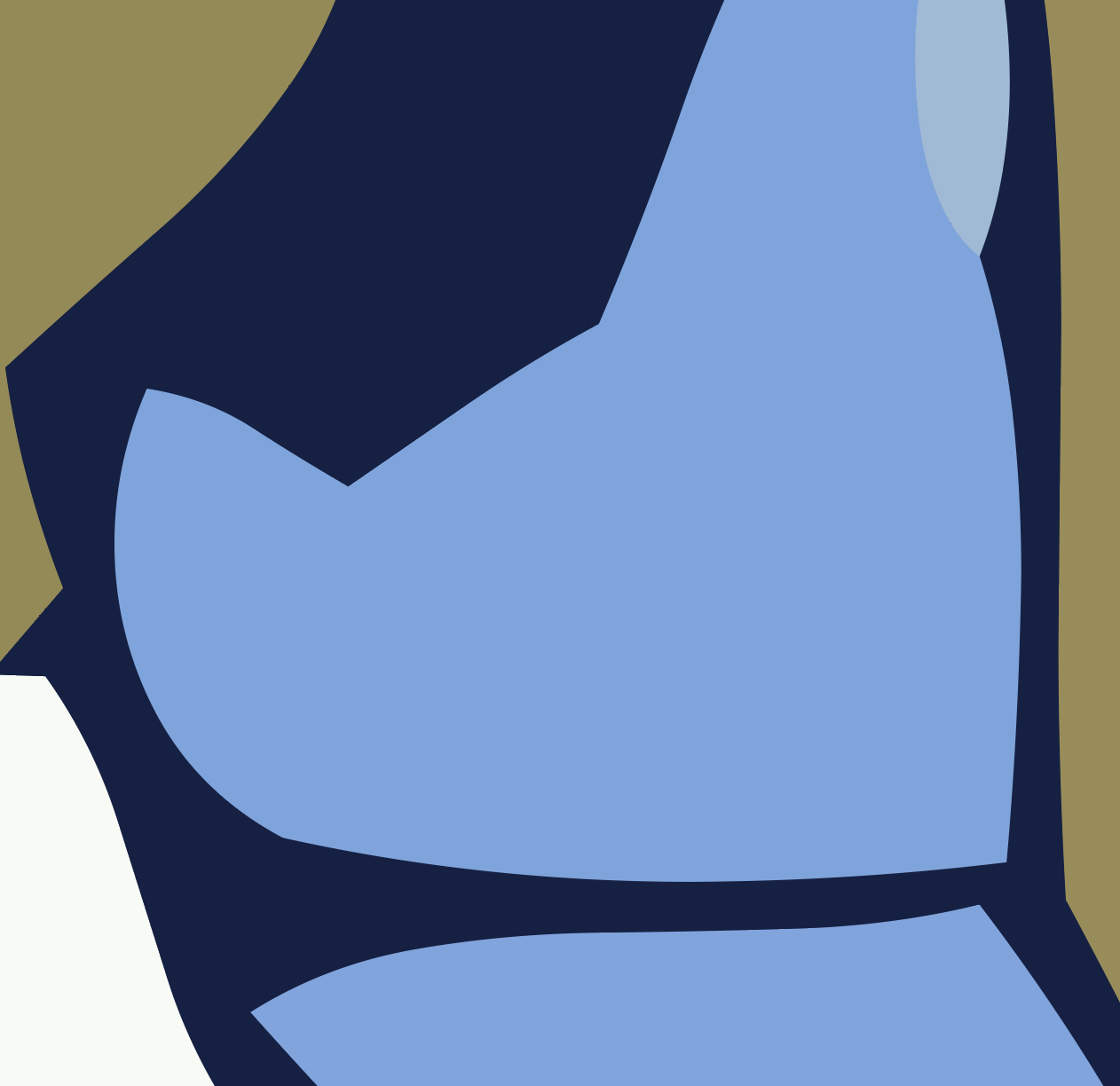}&
\includegraphics[width=0.18\textwidth]{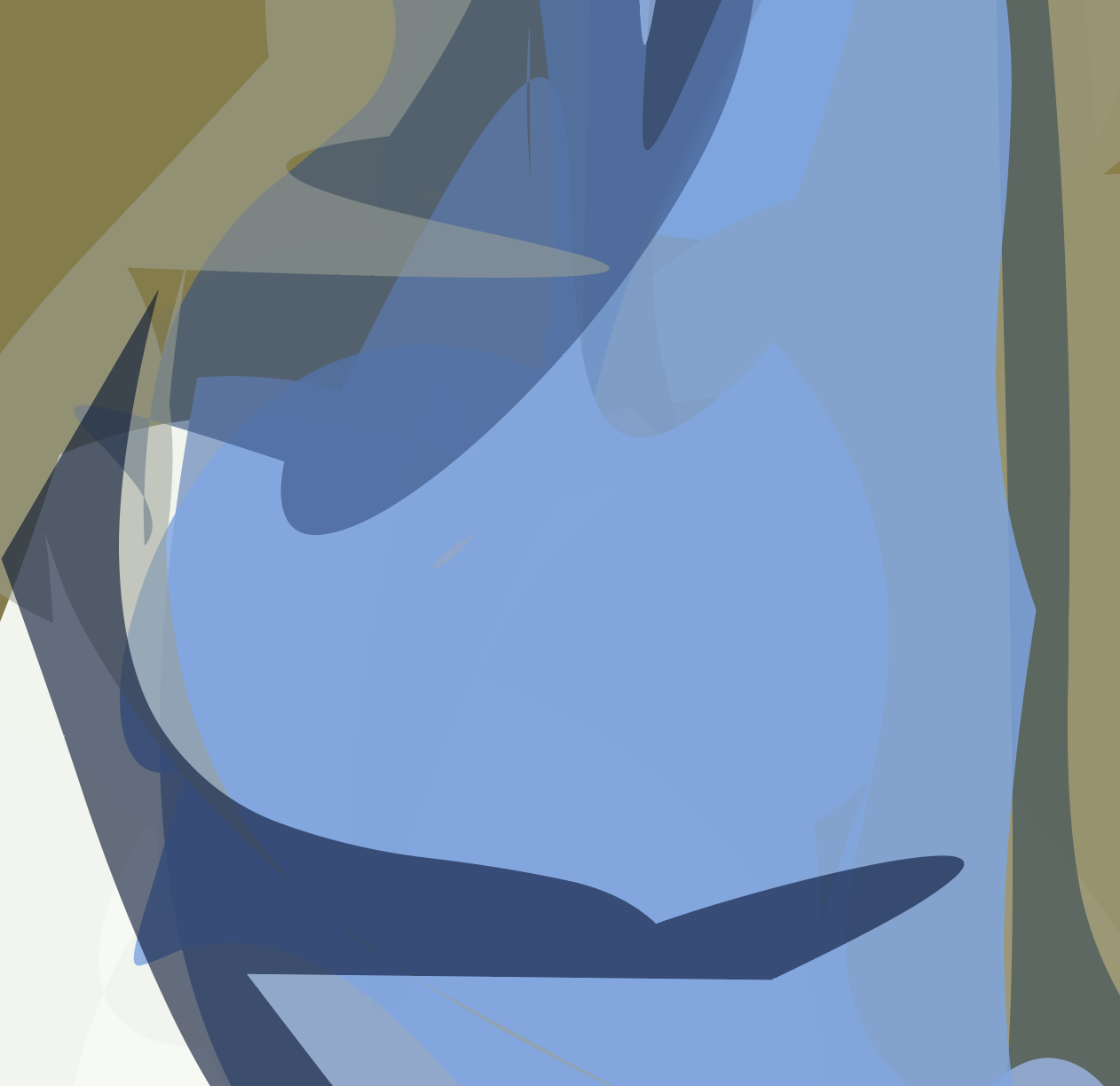}
\end{tabular}
\caption{Detailed comparisons among (a) the raster input, (b) Area, (c) AI with automatic simplication, (d) VM high, and (e) DiffVG. In this region, Area removes the pixelation effects in (a) while avoiding gaps as in (c), illusory shapes and deviating colors as in (d), and irregular overlapping shapes as in (e). 
}\label{fig_detail_SOTA}
\end{figure}

\textbf{Qualitative comparison.} We tested the aforementioned methods and our proposed algorithm using different criteria on the  image in Figure~\ref{fig_gain_compare1} (a) and show the comparisons in a zoomed-in region in Figure~\ref{fig_SOTA_1_zoom}. We observe that Area region merging is more natural and faithful than the other results while using small number of regions. Although BG keeps many details, it does not preserve geometric regularity such as the continuity of  strokes. Both Scale and MS region merging lead to oversimplification where many elongated strokes are missed. For VM High and Medium, we see that contour strokes are sharp and visually appealing; however, their colors severely deviate from those of the input. We note that VM Medium also wrongly captures the eye as  a disk. In VM Low, many details are lost, and some regions are filled with wrong colors, similarly as Scale and MS.   The smoothness of AI's result is close to that of Area, yet many details such as textures of the hair, pupil of the left eye, and the contour line along the cheek are missed. Containing more regions, VAI renders sharper and more faithful line drawings compared to VM; however,  the regions with smooth variation are not well represented.  The results by VZ with different levels of details are less satisfying. They can create gaps between regions, colors are inaccurate, and fine-scale details are generally not well preserved. The result by DiffVG is visually close to AI, and the result by TOP exhibits false colors due to color quantization. Finally, both IS and VZ max present noticeable gaps among regions, yielding unpleasant results.

To further highlight the strength of our proposed Area region merging method, we compare Area with some representative methods in more details. Figure~\ref{fig_detail_SOTA} (a) shows a zoomed-in region of Figure~\ref{fig_gain_compare1} (a), and (b)-(e) show results from Area, AI, VM, and DiffVG. For AI, we applied the automatic simplification, and for VM, we chose the mode High.  We have observed that IS and VZ max create undesirable gaps among regions in Figure~\ref{fig_SOTA_1_zoom}. In (c), we show that such gaps can also appear in AI if simplification or stronger smoothing is employed. In (d), we see that VM renders false geometries and colors which differ from (a). In (e), we see that DiffVG produces transparent and overlapping shapes with irregular contours to minimize overall $L_2$ loss; however, the resulting elements generally do not align with the objects. Area, as shown in (a), avoids the aforementioned defects while successfully removing pixelation effects and faithfully preserving key features of the input image. 

\begin{figure}
\centering
\begin{tabular}{c@{\hspace{2pt}}c@{\hspace{2pt}}c}
(a)&(b)&(c)\\
\includegraphics[width=0.3\textwidth]{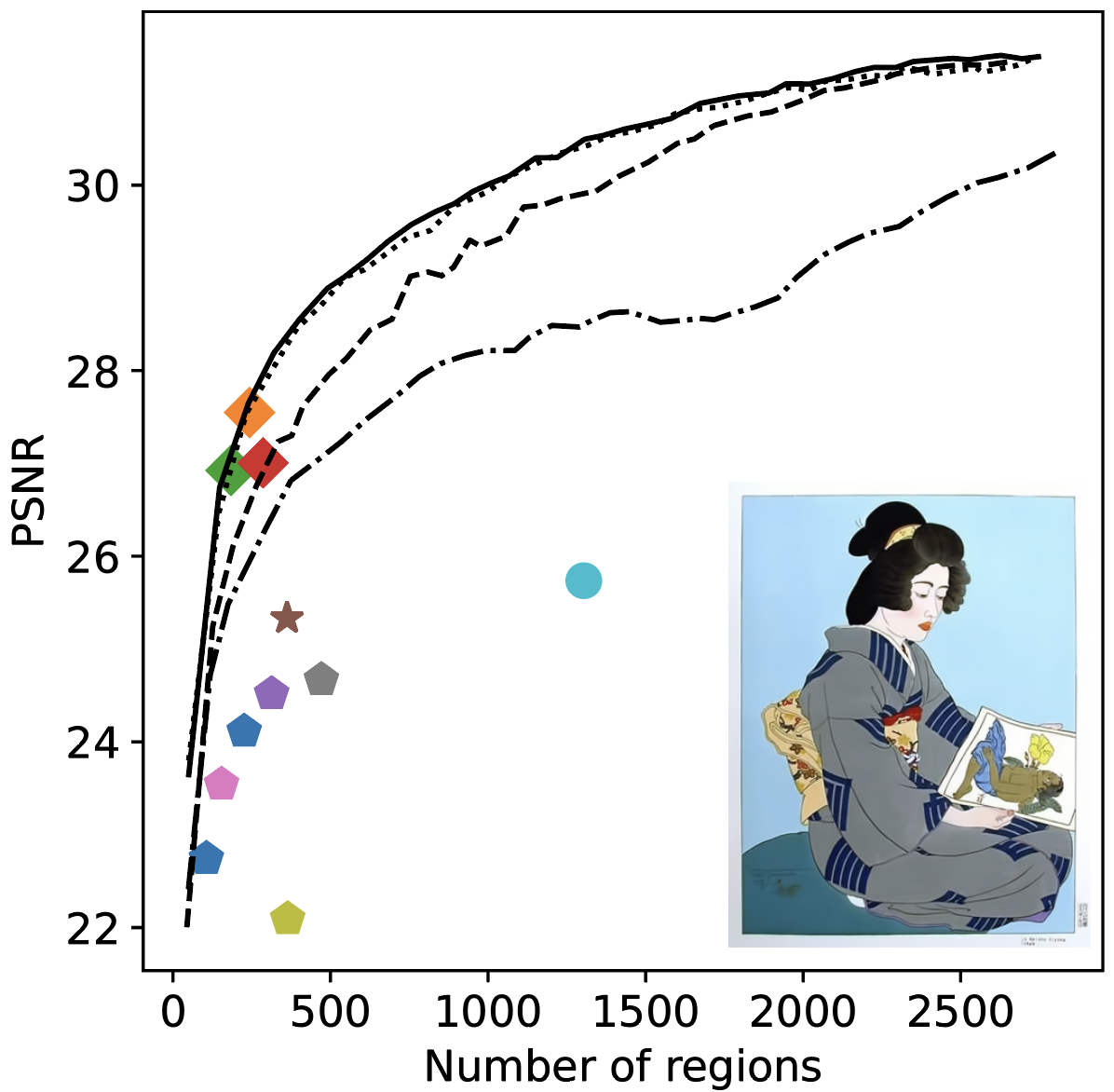}&\includegraphics[width=0.3\textwidth]{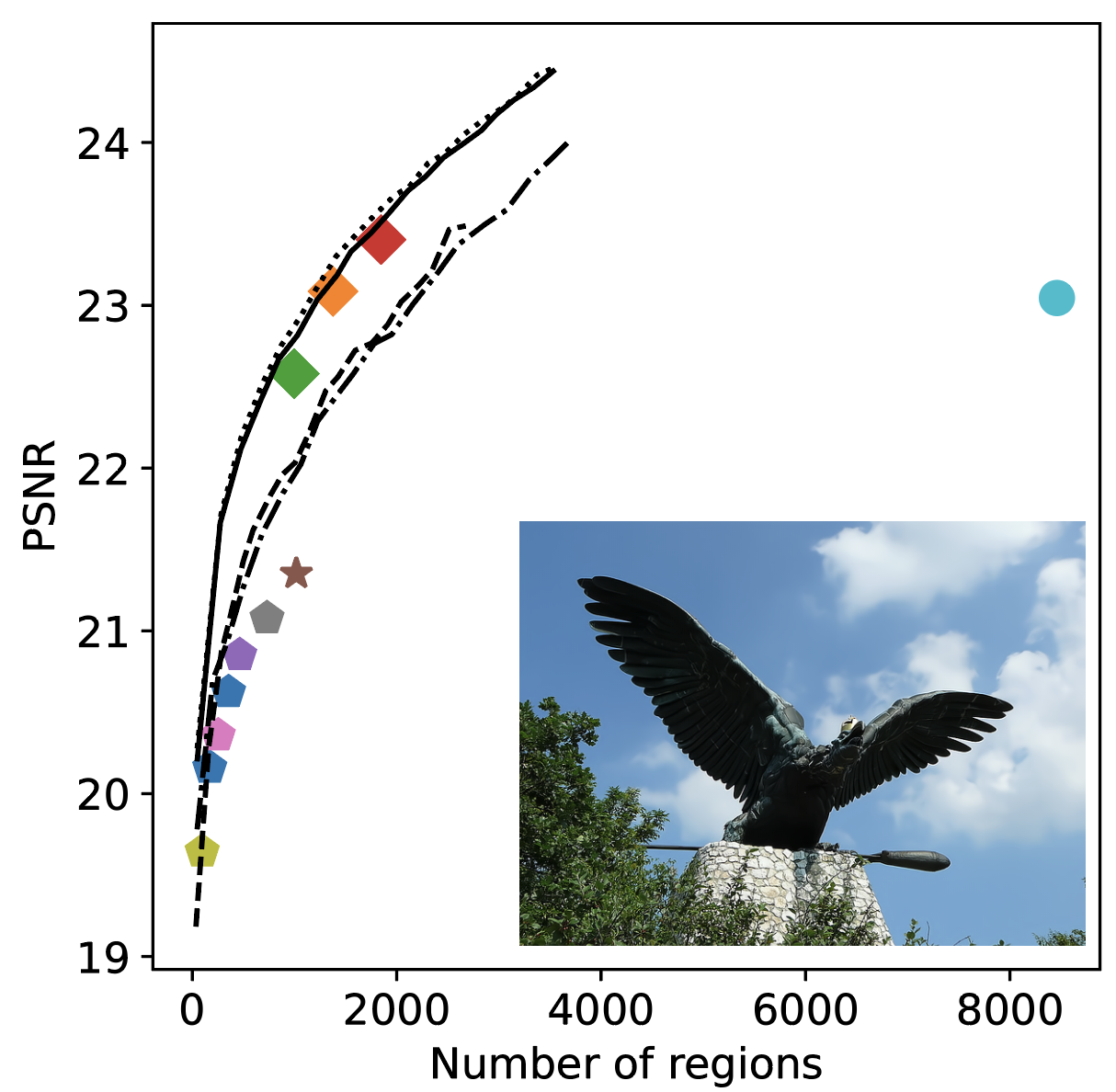}&\includegraphics[width=0.31\textwidth]{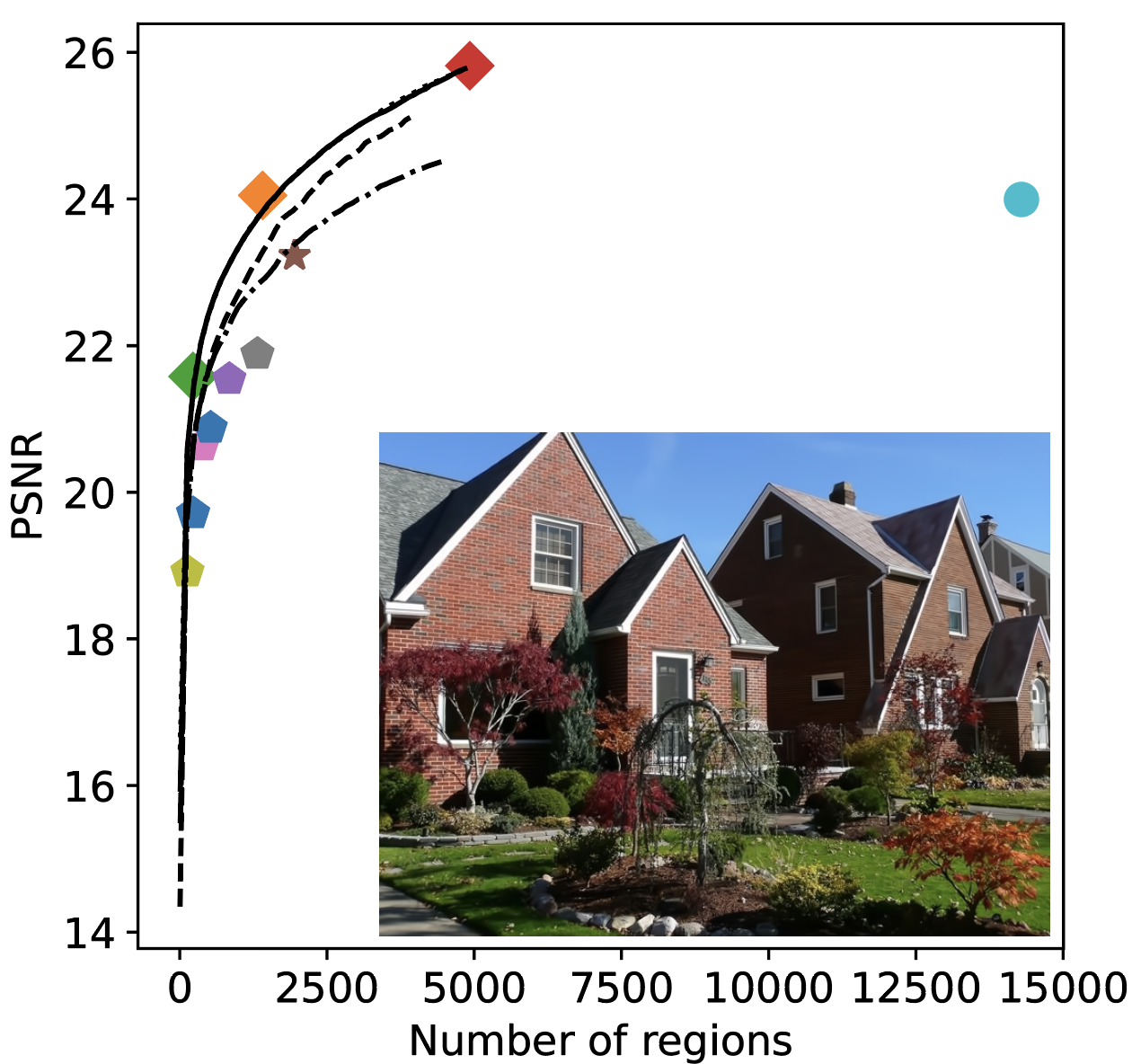}\\
\multicolumn{3}{c}{\includegraphics[width=0.6\textwidth]{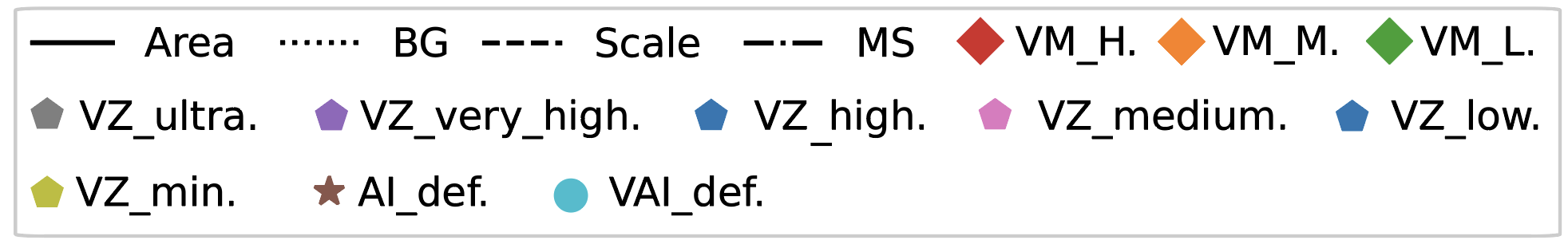}}
\end{tabular}
\caption{Quantitative comparison between the proposed methods (MS, BG, Scale, and Area) with state-of-the-art software. 
}\label{fig_SOTA_quant}
\end{figure}

\textbf{Quantitative comparison: similarity and complexity.} We quantitatively examine these methods by focusing on the similarity between the rasterized vector graphic and the raster input, and the complexity, i.e., the number of regions ($N$) in the result. 
Figure~\ref{fig_SOTA_quant} collectively shows three sets of comparison where the $x$-axis and $y$-axis correspond to the number of regions and the PNSR of the output, respectively. For these examples, results from VZ, AI, and VAI are consistently worse compared to the others. In particular, VAI always invokes great numbers of regions, but the results are not faithful to the inputs.  For the examples in (a) and (b),  Area region merging reaches the highest PSNR with the same number of regions;  VM High can achieve similar levels of accuracy, and they are generally higher than Scale region merging and MS region merging. In (c), we show an image where VM consistently yield vector graphics with high qualities which are comparable with Area and BG. We remark that although PSNR is used in this work for evaluating the accuracy, rasterizing vector graphics can underestimate the significance of the Area region merging. For example, removing small regions near boundaries helps to produce visually appealing vector graphics, yet this feature cannot be correctly evaluated after rasterization due to antialiasing. 

\section{Conclusion}~\label{sec::conclude}
In this work, we explored a generic image vectorization approach combining region merging and curve smoothing. We endowed each image partition with a dual graph linking all pairs of neighboring regions,  and a primal graph  linking pairs of vertices separated by a single region boundary. Region merging, acting on the dual graph, leverages local color and geometric regularity that are crucial for establishing relations among otherwise unorganized pixel data. Curve smoothing, acting on the  primal graph, yields resolution-independent representations for the perceived boundaries. We introduced four merging criteria for the dual step and analyzed the topological consistency issue related to networks of curves evolving under the affine shortening flow. We conducted extensive experiments to validate our models including qualitative and comparison studies. Our results show that the proposed method exhibits interpretable behaviors which can be fine-tuned by user-friendly hyperparameters. Moreover, we concluded from the comparison with color quantization-based approaches that local regularity is critical to the simplicity and visual quality of the vectorized representation. We also illustrated by comparison with state-of-the-art software that the proposed approach is more robust, general, and consistent. Combining the evidence from all experiments, we find that among the four merging schemes, the Area merging method  yields the most reliable and faithful results.

\appendix

\section{Proof of Proposition~\ref{prop_var_equation}}\label{proof_var_equation}
We prove Proposition~\ref{prop_var_equation} in the case of one-dimension, and the general cases are trivially generalized. First we note that
 \begin{align}
     \int_{O_i\cup O_j}f(\by)\,d\by=|O_i|\overline{f}(O_i)+|O_j|\overline{f}(O_j) \; .  \label{eq_temp1}
 \end{align}
 Hence
 \begin{align}
 \text{Var}(f,O_i\cup O_j)&=\int_{O_i\cup O_j}\left(f(\bx)-|O_i\cup O_j|^{-1}\int_{O_i\cup O_j}f(\by)\,d\by\right)^2\,d\bx\nonumber\\
 &=\int_{O_i\cup O_j}\left(f(\bx)-\frac{|O_i|}{|O_i|+|O_j|}\overline{f}(O_i)-\frac{|O_j|}{|O_i|+|O_j|}\overline{f}(O_j)\right)^2\,d\bx\nonumber\\
 &=\int_{O_i\cup O_j} f^2(\bx)\,d\bx-2\frac{|O_i|\overline{f}(O_i)}{|O_i|+|O_j|}\int_{O_i\cup O_j}f(\bx)\,d\bx\nonumber\\
 &-2\frac{|O_j|\overline{f}(O_j)}{|O_i|+|O_j|}\int_{O_i\cup O_j}f(\bx)\,d\bx\nonumber\\
 &+2\frac{|O_i|\cdot|O_j|}{|O_i|+|O_j|}\overline{f}(O_i)\overline{f}(O_j)+\frac{|O_i|^2}{|O_i|+|O_j|}\overline{f}^2(O_i)+\frac{|O_j|^2}{|O_i|+|O_j|}\overline{f}^2(O_j) \; .\label{eq_temp2}
 \end{align}
By~\eqref{eq_temp1}, we observe that
\begin{align*}
\frac{|O_i|\overline{f}(O_i)}{|O_i|+|O_j|}\int_{O_i\cup O_j}f(\bx)\,d\bx=\frac{|O_i|^2\overline{f}^2(O_i)+|O_i|\cdot|O_j|\overline{f}(O_i)\overline{f}(O_j)}{|O_i|+|O_j|},
\end{align*}
and by symmetry, we have
\begin{align*}
&\frac{|O_i|\overline{f}(O_i)}{|O_i|+|O_j|}\int_{O_i\cup O_j}f(\bx)\,d\bx+\frac{|O_j|\overline{f}(O_j)}{|O_i|+|O_j|}\int_{O_i\cup O_j}f(\bx)\,d\bx\\
&=\frac{|O_i|^2\overline{f}^2(O_i)+|O_j|^2\overline{f}^2(O_j)+2|O_i|\cdot|O_j|\overline{f}(O_i)\overline{f}(O_j)}{|O_i|+|O_j|} .
\end{align*}
Plugging this back to~\eqref{eq_temp2} gives
\begin{align*}
\text{Var}(f,O_i\cup O_j)&=\int_{O_i\cup O_j} f^2(\bx)\,d\bx-\frac{|O_i|^2\overline{f}^2(O_i)+|O_j|^2\overline{f}^2(O_j)+2|O_i|\cdot|O_j|\overline{f}(O_i)\overline{f}(O_j)}{|O_i|+|O_j|}.
\end{align*}
On the other hand, we have
\begin{align*}
    \text{Var}(f,O_i)&= \int_{O_i}\left(f(\bx)-\overline{f}(O_i)\right)^2\,d\bx\\
    &=\int_{O_i}f^2(\bx)\,dx-2\overline{f}(O_i)\int_{O_i}f(\bx)\,dx+|O_i|\overline{f}^2(O_i)\\
    &=\int_{O_i}f^2(\bx)\,dx-|O_i|\overline{f}^2(O_i) .
\end{align*}
By symmetry, we have
\begin{align*}
    \text{Var}(f,O_i)+\text{Var}(f,O_j)=\int_{O_i\cup O_j}f^2(\bx)\,dx-|O_i|\overline{f}^2(O_i)-|O_j|\overline{f}^2(O_j).
\end{align*}
Therefore,
\begin{align*}
&\text{Var}(f,O_i\cup O_j)-\text{Var}(f,O_i)-\text{Var}(f,O_j)\\
&=|O_i|\overline{f}^2(O_i)+|O_j|\overline{f}^2(O_j)-\frac{|O_i|^2\overline{f}^2(O_i)+|O_j|^2\overline{f}^2(O_j)+2|O_i|\cdot|O_j|\overline{f}(O_i)\overline{f}(O_j)}{|O_i|+|O_j|}\\
&=\frac{|O_i|\cdot|O_j|\overline{f}^2(O_i)+|O_i|\cdot|O_j|\overline{f}^2(O_j) -2|O_i|\cdot|O_j|\overline{f}(O_i)\overline{f}(O_j)}{|O_i|+|O_j|}\\
&=\frac{|O_i|\cdot|O_j|\left(\overline{f}(O_i)-\overline{f}(O_j)\right)^2}{|O_i|+|O_j|} 
\end{align*}
which completes the proof. $\square$

\section{Proof of Proposition~\ref{prop_scale_neighbor}}\label{proof_scale_neighbor}
A partition $\mathcal{P}$ is 2-normal for the Scale region merging if for any adjacent regions, say $O_1,O_2$, we have
\begin{align}
	\frac{|O_1|\cdot|O_2|}{|O_1|+|O_2|}\omega_f^2>\lambda \left(\frac{|\partial O_1|}{|O_1|}+\frac{|\partial O_2|}{|O_2|}-\frac{|\partial (O_1\cup O_2)|}{|O_1\cup O_2|}\right)
\end{align}
where we used Proposition~\ref{prop_var_equation} for simplifying the numerator of~\eqref{eq_Delta_E} and defined $\omega^2_f = \sum_{i=1}^d\left(\sup(f_i)-\inf(f_i)\right)^2$ as the oscillation of $f$ over $\Omega$, where $f_i:\Omega \to\mathbb{R}$ is the $i$-th channel. By the following elementary inequality
$$\frac{a}{b}\leq \frac{a+d}{b+c}\leq \frac{d}{c}$$
for any positive numbers $a,b,c,d$ with $a/b\leq d/c$, with the equality holding if and only if $a/b=d/c$, we have
\begin{align*}
\frac{|\partial O_1|}{|O_1|}+\frac{|\partial O_2|}{|O_2|}-\frac{|\partial (O_1\cup O_2)|}{|O_1\cup O_2|}&\geq \min(\frac{|\partial O_1|}{|O_1|},\frac{|\partial O_2|}{|O_2|})+\frac{|\partial O_1|+|\partial O_2|-|\partial (O_1\cup O_2)|}{|O_1\cup O_2|}\\
&=\min(\frac{|\partial O_1|}{|O_1|},\frac{|\partial O_2|}{|O_2|})+\frac{2|\partial O_1\cap \partial O_2|}{|O_1\cup O_2|} .
\end{align*}

Now consider an arbitrary region $O$ and any of its neighbor $O'$. From what we have proved above, we see that
\begin{align}
	\frac{\omega_f^2}{\lambda}|O|>\min(\frac{|\partial O|}{|O|},\frac{|\partial O'|}{|O'|}) +\frac{2|\partial O\cap \partial O')|}{|O\cup O'|} .
\end{align}
Denote  the set of neighboring regions of $O$ by $\mathcal{N}(O)$ and $N(O)=|\mathcal{N}(O)|$ as the number of neighbors. Summing up for all the neighbors of $O$ gives
\begin{align}
	\frac{\omega_f^2}{\lambda}N(O)|O|&>\sum_{O'\in\mathcal{N}(O)}\min(\frac{|\partial O|}{|O|},\frac{|\partial O'|}{|O'|})+\sum_{O'\in\mathcal{N}(O)}\frac{2|\partial O\cap \partial O')|}{|O\cup O'|} \nonumber\\
	&\geq\frac{\sum_{O'\in\mathcal{N}(O)}\min(|\partial O|,|\partial O'|)}{|\Omega|}+\frac{2|\partial O|}{|\Omega|}\nonumber\\
	&\geq \frac{\sum_{O'\in\mathcal{N}(O)}|\partial O\cap \partial O'|}{|\Omega|}+\frac{2|\partial O|}{|\Omega|} =\frac{3|\partial O|}{|\Omega|}\;,
 \label{eq_temp2}
\end{align}  
which concludes the proof. $\square$

\section{Proof of Theorem~\ref{thm_scale_n_region}}\label{proof_scale_n_region}
Applying the isoperimetric inequality, the relation~\eqref{eq_temp2} implies that the number of neighbors of $O$ is bounded from below
	$$N(O)>\frac{3C\lambda}{|\Omega|\omega_f^2\sqrt{|O|}}$$
	where $C$ denotes the isoperimetric constant for plane. Observe that the number of  regions with area smaller than $3|\Omega|/(\# \mathcal{P})$ must exceed $(\# \mathcal{P})/3$, and each of which has at least 
	$$\frac{C\lambda\sqrt{3(\# \mathcal{P})}}{|\Omega|^{3/2}\omega_f^2}$$
	neighboring regions. This implies that the number of edges in $\mathcal{P}$ is greater than
	$$\frac{(\# \mathcal{P})}{4}\cdot\frac{C\lambda\sqrt{3(\# \mathcal{P})}}{|\Omega|^{3/2}\omega_f^2}$$
	which is smaller than $3(\# \mathcal{P})$ according to Lemma 3.4 in~\cite{koepfler1994multiscale}. Therefore, we deduce that 
	$$\# \mathcal{P}<\frac{48|\Omega|^3\omega_f^4}{C^2\lambda^2}$$
	which proves the theorem. $\square$

\section{Proof of Proposition~\ref{prop_max_time}}\label{proof_max_time}

\begin{figure}
\centering
\begin{tabular}{cccc}
(a)&(b)&(c)&(d)\\
\includegraphics[width=0.22\textwidth]{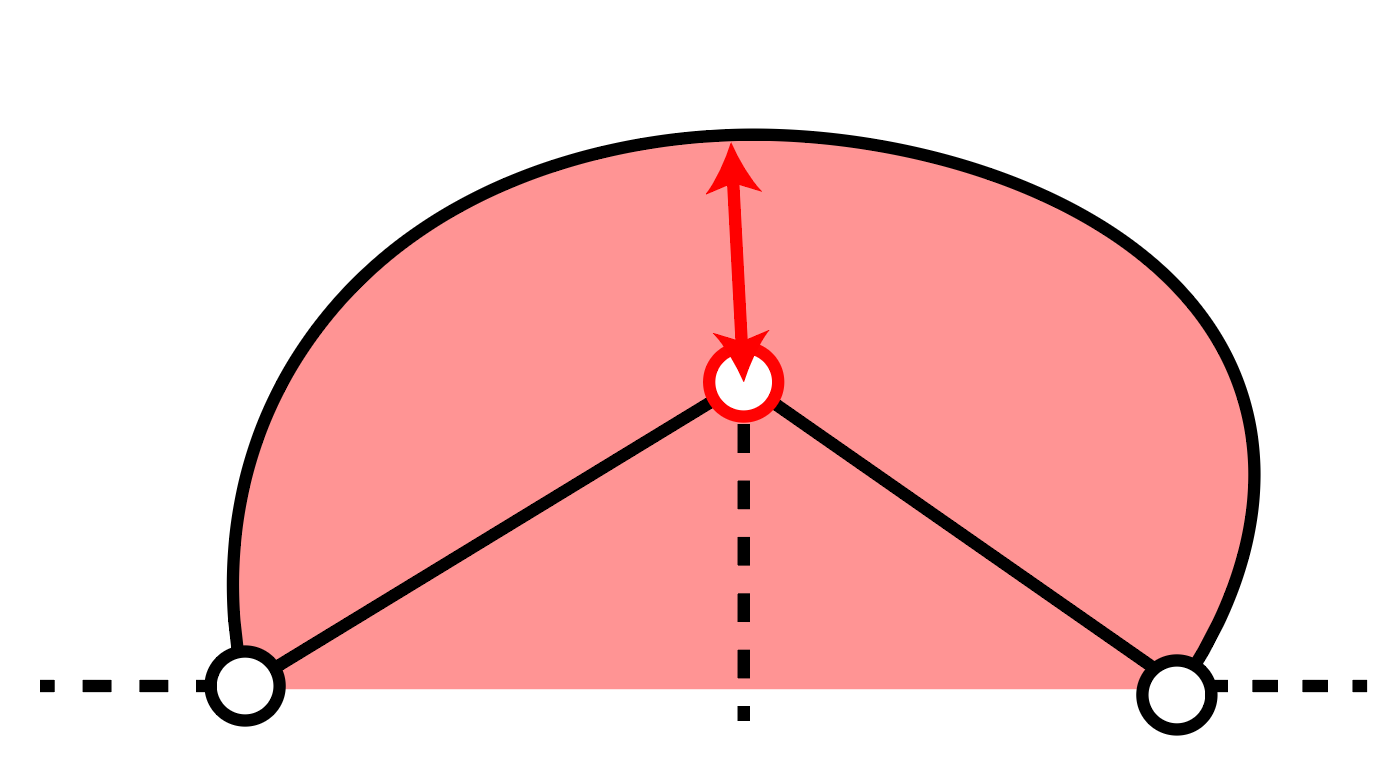}&
\includegraphics[width=0.22\textwidth]{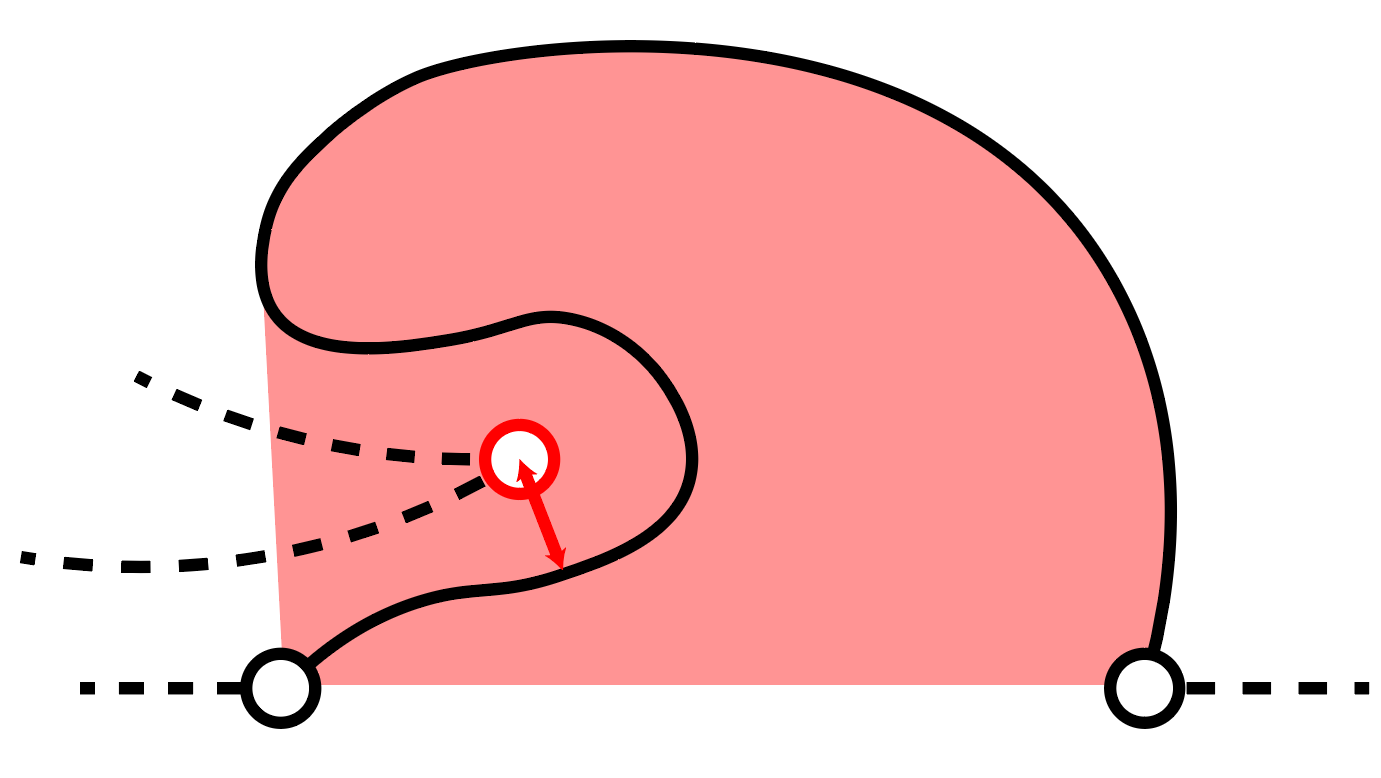}&
\includegraphics[width=0.22\textwidth]{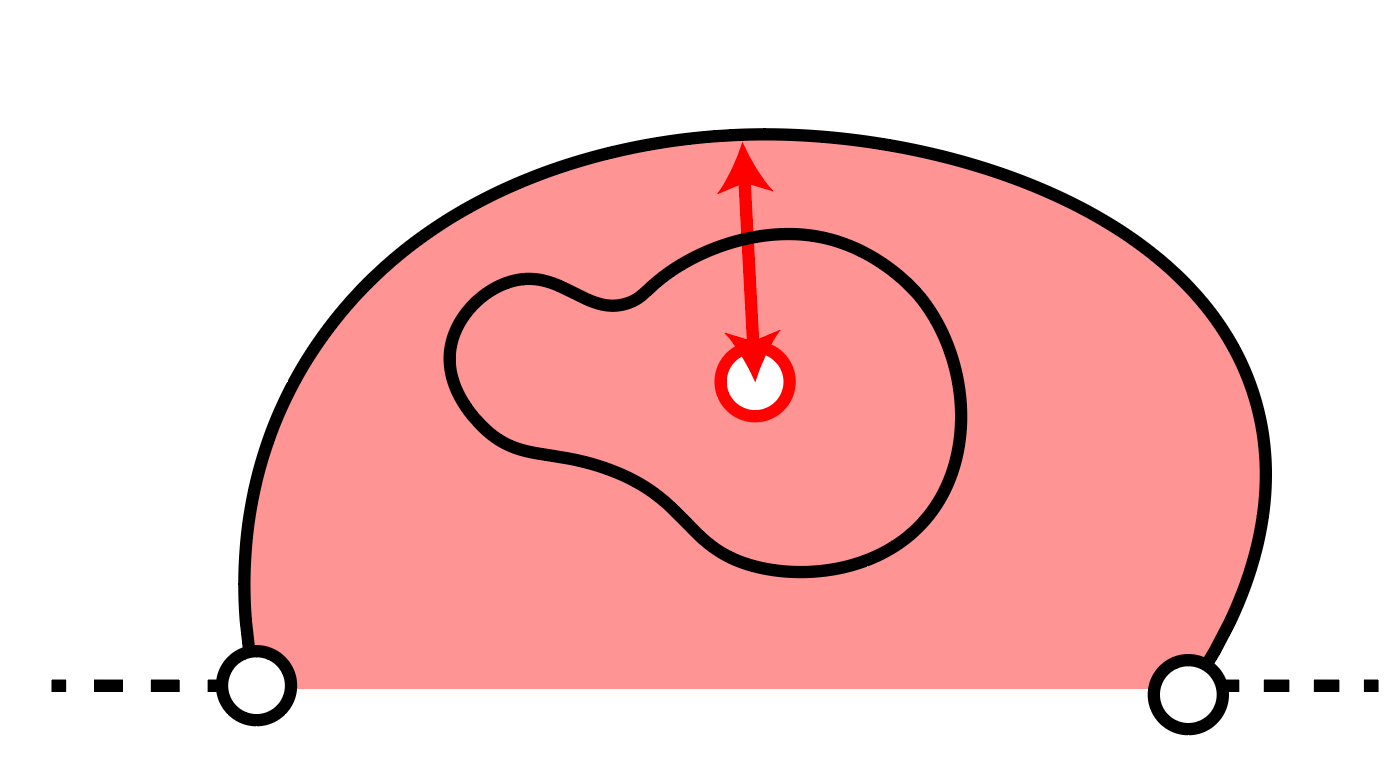}&
\includegraphics[width=0.22\textwidth]{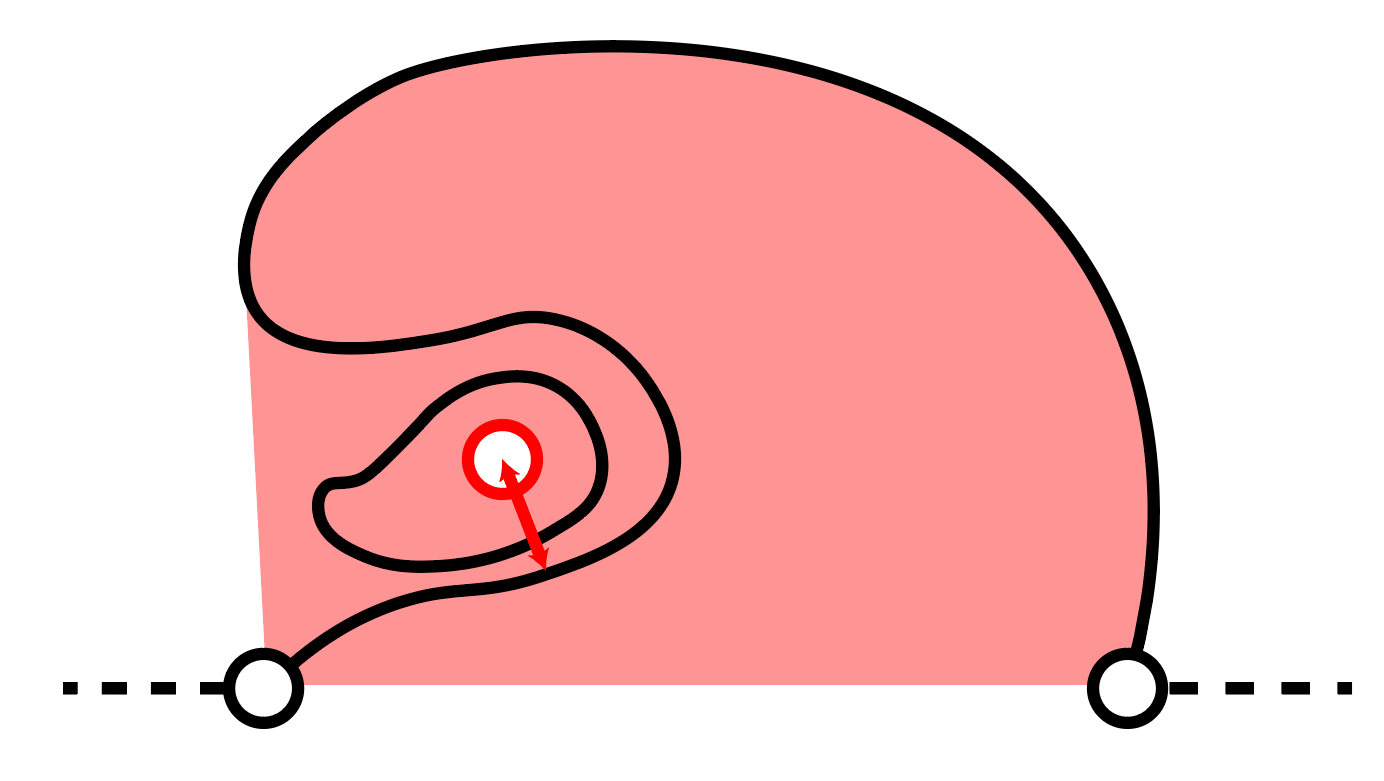}
\end{tabular}
\caption{Illustration of points (red circled dots) obstructing the long-time existence of networks of curves evolving according to the affine shortening flow. Circled dots connected to curves represent junction points, and the isolated circled dots in (c) and (d) represent contraction points. In each plot, the red region is the convex hull of the area enclosed by the open curve ending at two junction points in the bottom and the line segments between them. The Hausdorff distance between the red circled dot $\mathbf{P}$ and the solid open curve $\mathcal{C}$ affects the longest possible evolution time for $\mathcal{C}$ so that it does not touch $\mathbf{P}$.  }\label{fig_dist}
\end{figure}

We show that there exists a $T>0$ such that $\Gamma_t$ remains a network of curves for $t\in [0,T)$.  Let $\mathcal{P}$ be a partition of $\Omega$ and the induced network of curves is $\Gamma=\{\cC_1,\dots,\cC_N\}$. Let $\mathcal{J}$ be the set of intersection points of pairs of curves in $\Gamma$, i.e., the set of junction points not on $\partial\Omega$. For each $n=1,2,\dots,N$, define $\overline{\cC}_n$ as the region enclosed by the Jordan curve formed by closing $\cC_n$ with a line segment joining its  endpoints. Let $\mathcal{K}$ denote the set of contraction points for all the closed curves in $\Gamma$. We define the \textit{critical distance} for the partition as
 \begin{align}
\rho(\Gamma) = \min_{n=1,2,\dots,N}\min_{\bP\in\left(\mathcal{J}\cup\mathcal{K}\right)\cap\text{conv}(\overline{\cC}_n)}\text{dist}(\bP,\cC_n)
\end{align}
where $\text{conv}(\overline{\cC}_n)$ is the convex hull for $\overline{\cC}_n$. In case $(\mathcal{J}\cup \mathcal{K})\cap\text{conv}(\overline{\cC}_n)=\varnothing$ for some $n$,   we set  $\min_{\bP\in\mathcal{J}\cap\text{conv}(\overline{\cC}_n)}\text{dist}(\bP,\cC_n)=+\infty$. See Figure~\ref{fig_dist} for illustrations.

Suppose $\Gamma$ is a network of curves, and we evolve it according to the affine shortening flow by running the Moisan's scheme~\cite{moisan1998affine}.
Note that  one step of affine erosion $E_\sigma$ that removes chordal regions with area $\sigma>0$ shifts the curve by a distance of approximately $\omega\sigma^{2/3}|\kappa|^{1/3}$~\cite{moisan1998affine}, where $\omega$ is an absolute constant, and $\kappa$ is the curvature. Moreover, affine shortening flow with a duration of $T>0$ can be approximated by a composition of $n$ affine erosions $E_{\left(T/(n\omega)\right)^{3/2}}$ for positive integer $n$~\cite{lisani2003theory}. Therefore, using Moisan's scheme approximating the affine shortening flow with a duration of $T$ shifts the curve by a distance of approximately $T|\kappa|^{1/3}$.  From this, we deduce that if
 \begin{align}T< \frac{\rho(\Gamma)}{(K(\Gamma))^{1/3}}\label{eq_time_bound1}
 \end{align}
where $K(\Gamma)$ is the maximal absolute curvature of the curves in $\Gamma$, then $\Gamma_T$ has the same topology as $\Gamma_0=\Gamma$. Moreover,  since the curvature does not increase during the evolution~\cite{sapiro1993affine}, if $\Gamma$ is pixelated, then $\rho(\Gamma)\geq 1$ and $K(\Gamma)\leq 2\sqrt{2}$  result from the three-point-formula; thus the proposition is proved. $\square$

\section{Analysis on the discrepancy of the number of regions}\label{sec_discrepancy}

\begin{figure}
\centering
\begin{tabular}{c@{\hspace{2pt}}c@{\hspace{2pt}}c}
\multicolumn{3}{c}{(a)}\\
\multicolumn{3}{c}{\includegraphics[width=0.5\textwidth]{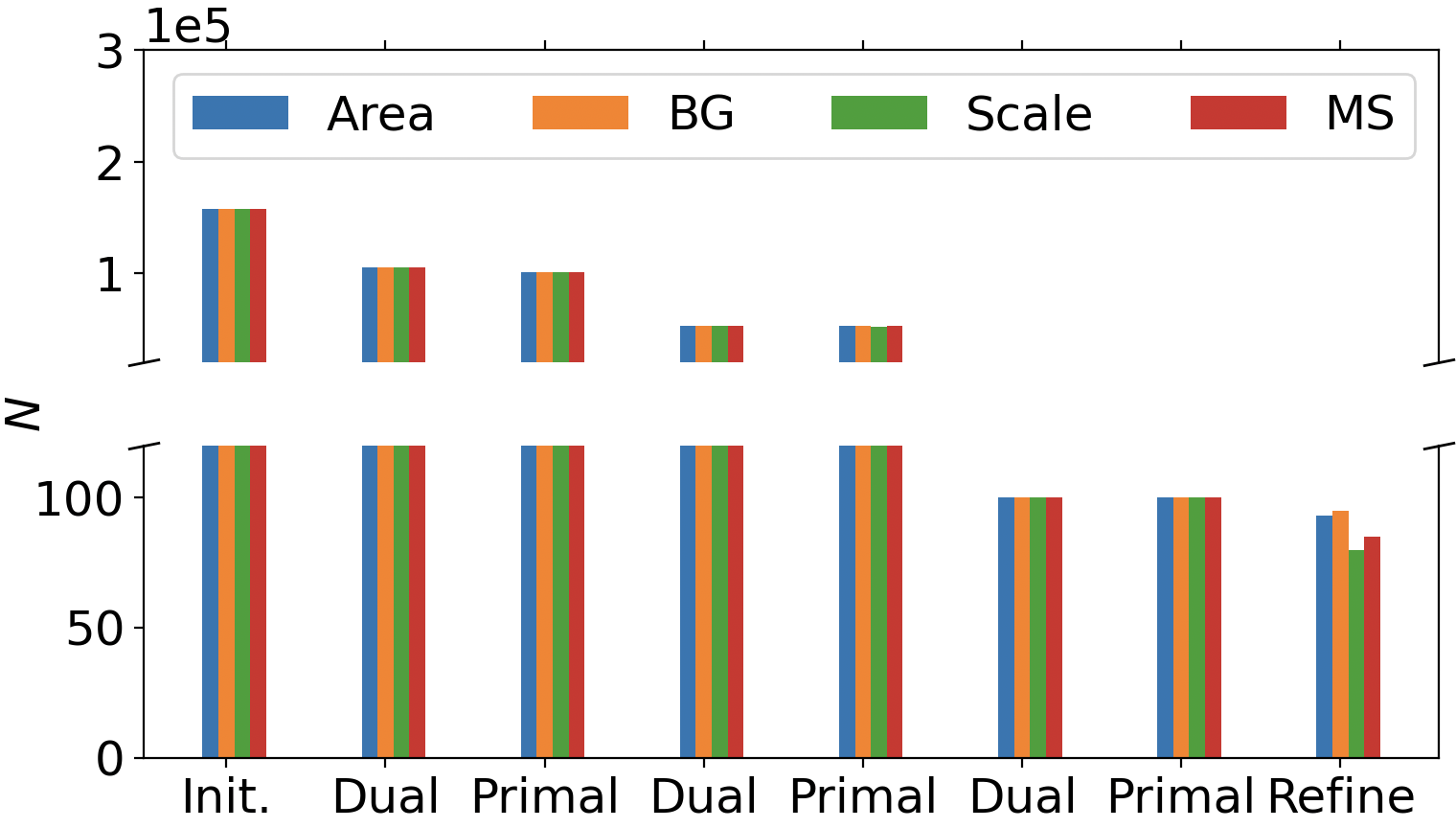}}
\\
(b)&(c)&(d)\\
\includegraphics[width=0.3\textwidth]{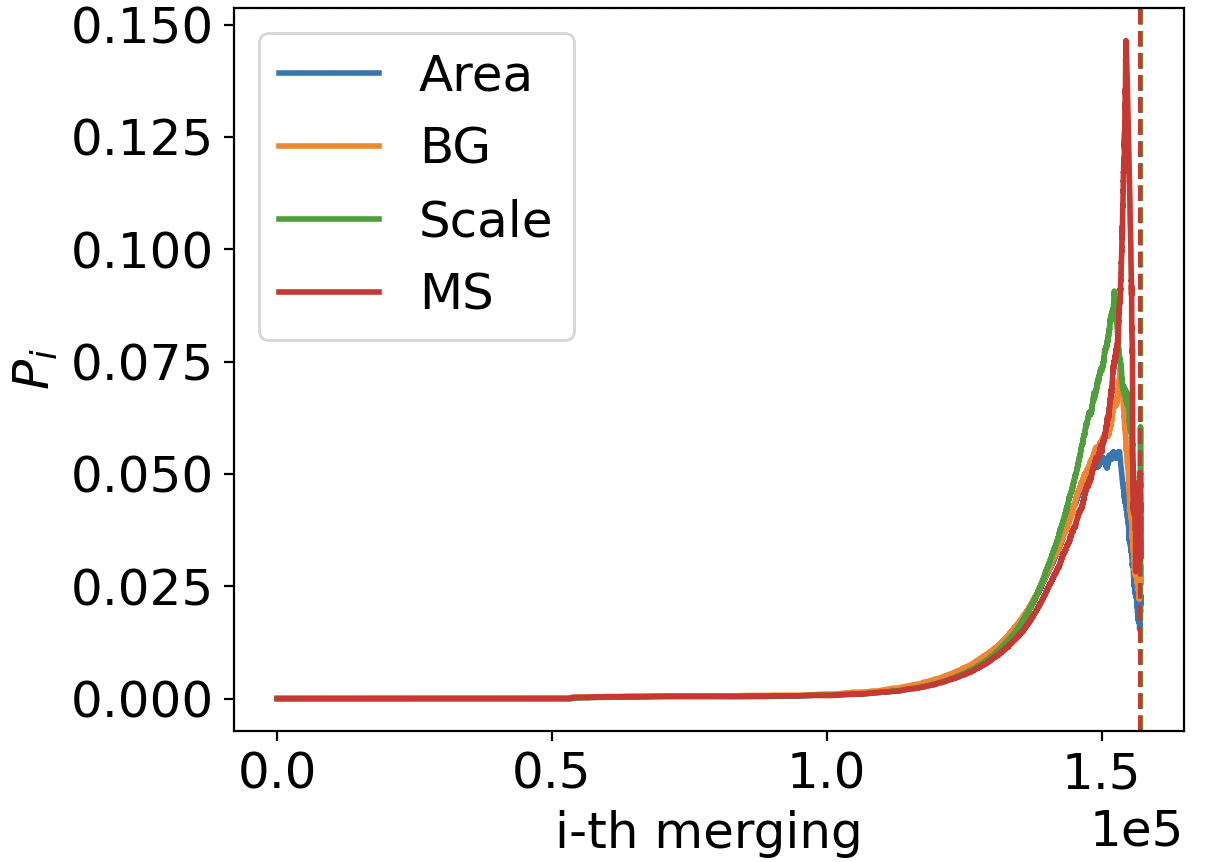}&
\includegraphics[width=0.3\textwidth]{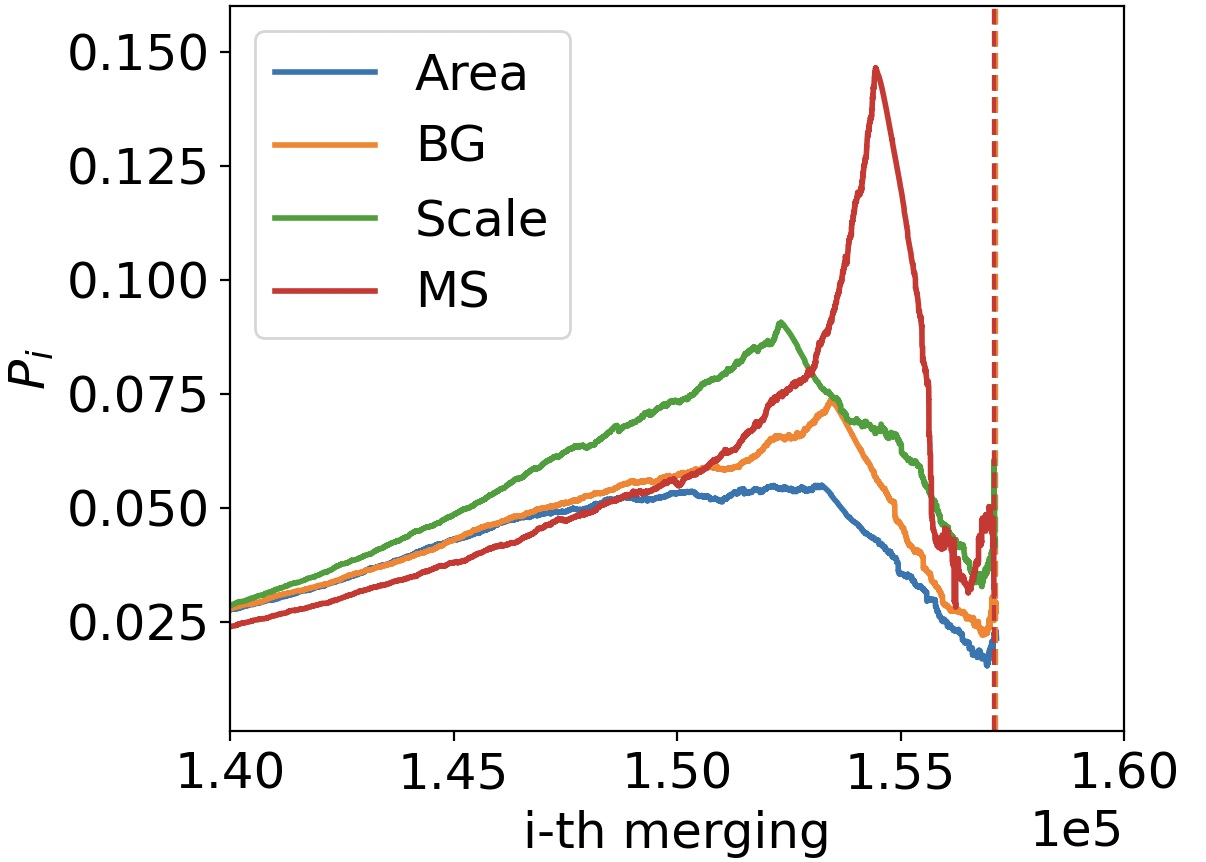}&
\includegraphics[width=0.3\textwidth]{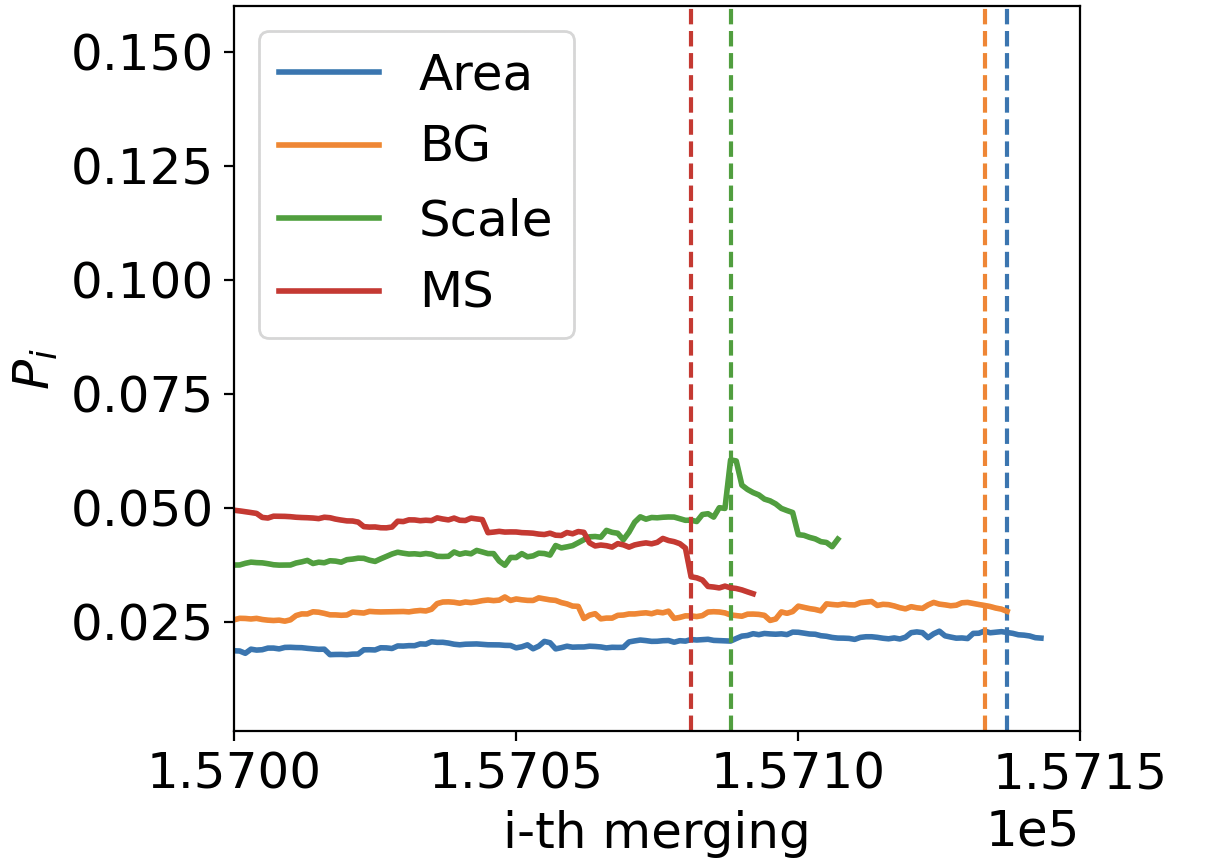}\\
\multicolumn{3}{c}{(e)}\\
\multicolumn{3}{c}{\begin{tabular}{c|cc|ccc}\hline
&\multicolumn{2}{c|}{before refine}& \multicolumn{3}{c}{after refine}\\\hline
Method&$N$& PSNR& $N$ (change)&PSNR (change) & Region Importance~\eqref{eq_sensitivity}\\\hline
Area&$100$&$\mathbf{25.82}$ &$93 (-7)$&$\mathbf{25.67}(-0.15)$& $2.14\times 10^{-2}$\\
BG&$100$& $25.62$&$95(-5)$&$25.52(-0.1)$& $2.00\times 10^{-2}$\\
Scale&$100$&$24.27$&$80(-20)$&$23.81(-0.46)$& $2.30\times 10^{-2}$\\
MS&$100$& $24.35$&$88(-12)$&$24.01(-0.34)$& $2.83\times 10^{-2}$\\\hline
\end{tabular}}
\end{tabular}
\caption{Analysis on the discrepancy of the number of regions $N$. Figure~\ref{fig_gain_compare1} (a) shows the input image and we set $N^*=100$ for all methods. (a) Variation of the number of regions ($N$) when alternating the Dual and Primal steps followed by the Refine step. See Algorithm~\ref{alg_pseudo}. (b) Evolution of the merging potential~\eqref{eq_potential} during the iteration. (c) Zoom-in the last thousands iterations of (b). (d)  Zoom-in the last hundreds iterations of (b). The dashed lines indicate when the number of regions $N=100$. (e) The compairson on PSNR and $N$  before and after the Refine step. The region importance~\eqref{eq_sensitivity} reflects the information carried by each region in the partition before the Refine step.}\label{fig_gain_compare3}
\end{figure}

We further investigate the  variations in $N$ when using different merging methods. In Figure~\ref{fig_gain_compare3} (a), we show   how $N$ changes after each major step of Algorithm~\ref{alg_pseudo} when fixing $N^*=100$. The results indicate that the Refine step is primarily responsible for the observed discrepancies in $N$. The Dual step reduces the number of regions by a fixed amount. The Primal step modifies the geometry of partition,  which may lead to small regions disappearing, although it does not result in significant changes in $N$ overall.  In the Refine step, any pairs of adjacent regions with merging costs smaller than $\lambda^*$ will be merged; here $\lambda^*$ is the largest cost incurred during the Dual steps. Therefore, the degree of reduction in $N$ depends on merging behaviors prior to the Refine step. 

To numerically characterize this effect, we define the \textbf{merging potential} at the $i$-th merging as follows:
\begin{equation}
P_i = \frac{\max_{j\leq i}\{\Delta E_j\}}{M_i}\;.\label{eq_potential}
\end{equation}
Here $\Delta E_j$ is the merging cost for the $j$-th merging, and $M_i$ is the average of the merging costs of all pairs of adjacent regions right after the $i$-th merging occurs. Particularly, suppose $i^*$ is the number of merging right before the Refine step, then $\max_{j\leq i^*}\{\Delta E_j\} = \lambda^*$. If $P_i$ is large, either the average merging cost becomes low or the maximal merging cost grows higher, and both scenarios lead to potentially more regions to be merged during the Refine step. In Figure~\ref{fig_gain_compare3} (b), we see that the merging potentials for all methods generally grow as the merging proceeds. After around $1.40\times10^{5}$ merging, we find in (c) that the merging potential for Area and BG grows slowly, that for Scale grows faster, and that for MS grows exponentially fast. Around the $1.53\times 10^{5}$-th merging, the merging potential for Scale reaches a peak; around the $1.54\times 10^{5}$-th merging, the potentials for BG and Area reach their respective peaks; and around the $1.55\times 10^{5}$-th merging, MS attains its highest peak. The potential drops following these peaks indicate that the average merging costs increase significantly. They correspond to phases when groups of regions with similar merging costs are being merged. Associated with different choices of gains, these merged regions have different geometric features compared to those merged before the peaks. Notice that the merging potentials slightly increase after the drops, indicating another phase of merging.  In (d), we focus on the last hundreds of merging, and the dashed lines indicate when the Refine steps start. In particular, we see that the merging potential of Scale is the highest,  MS is the second, BG is the third, and Area is the lowest. Except for BG, this aligns well with the rank of the numbers of regions reduced after refining. See Figure~\ref{fig_gain_compare3} (d).  the dashed lines indicate that MS reaches $N=100$ first, followed by Scale in second place, BG in third, and Area last. Given that the number of reduced regions from the Dual steps are fixed, these variances arise from the Primal steps where regions disappear due to the affine shortening flow. This is consistent with the fact that MS favors regions with shorter perimeters, which are more likely to vanish during the Primal steps.

In Figure~\ref{fig_gain_compare3} (d), we report $N$ and PSNR of the results before and after the Refine step for different merging criteria. We see that before refining, all methods yield the identical number of regions specified by $N^*$; and Area achieves the highest PSNR. After refining, both Scale and MS removes over $10$ regions resulting more reduction in the approximation accuracy. In contrast, both Area and BG retain better quality while eliminating fewer regions. To quantify the impact of the Refine step on the representation quality, we define the \textbf{region importance} as 
\begin{equation}
S = \frac{\text{Reduction in PSNR after Refine}}{\text{Reduction in }N\text{ after Refine}}\;.\label{eq_sensitivity}
\end{equation}
It can be interpreted as the reduced PSNR per removed region. We see that MS has the highest region importance among these criteria. It implies that each region produced by MS carries much information, thus removing any of them leads to significant quality degradation. With less restrictions on the shapes of contours, both BG and Area have higher tolerance on the region reduction.

\bibliographystyle{abbrv}
\bibliography{references}

\end{document}